\documentclass{article} 

\usepackage{minnesotanlp}
\makeatletter\let\evoduet@addcontentsline\addcontentsline\makeatother
\usepackage{iclr2027_conference,times}
\makeatletter\let\addcontentsline\evoduet@addcontentsline\makeatother
\usepackage[T1]{fontenc}

\usepackage{amsmath,amsfonts,bm}

\def\eqref#1{equation~\ref{#1}}

\def\1{\bm{1}}

\DeclareMathAlphabet{\mathsfit}{\encodingdefault}{\sfdefault}{m}{sl}
\SetMathAlphabet{\mathsfit}{bold}{\encodingdefault}{\sfdefault}{bx}{n}

\usepackage{hyperref}
\usepackage{url}
\usepackage{algorithm}
\usepackage{algpseudocode}
\usepackage{booktabs}
\usepackage{colortbl}      
\usepackage{adjustbox}
\usepackage{capt-of}       
\usepackage{arydshln}      
\usepackage{graphicx}
\usepackage{wrapfig}
\usepackage{xspace}
\usepackage{xcolor}        
\usepackage{tcolorbox}
\tcbuselibrary{breakable,listings,skins,raster}
\usepackage{nicematrix}    
\usepackage{enumitem}
\usepackage{subcaption}
\usepackage{needspace}

\definecolor{TaskBlue}{HTML}{315A85}
\definecolor{TaskTeal}{HTML}{267467}
\definecolor{TaskPlum}{HTML}{755184}
\definecolor{TaskAmber}{HTML}{976027}
\newtcolorbox{benchmarkexample}[2]{
    colback=#1!3!white,
    colframe=#1!70!white,
    colbacktitle=#1!12!white,
    coltitle=#1!75!black,
    title={#2},
    fonttitle=\bfseries\small,
    fontupper=\small,
    arc=2mm,
    boxrule=0.5pt,
    left=2mm, right=2mm, top=1.5mm, bottom=1.5mm,
    before skip=6pt, after skip=9pt
}

\definecolor{takeawaycolor}{HTML}{3F78C0}
\definecolor{takeawaypurple}{HTML}{7A5FC4}
\definecolor{takeawaybackground}{HTML}{EAF2FC}
\tcbset{
  takeawaybox/.style={
    width=\linewidth,
    boxrule=0.8pt,
    top=6pt, bottom=2pt, left=3pt, right=3pt,
    before skip=12pt, after skip=8pt,
    colback=takeawaybackground,
    colframe=takeawaycolor,
    colbacktitle=takeawaycolor,
    fonttitle=\bfseries\small,
    fontupper=\small,
    enhanced, breakable,
    attach boxed title to top left={yshift=-0.1in, xshift=0.15in},
    boxed title style={
      enhanced, boxrule=0pt, colframe=white,
      interior style={left color=takeawaycolor, right color=takeawaypurple},
    },
  }
}
\newtcolorbox{takeawaybox}[2][]{takeawaybox, title={#2}, #1}

\definecolor{LightBlue}{RGB}{220,230,250}
\definecolor{LightRed}{RGB}{250,220,220}

\newcommand{\dpos}[1]{\textcolor{green!55!black}{#1}}
\newcommand{\dneg}[1]{\textcolor{red!70!black}{#1}}

\colorlet{evoduetcolor}{takeawaycolor}
\colorlet{evoduetpurple}{takeawaypurple}
\colorlet{ourscolor}{evoduetcolor}

\newcommand{\evoduetcolor}{\rowcolor{evoduetcolor!20}}
\newcommand{\evoduetgradientrows}[2]{%
  \begin{tikzpicture}
    \foreach \gradientrow in {#1} {%
      \pgfmathtruncatemacro{\gradientnextrow}{\gradientrow+1}%
      \shade[left color=evoduetcolor!20, right color=evoduetpurple!20]
        (\gradientrow-|1) rectangle (\gradientnextrow-|#2);
    }
  \end{tikzpicture}%
}

\definecolor{Red}{rgb}{0.768, 0.054, 0.054}
\definecolor{Blue}{rgb}{0.152, 0.294, 0.925}
\definecolor{Green}{rgb}{0,0.4,0.7}

\usepackage{cleveref}
\crefname{section}{§}{§§}

\hypersetup{
    colorlinks=true,
    citecolor=teal,
    linkcolor=Red,
    urlcolor=Green,
}

\newtcolorbox{behaviorbox}{colback=gray!4!white, colframe=gray!55!white, boxrule=0.5pt, arc=1.5mm,
    left=2mm, right=2mm, top=1.2mm, bottom=1.2mm, before skip=6pt, after skip=8pt, fontupper=\small}
    
\DeclareMathOperator*{\argopt}{arg\,opt}

\newcommand{\eg}{e.g.,\xspace}
\newcommand{\ie}{i.e.,\xspace}

\newcommand{\methodName}{\textsc{EvoDuet}\xspace}
\definecolor{NoopViolet}{HTML}{7A5FC4}
\definecolor{LookupAmber}{HTML}{F0A32A}
\definecolor{RetrieveBlue}{HTML}{4F86CF}
\definecolor{NoopGray}{HTML}{6B7280}
\definecolor{LookupIndigo}{HTML}{3F51B5}
\definecolor{RetrieveChocolate}{HTML}{B5651D}
\newcommand{\NOOP}{{\protect\color{NoopViolet}\ensuremath{\overline{\underline{\textsc{no-op}}}}}\xspace}
\newcommand{\LOOKUP}{{\protect\color{LookupAmber}\ensuremath{\overline{\underline{\textsc{look-up}}}}}\xspace}
\newcommand{\RETRIEVE}{{\protect\color{RetrieveBlue}\ensuremath{\overline{\underline{\textsc{retrieve}}}}}\xspace}

\title{\methodName: Bilevel Co-Evolution of Web Searching and \\ Task Solving for Scientific Discovery}

\author{
\textbf{Young-Jun Lee}\textsuperscript{1} \quad
\textbf{Jinheon Baek}\textsuperscript{2}\thanks{Co-second authors with equal contribution, listed alphabetically by surname.} \quad
\textbf{Soyeong Jeong}\textsuperscript{2}\footnotemark[1] \quad
\textbf{Minki Kang}\textsuperscript{2}\footnotemark[1] \quad
\textbf{Seungyeon Jwa}\textsuperscript{1,3} \quad \\
\textbf{Jonghyun Choi}\textsuperscript{3} \quad
\textbf{Seungho Han}\textsuperscript{4} \quad
\textbf{Dongyeop Kang}\textsuperscript{1}
\\[0.6em]
\textsuperscript{1}University of Minnesota \quad
\textsuperscript{2}KAIST \quad
\textsuperscript{3}Seoul National University \quad
\textsuperscript{4}Hanyang University \\
\texttt{passing2961@gmail.com} \quad \texttt{dongyeop@umn.edu} 
\\[0.6em]
\href{https://open-galapagos.github.io/evoduet_project_page/}{%
    \raisebox{\dimexpr0.5ex-0.5\height\relax}{%
    \includegraphics[height=0.5cm]{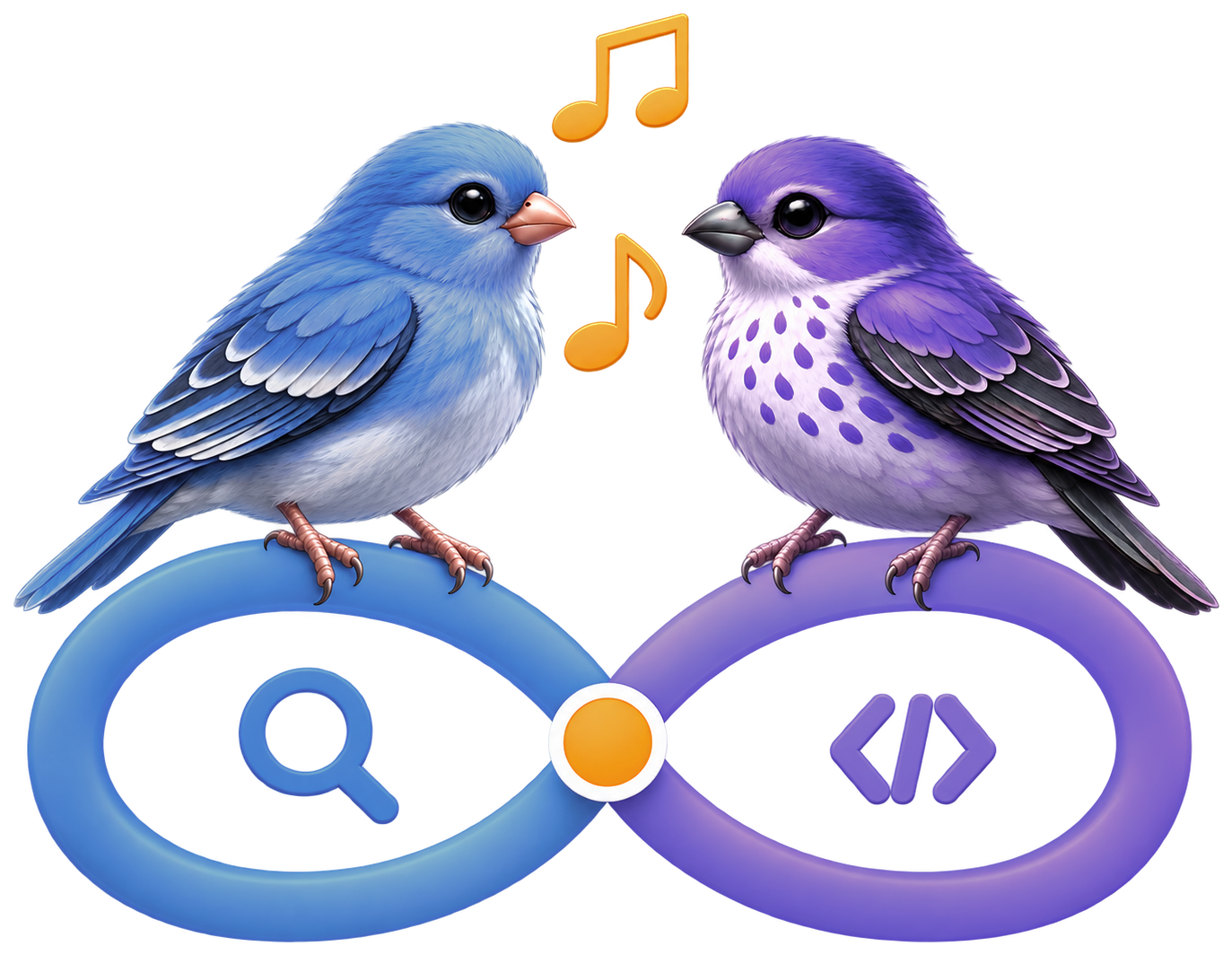}%
    }%
    \hspace{0.35em}%
    \raisebox{-0.15em}{\texttt{EvoDuet Project Page}}%
}}

\begin{document}

\addtocontents{toc}{\protect\setcounter{tocdepth}{-1}}

\maketitle

\begin{abstract}
    Evolutionary search with large language models (LLMs) can stall when progress requires external knowledge the model lacks.
    Supplying relevant documents helps, but simply adding web search tool can keep returning the same pages as solutions change.
    We introduce \textbf{\methodName}, a bi-level optimization method that co-evolves \textit{solutions} and \textit{search queries} with fixed model parameters.
    At each iteration, a retrieval gate lets the LLM assess its knowledge gap and choose to retrieve new documents, reuse stored ones, or proceed without them.
    An inner loop refines queries and ranks documents by the solution scores they are predicted to yield; an outer loop generates candidates in parallel from these documents and records the evaluated outcomes for later searches.
    Across 21 optimization tasks with one candidate per iteration, \methodName raises OpenEvolve's normalized discovery gain from 74.1\% to 78.0\% with GPT-5.6-Luna and from 61.3\% to 82.3\% with Gemini-3.8-Flash, whereas Qwen3.5-9B does not benefit.
    Our best runs surpass the previously reported best scores on eight tasks, including Swap Reduction on Q20 and Rosetta, and match them on three more.
    \methodName also improves with other scaffolds (e.g., Top-$K$, EvoX) on Sums/Diffs and Denoising, demonstrating its applicability across evolutionary search scaffolds.
\end{abstract}

\section{Introduction} \label{main_sec:intro}

Evolutionary search scaffolds driven by large language models (LLMs)~\citep{romera2024mathematical, novikov2025alphaevolve} have begun to discover new solutions to optimization tasks such as the Erd\H{o}s minimum-overlap problem~\citep{erdHos1955some}, GPU kernel design~\citep{ouyang2025kernelbench}, and single-cell RNA-seq denoising~\citep{yuksekgonul2026learning}. In these tasks, an evaluator can score any candidate solution, but the optimal solution cannot be computed directly~\citep{kirkpatrick1983optimization}.
Scaffolds therefore iterate: the LLM proposes candidates, the evaluator scores them, and high-scoring candidates seed the next iteration.

\begin{wrapfigure}{r}{0.33\linewidth}
    \vspace{-1.4em}
    \centering
    \includegraphics[width=0.99\linewidth]{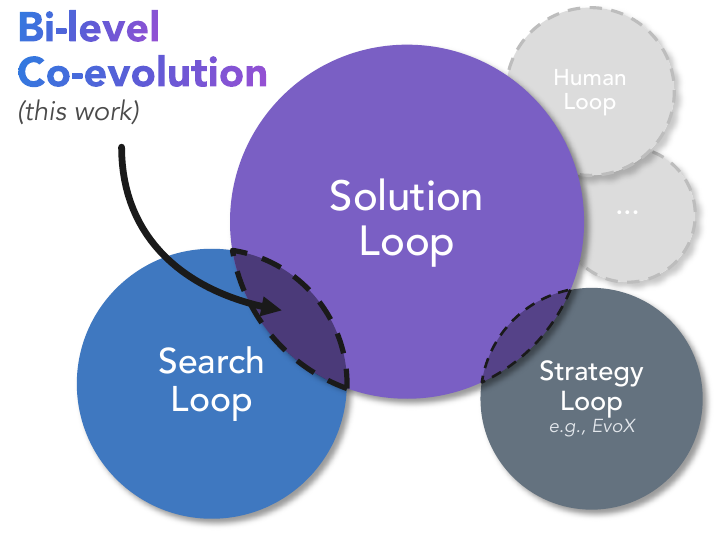}
    \caption{Packing loops of multiple optimization scaffolds}
    \label{main_fig:pack_loop}
    \vspace{-1.4em}
\end{wrapfigure}

Most scaffolds~\citep{openevolve,lange2025shinkaevolve,cemri2026adaevolve} run only this \emph{solution loop}.
Others keep the solution loop from saturating by wrapping additional loops around it, a design we call \textit{Loop Packing} (Figure~\ref{main_fig:pack_loop}).
For example, EvoX~\citep{liu2026evox} adds a strategy loop that revises the search strategy when progress stagnates. 
However, every loop packed so far draws on the same two sources of knowledge: the run's evolutionary history and the LLM's parametric knowledge.
The search is therefore \textbf{closed}, and it stalls when progress requires external knowledge that neither source contains (\eg how to use a newer library version). 
Human scientists, by contrast, consult the literature and search again as new knowledge gaps emerge.

Our preliminary analysis (\cref{main_subsec:web_search_helpful_analysis}) shows that external knowledge helps LLM-driven search, but not automatically.
Supplying task-relevant and helpful documents at every iteration (oracle documents) raises OpenEvolve's average Normalized Discovery Gain (NDG) on 31 tasks by 10.3\% with Qwen3.5-9B and 4.0\% with GPT-5.6-Luna. 
Two further observations show that turning such knowledge into progress takes more than a search tool: 
First, the LLM needs room to explore: on Sums/Diffs, oracle documents help only when the LLM generates eight candidates per iteration rather than one.
Second, queries must evolve with the solution: on Denoising, an LLM that searches inside the solution loop mostly retrieves pages it has already seen (88.1\% of returned URLs) and stops improving after 50 iterations. We therefore co-evolve web search queries with solutions (Figure~\ref{main_fig:pack_loop}), making the search \textbf{open}.

\begin{figure}[t]
    \centering
    \includegraphics[width=\linewidth]{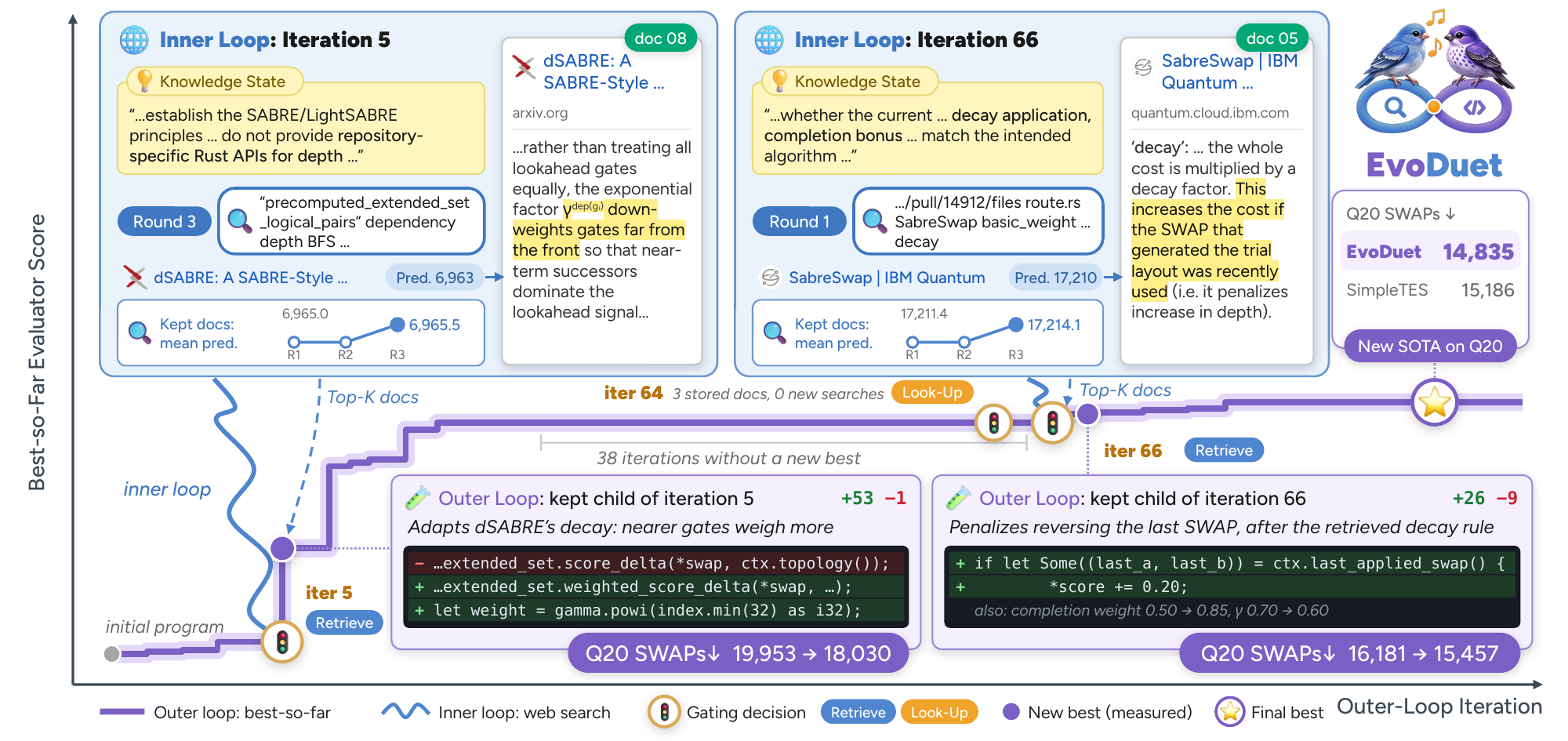}
    \caption{\textbf{\methodName co-evolves solutions and web queries.} The inner loop evolves queries; the outer loop uses retrieved documents to revise and evaluate solutions. The Swap Reduction example illustrates \emph{method transfer} (the most common of the six behaviors analyzed in Section~\ref{main_sec:behavior}): the model applies retrieved routing principles to prioritize gates closer to execution (iteration~5) and penalize reversing the last SWAP (iteration~66). At iteration~64, the gate reuses stored documents without a new web search. The best router found in this run uses 14,835 Q20 SWAPs, 2.31\% fewer than the released SimpleTES program~\citep{simpletes2026} (15,186) under the same evaluator.}
    \label{main_fig:teaser}
\end{figure}

We introduce \textbf{\methodName}, a bi-level optimization method that co-evolves solutions in an outer loop and web search queries in an inner loop, with the LLM's parameters fixed. A knowledge-gap-based retrieval gate connects the loops: at each iteration, the LLM assesses what it still needs to know to improve the current solution (its knowledge gap) and decides whether to retrieve new documents, reuse stored ones, or proceed without documents. When it retrieves, the inner loop constructs queries targeting the gap, searches the web, and ranks documents by \textit{hypothetical evidence scoring}, which predicts the evaluator score of a candidate built on each document. These predictions and the updated knowledge state steer the next round of queries without evaluating any candidate. Given documents, the outer loop generates candidates in parallel and records the documents with the evaluated outcome to guide later searches. Figure~\ref{main_fig:teaser} illustrates this interplay on Swap Reduction (quantum compilation).

We evaluate \methodName with OpenEvolve on 21 optimization tasks. With one candidate per iteration, it raises overall NDG from 74.1\% to 78.0\% with GPT-5.6-Luna and from 61.3\% to 82.3\% with Gemini-3.8-Flash, and parallel generation improves it further. 
Qwen3.5-9B does not benefit, even though oracle documents help it: it often leaves retrieved methods unused or misimplemented (26\% of sampled revisions), suggesting that co-evolution requires a model that can both find and apply evidence.
The best runs of \methodName surpass the previously reported best scores on eight tasks and match them on three more; on Denoising, \methodName surpasses SimpleTES~\citep{simpletes2026} at $6.9\times$ lower estimated cost.
It also improves other scaffolds, Top-$K$ and EvoX, on Sums/Diffs and Denoising, showing that the search loop can be packed alongside existing loops.

Finally, we analyze how retrieved documents shape the search. Candidates improve on their parent in 67.1\% of iterations that retrieve a document predicted to beat the parent, compared with 52.1\% of iterations without documents. The most common use of documents is \emph{method transfer}, in which the model applies a retrieved method to its solution (55 of 82 runs). Yet documents predicted to help can also yield much worse candidates (29 of 82 runs), so predicted gains must still be validated by the evaluator.

\section{Analysis of Web Search in Scientific Discovery} \label{main_sec:analysis_web_search}

\begin{figure}[!t]
    \centering
    \captionsetup[subfigure]{font={scriptsize},labelfont={bf},skip=2pt,position=bottom,justification=centering,singlelinecheck=false}
    \begin{subfigure}[t]{0.22\textwidth}
        \centering
        \includegraphics[width=0.9\linewidth]{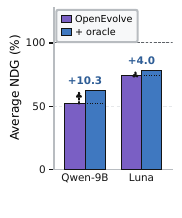}
        \caption{Average NDG ($\uparrow$)}
        \label{main_fig:avg_ndg_a}
    \end{subfigure}\hfill
    \begin{subfigure}[t]{0.36\textwidth}
        \centering
        \includegraphics[width=0.9\linewidth]{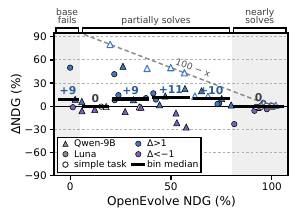}
        \caption{Gain vs.\ Headroom}
        \label{main_fig:gain_headroom_b}
    \end{subfigure}\hfill
    \begin{subfigure}[t]{0.38\textwidth}
        \centering
        \includegraphics[width=0.9\linewidth]{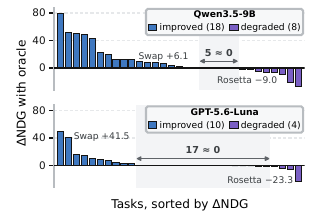}
        \caption{$\Delta$NDG per task ($\uparrow$)}
        \label{main_fig:delta_ndg_per_task_c}
    \end{subfigure}
    \vspace{-0.4em}
    \caption{\textbf{OpenEvolve with and without oracle documents} on 31 optimization tasks using Qwen3.5-9B and GPT-5.6-Luna. (a) Average NDG across tasks. (b) Per-task $\Delta$NDG versus baseline NDG, with bin medians pooled across models. Shading marks baseline NDG below 5\% or at least 80\%; hollow markers denote simple tasks. The dashed line shows the gain needed to reach 100\% NDG. (c) Per-task $\Delta$NDG in descending order. Legends count tasks with NDG increases (improved) or decreases (degraded) of more than 1\%; shading indicates changes within $\pm1\%$.}
    \label{main_fig:default_vs_default_oracle}
\end{figure}

\subsection{Preliminaries: Evolutionary Search Scaffold}
\label{main_sec:prelim}

An optimization task $\tau=(I,x_0,\mathcal{E})$ consists of an instruction $I$, an initial solution $x_0$, and a deterministic, verifiable evaluator $\mathcal{E}(x)\in\mathbb{R}$. The goal of an evolutionary search scaffold is to find $x^{\star}=\argopt_{x\in\mathcal{X}}\mathcal{E}(x)$, where $\mathrm{opt}\in\{\max,\min\}$ and $\mathcal{X}=\{x_0,\ldots,x_T\}$ is the set of solutions explored within a search budget of $T$. The task determines the optimization direction (\eg minimizing the $c_5$ bound in the Erdős minimum overlap problem). At each iteration, the evolutionary search scaffold~\citep{romera2024mathematical, novikov2025alphaevolve, openevolve, liu2026evox} selects a parent solution $x_t$ and its evolutionary history $\mathcal{H}_t$ from its solution database (\textit{a.k.a.} population) $\mathcal{D}_p$ under a policy $\phi$, forming the context $c_t=[\mathcal{H}_t;x_t]$. A frozen LLM $\mathcal{M}_\theta$, serving as the mutation operator, generates a candidate solution $x_{t+1}\sim\mathcal{M}_\theta(I,c_t)$, which the evaluator $\mathcal{E}$ then scores. Based on the selection policy $\phi$, the candidate solution $x_{t+1}$ is then either added to the population $\mathcal{D}_p$ or discarded. We call $x^{\star}$ a \textit{discovery} if it improves upon the previous best-known solution $x_{\mathrm{sota}}$, \ie $\mathcal{E}(x^{\star})>\mathcal{E}(x_{\mathrm{sota}})$ for maximization tasks, with the inequality reversed for minimization tasks.

\subsection{Experimental Setup}

\noindent \textbf{Tasks.} We evaluate on 31 optimization tasks spanning simple scientific optimization (10), quantum compilation (1), astrodynamics (5), scientific algorithms (1), AI foundations (4), mathematics (8), and algorithm engineering (2). Detailed task descriptions are provided in Appendix~\ref{supp_sec:eval_task_details}.

\noindent \textbf{Baselines.} We use Qwen3.5-9B~\citep{qwen35blog} and GPT-5.6-Luna~\citep{openai2026gpt56} as the LLM $\mathcal{M}_{\theta}$ in OpenEvolve. For each model, we compare (1) OpenEvolve and (2) OpenEvolve with oracle documents supplied to the LLM at every iteration $t$. Appendix~\ref{supp_sec:oracle_setup} describes how we construct the oracle documents. For robustness, we run each baseline with four different random seeds.

\noindent \textbf{Evaluation Metric.} Since evaluator score ranges vary across tasks, we report \textit{Normalized Discovery Gain} (NDG) to compare optimization progress on a common scale. NDG measures the percentage of the gap between the initial and SOTA scores closed by a run: $\mathrm{NDG}=\frac{\mathcal{E}(x_{\mathrm{best}})-\mathcal{E}(x_0)}{\mathcal{E}(x_{\mathrm{SOTA}})-\mathcal{E}(x_0)}\times 100$, where $x_{\mathrm{best}}$ is the best-scoring solution in $\mathcal{X}$, and $x_{\mathrm{SOTA}}$ is the best program reported in prior work~\citep{simpletes2026}. Higher NDG indicates greater progress toward SOTA performance.

\subsection{Does web search help scientific discovery?} \label{main_subsec:web_search_helpful_analysis}

\noindent \textbf{Oracle documents improve optimization on average, but their benefits vary across models and tasks.} As shown in Figure~\ref{main_fig:avg_ndg_a}, oracle documents raise the average NDG of Qwen3.5-9B by 10.3\% and that of GPT-5.6-Luna by 4.0\%. The smaller gain for GPT-5.6-Luna likely reflects limited headroom: OpenEvolve alone already reaches at least 95\% NDG on 16 of the 31 tasks. On these tasks, oracle documents change NDG by only 0.1\% on average (Figure~\ref{main_fig:gain_headroom_b}). Although oracle documents lead to 100\% NDG primarily on simple optimization tasks, their benefits also extend to challenging tasks. For example, on Swap Reduction (quantum compilation), oracle documents increase NDG by 6.1\% for Qwen3.5-9B and 41.5\% for GPT-5.6-Luna (Figure~\ref{main_fig:delta_ndg_per_task_c}). However, these benefits are not universal: oracle documents lower NDG by more than 1\% on 8 and 4 of the 31 tasks for Qwen3.5-9B and GPT-5.6-Luna, respectively, including decreases of 9.0\% and 23.3\% on Rosetta. Full results are provided in Appendix~\ref{supp_sec:oracle_results}.

\begin{figure}[!t]
    \centering
    \captionsetup[subfigure]{font={scriptsize},labelfont={bf},skip=2pt,position=bottom,justification=centering,singlelinecheck=false}
    \begin{subfigure}[t]{0.225\textwidth}
        \centering
        \includegraphics[width=0.9\linewidth]{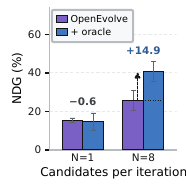}
        \caption{Parallel generation ($\uparrow$)}
        \label{main_fig:search_analysis_a}
    \end{subfigure}\hfill
    \begin{subfigure}[t]{0.225\textwidth}
        \centering
        \includegraphics[width=0.9\linewidth]{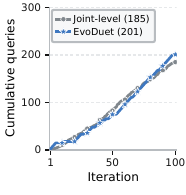}
        \caption{Cumulative search queries}
        \label{main_fig:search_analysis_b}
    \end{subfigure}\hfill
    \begin{subfigure}[t]{0.225\textwidth}
        \centering
        \includegraphics[width=0.9\linewidth]{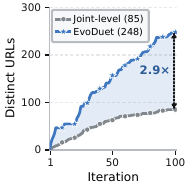}
        \caption{Distinct URLs returned ($\uparrow$)}
        \label{main_fig:search_analysis_c}
    \end{subfigure}\hfill
    \begin{subfigure}[t]{0.225\textwidth}
        \centering
        \includegraphics[width=0.9\linewidth]{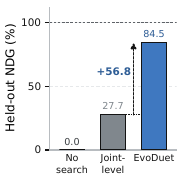}
        \caption{Held-out NDG ($\uparrow$)}
        \label{main_fig:search_analysis_d}
    \end{subfigure}
    \vspace{-0.4em}
    \caption{\textbf{Effects of parallel generation and query evolution with GPT-5.6-Luna.} (a) NDG of OpenEvolve on Sums/Diffs (mathematics) with and without oracle documents, using $N=1$ or $N=8$ candidates per iteration. (b) Cumulative number of search queries. (c) Number of distinct URLs returned. (d) Held-out NDG of the best program for each method.}
    \label{main_fig:search_analysis}
\end{figure}

\noindent \textbf{LLMs may need more opportunities to benefit from oracle documents.} The mixed effects of oracle documents raise a question: does the LLM lack the ability to use oracle documents, or does it lack sufficient opportunities to act on them? To examine this, we generate $N$ candidates in parallel at each iteration $t$ from the same context $c_t$. On Sums/Diffs, oracle documents offer no improvement at $N=1$ (14.8\% NDG with oracle documents versus 15.4\% without), but increase NDG from 25.8\% to 40.7\% at $N=8$ (Figure~\ref{main_fig:search_analysis_b}). This suggests that the LLM can benefit from oracle documents when given more opportunities to explore strategies grounded in them. Additional results are provided in Appendix~\ref{supp_sec:parallel_generation_results}.

\noindent \textbf{Without query evolution, in-loop web search repeatedly retrieves the same pages and stalls.} On Denoising with GPT-5.6-Luna, we equip OpenEvolve with Tavily~\footnote{\url{https://www.tavily.com/}} as an in-loop search tool (\textit{joint-level}), allowing the LLM to construct its own queries during solution optimization. We compare this baseline with \methodName, which evolves queries based on the LLM's knowledge gap through bi-level co-evolution. Over 100 iterations, joint-level search uses the tool in 99 iterations and produces 185 queries, compared with 201 for \methodName (Figure~\ref{main_fig:search_analysis_b}). However, it retrieves only 85 distinct URLs, compared with 248 for \methodName (Figure~\ref{main_fig:search_analysis_c}); 88.1\% of its returned URLs have appeared before, compared with 62.4\% for \methodName. Moreover, joint-level search stops improving after iteration 50, with its best program achieving a held-out NDG of 27.7\%, compared with 84.5\% for \methodName (Figure~\ref{main_fig:search_analysis_d}). These results suggest that evolving what to ask based on the LLM's knowledge gap helps it search for external knowledge more effectively, enabling it to find better solutions.

\begin{takeawaybox}{Takeaways: Web search benefits from broader exploration and query evolution}
\begin{itemize}[leftmargin=*, itemsep=1pt, topsep=2pt]
    \item Oracle documents increase average NDG across the 31 tasks, with a larger gain for the smaller model (+10.3\% vs.\ +4.0\%).
    \item More opportunities for exploration can help LLMs use oracle documents: on Sums/Diffs, oracle documents improve NDG by 14.9\% with eight parallel candidates per iteration.
    \item Query evolution helps sustain progress: with comparable query counts on Denoising, joint-level search repeatedly retrieves the same pages and stalls, whereas bi-level search, which evolves queries based on the LLM's knowledge gap, retrieves about $2.9\times$ as many distinct pages and increases NDG by 56.8\%.
\end{itemize}
\end{takeawaybox}

\section{\methodName} \label{main_sec:method}

\begin{figure}[!t]
    \centering
    \includegraphics[width=\linewidth]{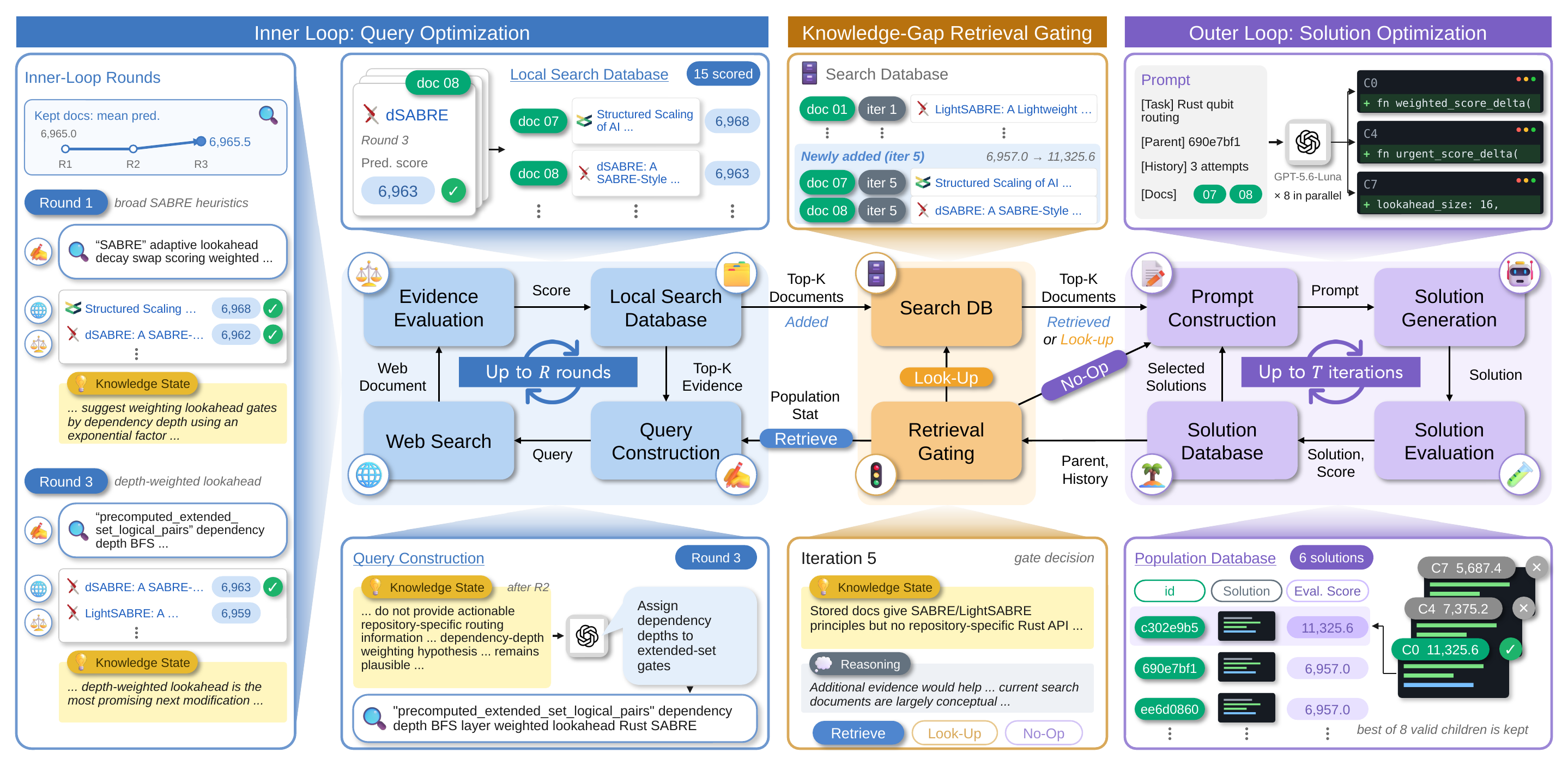}
    \caption{\textbf{Overview of \methodName.} The outer loop evolves solutions, while the inner loop evolves web search queries using predicted solution scores. A knowledge-gap-based retrieval gate links the two loops by deciding whether to retrieve new documents (\RETRIEVE), reuse stored documents (\LOOKUP), or proceed without documents (\NOOP).}
    \label{main_fig:evoduet_method_overview}
\end{figure}

\subsection{An Overview of \methodName} \label{main_sec:method_overview}

We formulate discovery as a bi-level optimization problem with an outer loop that evolves solutions and an inner loop that evolves queries for web search.  As shown in Figure~\ref{main_fig:evoduet_method_overview}, \methodName has three main components:
\begin{itemize}[leftmargin=2em, itemsep=1pt, parsep=0pt, topsep=1pt]
    \item \textbf{Solution Optimization Loop (outer, \cref{main_sec:outer_loop}):} The outer loop follows the scaffold's evolutionary loop (\cref{main_sec:prelim}) to evolve solutions. At iteration $t$, the selection policy $\phi$ selects a parent solution $x_t$ and its evolutionary history $\mathcal{H}_t$ from the population $\mathcal{D}_e$, and the LLM $\mathcal{M}_\theta$ generates a candidate solution $x_{t+1}\sim\mathcal{M}_\theta(I,c_t,\mathcal{S}_t)$, where $\mathcal{S}_t$ is either empty or a set of retrieved web documents. 
    
    \item \textbf{Knowledge Gap-based Retrieval Gating (\cref{main_sec:retrieval_gating}):} This component connects the outer and inner loops, enabling their co-evolution. Based on gaps in the LLM $\mathcal{M}_\theta$'s current knowledge status, it determines whether to invoke the inner loop for web search.
    
    \item \textbf{Query Optimization Loop (inner, \cref{main_sec:inner_loop}):} The inner loop evolves web search queries to retrieve documents $\mathcal{S}_t$ that provide the external knowledge needed to generate an improved candidate solution $x_{t+1}$.
\end{itemize}
Both loops share the task objective $\mathcal{E}$, with query quality defined by the performance of the candidate solution generated from the retrieved documents. Let $\mathcal{S}_t(q)$ denote the documents retained from query $q$ and $x_{t+1}(q)\sim\mathcal{M}_\theta(I,c_t,\mathcal{S}_t(q))$ the resulting candidate. The ideal bi-level optimization is formulated as
\begin{equation} \label{eq:bilevel}
    \underbrace{x^{\star}=\argopt_{x\in\mathcal{X}}\mathcal{E}(x)}_{\text{outer: solution optimization}}
    \quad\text{s.t.}\quad
    \underbrace{q_t^{\star}=\argopt_{q\in\mathcal{Q}_t}\mathcal{E}\bigl(x_{t+1}(q)\bigr)}_{\text{inner: query optimization}},
    \qquad
    x_{t+1}\sim\mathcal{M}_\theta\bigl(I,c_t,\mathcal{S}_t(q_t^{\star})\bigr),
\end{equation}
where $\mathcal{X}=\{x_0,\ldots,x_T\}$ is the set of solutions explored within budget $T$, and $\mathcal{Q}_t$ is the set of admissible queries at iteration $t$. However, the inner objective $\mathcal{E}(x_{t+1}(q))$ is observable only after the corresponding candidate has been generated and evaluated. Repeating this process for every query within the inner loop would be costly. We therefore approximate the inner optimization in Eq.~\ref{eq:bilevel} using $\mathcal{M}_\theta$ to \textbf{predict the candidate score} that the retrieved documents would yield. Denoting this score predictor by $\widehat{\mathcal{E}}_\theta$, the approximate inner optimization can then be reformulated as $\hat q_t=\argopt_{q\in\mathcal{Q}_t}\widehat{\mathcal{E}}_\theta\bigl(I,c_t,\mathcal{S}_t(q)\bigr)$. The detailed algorithm is presented in Algorithm~\ref{alg:method}.

\subsection{Outer Loop: Solution Optimization} \label{main_sec:outer_loop}
The outer loop follows the scaffold's typical evolutionary procedure (\cref{main_sec:prelim}), but differs in the inputs provided to the LLM $\mathcal{M_\theta}$ during candidate generation. When web documents $\mathcal{S}_t$ are provided, the LLM $\mathcal{M_\theta}$ generates a candidate solution $x_{t+1}\sim\mathcal{M}_\theta(I,c_t,\mathcal{S}_t)$, which is then evaluated by $\mathcal{E}$ and retained or discarded according to $\phi$. The web documents $\mathcal{S}_t$ are stored in the search database $\mathcal{D}_{\mathrm{s}}$ together with the candidate's actual evaluator score $\mathcal{E}(x_{t+1})$. This stored information is used to decide whether to reuse existing documents in $\mathcal{D}_{\mathrm{s}}$ through a search database $\mathcal{D}_s$ lookup, retrieve additional knowledge from the web, or skip the inner loop.

\paragraph{Gated parallel candidate solution generation.}
In Section~\ref{main_sec:analysis_web_search}, we observe that parallel scaling with rich oracle documents enables the LLM $\mathcal{M}_\theta$ to explore diverse approaches and find better solutions. Motivated by this observation, \methodName generates $N_t$ candidates in parallel from the same input prompt, with $N_t=N$ for \LOOKUP or \RETRIEVE and $N_t=1$ for \NOOP. Candidate generation and selection are formulated as
\begin{equation}
x_{t+1}^{(n)} \sim \mathcal{M}_\theta\bigl(I,c_t,\mathcal{S}_t\bigr),
\quad n=1,\ldots,N_t,
\qquad
x_{t+1} = \argopt_{x\in\mathcal{C}_t^{\mathrm{valid}}}\mathcal{E}(x),
\end{equation}
where $\mathcal{C}_t^{\mathrm{valid}}$ is the set of valid candidates (\ie those that encounter no errors, such as evaluator errors) among the $N_t$ generated solutions. Each candidate is evaluated, and only the best valid solution is passed to $\phi$ as $x_{t+1}$.

\subsection{Knowledge-Gap-Based Retrieval Gating} \label{main_sec:retrieval_gating}

The gate allows $\mathcal{M}_\theta$ to determine \textit{when} to search by assessing whether it needs new external knowledge to improve the current solution $x_t$, can reuse information stored in $\mathcal{D}_{\mathrm{s}}$, or can evolve the solution using only its internal knowledge. At each iteration, given the context $c_t$ and the documents most recently stored in $\mathcal{D}_{\mathrm{s}}$, $\mathcal{M}_\theta$ outputs its current knowledge state $K_t$ and a retrieval decision $g_t$:
\begin{equation}
\label{eq:state_gate}
(K_t,g_t)=\textsc{Gate}_{\mathcal{M}_\theta}\bigl(c_t,\mathcal{D}_{\mathrm{s}}\bigr),
\qquad
g_t\in\{\NOOP,\LOOKUP,\RETRIEVE\}.
\end{equation}
The knowledge state $K_t$ distinguishes the model's existing knowledge, findings from previous searches and experiments, and unresolved questions about $x_t$ that define the current knowledge gap. The gate selects \NOOP when the model's knowledge suffices, \LOOKUP when stored documents provide the missing knowledge, and \RETRIEVE when neither source is sufficient. Under \LOOKUP, the selected documents are included in the solution prompt without a new web search. This makes retrieval responsive to the current knowledge gap and enables the reuse of previously retrieved evidence. The results in Table~\ref{main_tab:ablation_gating} show that knowledge-gap-based gating is more effective than the heuristic gating based on stalled progress used in EvoX~\citep{liu2026evox}.

\subsection{Inner Loop: Query Optimization} \label{main_sec:inner_loop}

When $g_t=\RETRIEVE$, the inner loop approximates the query optimization in Eq.~\ref{eq:bilevel} over $R$ inner rounds within the current outer iteration $t$, without generating or evaluating candidate solution.

\noindent \textbf{Population state descriptor.}
Following EvoX~\citep{liu2026evox}, the inner loop first computes a population state descriptor, $\mathrm{stats}(\mathcal{D}_p)$, comprising deterministic statistics of the retained population, its score distribution, recent parent-to-child outcomes, and parent-selection frequencies. $\mathcal{M}_\theta$ summarizes these statistics into factual observations $A_t=\mathcal{M}_\theta(\mathrm{stats}(\mathcal{D}_p))$ without additional interpretation. These observations complete the initial inner-loop context $\tilde{c}_t^{0}=(c_t,K_t,A_t)$, which the inner rounds condition on and update.

\noindent \textbf{Iterative query optimization.}
\label{sec:evidence}
Let $\mathcal{S}_t^r$ denote the documents retained after inner round $r$, and $\hat{s}_t(d)$ the predicted evaluator score of a candidate $x_{t+1}$ obtained by improving $x_t$ using document $d$ alone. Starting with $\mathcal{S}_t^{0}=\emptyset$ and $K_t^{0}=K_t$, each round $r=1,\ldots,R$ performs four operations:
\begin{enumerate}[label=(\arabic*), leftmargin=2em, itemsep=1pt, parsep=0pt, topsep=1pt]
    \item \textbf{Query Construction:} The LLM $\mathcal{M_\theta}$ constructs $J$ queries targeting the remaining knowledge gaps. The query batch is $Q_r=(q_r^{(j)})_{j=1}^{J}$, with each query sampled as $q_r^{(j)}\sim\mathcal{M}_\theta(\tilde{c}_t^{r-1},\mathcal{S}_t^{r-1},\{\hat{s}_t(d)\}_{d\in\mathcal{S}_t^{r-1}})$. At $r=1$, the document and score inputs are empty, so query construction uses only the initial context $\tilde{c}_t^0$. Later rounds also use the retained documents and their predicted scores.

    \item \textbf{Web Search:} Each distinct query in $Q_r$ is executed once per round. The returned documents are combined with the previously retained documents to form the pool (\ie local search database in Figure~\ref{main_fig:evoduet_method_overview}) $\mathcal{P}_r=\mathcal{S}_t^{r-1}\cup\textsc{Search}(Q_r)$.

    \item \textbf{Hypothetical Evidence Scoring:} When unscored documents are available, a single call to $\mathcal{M}_\theta$ predicts $\hat{s}_t(d)=\mathcal{M}_\theta(\tilde{c}_t^{r-1},\mathcal{E}(x_t),d)$ for each such $d\in\mathcal{P}_r$. These \textit{hypothetical absolute scores} are expressed on the evaluator's scale and provide a surrogate signal for the inner objective in Eq.~\ref{eq:bilevel} (Figure~\ref{main_fig:behavior_audit_a} shows the model's ability to predict these scores). Each document is scored only once within the inner loop, and its score is reused in later rounds.

    \item \textbf{Knowledge State Update:} The same call used for hypothetical evidence scoring also updates the knowledge state to $K_t^r$ using the new documents, yielding the next round's context $\tilde{c}_t^r=(c_t,K_t^r,A_t)$. If no documents require scoring, the call is skipped and $K_t^r=K_t^{r-1}$. The loop then retains up to $D$ documents with the highest predicted scores as $\mathcal{S}_t^{r}=\mathrm{TopD}_{\hat{s}_t}(\mathcal{P}_r)$.
\end{enumerate}
After all $R$ rounds, the retained Top-$D$ documents $\mathcal{S}_t=\mathcal{S}_t^{R}$ are passed to the outer loop.

\section{Experimental Results}

\begin{figure*}[t]
    \centering
    \captionsetup[subfigure]{font={scriptsize},labelfont={bf},skip=2pt,position=bottom,justification=centering,singlelinecheck=false}
    \begin{subfigure}[t]{0.32\textwidth}
        \centering
        \includegraphics[width=0.9\linewidth]{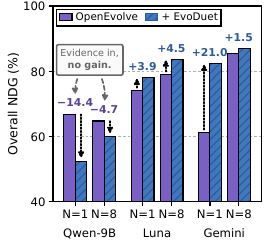}
        \caption{Overall NDG ($\uparrow$)}
        \label{main_fig:ndg_panel_a}
    \end{subfigure}\hfill
    \begin{subfigure}[t]{0.32\textwidth}
        \centering
        \includegraphics[width=0.9\linewidth]{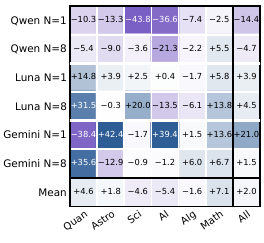}
        \caption{Group $\Delta$NDG ($\uparrow$)}
        \label{main_fig:ndg_panel_b}
    \end{subfigure}\hfill
    \begin{subfigure}[t]{0.32\textwidth}
        \centering
        \includegraphics[width=0.9\linewidth]{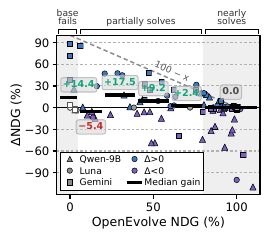}
        \caption{Gain vs. Headroom}
        \label{main_fig:ndg_panel_c}
    \end{subfigure}
    \vspace{-0.2em}
    \caption{\textbf{Discovery performance of OpenEvolve with and without \methodName.} Results cover 21 paired tasks with $N=1$ or $8$ candidates per iteration. For each model, method, task, and $N$, we report the best NDG across four runs. \textbf{(a)}~Mean best NDG across tasks and the gain from adding \methodName ($\Delta$NDG). \textbf{(b)}~Mean $\Delta$NDG within each task group. \textbf{(c)}~Task-level $\Delta$NDG versus OpenEvolve's NDG; black bars and labels show the median gain within each NDG range.}
    \label{main_fig:ndg_main_panels}
\end{figure*}

\noindent \textbf{\methodName improves discovery, but only for LLMs that can exploit retrieved evidence.}
At $N=1$, Figure~\ref{main_fig:ndg_panel_a} shows that \methodName improves OpenEvolve's overall NDG from 74.1\% to 78.0\% (+3.9\%) with GPT-5.6-Luna and from 61.3\% to 82.3\% (+21.0\%) with Gemini-3.8-Flash. Qwen3.5-9B, however, shows a 14.4\% decline at $N=1$ and still loses 4.7\% with parallel generation at $N=8$, despite gaining 10.3\% when given oracle documents (Figure~\ref{main_fig:avg_ndg_a}). Unlike the oracle setting, \methodName requires the model to decide when and what to search for based on its own knowledge gaps. This contrast suggests that these additional demands may limit the benefits of retrieval for weaker backbones such as Qwen3.5-9B. Inspection of 50 randomly sampled Qwen3.5-9B revisions with documents at $N=1$ identifies unused methods in 6\% and incorrect implementations in 20\% (26\% combined; Appendix~\ref{supp_sec:model_failures}). These findings suggest that effective bi-level co-evolution depends on the model's ability to both retrieve and apply relevant evidence.

\begin{wrapfigure}{r}{0.275\textwidth}
    \vspace{-0.8em}
    \centering
    \includegraphics[width=0.9\linewidth]{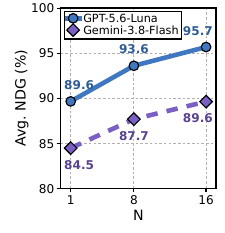}
    \captionsetup{font=small,skip=3pt}
    \caption{\textbf{Scaling effect of $N$.}}
    \label{main_fig:evoduet_candidate_budget}
    \vspace{-1em}
\end{wrapfigure}

\noindent \textbf{Parallel generation consistently improves discovery across models.}
Figure~\ref{main_fig:ndg_panel_a} shows that increasing the candidate budget from $N=1$ to $N=8$ improves \methodName's overall NDG for all three models. For Qwen3.5-9B, although \methodName remains below the corresponding OpenEvolve baseline at both budgets, its overall NDG improves from $N=1$ to $N=8$, indicating that parallel generation also benefits this model. We further compare $N\in\{1,8,16\}$ for GPT-5.6-Luna and Gemini-3.8-Flash on 8 tasks in Figure~\ref{main_fig:evoduet_candidate_budget}. We observe that the average NDG increases monotonically with $N$ for both models, rising from 89.6\% to 95.7\% for GPT-5.6-Luna and from 84.5\% to 89.6\% for Gemini-3.8-Flash. These results suggest that broader exploration guided by web documents helps LLMs discover better solutions.

\noindent \textbf{\methodName achieves its largest average gain in mathematics.}
As shown in Figure~\ref{main_fig:ndg_panel_b}, \methodName improves NDG over OpenEvolve on mathematics tasks in five of the six model/budget settings, with an average gain of 7.1\% across all six. At $N=8$, Qwen3.5-9B gains 5.5\% despite its negative overall gain, and Gemini-3.8-Flash gains 6.7\%.

\noindent \textbf{\methodName discovers novel solutions across 11 different tasks.}
Table~\ref{main_tab:new_sota_result} shows that \methodName surpasses the previous SOTA program scores on eight tasks across five domains and matches them on three more. The mean cost across the eleven reported runs is \$45.56. Interestingly, on Rosetta and Erd\H{o}s, the model initially evolves solutions, then retrieves public constructions at a later iteration and uses them as starting points for further optimization: refining a published trajectory and locally optimizing a published construction ($C_5: 0.380859 \rightarrow 0.380859$), respectively. We call this behavior \textit{public artifact reuse}, one of six behaviors reported in Appendix~\ref{supp_sec:behavior_examples}. This behavior occurs in only 6 of 82 GPT-5.6-Luna runs (7.3\%), compared with \textit{method transfer} (55 of 82), in which the model applies principles or methods retrieved by searching relevant literature (Figure~\ref{main_fig:behavior_audit_c}). Only two of these 11 SOTA results reuse public artifacts in their best programs (in Appendix~\ref{supp_sec:best_programs}).

\noindent \textbf{The largest median gain occurs on partially solved tasks.}
Figure~\ref{main_fig:ndg_panel_c} compares OpenEvolve and \methodName across 126 combinations of tasks, models, and candidate counts. When OpenEvolve's NDG is 5--20\%, 20--40\%, 40--60\%, and 60--80\%, the median gains from \methodName are -5.4\%, +17.5\%, +9.2\%, and +2.4\%, respectively. When OpenEvolve already performs well (NDG $\geq$80\%), the median gain is 0.0\% across 75 comparisons. Of the 52 comparisons with OpenEvolve NDG $\geq$95\%, six show declines greater than 5\%. When OpenEvolve makes little progress (NDG below 5\%), \methodName raises five of the ten cases above 5\%, with a median gain of +71.8\% among these five. In the remaining five cases, both methods stay near the initial program's performance. Overall, \methodName tends to provide smaller gains when OpenEvolve's NDG is higher (Pearson $r=-0.41$).

\section{Analysis and Discussions}

\subsection{How Does Web Search Help?}
\label{main_sec:behavior}

\begin{figure}[t]
    \centering
    \captionsetup[subfigure]{font={scriptsize},labelfont={bf},skip=2pt,position=bottom,justification=centering,singlelinecheck=false}
    \begin{subfigure}[t]{0.25\linewidth}
        \centering
        \includegraphics[width=0.9\linewidth]{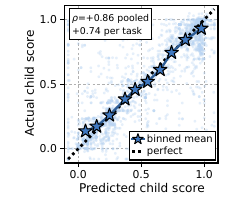}
        \caption{Predicted vs.\ evaluated scores}
        \label{main_fig:behavior_audit_a}
    \end{subfigure}\hfill
    \begin{subfigure}[t]{0.365\linewidth}
        \centering
        \includegraphics[width=0.9\linewidth]{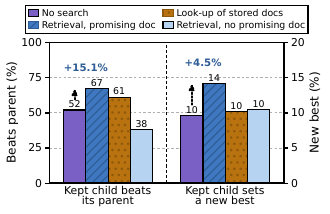}
        \caption{Iteration improvements}
        \label{main_fig:behavior_audit_b}
    \end{subfigure}\hfill
    \begin{subfigure}[t]{0.365\linewidth}
        \centering
        \includegraphics[width=0.9\linewidth]{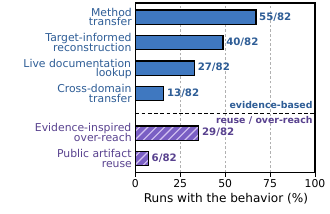}
        \caption{Six patterns of document use}
        \label{main_fig:behavior_audit_c}
    \end{subfigure}
    \vspace{-0.4em}
    \caption{ \textbf{Program improvement with web documents and patterns of document use.} Results cover 8,200 iterations from 82 GPT-5.6-Luna runs across 21 tasks. \textbf{(a)}~Predicted versus evaluated candidate scores, scaled to $[0,1]$ using each task's 2nd and 98th percentiles. \textbf{(b)}~Rates at which the selected candidate improves on its parent (left axis) or sets a new best score within the run (right axis), grouped by gate decision. Arrows compare \NOOP iterations with \RETRIEVE iterations that retain at least one promising document. \textbf{(c)}~Percentage of runs exhibiting each behavior. Each run is counted at most once per category and may appear in multiple categories.}
    \label{main_fig:behavior_audit}
\end{figure}

\noindent \textbf{Hypothetical evidence scores track actual program performance.}
Across 4,324 retrievals, hypothetical evidence scores correlate strongly with the selected candidates' evaluated scores (Spearman $\rho=+0.86$; Figure~\ref{main_fig:behavior_audit_a}). Correlations are positive on all 21 tasks, with a median of $+0.74$. This supports using these scores as a surrogate signal for inner-loop query optimization (\cref{main_sec:inner_loop}).

\noindent \textbf{Programs improve more often with promising documents.}
A document is \textit{promising} when its hypothetical evidence score exceeds the parent's actual evaluator score, \ie $\hat{s}_t(d)>\mathcal{E}(x_t)$. When \RETRIEVE retains at least one such document, the selected candidate improves on its parent in 67.1\% of iterations and sets a new run best in 14.1\%, compared with 52.1\% and 9.6\% under \NOOP (Figure~\ref{main_fig:behavior_audit_b}). Under \LOOKUP, which reuses stored documents, these rates are 61.2\% and 10.1\%, respectively. On 17 of 21 tasks, parent improvement rates are higher with promising documents than under \NOOP (medians: 69\% vs.\ 54\%). However, 36.5\% of retrievals yield no promising document; the parent improvement rate in these cases is only 38.3\%. These results suggest that hypothetical evidence scoring helps identify documents useful for improving the current solution.

\noindent \textbf{Web documents mainly provide methods and performance targets.} Figure~\ref{main_fig:behavior_audit_c} summarizes six behavior categories across 82 GPT-5.6-Luna runs (8,200 iterations), with each run counted once in every applicable category. The most common behaviors are applying principles and methods from retrieved documents (55 runs) and using published best scores as reference targets (40 runs). These targets guide program development, while the task evaluator measures the resulting programs' performance. We also observe reuse of published solutions (6 runs) and unsuccessful revisions inspired by retrieved ideas (29 runs). Detailed explanations are presented in Appendix~\ref{supp_sec:behavior_examples}.

\subsection{Additional Analysis} \label{main_sec:additional_analysis}

\begin{table*}[t]
\centering
\small
\setlength{\tabcolsep}{3pt}
\renewcommand{\arraystretch}{1.08}
\captionsetup[subtable]{font=small, skip=2pt, justification=centering}

\caption{\textbf{Additional analyses of \methodName.} For (b)-(e), we use GPT-5.6-Luna.}
\label{main_tab:ablation_row}

\vspace{2pt}

\begin{minipage}[t]{0.40\linewidth}
\centering
\begin{subtable}[t]{\linewidth}
    \centering
    \caption{Best programs of \methodName.}
    \label{main_tab:new_sota_result}
    \scriptsize
    \setlength{\tabcolsep}{2.4pt}
    \renewcommand{\arraystretch}{1.05}
    \begin{adjustbox}{max width=\linewidth}
    \begin{NiceTabular}{@{}lrrr@{}}
        \CodeBefore
        \begin{tikzpicture}
            \shade[left color=evoduetcolor!20, right color=evoduetpurple!20]
                (1-|3) rectangle (13-|4);
        \end{tikzpicture}
        \Body
        \toprule
        \textbf{Task} & \textbf{Prev. SOTA} & \textbf{\methodName} & \textbf{Cost (USD)} \\
        \midrule
        Swap Reduction ($\downarrow$) & 15{,}186 & \textbf{14{,}835} & 50.90 \\
        Rosetta ($\downarrow$) & 1.552968 & \textbf{1.396424} & 44.12 \\
        Voyager 2 ($\downarrow$) & 3.430214 & \textbf{3.430206} & 43.02 \\
        Denoising ($\uparrow$) & 0.722690 & \textbf{0.722906} & 38.45 \\
        Domain mix. ($\uparrow$) & 0.996922 & \textbf{0.997062} & 15.84 \\
        Parallel ($\uparrow$) & 0.999970 & \textbf{0.999975} & 11.93 \\
        Erd\H{o}s ($\downarrow$) & 0.380868 & \textbf{0.380859} & 29.55 \\
        Hadamard ($\uparrow$) & \textbf{0.935673} & \textbf{0.935673} & 27.33 \\
        Sums/Diffs ($\uparrow$) & 1.144887 & \textbf{1.144999} & 161.33 \\
        CP ($n$=26) ($\uparrow$) & \textbf{2.635983} & \textbf{2.635983} & 36.33 \\
        CP ($n$=32) ($\uparrow$) & \textbf{2.939573} & \textbf{2.939573} & 42.32 \\
        \bottomrule
    \end{NiceTabular}
    \end{adjustbox}
\end{subtable}

\end{minipage}%
\hfill
\begin{minipage}[t]{0.30\linewidth}
\centering
\begin{subtable}[t]{\linewidth}
\centering
\caption{Retrieval gating.}
\label{main_tab:ablation_gating}
\scriptsize
\renewcommand{\arraystretch}{1.05}
\begin{adjustbox}{max width=\linewidth}
\begin{NiceTabular}{lcc}
\CodeBefore
\evoduetgradientrows{4}{4}
\Body
\toprule
\textbf{Gating} & \textbf{Denoising} & \textbf{Erd\H{o}s} \\
\midrule
Random & 58.5\% & 93.9\% \\
Heuristic & 60.6\% & 99.5\% \\
Knowledge Gap & \textbf{84.5\%} & \textbf{100.0\%} \\
\bottomrule
\end{NiceTabular}
\end{adjustbox}
\end{subtable}

\vspace{6pt}

{\scriptsize\renewcommand{\arraystretch}{1.05}%
\begin{subtable}[t]{\linewidth}
\centering
\caption{Search-method comparison}
\label{main_tab:deepevolve_comparison}
\label{main_tab:search_method_comparison}
\setlength{\tabcolsep}{3pt}
\begin{adjustbox}{max width=\linewidth}
\begin{NiceTabular}{lrrr}
\CodeBefore
\evoduetgradientrows{5}{5}
\Body
\toprule
\textbf{Method} & \textbf{Molecule ($\uparrow$)} & \textbf{Burgers ($\uparrow$)} & \textbf{CP ($n$=26)} \\
\midrule
OpenEvolve & 0.8496 & 0.6937 & \textbf{2.635983} \\
Joint-level & 0.8474 & 0.6919 & 2.635980 \\
DeepEvolve & 0.8149 & 0.6666 & 2.581971 \\
\methodName & \textbf{0.8524} & \textbf{0.7846} & \textbf{2.635983} \\
\bottomrule
\end{NiceTabular}
\end{adjustbox}
\end{subtable}
\par}
\end{minipage}%
\hfill
\begin{minipage}[t]{0.28\linewidth}
\centering
\begin{subtable}[t]{\linewidth}
\centering
\caption{Cross-scaffold $\Delta$NDG.}
\label{main_tab:scaffold_native_diff_full}
\scriptsize
\renewcommand{\arraystretch}{1.05}
\begin{adjustbox}{max width=\linewidth}
\begin{tabular}{lcc}
\toprule
\textbf{Scaffold} & \textbf{Sums/Diffs} & \textbf{Denoising} \\
\midrule
OpenEvolve & +23.5\% & +84.5\% \\
Top-$K$ & +46.7\% & +63.6\% \\
EvoX & +55.1\% & +37.8\% \\
\bottomrule
\end{tabular}
\end{adjustbox}
\end{subtable}

\vspace{6pt}

\begin{subtable}[t]{\linewidth}
\centering
\caption{Denoising Pareto frontier.}
\label{main_fig:cost_pareto}
\includegraphics[width=0.95\linewidth]{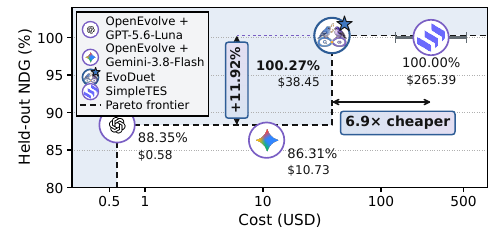}
\end{subtable}
\end{minipage}
\vspace{-1em}
\end{table*}

\noindent \textbf{The knowledge-gap-based gate achieves the highest NDG.}
Table~\ref{main_tab:ablation_gating} compares random gating ($p=0.5$), a stagnation heuristic~\citep{liu2026evox}, and our knowledge-gap-based gate. Our gate performs best on both Denoising and Erd\H{o}s, with a particularly large gain of 23.9\% over the stagnation heuristic on Denoising. This highlights the importance of the model's assessment of its own knowledge state when deciding whether to search the web.

\noindent \textbf{Bi-level optimization is an effective way to couple web search with solution evolution.}
We compare four variants: OpenEvolve without web search, OpenEvolve with an in-loop search tool (joint-level), DeepEvolve (sequential), and \methodName (bi-level). A detailed comparison of these variants is provided in Appendix~\ref{supp_sec:loop_method_comparison}. In Table~\ref{main_tab:search_method_comparison}, \methodName outperforms all other methods on three tasks, highlighting the importance of the bi-level design.

\noindent \textbf{\methodName integrates effectively with existing scaffolds.}
Table~\ref{main_tab:scaffold_native_diff_full} compares OpenEvolve, Top-$K$, and EvoX with and without \methodName.  \methodName improves NDG on Sums/Diffs and Denoising across all three scaffolds, with respective gains of 23.5 and 84.5 for OpenEvolve, 46.7 and 63.6 for Top-$K$, and 55.1 and 37.8 for EvoX. Although EvoX already includes a solution loop and a meta-level strategy loop, adding \methodName's web search loop yields substantial gains, highlighting the benefits of \textit{packing the loop} (Figure~\ref{main_fig:pack_loop}).

\noindent \textbf{Cost efficiency.}
Table~\ref{main_fig:cost_pareto} compares SimpleTES~\citep{simpletes2026}, the best observed OpenEvolve runs, and \methodName on Denoising. \methodName achieves 100.27\% NDG at \$38.45, a $6.9\times$ reduction in cost compared with SimpleTES (100\% NDG at an estimated API-equivalent cost of \$265.39). 
These results suggest that bi-level co-evolution enables cost-efficient performance by retrieving relevant external knowledge when the model needs it, despite the additional cost of web search.

\section{Related Work} \label{main_sec:related_work}

LLM-driven evolutionary scaffolds use prior candidates and evaluation feedback to improve programs~\citep{novikov2025alphaevolve,openevolve,lange2025shinkaevolve,wang2025thetaevolve,qu2026coral,yuksekgonul2026learning,lee2026evolution}.
GEPA~\citep{agrawal2025gepa} optimizes prompts, while other methods evolve agent or harness code~\citep{zhang2025darwin,zhang2026hyperagents,lee2026meta}.
Most of these methods operate within a single solution loop.
EvoX~\citep{liu2026evox} adds a meta-level loop that evolves search strategies alongside solutions, but both loops draw on the LLM's existing knowledge and the run's evolutionary history. DeepEvolve~\citep{liu2025deepevolve} adds a web search loop, but couples search and solution optimization sequentially rather than through a bi-level formulation.

\section{Conclusion} \label{main_sec:conclusion}

We introduce \methodName, a bi-level method that co-evolves solutions and web search queries, letting the LLM's knowledge state decide \textit{when} to search and \textit{what} to ask. Across 21 tasks, \methodName improves OpenEvolve with GPT-5.6-Luna and Gemini-3.8-Flash (but not Qwen3.5-9B), surpasses the best previously reported scores on eight tasks, and improves Top-$K$ and EvoX on Sums/Diffs and Denoising.

\subsubsection*{Acknowledgments}
This work was partly supported by the IITP grants (RS-2022-II220077, RS-2022-II220113, RS-2022-II220959, RS-2022-II220871) funded by the Korea government (MSIT) and the BK21 FOUR program, SNU in 2025.
This research was supported by the ``Advanced GPU Utilization Support Program'' funded by the Government of the Republic of Korea (Ministry of Science and ICT). 

\subsection*{AI use statement}

In this work, we used generative AI tools for trajectory analysis. We have not used generative AI tools for idea proposal or method development, and proof-related tasks are not applicable to this work. Additionally, we used generative AI tools for paper writing and figure preparation.
We also used generative AI tools to retrieve and summarize oracle documents,
as described in Appendix~\ref{supp_sec:oracle_setup}.

We have reviewed all AI-assisted work. We checked the AI-assisted text and figures
for consistency with the reported experimental results. For trajectory analysis,
we checked the interpretations against recorded search queries, retrieved documents, parent and child programs, and evaluator outputs, with worked cases provided in Appendix~\ref{supp_sec:trajectories}. We also inspected cases of public artifact reuse and documented the sources of reused solutions, as detailed in Appendices~\ref{supp_sec:behavior_examples} and~\ref{supp_sec:best_programs}.

We take responsibility for the final content of this work, including text,
claims or artifacts produced with the aid of generative AI.

\subsection*{Ethics statement}

This work evaluates LLM-guided computational optimization across a set of tasks, using publicly available web sources to inform the search. To support proper attribution, we document cases of public artifact reuse, including the original sources and subsequent optimization steps, in Appendices~\ref{supp_sec:behavior_examples} and~\ref{supp_sec:best_programs}. The reported gains reflect performance under the task evaluators used in this study. The resulting programs require independent validation before use in other settings. Appendix~\ref{supp_sec:broader_impact} discusses source attribution, the reliability of retrieved information, program validation, and human oversight of downstream applications.

\subsection*{Reproducibility statement}

We describe \methodName in Section~\ref{main_sec:method} and provide pseudocode,
inference settings, and retrieval budgets in Appendix~\ref{supp_sec:method_details}.
Task definitions, evaluation procedures, and oracle-document construction are
documented in Appendix~\ref{supp_sec:experimental_setup_details}, and the prompt
templates are provided in Appendix~\ref{supp_sec:prompt_templates}.
Detailed experimental results appear in
Appendix~\ref{supp_sec:additional_experimental_results}.
Appendix~\ref{supp_sec:trajectories} links selected search trajectories to retrieved
sources, code revisions, and evaluation results. Appendix~\ref{supp_sec:best_programs}
provides complete source listings and identifies the runs underlying
Table~\ref{main_tab:new_sota_result}.
Detailed trajectories, prompt templates, and complete program listings
are provided in the supplementary materials
(Appendices~\ref{supp_sec:trajectories}--\ref{supp_sec:best_programs}).

\bibliography{iclr2027_conference}
\bibliographystyle{iclr2027_conference}

\clearpage
\appendix

\addtocontents{toc}{\protect\setcounter{tocdepth}{2}}
\begingroup
    \renewcommand{\contentsname}{Appendix Contents}
    \hypersetup{linkcolor=black,linktoc=all}
    \small
    \setlength{\parskip}{0pt}
    \tableofcontents
\endgroup
\clearpage

\section{Extended Related Work} \label{supp_sec:related_work}

\paragraph{Evolving solutions and search strategies.}
LLM-driven discovery methods differ in whether they optimize candidate programs, search procedures, prompts, or model parameters.
FunSearch~\citep{romera2024mathematical}, AlphaEvolve~\citep{novikov2025alphaevolve}, OpenEvolve~\citep{openevolve}, and CodeEvolve~\citep{assumpccao2025codeevolve} use execution feedback to evolve programs.
EoH~\citep{liu2024evolution} jointly evolves heuristic descriptions and code.
At the search level, ShinkaEvolve~\citep{lange2025shinkaevolve}, PACEvolve~\citep{yan2026pacevolve}, and AdaEvolve~\citep{cemri2026adaevolve} adapt selection, context management, and exploration, respectively, while EvoX~\citep{liu2026evox} jointly evolves solutions and search strategies.
The Darwin G{\"o}del Machine~\citep{zhang2025darwin}, Hyperagents~\citep{zhang2026hyperagents}, and Meta-Harness~\citep{lee2026meta} optimize agent or harness code.
CORAL~\citep{qu2026coral} coordinates collaborative evolution through shared memory, and SkyDiscover~\citep{liu2026skydiscover} provides modular discovery infrastructure.
OPRO~\citep{yang2023opro}, APO~\citep{pryzant2023apo}, TextGrad~\citep{yuksekgonul2024textgrad}, and GEPA~\citep{agrawal2025gepa} optimize prompts or textual variables using performance feedback or textual critiques.
ThetaEvolve~\citep{wang2025thetaevolve} and TTT-Discover~\citep{yuksekgonul2026learning} update model parameters through reinforcement learning during search, while Kernel-Smith~\citep{du2026kernel} and Evolution Fine-Tuning~\citep{lee2026evolution} train on evolutionary trajectories.

\paragraph{Web-augmented scientific discovery.}
DeepEvolve~\citep{liu2025deepevolve} sequentially couples deep research with program revision and evaluation, using prior evaluation outcomes to guide subsequent research.
KernelEvolve~\citep{liao2025kernelevolve} uses runtime diagnostics to guide retrieval for kernel optimization.
AI co-scientist~\citep{gottweis2025coscientist} integrates literature search into hypothesis generation and scientific review.
Robin~\citep{ghareeb2026multi} uses literature search to propose and assess candidate hypotheses, then incorporates experimental feedback into subsequent refinement.
The AI Scientist~\citep{lu2026automatingresearch}, AI-Researcher~\citep{tang2025airesearcher}, and Agent Laboratory~\citep{schmidgall2025agentlaboratory} connect literature search, computational experiments, and reporting.
AgentRxiv~\citep{schmidgall2025agentrxiv} and AgenticSciML~\citep{jiang2026agenticsciml} reuse accumulated research reports or method memories.
Coscientist~\citep{boiko2023autonomous}, ChemCrow~\citep{bran2024chemcrow}, and Biomni~\citep{huang2026biomni} combine external information with scientific tools.

\paragraph{Retrieval control and query refinement.}
Retrieval methods address when to search and how to refine queries.
ReAct~\citep{yao2023react}, IRCoT~\citep{trivedi2023ircot}, Iter-RetGen~\citep{shao2023iterretgen}, and Auto-RAG~\citep{yu2024autorag} guide successive retrieval steps using intermediate reasoning, answers, or observations.
Self-RAG~\citep{asai2023selfrag}, FLARE~\citep{jiang2023active}, Adaptive-RAG~\citep{jeong2024adaptive}, and CRAG~\citep{yan2024crag} base retrieval decisions on reflection, uncertainty, question complexity, and evidence quality, respectively.
Rewrite-Retrieve-Read~\citep{ma2023query} learns query reformulation, while RQ-RAG~\citep{chan2024rqrag} learns rewriting, decomposition, and disambiguation.
Search-R1~\citep{jin2025searchr1}, R1-Searcher~\citep{song2025r1searcher}, and ReSearch~\citep{chen2025research} learn search policies from outcome rewards.
REPLUG~\citep{shi2024replug} uses feedback from a frozen LM to train a retriever, while ITER~\citep{chen2026iter} trains a retriever conditioned on reasoning and search histories.
\methodName couples solution and query evolution through a shared task objective. Its knowledge-state gate chooses whether to retrieve new documents, reuse stored documents, or skip retrieval. The inner loop selects evidence using predicted candidate scores and refines queries without generating or evaluating programs. The outer loop records measured outcomes to inform subsequent retrieval decisions. Model parameters and prompt templates remain fixed throughout.

\section{Method Details}
\label{supp_sec:method_details}

Algorithm~\ref{alg:method} presents the overall \methodName procedure, and Algorithm~\ref{alg:inner_loop} details its query optimization loop. The inner loop refines queries using predicted candidate scores; the outer loop evaluates solutions and records their actual scores. All model calls use the same frozen $\mathcal{M}_\theta$.

\makeatletter
\providecommand{\theHALG@line}{}
\renewcommand{\theHALG@line}{\thealgorithm.\arabic{ALG@line}}
\makeatother

\begin{algorithm}[t]
\small
\caption{\methodName: overall procedure.}
\label{alg:method}
\begin{algorithmic}[1]
\Require Task $(I,x_0,\mathcal{E})$, selection policy $\phi$, outer iterations $T$
\Require Candidate count $N$, retrieval settings $(R,J,M,D,K)$
\State $\mathcal{D}_e\gets\{(x_0,\mathcal{E}(x_0))\}$; $\mathcal{D}_{\mathrm{s}}\gets\emptyset$
\For{$t=0,\ldots,T-1$} \Comment{\textcolor{NoopViolet}{Outer loop: solution optimization}}
    \State $(x_t,\mathcal{H}_t)\sim\phi(\mathcal{D}_e)$; $c_t\gets[\mathcal{H}_t;x_t]$
    \State $(K_t,g_t)\gets\textsc{Gate}_{\mathcal{M}_\theta}(c_t,\mathcal{D}_{\mathrm{s}})$ \Comment{Retrieval gating}
    \State $\mathcal{S}_t\gets\emptyset$ \Comment{\NOOP uses no documents}
    \If{$g_t=\LOOKUP$}
        \State $\mathcal{S}_t\gets$ stored documents identified by the gate
    \ElsIf{$g_t=\RETRIEVE$}
        \State \tcbox[
            on line, colframe=RetrieveBlue, colback=RetrieveBlue!6!white,
            arc=1.3mm, boxrule=0.7pt, boxsep=0pt,
            left=4pt, right=4pt, top=3pt, bottom=3pt
        ]{$\mathcal{S}_t\gets$ \Call{OptimizeQueries}{$c_t,K_t,\mathcal{D}_e,\mathcal{E}(x_t)$}}
        \Comment{\hyperref[alg:inner_loop]{\textcolor{RetrieveBlue}{Inner loop: query optimization}}}
    \EndIf
    \State $N_t\gets N$ if $g_t\in\{\LOOKUP,\RETRIEVE\}$, otherwise $1$
    \State Generate $x_{t+1}^{(n)}\sim\mathcal{M}_\theta(I,c_t,\mathcal{S}_t)$, $n=1,\ldots,N_t$, in parallel
    \State Evaluate candidates with $\mathcal{E}$ and collect valid candidates in $\mathcal{C}_t^{\mathrm{valid}}$
    \If{$\mathcal{C}_t^{\mathrm{valid}}\neq\emptyset$}
        \State $x_{t+1}\gets\argopt_{x\in\mathcal{C}_t^{\mathrm{valid}}}\mathcal{E}(x)$
        \State $\mathcal{D}_e\gets\phi(\mathcal{D}_e\cup\{(x_{t+1},\mathcal{E}(x_{t+1}))\})$
        \State Update $\mathcal{D}_{\mathrm{s}}$ with evidence and the measured outcome if $g_t\neq\NOOP$
    \Else
        \State Apply the scaffold's retry policy
    \EndIf
\EndFor
\State \Return best evaluated solution
\end{algorithmic}
\end{algorithm}

\subsection{Outer Loop: Solution Optimization}
\label{supp_subsec:evidence}

\paragraph{Candidate generation and selection.}
The gate sets $N_t=1$ for \NOOP and $N_t=N$ for \LOOKUP or \RETRIEVE, including retrievals that return no usable documents. Candidates are sampled independently in parallel from the same prompt $(I,c_t,\mathcal{S}_t)$ and evaluated after successful generation and parsing. Only the best valid candidate under the task's optimization direction is passed to $\phi$; ties favor the earlier candidate. If all candidates fail, the scaffold applies its existing retry policy.

\paragraph{Search database update.}
The search database $\mathcal{D}_{\mathrm{s}}$ associates the selected candidate's outcome with the documents it received. A record contains the executed queries, retained documents and their predicted scores, document identifiers, actual parent and candidate scores $\mathcal{E}(x_t)$ and $\mathcal{E}(x_{t+1})$, and evaluator feedback. Each prediction assumes one document alone, whereas the candidate uses the retained set jointly; the measured outcome therefore belongs to that complete attempt. Intermediate rounds receive no separate measured outcome. Failed evaluations leave the outcome unknown. Candidate code, responses, evaluations, and intermediate retrieval results are retained in the iteration trace.

\subsection{Knowledge-Gap-Based Retrieval Gating}
\label{supp_subsec:knowledge_state}

\paragraph{Gate inputs and outputs.}
A single call reads the parent solution, the scaffold's evolutionary history, and recent search records. It returns the knowledge state $K_t$, a decision $g_t$, and document identifiers when $g_t=\LOOKUP$. The knowledge state distinguishes prior model knowledge, findings from searches and evaluations, and unresolved questions about the current solution. The gate runs before population analysis, so \NOOP and \LOOKUP require no additional analysis call.

\paragraph{Stored search context.}
The gate, query construction, and evidence scoring share a snapshot of the ten most recent records in $\mathcal{D}_{\mathrm{s}}$, fixed throughout the outer iteration. Each record shows its queries, retained predictions, document identifiers, and measured outcomes, with up to three document bodies. Bodies are truncated to 10{,}000 characters, preferring full text over a snippet when available. This context lets the model relate previously retrieved evidence to actual optimization outcomes.

\paragraph{No-op and look-up.}
Under \NOOP, the solution prompt receives $\mathcal{S}_t=\emptyset$. Under \LOOKUP, the gate selects stored document identifiers, and their bodies are added to the solution prompt. Both decisions bypass the inner loop and make no web-search call.

\subsection{Inner Loop: Query Optimization}
\label{supp_subsec:inner_loop}

Algorithm~\ref{alg:inner_loop} is invoked only under \RETRIEVE and returns the documents used for solution generation.

\begin{algorithm}[t]
\small
\caption{Inner loop: query optimization.}
\label{alg:inner_loop}
\begin{algorithmic}[1]
\Require $R$ rounds, $J$ queries per round, $M$ documents per query, local pool capacity $D$
\Function{OptimizeQueries}{$c_t,K_t,\mathcal{D}_e,\mathcal{E}(x_t)$}
    \State $\mathcal{D}_p\gets$ retained solutions in $\mathcal{D}_e$; $A_t\gets\mathcal{M}_\theta(\mathrm{stats}(\mathcal{D}_p))$
    \State $K_t^0\gets K_t$; $\mathcal{S}_t^0\gets\emptyset$; initialize an empty score cache $\hat{s}_t$
    \State $\tilde{c}_t^0\gets(c_t,K_t^0,A_t)$
    \For{$r=1,\ldots,R$}
        \State $q_r^{(j)}\sim\mathcal{M}_\theta(\tilde{c}_t^{r-1},\mathcal{S}_t^{r-1},\{\hat{s}_t(d)\}_{d\in\mathcal{S}_t^{r-1}})$, $j=1,\ldots,J$, in parallel
        \State $Q_r\gets(q_r^{(j)})_{j=1}^{J}$ \Comment{Query construction}
        \State $\mathcal{W}_r\gets\textsc{Search}(Q_r)$ \Comment{Up to $M$ documents per distinct query}
        \State $\mathcal{P}_r\gets\textsc{Deduplicate}(\mathcal{S}_t^{r-1}\cup\mathcal{W}_r)$
        \State $\mathcal{U}_r\gets\{d\in\mathcal{P}_r:\hat{s}_t(d)\text{ is undefined}\}$
        \State $K_t^r\gets K_t^{r-1}$
        \If{$\mathcal{U}_r\neq\emptyset$} \Comment{Evidence scoring and knowledge update}
            \State $\{\hat{s}_t(d)\}_{d\in\mathcal{U}_r},K_t^r\gets\textsc{Predict}_{\mathcal{M}_\theta}(\tilde{c}_t^{r-1},\mathcal{E}(x_t),\mathcal{P}_r)$
            \State Cache validated predictions; preserve previously cached scores
        \EndIf
        \State $\mathcal{S}_t^r\gets\mathrm{TopD}_{\hat{s}_t}(\mathcal{P}_r)$ \Comment{Rank documents with validated scores}
        \State $\tilde{c}_t^r\gets(c_t,K_t^r,A_t)$
    \EndFor
    \State \Return $\mathcal{S}_t^R$ \Comment{Pass retained documents to the solution prompt}
\EndFunction
\end{algorithmic}
\end{algorithm}

\paragraph{Population state descriptor.}
After a \RETRIEVE decision, deterministic statistics summarize the retained population $\mathcal{D}_p$ in $\mathcal{D}_e$: its score distribution, recent parent-to-child outcomes, and parent-selection frequencies. Following EvoX, $\mathcal{M}_\theta$ converts these statistics into factual observations $A_t$ without additional interpretation. The call is skipped if the population is empty. The parent, its score, the evolutionary history, and $A_t$ remain fixed across all $R$ inner rounds.

\paragraph{Query construction.}
Each round generates $J$ independent responses to the same query prompt in parallel. A response contains one query, its keywords, target source type, intended discovery direction, and rationale; the query text is limited to 500 characters. Round 1 uses $\tilde{c}_t^0=(c_t,K_t,A_t)$. Later rounds additionally receive the retained documents and their predicted scores, and use the updated knowledge state $K_t^{r-1}$ to target the remaining gaps.

\paragraph{Web search and document pooling.}
Queries that differ only in whitespace are executed once per round. Each distinct valid query is sent to Tavily with advanced search depth and returns up to $M$ documents. The results are merged with $\mathcal{S}_t^{r-1}$ and deduplicated using the URL or provider identifier, title, and raw body. The pool $\mathcal{P}_r$ therefore contains at most $JM$ documents in the first round and $D+JM$ thereafter. A discarded document re-enters the pool only if a later search returns it again.

\paragraph{Hypothetical evidence scoring and knowledge update.}
When unscored documents are present, one model call predicts their absolute candidate scores $\hat{s}_t(d)$ on the evaluator's scale and updates the knowledge state. Each prediction estimates the score obtained by improving the fixed parent using document $d$ alone. Validated predictions are cached for the current inner loop and reused if the document appears again. Up to $D$ documents with the best predicted scores are retained, with ties resolved in pool order. If no document needs scoring, the call is skipped and $K_t^r=K_t^{r-1}$. After round $R$, the final local pool of up to $D$ documents is passed to the outer loop. Their bodies, titles, and URLs enter the solution prompt without a separate summarization call.

\paragraph{Validation and failures.}
Only finite predictions for valid, previously unscored document identifiers enter the ranking; existing predictions cannot be overwritten. A complete valid response updates the knowledge state. Partial responses retain valid predictions while preserving the previous knowledge state, and a failed scoring call preserves both the previous retained set and state. Failed searches consume their attempts, and malformed query samples are not replaced. All $R$ rounds run without early stopping. If no usable evidence is retained, the outer loop receives $\mathcal{S}_t=\emptyset$.

\subsection{Inference Settings and Call Budget}
\label{supp_subsec:inference}

\paragraph{Settings.}
The default retrieval budget is $R=3$ rounds, $J=1$ query per round, $M=5$ documents per query, and $D=3$ documents retained in the local pool. The entire final local pool enters the solution prompt. We evaluate $N=1$ and $N=8$ for candidate generation. All model calls use temperature 0.7, top-$p$ 0.95, medium reasoning effort, and up to 32{,}768 output tokens. Gate, query, and evidence-scoring calls use user-only prompts with tools disabled. Population analysis uses EvoX's system prompt and supplies the deterministic statistics in the user message.

\paragraph{Per-iteration budget.}
The following counts exclude retries. Under \RETRIEVE, the model-call budget includes one gate, at most one population analysis, $RJ$ query constructions, and up to $R$ evidence-scoring calls. With the default settings, the limits are eight retrieval-related model calls and three web searches. Candidate generation and evaluation follow separately.
\begin{center}
\small
\setlength{\tabcolsep}{7pt}
\begin{tabular}{lccc}
\toprule
\textbf{Decision} & \textbf{Retrieval LLM calls} & \textbf{Web searches} & \textbf{Generations} \\
\midrule
\NOOP & $1$ & $0$ & $1$ \\
\LOOKUP & $1$ & $0$ & $N$ \\
\RETRIEVE & $\leq RJ+R+2$ & $\leq RJ$ & $N$ \\
\bottomrule
\end{tabular}
\end{center}

\subsection{Coupling Web Search with Solution Evolution}
\label{supp_sec:loop_method_comparison}

\begin{figure}[!ht]
    \centering
    \includegraphics[width=\linewidth]{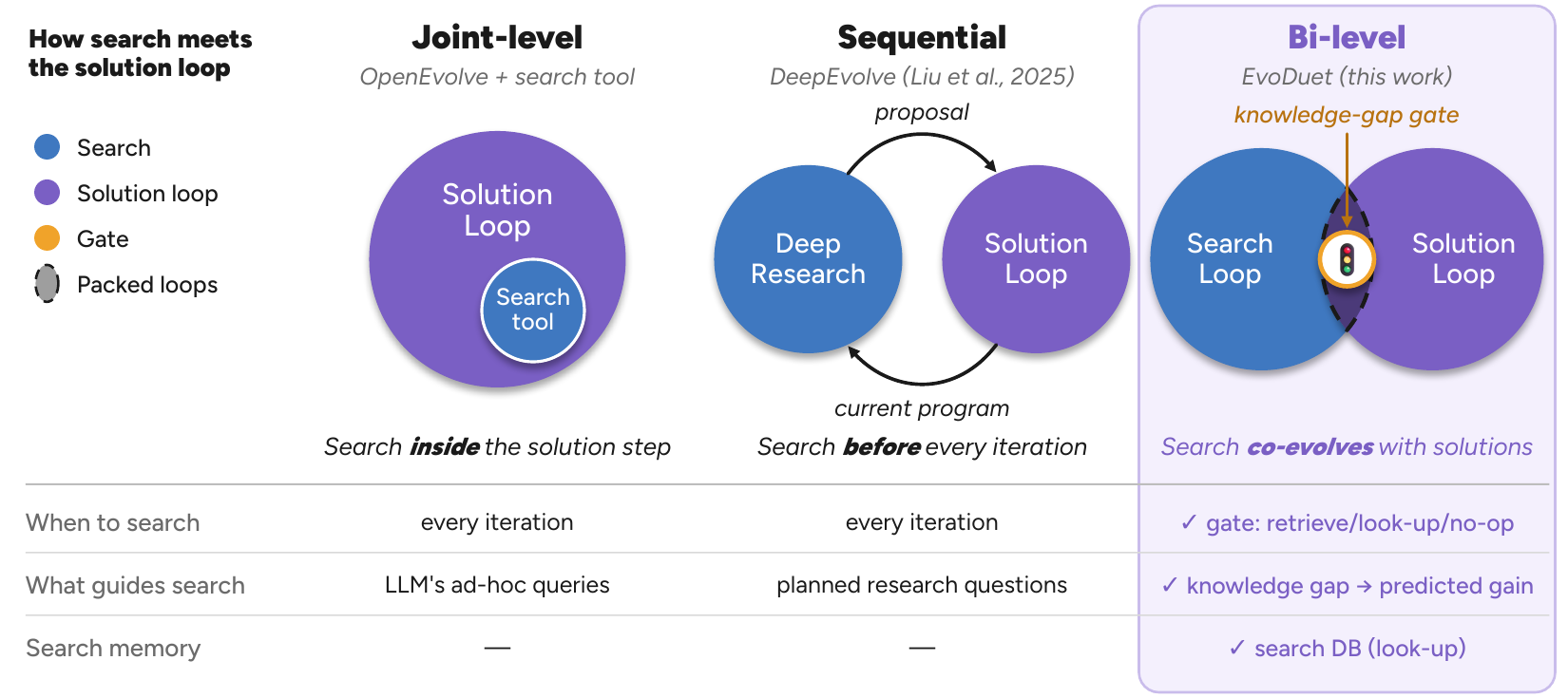}
    \caption{\textbf{Three ways to couple web search with solution evolution.} \emph{Top:} search runs within generation (joint-level), precedes generation (sequential), or co-evolves with solutions (bi-level). Blue denotes search, violet solution evolution, and amber the knowledge-gap gate; the dashed overlap marks the coupling between the two loops. \emph{Bottom:} search timing, query guidance, and memory for reuse across iterations. Checkmarks highlight \methodName's design choices.}
    \label{supp_fig:loop_method_comparison}
\end{figure}

Figure~\ref{supp_fig:loop_method_comparison} compares three ways to integrate web search into OpenEvolve: joint-level retrieval (Table~\ref{main_tab:ablation_bilevel}), sequential retrieval as in DeepEvolve~\citep{liu2025deepevolve}, and bi-level retrieval in \methodName. They differ in when they search, what guides their queries, and how they reuse search results.

\paragraph{Search placement changes how evidence enters the solution loop.}
\emph{Joint-level} retrieval runs during program generation: the LLM writes queries alongside program edits and must call Tavily at least once per iteration, with up to three tool rounds. \emph{Sequential} retrieval, as in DeepEvolve, runs a deep-research agent before generation at every iteration. Given the current program and programs selected as inspiration, the agent plans research questions, searches the web, and writes a proposal. Reflection can revisit planning or search when it judges the report incomplete; a coding agent then implements the proposal, followed by debugging and evaluation. Both joint-level and sequential retrieval are single-level: neither scores queries or documents against the task objective.

\methodName couples query and solution evolution through a knowledge-gap gate (\cref{main_sec:method}). The gate reads the current knowledge state and selects \RETRIEVE, \LOOKUP, or \NOOP. Under \RETRIEVE, the inner loop refines queries over $R$ rounds and selects documents using predicted program scores. The outer loop generates and evaluates programs, then stores the documents and measured outcomes in the search database to guide later gate decisions and queries.

\begin{table}[ht]
\centering
\small
\setlength{\tabcolsep}{6pt}
\renewcommand{\arraystretch}{1.08}
\caption{Joint-level versus bi-level retrieval (NDG, \%).}
\label{main_tab:ablation_bilevel}
\begin{adjustbox}{max width=\linewidth}
\begin{NiceTabular}{lcc}
\CodeBefore
\evoduetgradientrows{4}{4}
\Body
\toprule
\textbf{Method} & \textbf{Denoising} & \textbf{Sums/Diffs} \\
\midrule
No search & 0.0\% & 14.8\% \\
Joint-level & 43.2\% & 42.9\% \\
Bi-level & \textbf{84.5\%} & \textbf{76.0\%} \\
\bottomrule
\end{NiceTabular}
\end{adjustbox}
\end{table}

\paragraph{Joint-level search repeatedly returns the same pages on Denoising.}
In the GPT-5.6-Luna trajectory analyzed in Section~\ref{main_subsec:web_search_helpful_analysis}, joint-level retrieval returns 85 distinct URLs versus 248 for \methodName; 88.1\% of its returned URLs are repeats. Across eligible four runs, mean NDG at iteration 100 is 43.2\% on Denoising and 42.9\% on Sums/Diffs, compared with 84.5\% and 76.0\% for \methodName (Table~\ref{main_tab:ablation_bilevel}).

\paragraph{The DeepEvolve comparison uses published results.}
On Molecule and Burgers, DeepEvolve's released scores are 0.8149 and 0.6666, respectively, below our OpenEvolve baseline (0.8496 and 0.6937); \methodName reaches 0.8524 and 0.7846 (Table~\ref{main_tab:deepevolve_comparison}). The local scores are the best across available four runs with GPT-5.6-Luna at iteration 100; DeepEvolve was not rerun with the same model or search budget.

\section{Experimental Setup Details} \label{supp_sec:experimental_setup_details}

\definecolor{TaskOlive}{HTML}{5E7A2E}
\definecolor{TaskRose}{HTML}{A04A5E}
\definecolor{TaskIndigo}{HTML}{46508C}
\newtcolorbox{taskbox}[2]{enhanced, breakable, colback=#1!3!white, colframe=#1!65!white,
    colbacktitle=#1!12!white, coltitle=#1!80!black, title={#2}, fonttitle=\bfseries\small,
    fontupper=\small, arc=1.5mm, boxrule=0.5pt, left=2mm, right=2mm, top=1mm, bottom=1mm,
    toptitle=0.5mm, bottomtitle=0.5mm, before skip=6pt, after skip=8pt}

We evaluate on 31 tasks in two collections (Table~\ref{supp_tab:task_coverage}).
The descriptions below specify each task's optimization variables, constraints,
and evaluation procedure. Arrows indicate whether lower ($\downarrow$) or higher
($\uparrow$) values are better. Unless stated otherwise, we report the
evaluator's native objective.

\begin{table}[h]
\centering
\small
\setlength{\tabcolsep}{4pt}
\caption{Task coverage.}
\label{supp_tab:task_coverage}
\begin{tabular}{@{}lp{0.47\linewidth}r@{}}
\toprule
\textbf{Domain} & \textbf{Tasks} & \textbf{\#} \\
\midrule
Simple Scientific Optimization & Pressure Vessel, SINDy, Batch Reactor, LABS-27, LJ-13, Chwirut2, Perovskites, SRSD, GuacaMol, Summit & 10 \\
\midrule
Quantum compilation & Swap Reduction & 1 \\
Astrodynamics & Cassini, Galileo, Mariner~10, Rosetta, Voyager~2 & 5 \\
Scientific algorithms & Denoising & 1 \\
AI foundations & Domain Mixture, U Shape, LR \& BSZ, Parallel & 4 \\
Algorithm engineering & AHC039, AHC058 & 2 \\
Mathematics & Erd\H{o}s, AC1--AC3, CP ($n$=26, 32), Hadamard, Sums/Diffs & 8 \\
\midrule
Total & & 31 \\
\bottomrule
\end{tabular}
\end{table}

\subsection{Task Descriptions} \label{supp_sec:eval_task_details}

\subsubsection{Simple Scientific Optimization}
This collection consists of ten tasks constructed from public sources. For
each task, an oracle document was retrieved when the task was created
(Appendix~\ref{supp_sec:oracle_setup}). We describe the optimization problem
and report the scores of both the initial program and the solution stated in
the oracle document.

\begin{taskbox}{TaskBlue}{Pressure Vessel Design~\citep{task_pressure_vessel} ($\downarrow$ cost)}
The program specifies a vessel by its shell thickness $d_s$, head thickness
$d_h$, inner radius $r$, and length $L$. The two thicknesses must be multiples
of 0.0625\,in, and the length is restricted to $L\le200$\,in. The objective is
to minimize the fabrication cost
\[
    C(d_s,d_h,r,L)
    =0.6224d_srL+1.7781d_hr^2+3.1661d_s^2L+19.84d_s^2r,
\]
subject to the evaluator's thickness, volume, and dimension constraints.
The initial program has a cost of 10{,}905.34. The oracle solution, taken from
a SlideShare derivation of the optimum, has a cost of 6{,}059.71.
\end{taskbox}

\begin{taskbox}{TaskBlue}{SINDy Cubic 2D~\citep{task_sindy} ($\downarrow$ NRMSE)}
The input consists of 2{,}500 samples from a two-dimensional trajectory.
The program returns the coefficients of a polynomial vector field
$\dot{\mathbf{x}}=f(\mathbf{x})$ of degree at most three. The evaluator
measures how accurately the recovered field describes the dynamics and how
accurately its integrated trajectory reproduces the reference trajectory.
The objective is to minimize the mean of the vector-field and trajectory
normalized root mean squared errors (NRMSEs).

The initial program's field and trajectory NRMSEs are 0.670 and 1.197,
respectively. The cubic-oscillator equation in the PySINDy oracle document
gives corresponding errors of approximately zero and $1.7\times10^{-8}$.
\end{taskbox}

\begin{taskbox}{TaskBlue}{Batch Reactor Control~\citep{task_batch_reactor} ($\uparrow$ $B(1)$)}
The program controls the temperature of a batch reactor with consecutive
reactions $A\to B\to C$. It returns a schedule of 501 temperature values
over $t\in[0,1]$, each constrained to $[298,398]$\,K. The evaluator
integrates the reaction kinetics under this schedule, and the objective is
to maximize the terminal concentration $B(1)$ of the intermediate product.
The initial constant-temperature policy, $T(t)\equiv350$\,K, achieves
0.57706. The APMonitor GEKKO solution supplied by the oracle achieves 0.61076.
\end{taskbox}

\begin{taskbox}{TaskBlue}{LABS-27~\citep{task_labs} ($\downarrow$ energy)}
The low-autocorrelation binary sequence problem asks for a sequence
$s=(s_1,\ldots,s_{27})\in\{-1,+1\}^{27}$ with small correlations at nonzero
shifts. Its aperiodic autocorrelation at lag $k$ and its energy are
\[
    C_k(s)=\sum_{i=1}^{27-k}s_i s_{i+k},
    \qquad
    E(s)=\sum_{k=1}^{26}C_k(s)^2.
\]
The program returns the sequence, and the objective is to minimize $E(s)$.
The initial sequence has energy 177. The run-length-encoded optimum reported
by Packebusch and Mertens in the oracle document has energy 37.
\end{taskbox}

\begin{taskbox}{TaskBlue}{Lennard--Jones-13~\citep{task_lennard_jones} ($\downarrow$ energy)}
The program returns the positions $\mathbf{x}_1,\ldots,\mathbf{x}_{13}$
of 13 atoms in three-dimensional space. Writing
$r_{ij}=\|\mathbf{x}_i-\mathbf{x}_j\|_2$, the objective is to minimize the
reduced Lennard--Jones energy
\[
    E(\mathbf{x}_1,\ldots,\mathbf{x}_{13})
    =4\sum_{1\le i<j\le13}\left(r_{ij}^{-12}-r_{ij}^{-6}\right).
\]
Coincident atoms are invalid. The initial straight-chain configuration has
energy $-12.374$, while the configuration supplied by the Cambridge Cluster
Database oracle has energy $-44.327$.
\end{taskbox}

\begin{taskbox}{TaskBlue}{NIST Chwirut2~\citep{task_chwirut2} ($\downarrow$ RSS)}
Given NIST's 54 observations $(x_i,y_i)$, the program estimates the three
parameters of the nonlinear regression model
$\widehat{y}(x;b)=e^{-b_1x}/(b_2+b_3x)$. The objective is to minimize the
residual sum of squares
\[
    \operatorname{RSS}(b)
    =\sum_{i=1}^{54}\left(y_i-\frac{e^{-b_1x_i}}{b_2+b_3x_i}\right)^2.
\]
The initial program uses NIST Start~1 and has an RSS of 14{,}794.79.
The NIST-certified parameters in the oracle document give an RSS of 513.048.
\end{taskbox}

\begin{taskbox}{TaskBlue}{OLYMPUS Perovskites~\citep{task_gryffin,task_olympus} ($\downarrow$ bandgap)}
The search space contains 192 compositions, each specified by an organic
component, a cation, and a halide. The program selects one composition, and
the evaluator returns its published HSE06 bandgap from the dataset. The
objective is to minimize this bandgap. The initial composition has a bandgap
of 5.3704\,eV. The hydrazinium--Sn--I composition identified in the Gryffin
tutorial oracle has a bandgap of 1.5249\,eV.
\end{taskbox}

\begin{taskbox}{TaskBlue}{SRSD Feynman I.27.6~\citep{task_srsd} ($\downarrow$ error)}
The program returns a symbolic expression that predicts the focal distance
from the three inputs $(d_1,n,d_2)$. We use the official split of 8{,}000
training, 1{,}000 validation, and 1{,}000 test examples. The optimization
objective is the mean squared relative error on the validation set.
The initial expression has an error of 0.905. Equation~27.6 of the Feynman
Lectures, supplied by the oracle, gives zero error.
\end{taskbox}

\begin{taskbox}{TaskBlue}{GuacaMol C$_{11}$H$_{24}$~\citep{task_guacamol} ($\uparrow$ score)}
The program proposes molecular structures as SMILES strings, with
C$_{11}$H$_{24}$ as the target molecular formula. The evaluator canonicalizes
and deduplicates the proposals, then averages their formula-agreement scores
over the top 159 structures. Missing structures contribute zero, so the
objective rewards a collection of distinct matching structures.
The initial program supplies 5 of the 159 structures and scores 0.0315.
The PubChem isomer list supplied by the oracle gives a score of 1.0.
\end{taskbox}

\begin{taskbox}{TaskBlue}{Summit Baumgartner~\citep{task_summit,task_baumgartner} ($\uparrow$ yield)}
The program specifies reaction conditions for an aniline C--N coupling:
catalyst, base, base loading, temperature, and residence time. Five released
emulators predict the yield under these conditions. Each prediction is
clipped to $[0,1]$, and the objective is to maximize their mean.
The initial conditions score 0.655. The conditions in a row of the released
dataset, identified by the oracle, score 1.0.
\end{taskbox}

\subsubsection{\textsc{SimpleTES}}
We use 21 tasks from \textsc{SimpleTES}~\citep{simpletes2026}, together with their released initial programs and evaluators. For each task, the released best program or construction is re-evaluated with the same evaluator and serves as the SOTA anchor for NDG. The tasks cover the six domains below.

\paragraph{Quantum compilation.}
Swap Reduction searches over initial qubit mappings and routing decisions.

\begin{taskbox}{TaskPlum}{Swap Reduction ($\downarrow$ SWAPs)}
Given a logical circuit and a device connectivity graph, a Rust routing
policy chooses the initial assignment of logical qubits to physical qubits
and inserts SWAP operations as the circuit is processed. The resulting
mapping must allow each two-qubit gate to execute on a hardware edge.
The optimization target is to reduce the number of inserted SWAPs.

The evaluation scaffold contains 72 circuit--device cases across three
topologies. Its native score measures the CNOT overhead saved relative to
reference routings. In the paper, we report the number of added SWAPs on
the 24 circuits assigned to the 20-qubit device.
\end{taskbox}

\paragraph{Astrodynamics.}
Each program returns a spacecraft trajectory as a sequence of launch,
flyby, deep-space maneuver, and arrival events, with an epoch and a state
for each event. The evaluator checks the trajectory using DE430 ephemerides
and two-body propagation, enforces the task's time windows and minimum
flyby altitudes, and charges any speed mismatch at a flyby. The common
objective is to minimize the total velocity-change budget, in km/s,
\[
    \Delta v_{\mathrm{total}}
    =\Delta v_{\mathrm{launch}}+\Delta v_{\mathrm{flybys}}
     +\Delta v_{\mathrm{maneuvers}}+\Delta v_{\mathrm{arrival}}.
\]
The five tasks differ in their destinations, permitted flyby bodies,
and arrival conditions.

\begin{taskbox}{TaskTeal}{Cassini ($\downarrow$ $\Delta v$)}
The trajectory begins at Earth and ends with capture into a 115-day orbit
around Saturn. Flybys may use Venus, Earth, Mars, or Jupiter. A feasible
trajectory may contain at most six flybys and six deep-space maneuvers.
\end{taskbox}

\begin{taskbox}{TaskTeal}{Galileo ($\downarrow$ $\Delta v$)}
The trajectory begins at Earth and ends with capture into a 210-day orbit
around Jupiter. Venus and Earth are the permitted flyby bodies, and the
trajectory is limited to five flybys and five deep-space maneuvers.
\end{taskbox}

\begin{taskbox}{TaskTeal}{Mariner 10 ($\downarrow$ $\Delta v$)}
The trajectory travels from Earth to Mercury, with Venus as the only
permitted flyby body. The arrival contribution to $\Delta v$ is zero when
the arrival speed relative to Mercury is below 10\,km/s.
\end{taskbox}

\begin{taskbox}{TaskTeal}{Rosetta ($\downarrow$ $\Delta v$)}
The trajectory travels from Earth to a rendezvous with comet 67P at a fixed
arrival epoch. Flybys may use Earth or Mars, and each flyby must maintain
an altitude of at least 300\,km.
\end{taskbox}

\begin{taskbox}{TaskTeal}{Voyager 2 ($\downarrow$ $\Delta v$)}
The trajectory travels from Earth to a Neptune flyby, with intermediate
flybys permitted at Jupiter, Saturn, and Uranus. The arrival contribution
to $\Delta v$ is zero when the arrival speed relative to Neptune is below
30\,km/s.
\end{taskbox}

\paragraph{Scientific algorithms.}
These tasks evaluate algorithms for statistical estimation. Single-Cell RNA-seq Denoising maximizes reconstruction quality.

\begin{taskbox}{TaskAmber}{Single-Cell RNA-seq Denoising ($\uparrow$ score)}
The input is a matrix of noisy unique molecular identifier (UMI) counts.
The program returns a denoised matrix of the same shape. For each dataset,
the evaluator averages a normalized mean-squared-error score and a
normalized Poisson score, with the additional requirement that the Poisson
score be at least 0.97. The objective is to maximize this combined score.
Program search uses the Pancreas dataset, while the reported result is the
mean score on the held-out PBMC and Tabula datasets.
\end{taskbox}

\paragraph{AI foundations.}
Each task asks the program to define a parametric predictive law and a
routine for fitting it. The law is fitted separately for each evaluation
group, subject to a task-specific parameter budget. Search is guided by
training $R^2$, whereas the paper reports held-out $R^2$, clipped to
$[-1,1]$. The tasks differ in their input variables, prediction targets,
and held-out regimes.

\begin{taskbox}{TaskOlive}{Domain Mixture ($\uparrow$ $R^2$)}
The law predicts losses on five domains from the proportions of those
domains in a training mixture. Evaluation covers four model sizes, with
at most 35 fitted parameters per size. The test set contains 24 unseen
mixtures, measuring how well the fitted law predicts losses for mixtures
absent from training.
\end{taskbox}

\begin{taskbox}{TaskOlive}{U Shape ($\uparrow$ $R^2$)}
The law predicts Brier score as a function of log compute on nine
benchmarks. Each benchmark permits at most six fitted parameters.
The held-out set contains 127 observations at higher compute, so the
reported score measures extrapolation beyond the training regime.
\end{taskbox}

\begin{taskbox}{TaskOlive}{LR \& BSZ ($\uparrow$ $R^2$)}
The law predicts language-model loss from learning rate, batch size, data
size, and model parameter count, using at most 26 fitted parameters.
The test set contains 117 runs at a model size and data budget absent
from training, assessing transfer of the fitted relationship to that
held-out setting.
\end{taskbox}

\begin{taskbox}{TaskOlive}{Parallel ($\uparrow$ $R^2$)}
The law predicts loss from model parameter count and the number $P$ of
parallel streams. It is fitted on two datasets, with at most four
parameters per dataset. Training uses observations with $P\le4$, and
evaluation uses $P=8$, testing extrapolation to a larger number of streams.
\end{taskbox}

\paragraph{Algorithm engineering.}
We use two AtCoder Heuristic Contest problems from
ALE-Bench~\citep{imajuku2025ale}. For each task, a C++ solver is evaluated
on 150 public inputs with a 2\,s limit per case. A run receives the sum of
the official case scores, or zero if any case fails. The reported task
score is the average over four runs.

\begin{taskbox}{TaskRose}{AHC039: Purse Seine Fishing ($\uparrow$ score)}
Given the positions of mackerels and sardines, the solver returns an
axis-parallel polygon describing a fishing region. The polygon may have
at most 1{,}000 vertices and a perimeter of at most $4\times10^5$.
The objective is to maximize the official score by enclosing as many
mackerels and as few sardines as possible. We use ALE-Agent's fifth-place
solution as the initial program.
\end{taskbox}

\begin{taskbox}{TaskRose}{AHC058: Apple Incremental Game ($\uparrow$ score)}
The solver chooses a sequence of machine-upgrade decisions over 500 turns.
On each turn, it may perform one upgrade or take no action. If $S$ is the
final apple count, the objective is to maximize the official score
\[
    \operatorname{round}\!\left(10^5\log_2 S\right).
\]
The initial program performs no upgrades and receives a score of zero.
\end{taskbox}

\paragraph{Mathematics.}
These tasks search for finite representations of mathematical constructions:
step functions, circle packings, sign matrices, and integer sets. Each
evaluator checks the construction and computes the objective defined below.
Circle Packing at $n=26$ and $n=32$ constitutes two separate tasks.

\begin{taskbox}{TaskIndigo}{Erd\H{o}s Minimum Overlap ($\downarrow$ $C_5$)}
The program specifies a step function $h:[0,2]\to[0,1]$ satisfying
\[
    \int_0^2 h(x)\,dx=1.
\]
For this function, the evaluator computes $C_5$, the maximum overlap
between $h$ and its shifted complement. The objective is to minimize
$C_5$ while maintaining the range and integral constraints.
\end{taskbox}

\begin{taskbox}{TaskIndigo}{First Autocorrelation Inequality, AC1 ($\downarrow$ $C_1$)}
The program represents a nonnegative step function on $[-1/4,1/4]$ by
$n$ heights $h_1,\ldots,h_n$. Writing $(h*h)_k$ for the discrete
convolution of the height sequence with itself, the objective is
\[
    \min_h C_1(h),
    \qquad
    C_1(h)=\frac{2n\max_k(h*h)_k}{\left(\sum_{i=1}^{n}h_i\right)^2}.
\]
The heights must have a positive sum so that the normalized objective
is defined.
\end{taskbox}

\begin{taskbox}{TaskIndigo}{Second Autocorrelation Inequality, AC2 ($\uparrow$ $R_2$)}
The program specifies a nonzero, nonnegative step function $f$ supported
on $[-1/4,1/4]$. Let $f*f$ denote its self-convolution. The objective is
to maximize
\[
    R_2(f)=
    \frac{\|f*f\|_2^2}{\|f*f\|_1\,\|f*f\|_\infty}.
\]
\end{taskbox}

\begin{taskbox}{TaskIndigo}{Third Autocorrelation Inequality, AC3 ($\downarrow$ $C_3$)}
The program represents a step function $f$ on $[-1/4,1/4]$ using signed
heights $h_1,\ldots,h_n$. The objective is to minimize
\[
    C_3(f)=\frac{\|f*f\|_\infty}{\left(\int f\right)^2},
    \qquad
    C_3(h)=\frac{2n\max_k|(h*h)_k|}
                    {\left(\sum_{i=1}^{n}h_i\right)^2}.
\]
The second expression is the evaluator's form for the step-height
representation. Heights may be positive or negative, but their sum must
be nonzero for the objective to be defined.
\end{taskbox}

\begin{taskbox}{TaskIndigo}{Circle Packing, $n$=26 and 32 ($\uparrow$ $\sum_i r_i$)}
For each of the two values of $n$, the program returns centers
$\mathbf{c}_i=(x_i,y_i)$ and nonnegative radii $r_i$ for $n$ circles. Each circle
must lie inside the unit square, and no two circles may overlap:
\[
    \begin{gathered}
        r_i\le x_i\le1-r_i,\qquad r_i\le y_i\le1-r_i,\\
        \|\mathbf{c}_i-\mathbf{c}_j\|_2\ge r_i+r_j
        \quad (i\ne j).
    \end{gathered}
\]
The objective is to maximize $\sum_{i=1}^{n}r_i$. The evaluator checks
the geometric constraints with tolerance $10^{-12}$.
\end{taskbox}

\begin{taskbox}{TaskIndigo}{Hadamard, Order 29 ($\uparrow$ ratio)}
The program returns a matrix $H\in\{-1,+1\}^{29\times29}$. The objective
is to maximize $|\det H|$. The evaluator computes the determinant
exactly and reports its absolute value divided by a fixed normalization
constant.
\end{taskbox}

\begin{taskbox}{TaskIndigo}{Sums and Differences ($\uparrow$ $c$)}
The program returns a set $A$ containing between 2 and 512 distinct
integers in $[-10^6,10^6]$. Define its sumset and difference set as
$A+A=\{a+b:a,b\in A\}$ and $A-A=\{a-b:a,b\in A\}$. The objective is
to maximize
\[
    c(A)=\frac{\log(|A+A|/|A|)}{\log(|A-A|/|A|)}.
\]
The evaluator computes both sets from the submitted integers and scores
their cardinalities through this ratio.
\end{taskbox}

\subsection{Oracle Document Construction} \label{supp_sec:oracle_setup}

\paragraph{\textsc{SimpleTES}.}
For each task, we give Claude Code (Claude Opus~5 with maximum reasoning effort, restricted to its WebSearch and WebFetch tools) the task instructions, the initial program, the evaluator, and the best program released by SimpleTES. It identifies the functional differences that let the best program score higher, writes a web-search query for each, and returns the documents it fetched, each with a summary limited to what the document states. The resulting oracle documents are semantic rather than verbatim. Although the prompt asks for each document's text as fetched, Claude Code's WebFetch returns the output of a smaller model that reads the page rather than the page itself, so a document carries what its source says about an improvement, such as a method, a formula, or a parameter setting, but not necessarily its wording or its full content.

\paragraph{Simple Scientific Optimization.}
The oracle documents of these ten tasks were collected with Claude Code when the tasks were created. Each states the task's solution directly, for example NIST's certified parameters for Chwirut2 or the Cambridge Cluster Database coordinates for Lennard--Jones-13, and we validate the stated solution with the task's evaluator; the boxes in Appendix~\ref{supp_sec:eval_task_details} give its score.

\section{Additional Experimental Results}
\label{supp_sec:additional_experimental_results}

\subsection{OpenEvolve with and without Oracle Documents}
\label{supp_sec:oracle_results}

Table~\ref{supp_tab:openevolve_oracle_full_results} reports the per-task mean and best NDG for the runs underlying Figure~\ref{main_fig:default_vs_default_oracle}. We evaluate simple scientific optimization tasks at iteration 20 and the remaining tasks at iteration 100. Each condition reports the mean and maximum NDG over its available runs independently.

\begin{table*}[!htbp]
\centering
\definecolor{OracleSimple}{HTML}{315A85}
\definecolor{OracleQuantum}{HTML}{755184}
\definecolor{OracleAstro}{HTML}{267467}
\definecolor{OracleScience}{HTML}{976027}
\definecolor{OracleAI}{HTML}{536AA0}
\definecolor{OracleMath}{HTML}{937544}
\definecolor{OracleEngineering}{HTML}{9A5B70}
\small
\setlength{\tabcolsep}{2.5pt}
\renewcommand{\arraystretch}{1.05}
\caption{Full results for OpenEvolve with and without oracle documents. Entries report \textbf{Mean $\pm$ Std / Best} NDG (\%; higher is better) over four runs, where Std is the sample standard deviation and Best is the maximum run NDG. $\Delta$ reports oracle minus OpenEvolve for Mean / Best, computed before rounding. Bold marks the higher value for each statistic at the displayed precision. Row colors distinguish task groups. The overall row averages task means and task bests across all 31 tasks.}
\label{supp_tab:openevolve_oracle_full_results}
\vspace{3pt}
\begin{adjustbox}{max width=\linewidth}
\begin{tabular}{@{}lrrrrrr@{}}
\toprule
& \multicolumn{3}{c}{\textbf{Qwen3.5-9B}} & \multicolumn{3}{c}{\textbf{GPT-5.6-Luna}} \\
\cmidrule(lr){2-4}\cmidrule(lr){5-7}
\textbf{Task} & \textbf{OpenEvolve} & \textbf{+ Oracle} & $\Delta$ (\%) & \textbf{OpenEvolve} & \textbf{+ Oracle} & $\Delta$ (\%) \\
\midrule
\rowcolor{OracleSimple!18} \multicolumn{7}{@{}l}{\textbf{Simple scientific optimization (10)}} \\
\rowcolor{OracleSimple!5} Pressure Vessel & $49.9$\,{\scriptsize $\pm 57.6$}$\,/\,100.0$ & $\mathbf{99.9}$\,{\scriptsize $\pm 0.1$}$\,/\,100.0$ & $\dpos{+50.1}\,/\,0.0$ & $100.0$\,{\scriptsize $\pm 0.0$}$\,/\,100.0$ & $100.0$\,{\scriptsize $\pm 0.0$}$\,/\,100.0$ & $0.0\,/\,0.0$ \\
\rowcolor{OracleSimple!5} SINDy (2D) & $99.8$\,{\scriptsize $\pm 0.2$}$\,/\,100.0$ & $99.8$\,{\scriptsize $\pm 0.1$}$\,/\,100.0$ & $\dneg{-0.1}\,/\,0.0$ & $100.0$\,{\scriptsize $\pm 0.0$}$\,/\,100.0$ & $100.0$\,{\scriptsize $\pm 0.0$}$\,/\,100.0$ & $0.0\,/\,0.0$ \\
\rowcolor{OracleSimple!5} Batch Reactor & $38.1$\,{\scriptsize $\pm 46.3$}$\,/\,\mathbf{100.0}$ & $\mathbf{87.2}$\,{\scriptsize $\pm 20.0$}$\,/\,99.4$ & $\dpos{+49.1}\,/\,\dneg{-0.6}$ & $99.5$\,{\scriptsize $\pm 0.6$}$\,/\,100.1$ & $\mathbf{100.0}$\,{\scriptsize $\pm 0.0$}$\,/\,100.1$ & $\dpos{+0.5}\,/\,0.0$ \\
\rowcolor{OracleSimple!5} LABS (27) & $61.8$\,{\scriptsize $\pm 0.0$}$\,/\,61.8$ & $\mathbf{74.9}$\,{\scriptsize $\pm 30.6$}$\,/\,\mathbf{100.0}$ & $\dpos{+13.0}\,/\,\dpos{+38.2}$ & $100.0$\,{\scriptsize $\pm 0.0$}$\,/\,100.0$ & $100.0$\,{\scriptsize $\pm 0.0$}$\,/\,100.0$ & $0.0\,/\,0.0$ \\
\rowcolor{OracleSimple!5} LJ (13) & $96.0$\,{\scriptsize $\pm 7.9$}$\,/\,100.0$ & $\mathbf{100.0}$\,{\scriptsize $\pm 0.1$}$\,/\,100.0$ & $\dpos{+4.0}\,/\,0.0$ & $100.0$\,{\scriptsize $\pm 0.0$}$\,/\,100.0$ & $100.0$\,{\scriptsize $\pm 0.0$}$\,/\,100.0$ & $0.0\,/\,0.0$ \\
\rowcolor{OracleSimple!5} Chwirut2 & $100.0$\,{\scriptsize $\pm 0.0$}$\,/\,100.0$ & $100.0$\,{\scriptsize $\pm 0.0$}$\,/\,100.0$ & $0.0\,/\,0.0$ & $100.0$\,{\scriptsize $\pm 0.0$}$\,/\,100.0$ & $100.0$\,{\scriptsize $\pm 0.0$}$\,/\,100.0$ & $0.0\,/\,0.0$ \\
\rowcolor{OracleSimple!5} Perovskites & $68.3$\,{\scriptsize $\pm 22.4$}$\,/\,96.7$ & $\mathbf{81.3}$\,{\scriptsize $\pm 19.6$}$\,/\,96.7$ & $\dpos{+12.9}\,/\,0.0$ & $97.4$\,{\scriptsize $\pm 0.8$}$\,/\,98.1$ & $\mathbf{97.9}$\,{\scriptsize $\pm 0.6$}$\,/\,98.1$ & $\dpos{+0.5}\,/\,0.0$ \\
\rowcolor{OracleSimple!5} Feynman I.27.6 & $19.8$\,{\scriptsize $\pm 30.1$}$\,/\,63.5$ & $\mathbf{100.0}$\,{\scriptsize $\pm 0.0$}$\,/\,\mathbf{100.0}$ & $\dpos{+80.2}\,/\,\dpos{+36.5}$ & $100.0$\,{\scriptsize $\pm 0.0$}$\,/\,100.0$ & $100.0$\,{\scriptsize $\pm 0.0$}$\,/\,100.0$ & $0.0\,/\,0.0$ \\
\rowcolor{OracleSimple!5} GuacaMol & $56.7$\,{\scriptsize $\pm 50.4$}$\,/\,100.0$ & $\mathbf{100.0}$\,{\scriptsize $\pm 0.0$}$\,/\,100.0$ & $\dpos{+43.3}\,/\,0.0$ & $100.0$\,{\scriptsize $\pm 0.0$}$\,/\,100.0$ & $100.0$\,{\scriptsize $\pm 0.0$}$\,/\,100.0$ & $0.0\,/\,0.0$ \\
\rowcolor{OracleSimple!5} Baumgartner & $102.0$\,{\scriptsize $\pm 4.8$}$\,/\,106.8$ & $\mathbf{103.1}$\,{\scriptsize $\pm 2.7$}$\,/\,\mathbf{107.1}$ & $\dpos{+1.1}\,/\,\dpos{+0.3}$ & $\mathbf{94.7}$\,{\scriptsize $\pm 16.8$}$\,/\,\mathbf{107.1}$ & $94.3$\,{\scriptsize $\pm 11.6$}$\,/\,104.2$ & $\dneg{-0.5}\,/\,\dneg{-2.9}$ \\
\addlinespace[3pt]
\rowcolor{OracleQuantum!18} \multicolumn{7}{@{}l}{\textbf{Quantum compilation (1)}} \\
\rowcolor{OracleQuantum!5} Swap Reduction & $5.8$\,{\scriptsize $\pm 4.9$}$\,/\,10.7$ & $\mathbf{11.9}$\,{\scriptsize $\pm 16.8$}$\,/\,\mathbf{23.8}$ & $\dpos{+6.1}\,/\,\dpos{+13.1}$ & $21.9$\,{\scriptsize $\pm 29.3$}$\,/\,63.2$ & $\mathbf{63.4}$\,{\scriptsize $\pm 20.1$}$\,/\,\mathbf{77.6}$ & $\dpos{+41.5}\,/\,\dpos{+14.4}$ \\
\addlinespace[3pt]
\rowcolor{OracleAstro!18} \multicolumn{7}{@{}l}{\textbf{Astrodynamics (5)}} \\
\rowcolor{OracleAstro!5} Cassini & $31.4$\,{\scriptsize $\pm 12.9$}$\,/\,43.4$ & $\mathbf{39.8}$\,{\scriptsize $\pm 29.4$}$\,/\,\mathbf{60.5}$ & $\dpos{+8.4}\,/\,\dpos{+17.1}$ & $73.8$\,{\scriptsize $\pm 9.3$}$\,/\,81.7$ & $\mathbf{77.4}$\,{\scriptsize $\pm 14.3$}$\,/\,\mathbf{96.2}$ & $\dpos{+3.5}\,/\,\dpos{+14.5}$ \\
\rowcolor{OracleAstro!5} Galileo & $\mathbf{12.2}$\,{\scriptsize $\pm 8.7$}$\,/\,\mathbf{24.2}$ & $8.0$\,{\scriptsize $\pm 11.3$}$\,/\,16.0$ & $\dneg{-4.2}\,/\,\dneg{-8.2}$ & $37.6$\,{\scriptsize $\pm 1.0$}$\,/\,38.3$ & $\mathbf{48.0}$\,{\scriptsize $\pm 16.7$}$\,/\,\mathbf{59.8}$ & $\dpos{+10.4}\,/\,\dpos{+21.6}$ \\
\rowcolor{OracleAstro!5} Mariner 10 & $\mathbf{93.8}$\,{\scriptsize $\pm 8.9$}$\,/\,\mathbf{100.0}$ & $93.3$\,{\scriptsize $\pm 9.3$}$\,/\,99.9$ & $\dneg{-0.5}\,/\,\dneg{-0.1}$ & $100.0$\,{\scriptsize $\pm 0.0$}$\,/\,100.0$ & $100.0$\,{\scriptsize $\pm 0.0$}$\,/\,100.0$ & $0.0\,/\,0.0$ \\
\rowcolor{OracleAstro!5} Rosetta & $\mathbf{52.5}$\,{\scriptsize $\pm 22.3$}$\,/\,\mathbf{96.7}$ & $43.5$\,{\scriptsize $\pm 11.5$}$\,/\,51.6$ & $\dneg{-9.0}\,/\,\dneg{-45.0}$ & $\mathbf{81.4}$\,{\scriptsize $\pm 19.1$}$\,/\,\mathbf{97.8}$ & $58.1$\,{\scriptsize $\pm 4.1$}$\,/\,61.0$ & $\dneg{-23.3}\,/\,\dneg{-36.9}$ \\
\rowcolor{OracleAstro!5} Voyager 2 & $47.2$\,{\scriptsize $\pm 26.6$}$\,/\,\mathbf{84.7}$ & $\mathbf{60.4}$\,{\scriptsize $\pm 18.0$}$\,/\,73.1$ & $\dpos{+13.2}\,/\,\dneg{-11.6}$ & $99.3$\,{\scriptsize $\pm 1.3$}$\,/\,100.0$ & $\mathbf{100.0}$\,{\scriptsize $\pm 0.0$}$\,/\,100.0$ & $\dpos{+0.7}\,/\,0.0$ \\
\addlinespace[3pt]
\rowcolor{OracleScience!18} \multicolumn{7}{@{}l}{\textbf{Scientific algorithms (1)}} \\
\rowcolor{OracleScience!5} Denoising & $\mathbf{17.9}$\,{\scriptsize $\pm 32.1$}$\,/\,\mathbf{79.5}$ & $17.5$\,{\scriptsize $\pm 24.7$}$\,/\,35.0$ & $\dneg{-0.4}\,/\,\dneg{-44.5}$ & $22.1$\,{\scriptsize $\pm 44.2$}$\,/\,\mathbf{88.4}$ & $\mathbf{29.9}$\,{\scriptsize $\pm 22.2$}$\,/\,45.6$ & $\dpos{+7.8}\,/\,\dneg{-42.8}$ \\
\addlinespace[3pt]
\rowcolor{OracleAI!18} \multicolumn{7}{@{}l}{\textbf{AI foundations (4)}} \\
\rowcolor{OracleAI!5} Domain Mixture & $65.1$\,{\scriptsize $\pm 35.5$}$\,/\,\mathbf{90.2}$ & $\mathbf{85.1}$\,{\scriptsize $\pm 2.6$}$\,/\,86.9$ & $\dpos{+20.0}\,/\,\dneg{-3.3}$ & $\mathbf{97.5}$\,{\scriptsize $\pm 0.5$}$\,/\,\mathbf{98.1}$ & $93.6$\,{\scriptsize $\pm 5.8$}$\,/\,97.7$ & $\dneg{-4.0}\,/\,\dneg{-0.4}$ \\
\rowcolor{OracleAI!5} U Shape & $\mathbf{34.3}$\,{\scriptsize $\pm 53.2$}$\,/\,\mathbf{109.7}$ & $33.3$\,{\scriptsize $\pm 47.2$}$\,/\,66.7$ & $\dneg{-0.9}\,/\,\dneg{-43.0}$ & $0.0$\,{\scriptsize $\pm 0.0$}$\,/\,0.0$ & $\mathbf{50.0}$\,{\scriptsize $\pm 70.7$}$\,/\,\mathbf{100.0}$ & $\dpos{+50.0}\,/\,\dpos{+100.0}$ \\
\rowcolor{OracleAI!5} LR \& BSZ & $\mathbf{52.9}$\,{\scriptsize $\pm 33.1$}$\,/\,\mathbf{87.0}$ & $31.5$\,{\scriptsize $\pm 44.5$}$\,/\,62.9$ & $\dneg{-21.4}\,/\,\dneg{-24.0}$ & $23.9$\,{\scriptsize $\pm 30.1$}$\,/\,62.4$ & $\mathbf{38.3}$\,{\scriptsize $\pm 54.1$}$\,/\,\mathbf{76.5}$ & $\dpos{+14.3}\,/\,\dpos{+14.1}$ \\
\rowcolor{OracleAI!5} Parallel & $26.2$\,{\scriptsize $\pm 10.1$}$\,/\,40.2$ & $\mathbf{78.0}$\,{\scriptsize $\pm 10.4$}$\,/\,\mathbf{85.4}$ & $\dpos{+51.8}\,/\,\dpos{+45.1}$ & $\mathbf{91.3}$\,{\scriptsize $\pm 15.5$}$\,/\,\mathbf{99.5}$ & $85.4$\,{\scriptsize $\pm 0.0$}$\,/\,85.4$ & $\dneg{-6.0}\,/\,\dneg{-14.2}$ \\
\addlinespace[3pt]
\rowcolor{OracleMath!18} \multicolumn{7}{@{}l}{\textbf{Mathematics (8)}} \\
\rowcolor{OracleMath!5} Erd\H{o}s & $\mathbf{93.1}$\,{\scriptsize $\pm 5.8$}$\,/\,\mathbf{99.8}$ & $91.6$\,{\scriptsize $\pm 9.9$}$\,/\,98.5$ & $\dneg{-1.6}\,/\,\dneg{-1.3}$ & $95.8$\,{\scriptsize $\pm 7.1$}$\,/\,99.9$ & $\mathbf{99.9}$\,{\scriptsize $\pm 0.1$}$\,/\,\mathbf{100.0}$ & $\dpos{+4.1}\,/\,0.0$ \\
\rowcolor{OracleMath!5} AC1 & $\mathbf{27.3}$\,{\scriptsize $\pm 15.8$}$\,/\,\mathbf{50.8}$ & $20.3$\,{\scriptsize $\pm 17.9$}$\,/\,32.9$ & $\dneg{-7.0}\,/\,\dneg{-17.9}$ & $42.5$\,{\scriptsize $\pm 7.3$}$\,/\,52.8$ & $\mathbf{51.1}$\,{\scriptsize $\pm 15.4$}$\,/\,\mathbf{62.3}$ & $\dpos{+8.6}\,/\,\dpos{+9.5}$ \\
\rowcolor{OracleMath!5} AC2 & $\mathbf{35.1}$\,{\scriptsize $\pm 17.6$}$\,/\,\mathbf{54.8}$ & $33.6$\,{\scriptsize $\pm 19.8$}$\,/\,47.6$ & $\dneg{-1.5}\,/\,\dneg{-7.2}$ & $41.2$\,{\scriptsize $\pm 24.8$}$\,/\,\mathbf{72.0}$ & $\mathbf{57.9}$\,{\scriptsize $\pm 8.2$}$\,/\,66.3$ & $\dpos{+16.7}\,/\,\dneg{-5.7}$ \\
\rowcolor{OracleMath!5} AC3 & $79.8$\,{\scriptsize $\pm 11.1$}$\,/\,91.8$ & $\mathbf{81.5}$\,{\scriptsize $\pm 18.6$}$\,/\,\mathbf{94.6}$ & $\dpos{+1.7}\,/\,\dpos{+2.9}$ & $\mathbf{99.7}$\,{\scriptsize $\pm 0.1$}$\,/\,\mathbf{99.8}$ & $99.6$\,{\scriptsize $\pm 0.1$}$\,/\,99.7$ & $\dneg{-0.1}\,/\,\dneg{-0.1}$ \\
\rowcolor{OracleMath!5} CP ($n=26$) & $56.9$\,{\scriptsize $\pm 24.9$}$\,/\,81.9$ & $\mathbf{79.9}$\,{\scriptsize $\pm 13.3$}$\,/\,\mathbf{89.3}$ & $\dpos{+23.0}\,/\,\dpos{+7.4}$ & $100.0$\,{\scriptsize $\pm 0.0$}$\,/\,100.0$ & $100.0$\,{\scriptsize $\pm 0.0$}$\,/\,100.0$ & $0.0\,/\,0.0$ \\
\rowcolor{OracleMath!5} CP ($n=32$) & $75.1$\,{\scriptsize $\pm 4.1$}$\,/\,78.7$ & $\mathbf{85.0}$\,{\scriptsize $\pm 2.2$}$\,/\,\mathbf{86.6}$ & $\dpos{+9.9}\,/\,\dpos{+7.8}$ & $\mathbf{100.0}$\,{\scriptsize $\pm 0.0$}$\,/\,100.0$ & $99.9$\,{\scriptsize $\pm 0.2$}$\,/\,100.0$ & $\dneg{-0.1}\,/\,0.0$ \\
\rowcolor{OracleMath!5} Hadamard & $44.3$\,{\scriptsize $\pm 3.3$}$\,/\,\mathbf{47.9}$ & $\mathbf{46.1}$\,{\scriptsize $\pm 1.9$}$\,/\,47.5$ & $\dpos{+1.8}\,/\,\dneg{-0.4}$ & $73.1$\,{\scriptsize $\pm 17.5$}$\,/\,90.8$ & $\mathbf{75.8}$\,{\scriptsize $\pm 25.9$}$\,/\,\mathbf{98.2}$ & $\dpos{+2.7}\,/\,\dpos{+7.4}$ \\
\rowcolor{OracleMath!5} Sums/Diffs & $\mathbf{5.9}$\,{\scriptsize $\pm 13.0$}$\,/\,\mathbf{32.4}$ & $0.0$\,{\scriptsize $\pm 0.0$}$\,/\,0.0$ & $\dneg{-5.9}\,/\,\dneg{-32.4}$ & $\mathbf{15.4}$\,{\scriptsize $\pm 1.2$}$\,/\,16.0$ & $14.8$\,{\scriptsize $\pm 4.4$}$\,/\,\mathbf{20.5}$ & $\dneg{-0.6}\,/\,\dpos{+4.5}$ \\
\addlinespace[3pt]
\rowcolor{OracleEngineering!18} \multicolumn{7}{@{}l}{\textbf{Algorithm engineering (2)}} \\
\rowcolor{OracleEngineering!5} AHC039 & $2.5$\,{\scriptsize $\pm 6.1$}$\,/\,15.0$ & $\mathbf{11.5}$\,{\scriptsize $\pm 16.2$}$\,/\,\mathbf{22.9}$ & $\dpos{+8.9}\,/\,\dpos{+7.9}$ & $\mathbf{1.3}$\,{\scriptsize $\pm 1.8$}$\,/\,\mathbf{3.8}$ & $0.0$\,{\scriptsize $\pm 0.0$}$\,/\,0.0$ & $\dneg{-1.3}\,/\,\dneg{-3.8}$ \\
\rowcolor{OracleEngineering!5} AHC058 & $\mathbf{57.6}$\,{\scriptsize $\pm 31.7$}$\,/\,\mathbf{86.0}$ & $30.4$\,{\scriptsize $\pm 18.0$}$\,/\,43.2$ & $\dneg{-27.1}\,/\,\dneg{-42.8}$ & $\mathbf{92.1}$\,{\scriptsize $\pm 0.4$}$\,/\,92.6$ & $91.2$\,{\scriptsize $\pm 8.8$}$\,/\,\mathbf{97.5}$ & $\dneg{-0.9}\,/\,\dpos{+4.9}$ \\
\addlinespace[3pt]
\midrule
\rowcolor{gray!12} \textbf{Overall average (31 tasks)} & $51.9\,/\,\mathbf{75.3}$ & $\mathbf{62.2}\,/\,71.9$ & $\dpos{+10.3}\,/\,\dneg{-3.4}$ & $74.2\,/\,82.7$ & $\mathbf{78.3}\,/\,\mathbf{85.4}$ & $\dpos{+4.0}\,/\,\dpos{+2.7}$ \\
\bottomrule
\end{tabular}
\end{adjustbox}
\end{table*}

\subsection{OpenEvolve with and without Parallel Generation}
\label{supp_sec:parallel_generation_results}

Parallel generation yields larger gains with oracle documents for GPT-5.6-Luna, but without them for Qwen3.5-9B (Table~\ref{supp_tab:openevolve_parallel_oracle_results}). Increasing $N$ from 1 to 8 raises average NDG by 15.0\% with oracle documents and 9.8\% without on the nine GPT-5.6-Luna tasks; the corresponding gains on the 12 Qwen3.5-9B tasks are 2.2\% and 11.8\%.

\begin{table*}[!htbp]
\centering
\definecolor{OracleSimple}{HTML}{315A85}
\definecolor{OracleQuantum}{HTML}{755184}
\definecolor{OracleAstro}{HTML}{267467}
\definecolor{OracleScience}{HTML}{976027}
\definecolor{OracleAI}{HTML}{536AA0}
\definecolor{OracleMath}{HTML}{937544}
\definecolor{OracleEngineering}{HTML}{9A5B70}
\small
\setlength{\tabcolsep}{2.5pt}
\renewcommand{\arraystretch}{1.08}
\caption{Parallel generation with and without oracle documents. Entries show mean $\pm$ sample standard deviation / best NDG (\%) at iteration 100. $\Delta$ is $N=8$ minus $N=1$ within each method; bold marks the higher value.}
\label{supp_tab:openevolve_parallel_oracle_results}
\vspace{3pt}
\begin{adjustbox}{max width=\linewidth}
\begin{tabular}{@{}lrrr@{\hspace{9pt}}rrr@{}}
\toprule
& \multicolumn{3}{c}{\textbf{OpenEvolve}} & \multicolumn{3}{c}{\textbf{OpenEvolve + Oracle}} \\
\cmidrule(lr){2-4}\cmidrule(lr){5-7}
\textbf{Task} & $N=1$ & $N=8$ & $\Delta$ (\%) & $N=1$ & $N=8$ & $\Delta$ (\%) \\
\midrule
\rowcolor{gray!18} \multicolumn{7}{@{}l}{\textbf{Qwen3.5-9B (12 tasks)}} \\
\rowcolor{OracleQuantum!18} \multicolumn{7}{@{}l}{\textbf{Quantum compilation}} \\
\rowcolor{OracleQuantum!5} Swap Reduction & $5.8$\,{\scriptsize $\pm 4.9$}$\,/\,10.7$ & $5.8$\,{\scriptsize $\pm 6.8$}$\,/\,\mathbf{13.0}$ & $\dpos{+0.1}\,/\,\dpos{+2.3}$ & $11.9$\,{\scriptsize $\pm 16.8$}$\,/\,23.8$ & $\mathbf{41.2}$\,{\scriptsize $\pm 27.1$}$\,/\,\mathbf{78.2}$ & $\dpos{+29.3}\,/\,\dpos{+54.4}$ \\
\addlinespace[2pt]
\rowcolor{OracleAstro!18} \multicolumn{7}{@{}l}{\textbf{Astrodynamics}} \\
\rowcolor{OracleAstro!5} Cassini & $31.4$\,{\scriptsize $\pm 12.9$}$\,/\,43.4$ & $\mathbf{44.0}$\,{\scriptsize $\pm 29.3$}$\,/\,\mathbf{60.4}$ & $\dpos{+12.6}\,/\,\dpos{+17.0}$ & $\mathbf{39.8}$\,{\scriptsize $\pm 29.4$}$\,/\,\mathbf{60.5}$ & $17.2$\,{\scriptsize $\pm 29.0$}$\,/\,60.3$ & $\dneg{-22.5}\,/\,\dneg{-0.2}$ \\
\rowcolor{OracleAstro!5} Galileo & $12.2$\,{\scriptsize $\pm 8.7$}$\,/\,24.2$ & $\mathbf{32.8}$\,{\scriptsize $\pm 2.2$}$\,/\,\mathbf{35.9}$ & $\dpos{+20.6}\,/\,\dpos{+11.7}$ & $8.0$\,{\scriptsize $\pm 11.3$}$\,/\,16.0$ & $\mathbf{24.5}$\,{\scriptsize $\pm 13.0$}$\,/\,\mathbf{35.5}$ & $\dpos{+16.5}\,/\,\dpos{+19.5}$ \\
\rowcolor{OracleAstro!5} Mariner 10 & $\mathbf{93.8}$\,{\scriptsize $\pm 8.9$}$\,/\,100.0$ & $73.1$\,{\scriptsize $\pm 48.9$}$\,/\,100.0$ & $\dneg{-20.7}\,/\,0.0$ & $\mathbf{93.3}$\,{\scriptsize $\pm 9.3$}$\,/\,99.9$ & $48.1$\,{\scriptsize $\pm 55.6$}$\,/\,\mathbf{100.0}$ & $\dneg{-45.2}\,/\,\dpos{+0.1}$ \\
\rowcolor{OracleAstro!5} Rosetta & $52.5$\,{\scriptsize $\pm 22.3$}$\,/\,96.7$ & $\mathbf{72.8}$\,{\scriptsize $\pm 17.2$}$\,/\,\mathbf{98.6}$ & $\dpos{+20.4}\,/\,\dpos{+2.0}$ & $43.5$\,{\scriptsize $\pm 11.5$}$\,/\,51.6$ & $\mathbf{58.7}$\,{\scriptsize $\pm 27.0$}$\,/\,\mathbf{77.7}$ & $\dpos{+15.2}\,/\,\dpos{+26.1}$ \\
\rowcolor{OracleAstro!5} Voyager 2 & $47.2$\,{\scriptsize $\pm 26.6$}$\,/\,84.7$ & $\mathbf{95.5}$\,{\scriptsize $\pm 2.4$}$\,/\,\mathbf{97.4}$ & $\dpos{+48.3}\,/\,\dpos{+12.7}$ & $\mathbf{60.4}$\,{\scriptsize $\pm 18.0$}$\,/\,73.1$ & $46.0$\,{\scriptsize $\pm 53.2$}$\,/\,\mathbf{92.1}$ & $\dneg{-14.3}\,/\,\dpos{+19.0}$ \\
\addlinespace[2pt]
\rowcolor{OracleScience!18} \multicolumn{7}{@{}l}{\textbf{Scientific algorithms}} \\
\rowcolor{OracleScience!5} Denoising & $26.8$\,{\scriptsize $\pm 37.5$}$\,/\,79.5$ & $\mathbf{78.5}$\,{\scriptsize $\pm 8.6$}$\,/\,\mathbf{85.7}$ & $\dpos{+51.7}\,/\,\dpos{+6.3}$ & $17.5$\,{\scriptsize $\pm 24.7$}$\,/\,35.0$ & $\mathbf{58.7}$\,{\scriptsize $\pm 12.2$}$\,/\,\mathbf{67.3}$ & $\dpos{+41.2}\,/\,\dpos{+32.4}$ \\
\addlinespace[2pt]
\rowcolor{OracleAI!18} \multicolumn{7}{@{}l}{\textbf{AI foundations}} \\
\rowcolor{OracleAI!5} U Shape & $\mathbf{34.3}$\,{\scriptsize $\pm 53.2$}$\,/\,\mathbf{109.7}$ & $21.3$\,{\scriptsize $\pm 25.1$}$\,/\,48.8$ & $\dneg{-13.0}\,/\,\dneg{-61.0}$ & $\mathbf{33.3}$\,{\scriptsize $\pm 47.2$}$\,/\,66.7$ & $23.3$\,{\scriptsize $\pm 46.6$}$\,/\,\mathbf{93.2}$ & $\dneg{-10.0}\,/\,\dpos{+26.5}$ \\
\rowcolor{OracleAI!5} LR \& BSZ & $52.5$\,{\scriptsize $\pm 33.8$}$\,/\,87.0$ & $\mathbf{62.9}$\,{\scriptsize $\pm 42.7$}$\,/\,\mathbf{92.8}$ & $\dpos{+10.4}\,/\,\dpos{+5.8}$ & $31.5$\,{\scriptsize $\pm 44.5$}$\,/\,62.9$ & $\mathbf{46.1}$\,{\scriptsize $\pm 37.7$}$\,/\,\mathbf{86.5}$ & $\dpos{+14.6}\,/\,\dpos{+23.6}$ \\
\addlinespace[2pt]
\rowcolor{OracleMath!18} \multicolumn{7}{@{}l}{\textbf{Mathematics}} \\
\rowcolor{OracleMath!5} AC2 & $\mathbf{35.1}$\,{\scriptsize $\pm 17.6$}$\,/\,\mathbf{54.8}$ & $14.5$\,{\scriptsize $\pm 20.2$}$\,/\,44.4$ & $\dneg{-20.6}\,/\,\dneg{-10.4}$ & $\mathbf{33.6}$\,{\scriptsize $\pm 19.8$}$\,/\,47.6$ & $19.1$\,{\scriptsize $\pm 31.5$}$\,/\,\mathbf{66.0}$ & $\dneg{-14.5}\,/\,\dpos{+18.4}$ \\
\rowcolor{OracleMath!5} Sums/Diffs & $\mathbf{5.9}$\,{\scriptsize $\pm 13.0$}$\,/\,\mathbf{32.4}$ & $4.4$\,{\scriptsize $\pm 3.3$}$\,/\,7.6$ & $\dneg{-1.4}\,/\,\dneg{-24.8}$ & $0.0$\,{\scriptsize $\pm 0.0$}$\,/\,0.0$ & $\mathbf{12.2}$\,{\scriptsize $\pm 7.9$}$\,/\,\mathbf{16.5}$ & $\dpos{+12.2}\,/\,\dpos{+16.5}$ \\
\addlinespace[2pt]
\rowcolor{OracleEngineering!18} \multicolumn{7}{@{}l}{\textbf{Algorithm engineering}} \\
\rowcolor{OracleEngineering!5} AHC058 & $57.6$\,{\scriptsize $\pm 31.7$}$\,/\,86.0$ & $\mathbf{91.1}$\,{\scriptsize $\pm 1.9$}$\,/\,\mathbf{92.2}$ & $\dpos{+33.5}\,/\,\dpos{+6.2}$ & $30.4$\,{\scriptsize $\pm 18.0$}$\,/\,43.2$ & $\mathbf{34.4}$\,{\scriptsize $\pm 6.9$}$\,/\,43.2$ & $\dpos{+4.0}\,/\,0.0$ \\
\addlinespace[2pt]
\midrule
\rowcolor{gray!12} \textbf{Overall average (12 tasks)} & $37.9\,/\,\mathbf{67.4}$ & $\mathbf{49.7}\,/\,64.7$ & $\dpos{+11.8}\,/\,\dneg{-2.7}$ & $33.6\,/\,48.4$ & $\mathbf{35.8}\,/\,\mathbf{68.0}$ & $\dpos{+2.2}\,/\,\dpos{+19.7}$ \\
\addlinespace[5pt]
\midrule
\rowcolor{gray!18} \multicolumn{7}{@{}l}{\textbf{GPT-5.6-Luna (9 tasks)}} \\
\rowcolor{OracleAstro!18} \multicolumn{7}{@{}l}{\textbf{Astrodynamics}} \\
\rowcolor{OracleAstro!5} Galileo & $37.6$\,{\scriptsize $\pm 1.0$}$\,/\,38.3$ & $\mathbf{43.6}$\,{\scriptsize $\pm 11.1$}$\,/\,\mathbf{60.3}$ & $\dpos{+6.1}\,/\,\dpos{+22.0}$ & $\mathbf{48.0}$\,{\scriptsize $\pm 16.7$}$\,/\,\mathbf{59.8}$ & $40.1$\,{\scriptsize $\pm 2.6$}$\,/\,42.0$ & $\dneg{-7.9}\,/\,\dneg{-17.9}$ \\
\rowcolor{OracleAstro!5} Rosetta & $81.4$\,{\scriptsize $\pm 19.1$}$\,/\,97.8$ & $\mathbf{87.9}$\,{\scriptsize $\pm 13.7$}$\,/\,\mathbf{99.8}$ & $\dpos{+6.5}\,/\,\dpos{+2.0}$ & $58.1$\,{\scriptsize $\pm 4.1$}$\,/\,61.0$ & $\mathbf{83.1}$\,{\scriptsize $\pm 23.2$}$\,/\,\mathbf{99.5}$ & $\dpos{+25.0}\,/\,\dpos{+38.5}$ \\
\addlinespace[2pt]
\rowcolor{OracleScience!18} \multicolumn{7}{@{}l}{\textbf{Scientific algorithms}} \\
\rowcolor{OracleScience!5} Denoising & $22.1$\,{\scriptsize $\pm 44.2$}$\,/\,\mathbf{88.4}$ & $\mathbf{60.5}$\,{\scriptsize $\pm 31.5$}$\,/\,80.3$ & $\dpos{+38.4}\,/\,\dneg{-8.1}$ & $29.9$\,{\scriptsize $\pm 22.2$}$\,/\,45.6$ & $\mathbf{80.6}$\,{\scriptsize $\pm 8.7$}$\,/\,\mathbf{86.8}$ & $\dpos{+50.7}\,/\,\dpos{+41.2}$ \\
\addlinespace[2pt]
\rowcolor{OracleAI!18} \multicolumn{7}{@{}l}{\textbf{AI foundations}} \\
\rowcolor{OracleAI!5} U Shape & $0.0$\,{\scriptsize $\pm 0.0$}$\,/\,0.0$ & $\mathbf{42.6}$\,{\scriptsize $\pm 50.6$}$\,/\,\mathbf{100.0}$ & $\dpos{+42.6}\,/\,\dpos{+100.0}$ & $50.0$\,{\scriptsize $\pm 70.7$}$\,/\,\mathbf{100.0}$ & $\mathbf{72.1}$\,{\scriptsize $\pm 23.3$}$\,/\,88.6$ & $\dpos{+22.1}\,/\,\dneg{-11.4}$ \\
\addlinespace[2pt]
\rowcolor{OracleMath!18} \multicolumn{7}{@{}l}{\textbf{Mathematics}} \\
\rowcolor{OracleMath!5} Erd\H{o}s & $95.8$\,{\scriptsize $\pm 7.1$}$\,/\,99.9$ & $\mathbf{99.9}$\,{\scriptsize $\pm 0.1$}$\,/\,\mathbf{100.0}$ & $\dpos{+4.1}\,/\,0.0$ & $99.9$\,{\scriptsize $\pm 0.1$}$\,/\,\mathbf{100.0}$ & $99.9$\,{\scriptsize $\pm 0.0$}$\,/\,99.9$ & $0.0\,/\,\dneg{-0.1}$ \\
\rowcolor{OracleMath!5} AC1 & $\mathbf{42.5}$\,{\scriptsize $\pm 7.3$}$\,/\,\mathbf{52.8}$ & $21.4$\,{\scriptsize $\pm 7.5$}$\,/\,28.5$ & $\dneg{-21.1}\,/\,\dneg{-24.4}$ & $51.1$\,{\scriptsize $\pm 15.4$}$\,/\,62.3$ & $\mathbf{73.8}$\,{\scriptsize $\pm 13.1$}$\,/\,\mathbf{83.1}$ & $\dpos{+22.7}\,/\,\dpos{+20.8}$ \\
\rowcolor{OracleMath!5} AC2 & $\mathbf{41.2}$\,{\scriptsize $\pm 24.8$}$\,/\,\mathbf{72.0}$ & $40.5$\,{\scriptsize $\pm 27.2$}$\,/\,58.2$ & $\dneg{-0.8}\,/\,\dneg{-13.8}$ & $57.9$\,{\scriptsize $\pm 8.2$}$\,/\,\mathbf{66.3}$ & $\mathbf{58.9}$\,{\scriptsize $\pm 0.0$}$\,/\,58.9$ & $\dpos{+1.0}\,/\,\dneg{-7.4}$ \\
\rowcolor{OracleMath!5} Sums/Diffs & $15.4$\,{\scriptsize $\pm 1.2$}$\,/\,16.0$ & $\mathbf{25.8}$\,{\scriptsize $\pm 5.2$}$\,/\,\mathbf{31.3}$ & $\dpos{+10.4}\,/\,\dpos{+15.3}$ & $14.8$\,{\scriptsize $\pm 4.4$}$\,/\,20.5$ & $\mathbf{40.7}$\,{\scriptsize $\pm 5.1$}$\,/\,\mathbf{44.3}$ & $\dpos{+25.9}\,/\,\dpos{+23.8}$ \\
\addlinespace[2pt]
\rowcolor{OracleEngineering!18} \multicolumn{7}{@{}l}{\textbf{Algorithm engineering}} \\
\rowcolor{OracleEngineering!5} AHC058 & $92.1$\,{\scriptsize $\pm 0.4$}$\,/\,92.6$ & $\mathbf{93.9}$\,{\scriptsize $\pm 1.1$}$\,/\,\mathbf{94.5}$ & $\dpos{+1.8}\,/\,\dpos{+1.9}$ & $\mathbf{91.2}$\,{\scriptsize $\pm 8.8$}$\,/\,\mathbf{97.5}$ & $86.7$\,{\scriptsize $\pm 6.0$}$\,/\,90.9$ & $\dneg{-4.6}\,/\,\dneg{-6.6}$ \\
\addlinespace[2pt]
\midrule
\rowcolor{gray!12} \textbf{Overall average (9 tasks)} & $47.6\,/\,62.0$ & $\mathbf{57.3}\,/\,\mathbf{72.5}$ & $\dpos{+9.8}\,/\,\dpos{+10.6}$ & $55.7\,/\,68.1$ & $\mathbf{70.6}\,/\,\mathbf{77.1}$ & $\dpos{+15.0}\,/\,\dpos{+9.0}$ \\
\bottomrule
\end{tabular}
\end{adjustbox}
\end{table*}

\clearpage
\subsection{Inner-loop Budget Ablations}
\label{supp_sec:inner_loop_budget_ablations}

\begin{table*}[!htbp]
\centering
\small
\setlength{\tabcolsep}{4pt}
\renewcommand{\arraystretch}{1.12}
\captionsetup[subtable]{font=small, skip=2pt}

\caption{Inner-loop budget ablations with GPT-5.6-Luna.}
\label{supp_tab:inner_loop_deepevolve}

\vspace{2pt}

\begin{minipage}{\linewidth}
\noindent
\begin{subtable}[t]{0.48\linewidth}
\centering
\caption{Rounds $R$ on Denoising.}
\label{main_tab:ablation_rounds}
\label{main_tab:ablation_rounds_queries}
\begin{adjustbox}{max width=\linewidth}
\begin{tabular}{lc}
\toprule
\textbf{Setting} & \textbf{NDG (\%)} \\
\midrule
$R=1$ & 58.35\% \\
\evoduetcolor $R=3$ (default) & \textbf{81.77\%} \\
$R=5$ & 34.28\% \\
\bottomrule
\end{tabular}
\end{adjustbox}
\end{subtable}%
\hfill
\begin{subtable}[t]{0.48\linewidth}
\centering
\caption{Queries $J$ on Denoising.}
\label{main_tab:ablation_queries}
\begin{adjustbox}{max width=\linewidth}
\begin{tabular}{lc}
\toprule
\textbf{Setting} & \textbf{NDG (\%)} \\
\midrule
\evoduetcolor $J=1$ (default) & \textbf{81.77\%} \\
$J=3$ ($k{=}3$) & 80.15\% \\
$J=3$ ($k{=}5$) & 71.86\% \\
$J=5$ ($k{=}3$) & 57.30\% \\
\bottomrule
\end{tabular}
\end{adjustbox}
\end{subtable}

\vspace{4pt}
\end{minipage}
\end{table*}

\paragraph{More inner-loop search is not better, in either direction.}
The inner loop is controlled by two budgets: the number of search rounds $R$ within one iteration, and the number of queries $J$ issued per round. Table~\ref{main_tab:ablation_rounds} and Table~\ref{main_tab:ablation_queries} vary each one on Denoising with GPT-5.6-Luna. Rounds peak at the default: $R{=}1$ reaches 58.35\% NDG, $R{=}3$ reaches 81.77\%, and $R{=}5$ falls to 34.28\%, so a single round gathers too little to act on while five spend the iteration reading instead of editing. Queries are worse than flat: $J{=}1$ gives 81.77\%, $J{=}3$ gives 80.15\% at $k{=}3$ and 71.86\% at $k{=}5$, and $J{=}5$ gives 57.30\%. 

\subsection{Full OpenEvolve Results}
\label{supp_sec:native_scores}
Tables~\ref{supp_tab:mean-std-quantum}--\ref{supp_tab:mean-std-frontier} report OpenEvolve scores with and without \methodName, organized by task group.

\begingroup
\newcommand{\performancestd}[1]{{\scriptsize\color{gray}$\pm$ #1}}
\newcommand{\performancecell}[2]{#1\,/\,#2}

\begin{table*}[t]
\centering
\small
\setlength{\tabcolsep}{3.5pt}

\caption{Quantum compilation and astrodynamics.}
\vspace{3pt}
\label{supp_tab:mean-std-quantum}
\label{supp_tab:mean-std-astro}

\renewcommand{\arraystretch}{1.00}

\begin{adjustbox}{max width=\linewidth}
\begin{NiceTabular}{l*{6}{c}}
\CodeBefore
\evoduetgradientrows{5,8,11,14,17,20}{8}
\Body
\toprule
 & \multicolumn{1}{c}{\textbf{Quantum compilation}} & \multicolumn{5}{c}{\textbf{Astrodynamics}} \\
\cmidrule(lr){2-2} \cmidrule(lr){3-7}
\textbf{Model / Method} & \textbf{Swap} $\downarrow$ & \textbf{Cassini} $\downarrow$ & \textbf{Galileo} $\downarrow$ & \textbf{Mariner 10} $\downarrow$ & \textbf{Rosetta} $\downarrow$ & \textbf{Voyager 2} $\downarrow$ \\
\cmidrule(lr){2-2} \cmidrule(lr){3-3} \cmidrule(lr){4-4} \cmidrule(lr){5-5} \cmidrule(lr){6-6} \cmidrule(lr){7-7}
 & \textbf{Mean $\pm$ Std / Best} & \textbf{Mean $\pm$ Std / Best} & \textbf{Mean $\pm$ Std / Best} & \textbf{Mean $\pm$ Std / Best} & \textbf{Mean $\pm$ Std / Best} & \textbf{Mean $\pm$ Std / Best} \\
\midrule
Qwen3.5-9B ($N=1$) & \performancecell{\textbf{19{,}802} \performancestd{353}}{\textbf{19{,}487}} & \performancecell{4.4977 \performancestd{0.691}}{3.8518} & \performancecell{\textbf{2.4036} \performancestd{0.159}}{\textbf{2.1835}} & \performancecell{\textbf{0.5648} \performancestd{0.340}}{\textbf{0.3274}} & \performancecell{\textbf{5.0725} \performancestd{1.649}}{\textbf{1.7998}} & \performancecell{\textbf{3.7582} \performancestd{0.165}}{\textbf{3.5255}} \\
+ \methodName & \performancecell{20{,}279 \performancestd{393}}{20{,}001} & \performancecell{\textbf{3.3296} \performancestd{0.097}}{\textbf{3.2613}} & \performancecell{2.9228 \performancestd{0.872}}{2.3577} & \performancecell{0.6607 \performancestd{0.412}}{0.3691} & \performancecell{5.1472 \performancestd{0.836}}{4.3908} & \performancecell{3.8968 \performancestd{0.179}}{3.7227} \\
$\Delta$ & $\textcolor{red!70!black}{-478}\,/\,\textcolor{red!70!black}{-514}$ & $\textcolor{green!45!black}{+1.1681}\,/\,\textcolor{green!45!black}{+0.5905}$ & $\textcolor{red!70!black}{-0.5192}\,/\,\textcolor{red!70!black}{-0.1742}$ & $\textcolor{red!70!black}{-0.0959}\,/\,\textcolor{red!70!black}{-0.0418}$ & $\textcolor{red!70!black}{-0.0747}\,/\,\textcolor{red!70!black}{-2.5910}$ & $\textcolor{red!70!black}{-0.1386}\,/\,\textcolor{red!70!black}{-0.1972}$ \\
\midrule
Qwen3.5-9B ($N=8$) & \performancecell{19{,}829 \performancestd{482}}{\textbf{19{,}374}} & \performancecell{3.8227 \performancestd{1.573}}{\textbf{2.9411}} & \performancecell{\textbf{2.0271} \performancestd{0.040}}{1.9687} & \performancecell{1.3569 \performancestd{1.874}}{0.3274} & \performancecell{\textbf{3.5636} \performancestd{1.274}}{\textbf{1.6533}} & \performancecell{\textbf{3.4582} \performancestd{0.015}}{\textbf{3.4464}} \\
+ \methodName & \performancecell{\textbf{19{,}697} \performancestd{92}}{19{,}632} & \performancecell{\textbf{3.1084} \performancestd{0.221}}{2.9525} & \performancecell{2.1751 \performancestd{0.325}}{\textbf{1.9454}} & \performancecell{\textbf{0.3345} \performancestd{0.010}}{0.3274} & \performancecell{3.9347 \performancestd{1.581}}{2.8168} & \performancecell{3.8236 \performancestd{0.265}}{3.6363} \\
$\Delta$ & $\textcolor{green!45!black}{+132}\,/\,\textcolor{red!70!black}{-258}$ & $\textcolor{green!45!black}{+0.7143}\,/\,\textcolor{red!70!black}{-0.0114}$ & $\textcolor{red!70!black}{-0.1480}\,/\,\textcolor{green!45!black}{+0.0233}$ & $\textcolor{green!45!black}{+1.0224}\,/\,0.0000$ & $\textcolor{red!70!black}{-0.3711}\,/\,\textcolor{red!70!black}{-1.1635}$ & $\textcolor{red!70!black}{-0.3655}\,/\,\textcolor{red!70!black}{-0.1899}$ \\
\midrule
GPT-5.6-Luna ($N=1$) & \performancecell{18{,}950 \performancestd{1{,}414}}{16{,}958} & \performancecell{2.2221 \performancestd{0.500}}{1.8019} & \performancecell{1.9389 \performancestd{0.017}}{\textbf{1.9261}} & \performancecell{0.3274 \performancestd{0.000}}{0.3274} & \performancecell{\textbf{2.9331} \performancestd{1.416}}{1.7125} & \performancecell{\textbf{3.4343} \performancestd{0.008}}{3.4302} \\
+ \methodName & \performancecell{\textbf{17{,}894} \performancestd{1{,}579}}{\textbf{16{,}246}} & \performancecell{\textbf{1.5083} \performancestd{0.671}}{\textbf{0.8756}} & \performancecell{\textbf{1.9359} \performancestd{0.011}}{1.9264} & \performancecell{0.3274 \performancestd{0.000}}{0.3274} & \performancecell{3.0555 \performancestd{1.535}}{\textbf{1.5486}} & \performancecell{3.4383 \performancestd{0.009}}{3.4302} \\
$\Delta$ & $\textcolor{green!45!black}{+1{,}055}\,/\,\textcolor{green!45!black}{+712}$ & $\textcolor{green!45!black}{+0.7138}\,/\,\textcolor{green!45!black}{+0.9263}$ & $\textcolor{green!45!black}{+0.0030}\,/\,\textcolor{red!70!black}{-0.0003}$ & $0.0000\,/\,0.0000$ & $\textcolor{red!70!black}{-0.1224}\,/\,\textcolor{green!45!black}{+0.1640}$ & $\textcolor{red!70!black}{-0.0040}\,/\,0.0000$ \\
\midrule
GPT-5.6-Luna ($N=8$) & \performancecell{18{,}032 \performancestd{1{,}472}}{16{,}354} & \performancecell{2.8027 \performancestd{2.331}}{\textbf{0.8830}} & \performancecell{1.8276 \performancestd{0.204}}{1.5223} & \performancecell{1.2857 \performancestd{1.917}}{0.3274} & \performancecell{\textbf{2.4516} \performancestd{1.016}}{\textbf{1.5644}} & \performancecell{\textbf{3.4302} \performancestd{0.000}}{3.4302} \\
+ \methodName & \performancecell{\textbf{17{,}690} \performancestd{1{,}997}}{\textbf{14{,}835}} & \performancecell{\textbf{1.7475} \performancestd{0.548}}{1.0238} & \performancecell{\textbf{1.7617} \performancestd{0.213}}{\textbf{1.4863}} & \performancecell{\textbf{0.3274} \performancestd{0.000}}{0.3274} & \performancecell{2.5281 \performancestd{1.153}}{1.6152} & \performancecell{3.4343 \performancestd{0.008}}{3.4302} \\
$\Delta$ & $\textcolor{green!45!black}{+342}\,/\,\textcolor{green!45!black}{+1{,}519}$ & $\textcolor{green!45!black}{+1.0552}\,/\,\textcolor{red!70!black}{-0.1408}$ & $\textcolor{green!45!black}{+0.0659}\,/\,\textcolor{green!45!black}{+0.0361}$ & $\textcolor{green!45!black}{+0.9584}\,/\,0.0000$ & $\textcolor{red!70!black}{-0.0766}\,/\,\textcolor{red!70!black}{-0.0508}$ & $\textcolor{red!70!black}{-0.0040}\,/\,0.0000$ \\
\midrule
Gemini-3.8-Flash ($N=1$) & \performancecell{\textbf{17{,}274} \performancestd{602}}{\textbf{16{,}848}} & \performancecell{6.1800 \performancestd{0.000}}{6.1800} & \performancecell{2.6272 \performancestd{0.000}}{2.6272} & \performancecell{0.3274 \performancestd{0.000}}{0.3274} & \performancecell{8.9578 \performancestd{0.000}}{8.9578} & \performancecell{\textbf{3.7407} \performancestd{0.439}}{\textbf{3.4302}} \\
+ \methodName & \performancecell{18{,}829 \performancestd{187}}{18{,}697} & \performancecell{\textbf{3.8116} \performancestd{3.349}}{\textbf{1.4433}} & \performancecell{\textbf{1.9318} \performancestd{0.006}}{\textbf{1.9274}} & \performancecell{0.3274 \performancestd{0.000}}{0.3274} & \performancecell{\textbf{5.6932} \performancestd{4.617}}{\textbf{2.4285}} & \performancecell{3.7488 \performancestd{0.428}}{3.4464} \\
$\Delta$ & $\textcolor{red!70!black}{-1{,}555}\,/\,\textcolor{red!70!black}{-1{,}849}$ & $\textcolor{green!45!black}{+2.3684}\,/\,\textcolor{green!45!black}{+4.7367}$ & $\textcolor{green!45!black}{+0.6954}\,/\,\textcolor{green!45!black}{+0.6998}$ & $0.0000\,/\,0.0000$ & $\textcolor{green!45!black}{+3.2647}\,/\,\textcolor{green!45!black}{+6.5293}$ & $\textcolor{red!70!black}{-0.0081}\,/\,\textcolor{red!70!black}{-0.0162}$ \\
\midrule
Gemini-3.8-Flash ($N=8$) & \performancecell{17{,}363 \performancestd{0.000}}{17{,}363} & \performancecell{\textbf{0.8200} \performancestd{0.000}}{\textbf{0.8200}} & \performancecell{\textbf{0.7302} \performancestd{0.010}}{\textbf{0.7228}} & \performancecell{0.3274 \performancestd{0.000}}{0.3274} & \performancecell{\textbf{1.5168} \performancestd{0.000}}{1.5168} & \performancecell{\textbf{3.4302} \performancestd{0.000}}{3.4302} \\
+ \methodName & \performancecell{\textbf{17{,}131} \performancestd{1{,}009}}{\textbf{15{,}650}} & \performancecell{0.8375 \performancestd{0.024}}{0.8202} & \performancecell{2.1030 \performancestd{0.349}}{1.9261} & \performancecell{0.3274 \performancestd{0.000}}{0.3274} & \performancecell{2.3869 \performancestd{1.363}}{\textbf{1.4227}} & \performancecell{3.7407 \performancestd{0.439}}{3.4302} \\
$\Delta$ & $\textcolor{green!45!black}{+232}\,/\,\textcolor{green!45!black}{+1{,}713}$ & $\textcolor{red!70!black}{-0.0175}\,/\,\textcolor{red!70!black}{-0.0002}$ & $\textcolor{red!70!black}{-1.3728}\,/\,\textcolor{red!70!black}{-1.2033}$ & $0.0000\,/\,0.0000$ & $\textcolor{red!70!black}{-0.8700}\,/\,\textcolor{green!45!black}{+0.0941}$ & $\textcolor{red!70!black}{-0.3105}\,/\,0.0000$ \\
\bottomrule
\end{NiceTabular}
\end{adjustbox}
\end{table*}

\begin{table*}[t]
\centering
\small
\setlength{\tabcolsep}{3.5pt}

\caption{Scientific algorithms and AI foundations.}
\vspace{3pt}
\label{supp_tab:mean-std-science}
\label{supp_tab:mean-std-ai}

\renewcommand{\arraystretch}{1.00}

\begin{adjustbox}{max width=\linewidth}
\begin{NiceTabular}{l*{5}{c}}
\CodeBefore
\evoduetgradientrows{5,8,11,14,17,20}{7}
\Body
\toprule
 & \multicolumn{1}{c}{\textbf{Scientific algorithms}} & \multicolumn{4}{c}{\textbf{AI foundations}} \\
\cmidrule(lr){2-2} \cmidrule(lr){3-6}
\textbf{Model / Method} & \textbf{Denoising} $\uparrow$ & \textbf{Domain Mix} $\uparrow$ & \textbf{U Shape} $\uparrow$ & \textbf{LR \& BSZ} $\uparrow$ & \textbf{Parallel} $\uparrow$ \\
\cmidrule(lr){2-2} \cmidrule(lr){3-3} \cmidrule(lr){4-4} \cmidrule(lr){5-5} \cmidrule(lr){6-6}
 & \textbf{Mean $\pm$ Std / Best} & \textbf{Mean $\pm$ Std / Best} & \textbf{Mean $\pm$ Std / Best} & \textbf{Mean $\pm$ Std / Best} & \textbf{Mean $\pm$ Std / Best} \\
\midrule
Qwen3.5-9B ($N=1$) & \performancecell{0.6378 \performancestd{0.0453}}{\textbf{0.7060}} & \performancecell{0.9212 \performancestd{0.0821}}{\textbf{0.9768}} & \performancecell{\textbf{-0.5470} \performancestd{0.704}}{\textbf{0.4512}} & \performancecell{\textbf{0.0233} \performancestd{0.660}}{\textbf{0.6966}} & \performancecell{\textbf{0.999579} \performancestd{0.000054}}{\textbf{0.999654}} \\
+ \methodName & \performancecell{\textbf{0.6543} \performancestd{0.0190}}{0.6705} & \performancecell{\textbf{0.9322} \performancestd{0.0552}}{0.9713} & \performancecell{-1.0000 \performancestd{0.000}}{-1.0000} & \performancecell{-0.3085 \performancestd{0.422}}{0.1899} & \performancecell{0.999552 \performancestd{0.000067}}{0.999612} \\
$\Delta$ & $\textcolor{green!45!black}{+0.0165}\,/\,\textcolor{red!70!black}{-0.0356}$ & $\textcolor{green!45!black}{+0.0110}\,/\,\textcolor{red!70!black}{-0.0055}$ & $\textcolor{red!70!black}{-0.4530}\,/\,\textcolor{red!70!black}{-1.4512}$ & $\textcolor{red!70!black}{-0.3318}\,/\,\textcolor{red!70!black}{-0.5068}$ & $\textcolor{red!70!black}{-0.000027}\,/\,\textcolor{red!70!black}{-0.000042}$ \\
\midrule
Qwen3.5-9B ($N=8$) & \performancecell{\textbf{0.7052} \performancestd{0.0070}}{\textbf{0.7111}} & \performancecell{0.9119 \performancestd{0.0799}}{0.9566} & \performancecell{\textbf{-0.7185} \performancestd{0.332}}{\textbf{-0.3552}} & \performancecell{\textbf{0.2270} \performancestd{0.833}}{\textbf{0.8107}} & \performancecell{0.999750 \performancestd{0.000108}}{0.999883} \\
+ \methodName & \performancecell{0.6752 \performancestd{0.0466}}{0.7081} & \performancecell{\textbf{0.9317} \performancestd{0.0466}}{\textbf{0.9646}} & \performancecell{-1.0000 \performancestd{0.000}}{-1.0000} & \performancecell{-0.5298 \performancestd{0.387}}{-0.2563} & \performancecell{\textbf{0.999876} \performancestd{0.000119}}{\textbf{0.999960}} \\
$\Delta$ & $\textcolor{red!70!black}{-0.0301}\,/\,\textcolor{red!70!black}{-0.0029}$ & $\textcolor{green!45!black}{+0.0198}\,/\,\textcolor{green!45!black}{+0.0080}$ & $\textcolor{red!70!black}{-0.2815}\,/\,\textcolor{red!70!black}{-0.6448}$ & $\textcolor{red!70!black}{-0.7568}\,/\,\textcolor{red!70!black}{-1.0670}$ & $\textcolor{green!45!black}{+0.000126}\,/\,\textcolor{green!45!black}{+0.000077}$ \\
\midrule
GPT-5.6-Luna ($N=1$) & \performancecell{0.6065 \performancestd{0.0711}}{0.7132} & \performancecell{0.9919 \performancestd{0.0010}}{0.9930} & \performancecell{-1.0000 \performancestd{0.000}}{-1.0000} & \performancecell{-0.5333 \performancestd{0.587}}{0.2172} & \performancecell{\textbf{0.999924} \performancestd{0.000082}}{\textbf{0.999967}} \\
+ \methodName & \performancecell{\textbf{0.7019} \performancestd{0.0201}}{\textbf{0.7152}} & \performancecell{\textbf{0.9920} \performancestd{0.0007}}{0.9930} & \performancecell{-1.0000 \performancestd{0.000}}{-1.0000} & \performancecell{\textbf{-0.4481} \performancestd{0.652}}{\textbf{0.2740}} & \performancecell{0.999883 \performancestd{0.000089}}{0.999960} \\
$\Delta$ & $\textcolor{green!45!black}{+0.0953}\,/\,\textcolor{green!45!black}{+0.0020}$ & $\textcolor{green!45!black}{+0.0001}\,/\,0.0000$ & $0.0000\,/\,0.0000$ & $\textcolor{green!45!black}{+0.0852}\,/\,\textcolor{green!45!black}{+0.0568}$ & $\textcolor{red!70!black}{-0.000041}\,/\,\textcolor{red!70!black}{-0.000007}$ \\
\midrule
GPT-5.6-Luna ($N=8$) & \performancecell{0.6906 \performancestd{0.0256}}{0.7067} & \performancecell{0.9929 \performancestd{0.0016}}{0.9948} & \performancecell{\textbf{-0.4368} \performancestd{0.670}}{\textbf{0.3225}} & \performancecell{-0.8987 \performancestd{0.203}}{-0.5950} & \performancecell{\textbf{0.999971} \performancestd{0.000009}}{\textbf{0.999981}} \\
+ \methodName & \performancecell{\textbf{0.6995} \performancestd{0.0366}}{\textbf{0.7229}} & \performancecell{\textbf{0.9940} \performancestd{0.0021}}{\textbf{0.9971}} & \performancecell{-1.0000 \performancestd{0.000}}{-1.0000} & \performancecell{\textbf{-0.3040} \performancestd{0.662}}{\textbf{0.3224}} & \performancecell{0.999965 \performancestd{0.000006}}{0.999970} \\
$\Delta$ & $\textcolor{green!45!black}{+0.0089}\,/\,\textcolor{green!45!black}{+0.0162}$ & $\textcolor{green!45!black}{+0.0011}\,/\,\textcolor{green!45!black}{+0.0023}$ & $\textcolor{red!70!black}{-0.5632}\,/\,\textcolor{red!70!black}{-1.3225}$ & $\textcolor{green!45!black}{+0.5947}\,/\,\textcolor{green!45!black}{+0.9174}$ & $\textcolor{red!70!black}{-0.000006}\,/\,\textcolor{red!70!black}{-0.000011}$ \\
\midrule
Gemini-3.8-Flash ($N=1$) & \performancecell{\textbf{0.7112} \performancestd{0.0005}}{\textbf{0.7116}} & \performancecell{0.9923 \performancestd{0.0002}}{0.9924} & \performancecell{-1.0000 \performancestd{0.000}}{-1.0000} & \performancecell{-0.9403 \performancestd{0.084}}{-0.8806} & \performancecell{0.999875 \performancestd{0.000115}}{0.999956} \\
+ \methodName & \performancecell{0.7101 \performancestd{0.0002}}{0.7102} & \performancecell{\textbf{0.9928} \performancestd{0.0002}}{\textbf{0.9929}} & \performancecell{\textbf{-0.5252} \performancestd{0.671}}{\textbf{-0.0504}} & \performancecell{\textbf{0.3445} \performancestd{0.613}}{\textbf{0.7781}} & \performancecell{\textbf{0.999926} \performancestd{0.000048}}{\textbf{0.999960}} \\
$\Delta$ & $\textcolor{red!70!black}{-0.0011}\,/\,\textcolor{red!70!black}{-0.0014}$ & $\textcolor{green!45!black}{+0.0005}\,/\,\textcolor{green!45!black}{+0.0005}$ & $\textcolor{green!45!black}{+0.4748}\,/\,\textcolor{green!45!black}{+0.9496}$ & $\textcolor{green!45!black}{+1.2848}\,/\,\textcolor{green!45!black}{+1.6587}$ & $\textcolor{green!45!black}{+0.000051}\,/\,\textcolor{green!45!black}{+0.000004}$ \\
\midrule
Gemini-3.8-Flash ($N=8$) & \performancecell{\textbf{0.7100} \performancestd{0.000}}{\textbf{0.7100}} & \performancecell{\textbf{0.9962} \performancestd{0.000}}{\textbf{0.9962}} & \performancecell{\textbf{-0.9642} \performancestd{0.000}}{\textbf{-0.9642}} & \performancecell{-0.0528 \performancestd{1.340}}{\textbf{0.8945}} & \performancecell{0.999964 \performancestd{0.000007}}{0.999969} \\
+ \methodName & \performancecell{0.6915 \performancestd{0.0333}}{0.7093} & \performancecell{0.9928 \performancestd{0.0001}}{0.9929} & \performancecell{-1.0000 \performancestd{0.000}}{-1.0000} & \performancecell{\textbf{-0.0457} \performancestd{0.934}}{0.8666} & \performancecell{\textbf{0.999972} \performancestd{0.000004}}{\textbf{0.999975}} \\
$\Delta$ & $\textcolor{red!70!black}{-0.0186}\,/\,\textcolor{red!70!black}{-0.0007}$ & $\textcolor{red!70!black}{-0.0035}\,/\,\textcolor{red!70!black}{-0.0034}$ & $\textcolor{red!70!black}{-0.0358}\,/\,\textcolor{red!70!black}{-0.0358}$ & $\textcolor{green!45!black}{+0.0070}\,/\,\textcolor{red!70!black}{-0.0279}$ & $\textcolor{green!45!black}{+0.000008}\,/\,\textcolor{green!45!black}{+0.000006}$ \\
\bottomrule
\end{NiceTabular}
\end{adjustbox}
\end{table*}

\begin{table*}[t]
\centering
\small
\setlength{\tabcolsep}{3.5pt}

\caption{Mathematics.}
\vspace{3pt}
\label{supp_tab:mean-std-math}
\label{supp_tab:mean-std-packing}

\renewcommand{\arraystretch}{1.00}

\begin{adjustbox}{max width=\linewidth}
\begin{NiceTabular}{l*{8}{c}}
\CodeBefore
\evoduetgradientrows{5,8,11,14,17,20}{10}
\Body
\toprule
 & \multicolumn{8}{c}{\textbf{Mathematics}} \\
\cmidrule(lr){2-9}
\textbf{Model / Method} & \textbf{Erd\H{o}s} $\downarrow$ & \textbf{AC1} $\downarrow$ & \textbf{AC2} $\uparrow$ & \textbf{AC3} $\downarrow$ & \textbf{CP($n$=26)} $\uparrow$ & \textbf{CP($n$=32)} $\uparrow$ & \textbf{Hadamard} $\uparrow$ & \textbf{Sums/Diffs} $\uparrow$ \\
\cmidrule(lr){2-2} \cmidrule(lr){3-3} \cmidrule(lr){4-4} \cmidrule(lr){5-5} \cmidrule(lr){6-6} \cmidrule(lr){7-7} \cmidrule(lr){8-8} \cmidrule(lr){9-9}
 & \textbf{Mean $\pm$ Std / Best} & \textbf{Mean $\pm$ Std / Best} & \textbf{Mean $\pm$ Std / Best} & \textbf{Mean $\pm$ Std / Best} & \textbf{Mean $\pm$ Std / Best} & \textbf{Mean $\pm$ Std / Best} & \textbf{Mean $\pm$ Std / Best} & \textbf{Mean $\pm$ Std / Best} \\
\midrule
Qwen3.5-9B ($N=1$) & \performancecell{0.389821 \performancestd{0.007589}}{\textbf{0.381152}} & \performancecell{\textbf{1.5128} \performancestd{0.0019}}{\textbf{1.5099}} & \performancecell{\textbf{0.9253} \performancestd{0.0101}}{\textbf{0.9366}} & \performancecell{\textbf{1.7978} \performancestd{0.1889}}{\textbf{1.5940}} & \performancecell{2.0995 \performancestd{0.310}}{2.4103} & \performancecell{2.6367 \performancestd{0.050}}{2.6810} & \performancecell{\textbf{0.4946} \performancestd{0.0265}}{\textbf{0.5225}} & \performancecell{1.064772 \performancestd{0.011103}}{1.087358} \\
+ \methodName & \performancecell{\textbf{0.389166} \performancestd{0.003805}}{0.383647} & \performancecell{1.5305 \performancestd{0.0293}}{1.5159} & \performancecell{0.9184 \performancestd{0.0099}}{0.9282} & \performancecell{1.8663 \performancestd{0.2128}}{1.7158} & \performancecell{\textbf{2.6313} \performancestd{0.006}}{\textbf{2.6353}} & \performancecell{\textbf{2.9206} \performancestd{0.010}}{\textbf{2.9280}} & \performancecell{0.4373 \performancestd{0.0628}}{0.4786} & \performancecell{\textbf{1.086398} \performancestd{0.011955}}{\textbf{1.102270}} \\
$\Delta$ & $\textcolor{green!45!black}{+0.000655}\,/\,\textcolor{red!70!black}{-0.002495}$ & $\textcolor{red!70!black}{-0.0177}\,/\,\textcolor{red!70!black}{-0.0060}$ & $\textcolor{red!70!black}{-0.0068}\,/\,\textcolor{red!70!black}{-0.0085}$ & $\textcolor{red!70!black}{-0.0685}\,/\,\textcolor{red!70!black}{-0.1218}$ & $\textcolor{green!45!black}{+0.5318}\,/\,\textcolor{green!45!black}{+0.2250}$ & $\textcolor{green!45!black}{+0.2838}\,/\,\textcolor{green!45!black}{+0.2470}$ & $\textcolor{red!70!black}{-0.0573}\,/\,\textcolor{red!70!black}{-0.0438}$ & $\textcolor{green!45!black}{+0.021626}\,/\,\textcolor{green!45!black}{+0.014912}$ \\
\midrule
Qwen3.5-9B ($N=8$) & \performancecell{0.386232 \performancestd{0.003623}}{0.382699} & \performancecell{\textbf{1.5521} \performancestd{0.0725}}{1.5159} & \performancecell{0.9134 \performancestd{0.0117}}{\textbf{0.9306}} & \performancecell{1.6036 \performancestd{0.0202}}{1.5737} & \performancecell{1.9488 \performancestd{0.645}}{2.5171} & \performancecell{2.7595 \performancestd{0.069}}{2.8081} & \performancecell{0.2264 \performancestd{0.1662}}{0.4756} & \performancecell{\textbf{1.063549} \performancestd{0.002782}}{1.066260} \\
+ \methodName & \performancecell{\textbf{0.384427} \performancestd{0.003728}}{\textbf{0.381791}} & \performancecell{1.5768 \performancestd{0.0906}}{\textbf{1.5127}} & \performancecell{\textbf{0.9265} \performancestd{0.0011}}{0.9273} & \performancecell{\textbf{1.5920} \performancestd{0.0283}}{\textbf{1.5720}} & \performancecell{\textbf{2.5352} \performancestd{0.077}}{\textbf{2.5895}} & \performancecell{\textbf{2.8647} \performancestd{0.071}}{\textbf{2.9149}} & \performancecell{\textbf{0.5256} \performancestd{0.0203}}{\textbf{0.5399}} & \performancecell{1.063292 \performancestd{0.004949}}{\textbf{1.066792}} \\
$\Delta$ & $\textcolor{green!45!black}{+0.001804}\,/\,\textcolor{green!45!black}{+0.000908}$ & $\textcolor{red!70!black}{-0.0246}\,/\,\textcolor{green!45!black}{+0.0032}$ & $\textcolor{green!45!black}{+0.0131}\,/\,\textcolor{red!70!black}{-0.0034}$ & $\textcolor{green!45!black}{+0.0116}\,/\,\textcolor{green!45!black}{+0.0017}$ & $\textcolor{green!45!black}{+0.5864}\,/\,\textcolor{green!45!black}{+0.0724}$ & $\textcolor{green!45!black}{+0.1052}\,/\,\textcolor{green!45!black}{+0.1068}$ & $\textcolor{green!45!black}{+0.2992}\,/\,\textcolor{green!45!black}{+0.0643}$ & $\textcolor{red!70!black}{-0.000256}\,/\,\textcolor{green!45!black}{+0.000531}$ \\
\midrule
GPT-5.6-Luna ($N=1$) & \performancecell{0.386347 \performancestd{0.009176}}{0.380967} & \performancecell{\textbf{1.5109} \performancestd{0.0009}}{1.5097} & \performancecell{0.9288 \performancestd{0.0143}}{0.9465} & \performancecell{\textbf{1.4586} \performancestd{0.0018}}{\textbf{1.4571}} & \performancecell{\textbf{2.6360} \performancestd{0.000}}{2.6360} & \performancecell{\textbf{2.9396} \performancestd{0.000}}{2.9396} & \performancecell{\textbf{0.7221} \performancestd{0.1390}}{0.8626} & \performancecell{1.072913 \performancestd{0.001035}}{1.073430} \\
+ \methodName & \performancecell{\textbf{0.380937} \performancestd{0.000101}}{\textbf{0.380859}} & \performancecell{1.5111 \performancestd{0.0022}}{\textbf{1.5085}} & \performancecell{\textbf{0.9430} \performancestd{0.0040}}{0.9465} & \performancecell{1.4623 \performancestd{0.0020}}{1.4600} & \performancecell{2.6356 \performancestd{0.001}}{2.6360} & \performancecell{2.9369 \performancestd{0.005}}{2.9396} & \performancecell{0.6606 \performancestd{0.1737}}{\textbf{0.9211}} & \performancecell{\textbf{1.087436} \performancestd{0.010649}}{\textbf{1.098610}} \\
$\Delta$ & $\textcolor{green!45!black}{+0.005410}\,/\,\textcolor{green!45!black}{+0.000108}$ & $\textcolor{red!70!black}{-0.0002}\,/\,\textcolor{green!45!black}{+0.0011}$ & $\textcolor{green!45!black}{+0.0142}\,/\,0.0000$ & $\textcolor{red!70!black}{-0.0036}\,/\,\textcolor{red!70!black}{-0.0030}$ & $\textcolor{red!70!black}{-0.0004}\,/\,0.0000$ & $\textcolor{red!70!black}{-0.0027}\,/\,0.0000$ & $\textcolor{red!70!black}{-0.0615}\,/\,\textcolor{green!45!black}{+0.0585}$ & $\textcolor{green!45!black}{+0.014524}\,/\,\textcolor{green!45!black}{+0.025179}$ \\
\midrule
GPT-5.6-Luna ($N=8$) & \performancecell{\textbf{0.381008} \performancestd{0.000082}}{0.380917} & \performancecell{1.5135 \performancestd{0.0009}}{1.5127} & \performancecell{0.9281 \performancestd{0.0162}}{0.9386} & \performancecell{1.4604 \performancestd{0.0036}}{1.4581} & \performancecell{2.3246 \performancestd{0.623}}{2.6360} & \performancecell{2.6354 \performancestd{0.608}}{2.9396} & \performancecell{0.8381 \performancestd{0.1659}}{0.9211} & \performancecell{1.081721 \performancestd{0.004439}}{1.086459} \\
+ \methodName & \performancecell{0.382840 \performancestd{0.003888}}{\textbf{0.380859}} & \performancecell{\textbf{1.5082} \performancestd{0.0012}}{\textbf{1.5068}} & \performancecell{\textbf{0.9466} \performancestd{0.0014}}{\textbf{0.9481}} & \performancecell{\textbf{1.4569} \performancestd{0.0008}}{\textbf{1.4564}} & \performancecell{\textbf{2.6360} \performancestd{0.000}}{2.6360} & \performancecell{\textbf{2.9396} \performancestd{0.000}}{2.9396} & \performancecell{\textbf{0.9247} \performancestd{0.0073}}{\textbf{0.9357}} & \performancecell{\textbf{1.105613} \performancestd{0.021767}}{\textbf{1.124462}} \\
$\Delta$ & $\textcolor{red!70!black}{-0.001832}\,/\,\textcolor{green!45!black}{+0.000058}$ & $\textcolor{green!45!black}{+0.0053}\,/\,\textcolor{green!45!black}{+0.0058}$ & $\textcolor{green!45!black}{+0.0185}\,/\,\textcolor{green!45!black}{+0.0095}$ & $\textcolor{green!45!black}{+0.0036}\,/\,\textcolor{green!45!black}{+0.0017}$ & $\textcolor{green!45!black}{+0.3114}\,/\,0.0000$ & $\textcolor{green!45!black}{+0.3042}\,/\,0.0000$ & $\textcolor{green!45!black}{+0.0866}\,/\,\textcolor{green!45!black}{+0.0146}$ & $\textcolor{green!45!black}{+0.023892}\,/\,\textcolor{green!45!black}{+0.038004}$ \\
\midrule
Gemini-3.8-Flash ($N=1$) & \performancecell{0.380950 \performancestd{0.000004}}{0.380947} & \performancecell{1.5127 \performancestd{0.0028}}{1.5107} & \performancecell{0.9197 \performancestd{0.0184}}{0.9327} & \performancecell{1.4994 \performancestd{0.0201}}{1.4852} & \performancecell{2.6360 \performancestd{0.000}}{2.6360} & \performancecell{2.9396 \performancestd{0.000}}{2.9396} & \performancecell{0.8904 \performancestd{0.0434}}{0.9211} & \performancecell{1.089561 \performancestd{0.014543}}{1.099845} \\
+ \methodName & \performancecell{\textbf{0.380905} \performancestd{0.000005}}{\textbf{0.380902}} & \performancecell{\textbf{1.5069} \performancestd{0.0005}}{\textbf{1.5066}} & \performancecell{\textbf{0.9272} \performancestd{0.0275}}{\textbf{0.9466}} & \performancecell{\textbf{1.4597} \performancestd{0.0011}}{\textbf{1.4589}} & \performancecell{2.6360 \performancestd{0.000}}{2.6360} & \performancecell{2.9396 \performancestd{0.000}}{2.9396} & \performancecell{\textbf{0.9357} \performancestd{0.0000}}{\textbf{0.9357}} & \performancecell{\textbf{1.135834} \performancestd{0.006336}}{\textbf{1.140314}} \\
$\Delta$ & $\textcolor{green!45!black}{+0.000045}\,/\,\textcolor{green!45!black}{+0.000046}$ & $\textcolor{green!45!black}{+0.0057}\,/\,\textcolor{green!45!black}{+0.0041}$ & $\textcolor{green!45!black}{+0.0075}\,/\,\textcolor{green!45!black}{+0.0139}$ & $\textcolor{green!45!black}{+0.0397}\,/\,\textcolor{green!45!black}{+0.0263}$ & $0.0000\,/\,0.0000$ & $0.0000\,/\,0.0000$ & $\textcolor{green!45!black}{+0.0453}\,/\,\textcolor{green!45!black}{+0.0146}$ & $\textcolor{green!45!black}{+0.046272}\,/\,\textcolor{green!45!black}{+0.040469}$ \\
\midrule
Gemini-3.8-Flash ($N=8$) & \performancecell{0.380903 \performancestd{0.000009}}{0.380896} & \performancecell{1.5085 \performancestd{0.000}}{1.5085} & \performancecell{\textbf{0.9455} \performancestd{0.0075}}{\textbf{0.9508}} & \performancecell{1.4586 \performancestd{0.000}}{1.4586} & \performancecell{2.6360 \performancestd{0.000}}{2.6360} & \performancecell{2.9396 \performancestd{0.000}}{2.9396} & \performancecell{0.8596 \performancestd{0.000}}{0.8596} & \performancecell{1.124049 \performancestd{0.000}}{1.124049} \\
+ \methodName & \performancecell{\textbf{0.380898} \performancestd{0.000007}}{\textbf{0.380893}} & \performancecell{\textbf{1.5065} \performancestd{0.0015}}{\textbf{1.5055}} & \performancecell{0.9429 \performancestd{0.0066}}{0.9476} & \performancecell{\textbf{1.4579} \performancestd{0.0022}}{\textbf{1.4564}} & \performancecell{2.6360 \performancestd{0.000}}{2.6360} & \performancecell{2.9396 \performancestd{0.000}}{2.9396} & \performancecell{\textbf{0.9357} \performancestd{0.0000}}{\textbf{0.9357}} & \performancecell{\textbf{1.143770} \performancestd{0.001737}}{\textbf{1.144999}} \\
$\Delta$ & $\textcolor{green!45!black}{+0.000004}\,/\,\textcolor{green!45!black}{+0.000003}$ & $\textcolor{green!45!black}{+0.0020}\,/\,\textcolor{green!45!black}{+0.0030}$ & $\textcolor{red!70!black}{-0.0026}\,/\,\textcolor{red!70!black}{-0.0032}$ & $\textcolor{green!45!black}{+0.0006}\,/\,\textcolor{green!45!black}{+0.0022}$ & $0.0000\,/\,0.0000$ & $0.0000\,/\,0.0000$ & $\textcolor{green!45!black}{+0.0760}\,/\,\textcolor{green!45!black}{+0.0760}$ & $\textcolor{green!45!black}{+0.019721}\,/\,\textcolor{green!45!black}{+0.020950}$ \\
\bottomrule
\end{NiceTabular}
\end{adjustbox}
\end{table*}

\begin{table*}[t]
\centering
\small
\setlength{\tabcolsep}{3.5pt}

\caption{Algorithm engineering.}
\vspace{3pt}
\label{supp_tab:mean-std-ahc}
\label{supp_tab:mean-std-frontier}

\renewcommand{\arraystretch}{1.00}

\begin{adjustbox}{max width=\linewidth}
\begin{NiceTabular}{l*{2}{c}}
\CodeBefore
\evoduetgradientrows{5,8,11,14,17,20}{4}
\Body
\toprule
 & \multicolumn{2}{c}{\textbf{Algorithm engineering}} \\
\cmidrule(lr){2-3}
\textbf{Model / Method} & \textbf{AHC039} $\uparrow$ & \textbf{AHC058} $\uparrow$ \\
\cmidrule(lr){2-2} \cmidrule(lr){3-3}
 & \textbf{Mean $\pm$ Std / Best} & \textbf{Mean $\pm$ Std / Best} \\
\midrule
Qwen3.5-9B ($N=1$) & \performancecell{\textbf{554{,}499} \performancestd{2{,}471}}{\textbf{558{,}387}} & \performancecell{489{,}794{,}634 \performancestd{269{,}878{,}148}}{731{,}576{,}599} \\
+ \methodName & \performancecell{545{,}315 \performancestd{9{,}468}}{553{,}414} & \performancecell{\textbf{564{,}391{,}050} \performancestd{288{,}462{,}123}}{\textbf{732{,}606{,}033}} \\
$\Delta$ & $\textcolor{red!70!black}{-9{,}184}\,/\,\textcolor{red!70!black}{-4{,}973}$ & $\textcolor{green!45!black}{+74{,}596{,}416}\,/\,\textcolor{green!45!black}{+1{,}029{,}434}$ \\
\midrule
Qwen3.5-9B ($N=8$) & \performancecell{\textbf{554{,}619} \performancestd{2{,}448}}{\textbf{556{,}857}} & \performancecell{\textbf{774{,}976{,}672} \performancestd{16{,}261{,}216}}{\textbf{784{,}638{,}268}} \\
+ \methodName & \performancecell{450{,}025 \performancestd{137{,}654}}{547{,}361} & \performancecell{750{,}634{,}788 \performancestd{31{,}914{,}083}}{773{,}201{,}452} \\
$\Delta$ & $\textcolor{red!70!black}{-104{,}594}\,/\,\textcolor{red!70!black}{-9{,}496}$ & $\textcolor{red!70!black}{-24{,}341{,}884}\,/\,\textcolor{red!70!black}{-11{,}436{,}816}$ \\
\midrule
GPT-5.6-Luna ($N=1$) & \performancecell{\textbf{549{,}694} \performancestd{8{,}635}}{\textbf{556{,}948}} & \performancecell{784{,}199{,}030 \performancestd{3{,}491{,}251}}{787{,}862{,}153} \\
+ \methodName & \performancecell{547{,}362 \performancestd{1{,}702}}{548{,}660} & \performancecell{\textbf{784{,}263{,}233} \performancestd{7{,}949{,}405}}{\textbf{790{,}891{,}264}} \\
$\Delta$ & $\textcolor{red!70!black}{-2{,}332}\,/\,\textcolor{red!70!black}{-8{,}288}$ & $\textcolor{green!45!black}{+64{,}202}\,/\,\textcolor{green!45!black}{+3{,}029{,}111}$ \\
\midrule
GPT-5.6-Luna ($N=8$) & \performancecell{\textbf{552{,}132} \performancestd{5{,}993}}{\textbf{558{,}223}} & \performancecell{\textbf{799{,}284{,}884} \performancestd{9{,}407{,}424}}{804{,}359{,}003} \\
+ \methodName & \performancecell{536{,}812 \performancestd{12{,}818}}{546{,}430} & \performancecell{742{,}512{,}760 \performancestd{144{,}722{,}471}}{\textbf{816{,}714{,}532}} \\
$\Delta$ & $\textcolor{red!70!black}{-15{,}320}\,/\,\textcolor{red!70!black}{-11{,}793}$ & $\textcolor{red!70!black}{-56{,}772{,}124}\,/\,\textcolor{green!45!black}{+12{,}355{,}529}$ \\
\midrule
Gemini-3.8-Flash ($N=1$) & \performancecell{544{,}205 \performancestd{1{,}484}}{545{,}254} & \performancecell{801{,}414{,}461 \performancestd{46{,}248{,}047}}{834{,}116{,}769} \\
+ \methodName & \performancecell{\textbf{550{,}097} \performancestd{9{,}545}}{\textbf{556{,}846}} & \performancecell{\textbf{806{,}467{,}666} \performancestd{39{,}888{,}838}}{\textbf{834{,}673{,}334}} \\
$\Delta$ & $\textcolor{green!45!black}{+5{,}892}\,/\,\textcolor{green!45!black}{+11{,}592}$ & $\textcolor{green!45!black}{+5{,}053{,}206}\,/\,\textcolor{green!45!black}{+556{,}565}$ \\
\midrule
Gemini-3.8-Flash ($N=8$) & \performancecell{\textbf{562{,}175} \performancestd{0.000}}{562{,}175} & \performancecell{839{,}834{,}047 \performancestd{0.000}}{839{,}834{,}047} \\
+ \methodName & \performancecell{561{,}054 \performancestd{3{,}577}}{\textbf{563{,}583}} & \performancecell{\textbf{845{,}831{,}924} \performancestd{4{,}738{,}384}}{\textbf{849{,}182{,}468}} \\
$\Delta$ & $\textcolor{red!70!black}{-1{,}121}\,/\,\textcolor{green!45!black}{+1{,}408}$ & $\textcolor{green!45!black}{+5{,}997{,}878}\,/\,\textcolor{green!45!black}{+9{,}348{,}421}$ \\
\bottomrule
\end{NiceTabular}
\end{adjustbox}

\end{table*}
\endgroup

\clearpage
\section{Behavior and Search Analysis}
\label{supp_sec:behavior_analysis}

\subsection{Behavior Flags and Additional Examples}
\label{supp_sec:behavior_examples}

We define the six behavior flags in Figure~\ref{main_fig:behavior_audit}, report additional retrieval patterns, and examine what the flags capture in individual runs. Appendix~\ref{supp_sec:trajectories} provides the corresponding queries, sources, code edits, and evaluation results for seven worked cases.

\subsubsection{Audit Setup and Counting Rules}

\paragraph{Records and scope.}
The GPT-5.6-Luna audit covers 8{,}200 iterations from 82 runs on the 21 tasks in the main results. The Gemini-3.8-Flash audit covers 14 runs on seven tasks and uses iterations with recorded gate decisions. For each iteration, we inspect the gate decision, queries, retained documents, predicted child scores, and the selected child's changes relative to its parent. A document is \emph{promising} when its predicted child score exceeds the parent's score. We call a retrieval \emph{predicted to help} if it retains at least one such document.

\paragraph{Behavior definitions and flags.}
Each run is counted once per behavior, regardless of how often it occurs; categories can overlap. Regular expressions classify the final query as a known-result lookup, a method or implementation search, a problem-statement lookup, or other. The following rules identify candidate instances of each behavior. The case reviews below distinguish these flags from evidence that a retrieved idea was implemented and retained.

\begin{behaviorbox}
\begin{itemize}[leftmargin=*, itemsep=3pt, parsep=0pt, topsep=2pt, partopsep=0pt]
    \item \textbf{Method transfer:} A retrieved method or implementation detail becomes a program component. We flag a run when a method or implementation query retrieves a promising document and the child sets a new run best.
    \item \textbf{Cross-domain transfer:} The model adapts a method or artifact from another scientific field. The flag requires a new run best with a retained document from another field, identified by its title and host. General-purpose numerical tools, such as L-BFGS, trust-region methods, and SciPy, are excluded.
    \item \textbf{Live documentation lookup:} Retrieving documentation to check an interface. The flag requires an API, signature, version, release-note, or documentation query; a retained documentation page; and a child that beats its parent. It does not require a verified API correction.
    \item \textbf{Target-informed reconstruction:} A published value or construction guides development of a solver. We flag a run when a known-result query retrieves a promising document and the child sets a new run best, with no runtime download, copied code line, or pasted numerical solution.
    \item \textbf{Public artifact reuse:} A selected child downloads a published witness or solution at run time, or embeds numerical values from a published solution. Both forms are manually checked.
    \item \textbf{Evidence-inspired over-reach:} A retrieval predicted to help produces a child whose score falls below its parent's by more than 10\% of the run's total gain. Runs with no positive total gain do not receive this flag.
\end{itemize}
\end{behaviorbox}

\paragraph{Artifact checks.}
Direct code copying occurs in 19 GPT-5.6-Luna iterations (0.2\%): 16 copy a single line containing an import, path, or function signature, and three copy part of a Sums/Diffs integer set. The numeric-copy check flags added code containing at least three numbers with five or more significant digits that appear together, exactly or after rounding, on a page retrieved by that iteration. Manual review identifies 59 GPT-5.6-Luna iterations (0.7\%) in four runs: three reuse published Rosetta trajectories or flyby dates, and one reuses an AC2 step-function construction. Only one of these four runs retains a copied solution in its final best program. One Gemini-3.8-Flash iteration reuses an AC1 construction. We also inspect runtime network calls; an AHC039 identifier that matches the download detector is excluded as a false positive.

\paragraph{Additional patterns and denominators.}
\emph{Search miss with own progress} denotes a new run best after a retrieval with no promising document (57 of 82 GPT-5.6-Luna runs). \emph{Identifier fixation} denotes a PR or issue number, or an arXiv identifier, appearing in at least 30\% of a run's retrieval queries; this is assessed only for runs with at least ten retrievals (6 of 75). Figure~\ref{main_fig:behavior_audit}(b) separately measures whether the selected child beats its parent or the previous run best, including at iterations without search. Appendix~\ref{supp_sec:model_failures} compares program revision outcomes with and without documents across all three models.

\subsubsection{What the Worked Cases Establish}

\paragraph{Method transfer: Swap Reduction (Appendix~\ref{supp_sec:traj_swap}).}
At iteration~5, the gate describes the stored documents as ``largely conceptual''. The retrieved dSABRE paper supplies exponential depth weighting, which becomes a term in the SWAP scorer and remains in every later best program. The final router adds $14{,}835$ SWAPs on Q20, compared with $15{,}186$ for the released SimpleTES program under the same evaluator. This case verifies implementation and persistence of a retrieved mechanism, beyond the method-transfer flag assigned to 55 of 82 runs.

\paragraph{Cross-domain transfer: Erd\H{o}s minimum overlap (Appendix~\ref{supp_sec:traj_erdos}).}
The model retrieves a low-sidelobe radar codeword, \texttt{04CF5A2471657C6F}, and uses it to seed an optimizer for an additive number-theory problem. All four subsequent improvements to the run best retain the codeword, reducing $C_5$ from 0.381374 to 0.380915. The first improvement follows a look-up two iterations after retrieval. Inspection of the 13 flagged runs also finds weaker cases: the Anscombe transform is retrieved in three Denoising runs but appears in no candidate, while a quadratic-residue initialization appears in ten selected Erd\H{o}s children but is absent from that run's best program. By comparison, a variable-projection formulation introduced on Parallel Scaling at iteration~30, which accounts for 56\% of that run's gain, remains in its best program at iteration~84. A cross-domain source alone therefore does not establish lasting transfer.

\paragraph{Live documentation lookup: Voyager~2 (Appendix~\ref{supp_sec:traj_voyager}).}
The rule flags 72 retrievals in 27 runs; 13 of these iterations set a new best. Of the 72, 21 ask for an astrodynamics wrapper already documented in the task prompt, and 18 target a pull-request diff. In Voyager~2, 28 of 46 retrievals name the wrapper, but the retained pages describe public Lambert solvers. At iteration~41, a new tour reduces $\Delta v$ from 3.457 to 3.447\,km/s and explicitly passes \texttt{lowpath=True, M=0}. These are the existing defaults, so they do not change the solver branch; the final best program does not descend from this child. Across 300 wrapper-targeting retrievals, only two retain a page from the benchmark's source, and neither improves on its parent. The flag thus captures documentation-seeking behavior more broadly than verified API corrections.

\paragraph{Target-informed reconstruction: circle packing, $n$=26 (Appendix~\ref{supp_sec:traj_cp26}).}
The first query names a best-known sum of radii, 2.635977. The returned pages supply no coordinates, and the model writes an LP-plus-SLSQP solver that raises the objective from 1.390 to 2.628 in one iteration. At iteration~7, it reaches 2.635983, exceeding the queried target and matching the SimpleTES reference to six decimals. Later queries seek the record holder's source file, but the best program contains a generated solver with no retrieved coordinates or runtime download. Target-informed reconstruction is flagged in 40 of 82 GPT-5.6-Luna runs and 12 of 14 Gemini-3.8-Flash runs.

\paragraph{Public artifact reuse: Erd\H{o}s minimum overlap, second run (Appendix~\ref{supp_sec:traj_artifact}).}
At iteration~17, the child downloads a published witness and retains its optimizer as a fallback, reducing the run's best $C_5$ from 0.381017 to 0.380859. The best program without a runtime download reaches 0.381004. The program reported in the Erd\H{o}s row of Table~\ref{main_tab:new_sota_result} retains this download, so its result depends on access to the published artifact (Appendix~\ref{supp_sec:best_programs}). Two of 82 GPT-5.6-Luna runs download artifacts at run time; four others embed published numerical solutions, including three Rosetta trajectories or sets of flyby dates.

\paragraph{Evidence-inspired over-reach: Galileo (Appendix~\ref{supp_sec:traj_galileo}).}
At iteration~33, a document's predicted child score is 0.63701 (parent: 0.63692), but the rewrite scores 0.37713, raising $\Delta v$ from 1.926 to 3.873\,km/s. The run best remains 0.63692; iteration~34 recovers this score from a different parent without retrieval. Across GPT-5.6-Luna runs, 20\% of retrievals predicted to help produce a worse child, and 29 of 82 runs contain a drop exceeding 10\% of their total gain. Predicted gains therefore require validation by the evaluator.

\clearpage
\subsection{Language, Resources, and Reuse in Web Search}
\label{supp_sec:search_dynamics}

\begin{figure}[!ht]
    \centering
    \captionsetup[subfigure]{font={scriptsize},labelfont={bf},skip=2pt,position=bottom,justification=centering,singlelinecheck=false}
    \begin{subfigure}[t]{0.49\linewidth}
        \centering
        \includegraphics[width=\linewidth]{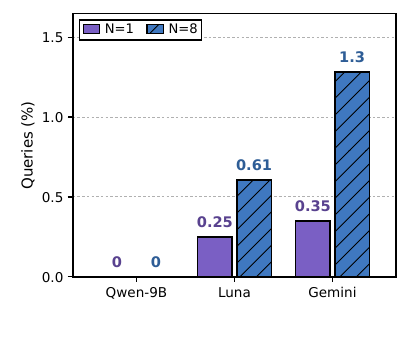}
        \caption{Queries with Japanese phrases}
    \end{subfigure}\hfill
    \begin{subfigure}[t]{0.49\linewidth}
        \centering
        \includegraphics[width=\linewidth]{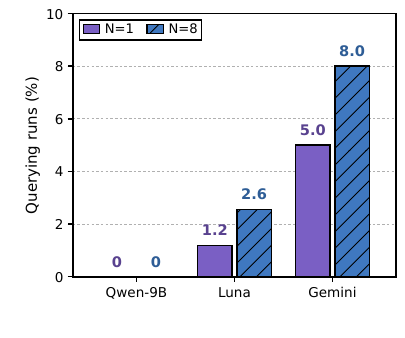}
        \caption{Runs with Japanese phrases}
    \end{subfigure}
    \vspace{-0.4em}
    \caption{\textbf{Japanese phrases in search queries.} (a) Share of parseable queries containing Japanese phrases, including repeated outputs. (b) Share of querying runs with at least one such query. Purple and hatched blue bars denote $N=1$ and $N=8$. Accented names such as Erd\H{o}s are excluded.}
    \label{supp_fig:multilingual_queries}
\end{figure}

\paragraph{Multilingual queries are rare and confined to contest tasks.}
Figure~\ref{supp_fig:multilingual_queries} shows Japanese phrases in 134 of 32{,}008 parseable queries (0.42\%), all on AHC039 or AHC058. These occur in 7 of 338 querying runs (2.1\%), with none from Qwen3.5-9B. We inspect non-English-script matches and Latin-script language-detector candidates; the latter contain only English terms, names, and identifiers. This conservative screen covers query text only.

\paragraph{Resource preferences vary across task domains.}
GitHub is the most exposed host in four of the six task domains, reaching 94.9\% of retrieval iterations in quantum compilation. arXiv leads AI foundations (69.6\%) and algorithm engineering (48.5\%; Figure~\ref{supp_fig:resource_heatmap}). These rates measure source exposure only.

\begin{figure}[!ht]
    \centering
    \captionsetup[subfigure]{font={scriptsize},labelfont={bf},skip=2pt,position=bottom,justification=centering,singlelinecheck=false}
    \begin{subfigure}[t]{0.49\linewidth}
        \centering
        \includegraphics[width=\linewidth]{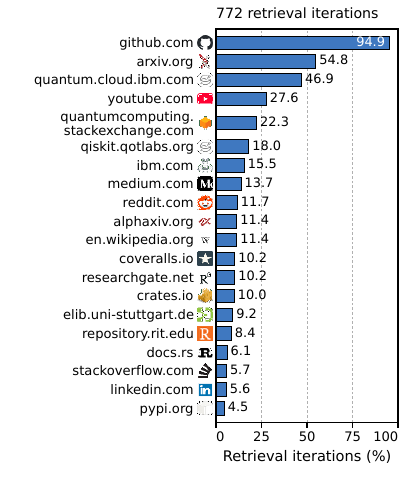}
        \caption{Quantum compilation}
    \end{subfigure}\hfill
    \begin{subfigure}[t]{0.49\linewidth}
        \centering
        \includegraphics[width=\linewidth]{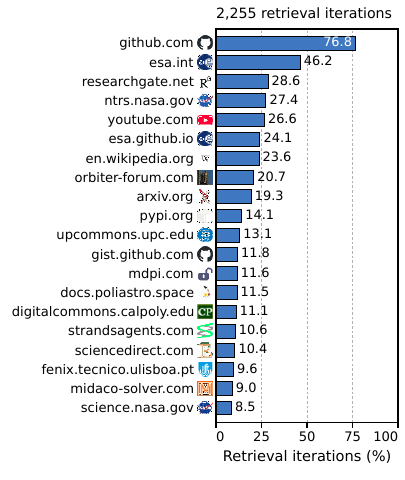}
        \caption{Astrodynamics}
    \end{subfigure}
    \par\medskip
    \begin{subfigure}[t]{0.49\linewidth}
        \centering
        \includegraphics[width=\linewidth]{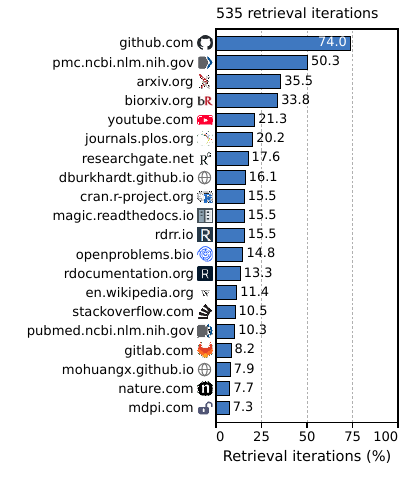}
        \caption{Scientific algorithms}
    \end{subfigure}\hfill
    \begin{subfigure}[t]{0.49\linewidth}
        \centering
        \includegraphics[width=\linewidth]{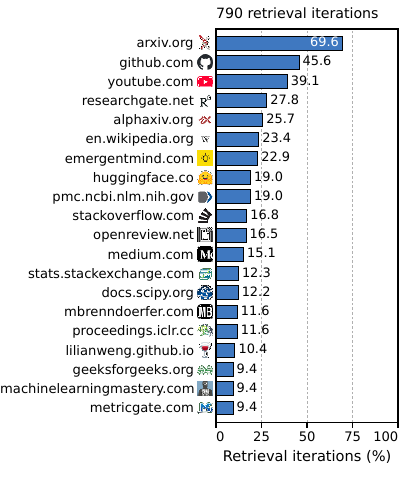}
        \caption{AI foundations}
    \end{subfigure}
    \vspace{-0.4em}
    \caption{\textbf{Top-20 web sources by task domain.} Hosts are ranked by the share of retrieval iterations. Each host counts once per iteration; hosts can co-occur. All panels use a common 0--100\% scale and report retrieval counts.}
    \label{supp_fig:resource_heatmap}
    \label{supp_fig:resource_distribution}
\end{figure}

\clearpage
\begin{figure}[!ht]
    \ContinuedFloat
    \centering
    \captionsetup[subfigure]{font={scriptsize},labelfont={bf},skip=2pt,position=bottom,justification=centering,singlelinecheck=false}
    \begin{subfigure}[t]{0.49\linewidth}
        \centering
        \includegraphics[width=\linewidth]{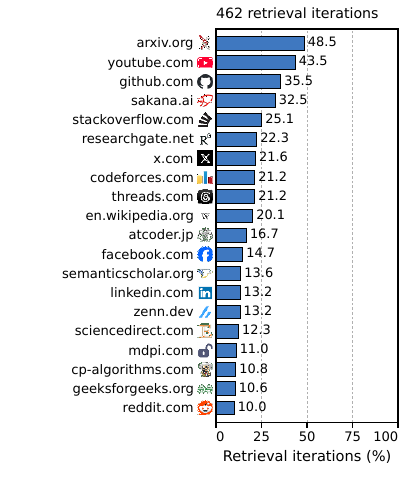}
        \caption{Algorithm engineering}
    \end{subfigure}\hfill
    \begin{subfigure}[t]{0.49\linewidth}
        \centering
        \includegraphics[width=\linewidth]{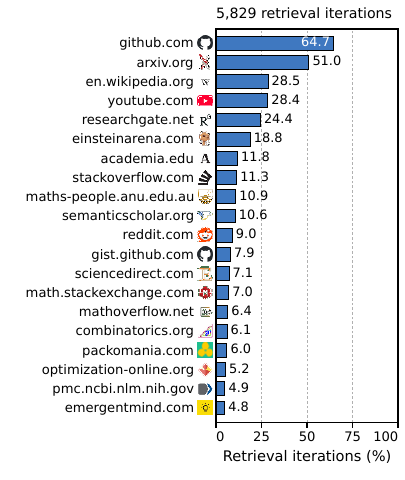}
        \caption{Mathematics}
    \end{subfigure}
    \vspace{-0.4em}
    \caption{\textbf{Top-20 web sources by task domain (continued).} Exposure rates pool all available runs within each domain, using the same counting rule and scale.}
\end{figure}

\begin{figure}[!ht]
    \centering
    \captionsetup[subfigure]{font={scriptsize},labelfont={bf},skip=2pt,position=bottom,justification=centering,singlelinecheck=false}
    \begin{subfigure}[t]{\linewidth}
        \centering
        \includegraphics[width=0.9\linewidth]{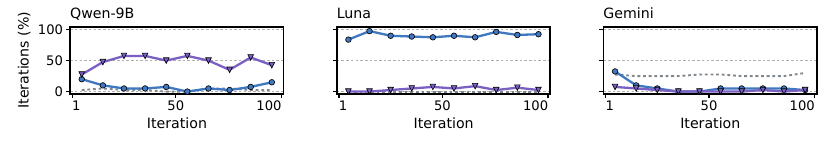}
        \caption{Quantum compilation}
    \end{subfigure}
    \par\smallskip
    \begin{subfigure}[t]{\linewidth}
        \centering
        \includegraphics[width=0.9\linewidth]{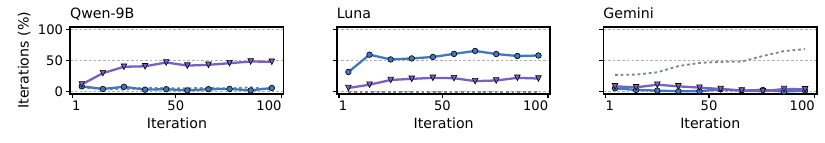}
        \caption{Astrodynamics}
    \end{subfigure}
    \par\smallskip
    \begin{subfigure}[t]{\linewidth}
        \centering
        \includegraphics[width=0.9\linewidth]{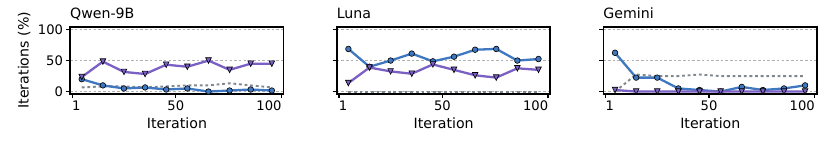}
        \caption{Scientific algorithms}
    \end{subfigure}
    \par\smallskip
    \begin{subfigure}[t]{\linewidth}
        \centering
        \includegraphics[width=0.9\linewidth]{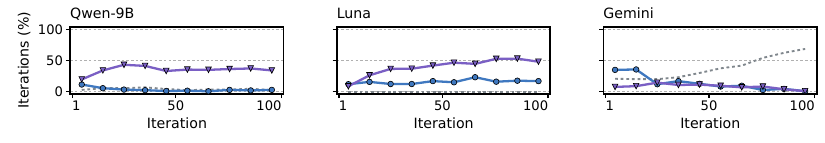}
        \caption{AI foundations}
    \end{subfigure}
    \par\smallskip
    \begin{subfigure}[t]{\linewidth}
        \centering
        \includegraphics[width=0.9\linewidth]{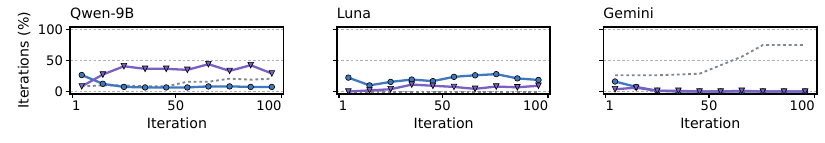}
        \caption{Algorithm engineering}
    \end{subfigure}
    \par\smallskip
    \begin{subfigure}[t]{\linewidth}
        \centering
        \includegraphics[width=0.9\linewidth]{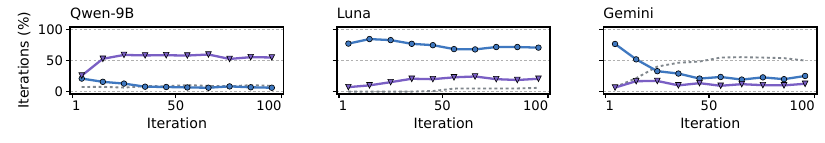}
        \caption{Mathematics}
    \end{subfigure}
    \includegraphics[width=0.9\linewidth]{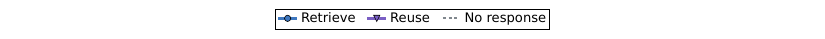}
    \vspace{-0.4em}
    \caption{\textbf{Aggregated search dynamics by task group and model.} Curves show retrieval (blue), reuse (purple), and missing gate responses (gray) in ten-iteration bins. Each panel pools all available runs. Percentages use all scheduled slots, including no-op and malformed decisions. All panels share the same axes.}
    \label{supp_fig:lookup_provenance}
\end{figure}

\clearpage
\paragraph{Models differ in how they balance retrieval and reuse.}
Figure~\ref{supp_fig:lookup_provenance} aggregates runs within each task group and model. In mathematics, Qwen's reuse share rises from 25.7\% in iterations 1--10 to 55.0\% in iterations 91--100, while Luna retrieves in 70.8\% of the final ten slots.

\begin{figure}[h]
    \centering
    \captionsetup[subfigure]{font={scriptsize},labelfont={bf},skip=2pt,position=bottom,justification=centering,singlelinecheck=false}
    \begin{subfigure}[t]{\linewidth}
        \centering
        \includegraphics[width=0.85\linewidth]{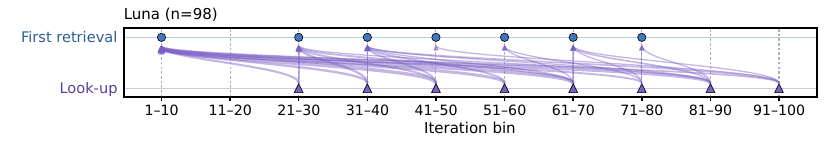}
        \caption{Quantum compilation}
    \end{subfigure}
    \par\smallskip
    \begin{subfigure}[t]{\linewidth}
        \centering
        \includegraphics[width=0.85\linewidth]{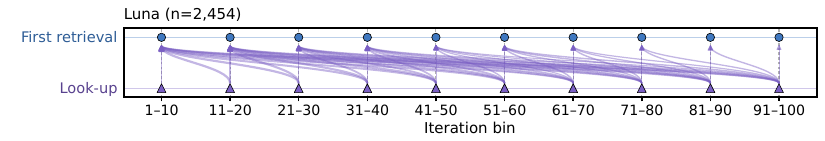}
        \caption{Astrodynamics}
    \end{subfigure}
    \par\smallskip
    \begin{subfigure}[t]{\linewidth}
        \centering
        \includegraphics[width=0.85\linewidth]{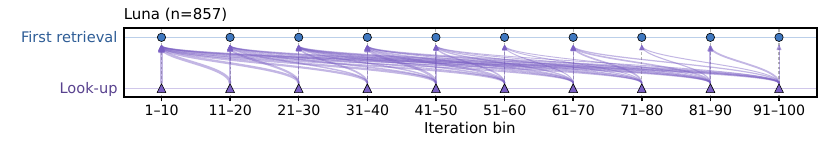}
        \caption{Scientific algorithms}
    \end{subfigure}
    \par\smallskip
    \begin{subfigure}[t]{\linewidth}
        \centering
        \includegraphics[width=0.85\linewidth]{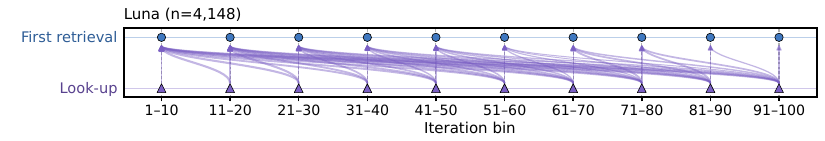}
        \caption{AI foundations}
    \end{subfigure}
    \par\smallskip
    \begin{subfigure}[t]{\linewidth}
        \centering
        \includegraphics[width=0.85\linewidth]{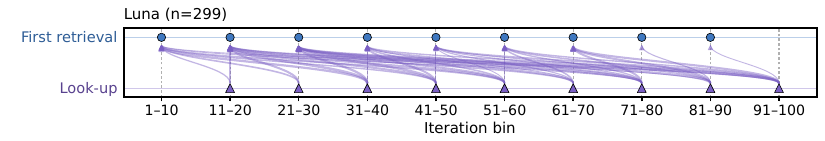}
        \caption{Algorithm engineering}
    \end{subfigure}
    \par\smallskip
    \begin{subfigure}[t]{\linewidth}
        \centering
        \includegraphics[width=0.85\linewidth]{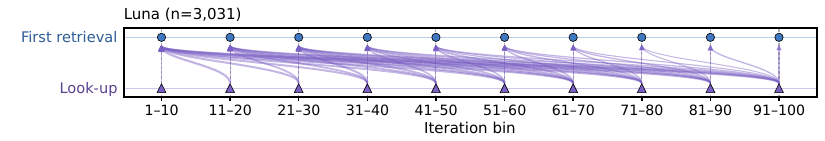}
        \caption{Mathematics}
    \end{subfigure}
    \includegraphics[width=0.85\linewidth]{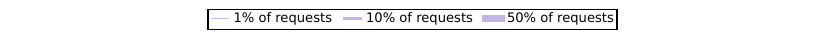}
    \vspace{-0.4em}
    \caption{\textbf{GPT-5.6-Luna's lookup provenance by task group.} Arrows point from each lookup (purple, bottom) back to the document's first recorded retrieval (blue, top), pooling all available runs. Both endpoints use ten-iteration bins; vertical arrows represent reuse within the same bin. Width encodes the within-panel share of resolved document ID--iteration pairs on a common square-root scale; $n$ is their total. All 10,887 resolved pairs are shown; 63 unresolved pairs are excluded.}
    \label{supp_fig:lookup_flow}
\end{figure}

\paragraph{Late lookups can reuse documents retrieved at the start.}
In quantum compilation, 12 of GPT-5.6-Luna's 21 resolved document requests during iterations 81--100 reuse documents first retrieved during iterations 1--10 (Figure~\ref{supp_fig:lookup_flow}).

\begin{figure}[!ht]
    \centering
    \captionsetup[subfigure]{font={scriptsize},labelfont={bf},skip=2pt,position=bottom,justification=centering,singlelinecheck=false}
    \begin{subfigure}[t]{0.49\linewidth}
        \centering
        \includegraphics[width=\linewidth]{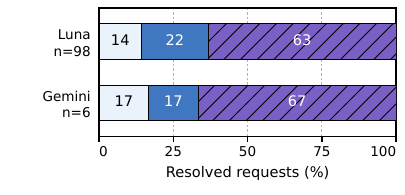}
        \caption{Quantum compilation}
    \end{subfigure}\hfill
    \begin{subfigure}[t]{0.49\linewidth}
        \centering
        \includegraphics[width=\linewidth]{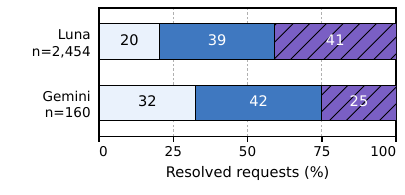}
        \caption{Astrodynamics}
    \end{subfigure}
    \par\medskip
    \begin{subfigure}[t]{0.49\linewidth}
        \centering
        \includegraphics[width=\linewidth]{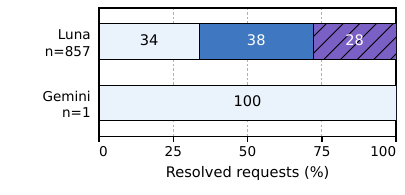}
        \caption{Scientific algorithms}
    \end{subfigure}\hfill
    \begin{subfigure}[t]{0.49\linewidth}
        \centering
        \includegraphics[width=\linewidth]{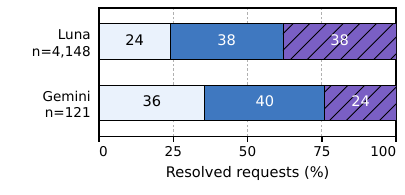}
        \caption{AI foundations}
    \end{subfigure}
    \par\medskip
    \begin{subfigure}[t]{0.49\linewidth}
        \centering
        \includegraphics[width=\linewidth]{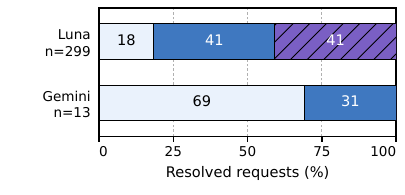}
        \caption{Algorithm engineering}
    \end{subfigure}\hfill
    \begin{subfigure}[t]{0.49\linewidth}
        \centering
        \includegraphics[width=\linewidth]{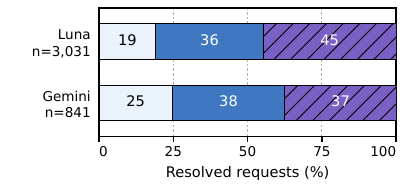}
        \caption{Mathematics}
    \end{subfigure}
    \includegraphics[width=\linewidth]{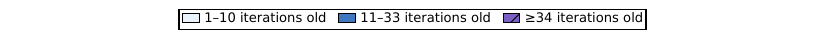}
    \vspace{-0.4em}
    \caption{\textbf{Source age at reuse by task group and model.} Bars aggregate resolved document requests from available GPT-5.6-Luna and Gemini-3.8-Flash checkpoints. Age is the reuse iteration minus the first retrieval iteration; $n$ counts document ID--iteration pairs. We exclude 63 unresolved Luna pairs. Qwen lacks saved checkpoint provenance and is omitted from this figure.}
    \label{supp_fig:lookup_source_age}
\end{figure}

\clearpage
\paragraph{Stored documents remain in use across many iterations.}
Luna reuses documents with a median age of 17 iterations in scientific algorithms and 44 in quantum compilation (Figure~\ref{supp_fig:lookup_source_age}). These distributions pool resolved requests within each task group and model.

\clearpage
\subsection{Failure Analysis of Retrieved Evidence Use}
\label{supp_sec:qwen_failure}
\label{supp_sec:model_failures}

We analyze 33{,}784 iterations from 362 \methodName runs across the 21 tasks with Qwen3.5-9B, GPT-5.6-Luna, and Gemini-3.8-Flash at $N\in\{1,8\}$. We consider each run's first 100 iterations and exclude iterations with recorded external service errors. Document exposure is verified from the generation prompt. Outcomes compare the selected child's recorded search score with its parent's.

\begin{figure}[!ht]
    \centering
    \captionsetup[subfigure]{font={scriptsize},labelfont={bf},skip=2pt,position=bottom,justification=centering,singlelinecheck=false}
    \begin{subfigure}[t]{0.49\linewidth}
        \centering
        \includegraphics[width=\linewidth]{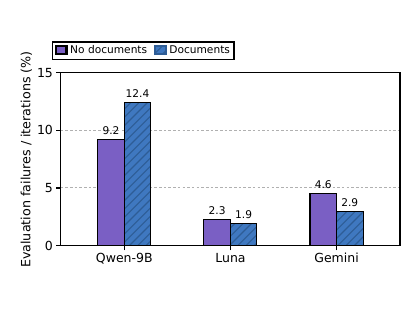}
        \caption{Evaluation failures at $N=1$}
    \end{subfigure}\hfill
    \begin{subfigure}[t]{0.49\linewidth}
        \centering
        \includegraphics[width=\linewidth]{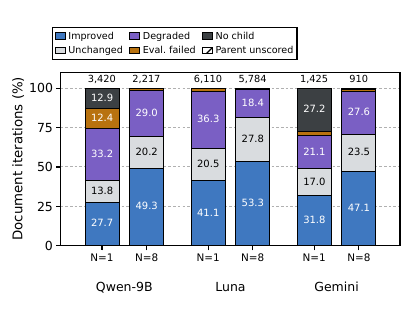}
        \caption{Outcomes with documents}
    \end{subfigure}
    \vspace{-0.3em}
    \caption{\textbf{Program revision outcomes with retrieved documents.} \textbf{(a)}~Evaluation failure rates with and without documents at $N=1$; both conditions use \methodName. \textbf{(b)}~Outcomes of 19{,}866 document iterations, with counts above the bars. All rates include unsuccessful generations in the denominator. The $N=8$ policy selects among eight candidates at document steps. Comparisons are observational because the gate selects document exposure.}
    \label{supp_fig:evidence_use}
    \label{supp_fig:qwen_failure_panels}
    \label{supp_fig:common_failures}
\end{figure}

\noindent \textbf{Qwen's failures with documents extend beyond evaluation errors.}
At $N=1$, Qwen fails evaluation in 12.4\% of document iterations versus 9.2\% without documents (Figure~\ref{supp_fig:evidence_use}(a)). Only 27.7\% improve on the parent, versus 41.1\% for Luna and 31.8\% for Gemini; 33.2\% produce a lower score and 12.9\% retain no child. Evaluation errors alone therefore miss revisions that execute but fail to improve.

\noindent \textbf{Qwen improves more often when selecting from parallel candidates.}
At $N=8$, Qwen's document iterations improve on the parent in 49.3\% of cases and fail evaluation in 1.1\%. Luna and Gemini improve in 53.3\% and 47.1\%, respectively. Since \RETRIEVE and \LOOKUP generate eight candidates per attempt while \NOOP generates one, these outcomes reflect both generation and selection.

\begin{table}[!ht]
    \centering
    \scriptsize
    \setlength{\tabcolsep}{4pt}
    \renewcommand{\arraystretch}{1.13}
    \caption{\textbf{Examples of unsuccessful document use by Qwen at $N=1$.} Supplied documents, parent and child code, and evaluator records are inspected together. Scores are recorded search scores ($\uparrow$). These cases illustrate mechanisms rather than their population frequencies.}
    \label{supp_tab:evidence_use_cases}
    \begin{tabular}{@{}p{0.14\linewidth}p{0.27\linewidth}p{0.44\linewidth}p{0.09\linewidth}@{}}
        \toprule
        Task & Supplied evidence & Observed revision & Outcome \\
        \midrule
        Sums/Diffs & A construction using the Chinese Remainder Theorem. & Adds random local search without implementing the supplied modular construction. & $1.05979$ $\rightarrow$ $1.05979$ \\
        Circle Packing (26) & Joint optimization of centers and radii using SLSQP. & Adopts SLSQP but mixes grouped and interleaved parameter layouts; a returned circle extends outside the square. & Invalid \\
        Denoising & The Anscombe transform $2\sqrt{x+3/8}$ and its inverse. & Omits the factor of two in the forward transform, while the inverse still divides by two before squaring. & $0.28849$ $\rightarrow$ $0.28607$ \\
        \bottomrule
    \end{tabular}
\end{table}

\noindent \textbf{Retrieved methods are unused or incorrectly implemented in 26\% of sampled Qwen revisions.}
Codex inspected 50 revisions sampled uniformly from 2{,}980 eligible Qwen revisions with documents and retained code at $N=1$. Methods were unused in 3 (6\%) and incorrectly implemented in 10 (20\%); 22 incorporated or retained a method, 7 lacked actionable evidence, and 8 were uncertain. These model-assessed sample proportions exclude external errors and count only specific new discrepancies with supplied methods. Table~\ref{supp_tab:evidence_use_cases} provides separate examples; Denoising's score decrease cannot be attributed to its transform error alone.

\section{Limitations}
\label{supp_sec:limitations}

\paragraph{Generalization Across Backbone Models.}
Our experiments cover GPT-5.6-Luna, Gemini-3.8-Flash, and Qwen3.5-9B. The benefits of \methodName vary across these models: Qwen3.5-9B has lower overall NDG with \methodName than with OpenEvolve alone at both $N=1$ and $N=8$. Our analysis also identifies cases in which this model ignores retrieved methods or implements them incorrectly. Evaluating more model families and sizes, and examining whether training for evidence selection and implementation reduces these failures, would help establish when a backbone can benefit from bi-level co-evolution.

\paragraph{Limited Gains on Algorithm Engineering Tasks.}
\methodName does not consistently improve performance on algorithm engineering tasks such as AHC039 and AHC058. Figure~\ref{main_fig:ndg_panel_b} shows lower group-average NDG than OpenEvolve in four of the six model/budget settings, with an average decline of 1.6\% across all six. These losses occur with both GPT-5.6-Luna and Qwen3.5-9B at $N=1$ and $N=8$. Further work is needed to determine how evidence selection, adaptation to task-specific heuristics, and the allocation of search effort contribute to these failures, and to improve the framework on these tasks.

\paragraph{Toward Multimodal and Multi-Turn Discovery.}

Scientific discovery often requires interpreting visual observations from experiments~\citep{sun2025scienceboard,ma2026orion}, motivating the use of vision-language models~\citep{lee2024collavo,lee2024moai,lee2024meteor,lee2024phantom,lee2024trol,lee2025genrecal,lee2025vlsi,lee2026unified,kang2026agent,lee2026masking,lee2026recursive,cho2026spatialclaw,yu2026hide,kim2026and,lee2026zone} as mutation operators in our solution optimization loop. In our current framework, the inner loop evolves search queries, while the outer loop generates each candidate solution in a single turn, conditioned on the parent program, evolutionary history, and any retrieved evidence. Extending this bi-level co-evolution paradigm to multi-turn interactions~\citep{lee2024dialogcc,lee2024stark,lee2024thanos,lee2024large,lee2025multiverse,lee2025refinebench} could enable the model to revisit explored lineages and iteratively refine both queries and solutions within a shared conversational context.

\section{Discussions}
\label{supp_sec:further_discussion}

The benefits of oracle documents vary across models and tasks. On Swap Reduction, the same documents raise NDG by 6.1\% for Qwen3.5-9B and 41.5 for GPT-5.6-Luna, while both models lose performance on Rosetta (Figure~\ref{main_fig:delta_ndg_per_task_c}). Qwen's positive average oracle gain also shows that its losses with \methodName do not imply a general inability to use documents. Those losses may reflect difficulties in identifying knowledge gaps, retrieving relevant evidence, or implementing it. The unused methods and incorrect implementations in Appendix~\ref{supp_sec:model_failures} provide examples of the last difficulty. These observations motivate training and evaluating evidence selection and implementation as separate skills, while accounting for baseline performance and search budget.

\section{Broader Impact}
\label{supp_sec:broader_impact}

\methodName may make computational scientific discovery more efficient by helping researchers find and apply relevant knowledge from the web. However, unreliable sources can lead to incorrect conclusions, while reusing published solutions can result in overstated claims of novelty. To support auditing, we record retrieval and optimization trajectories. We also recommend explicit source attribution, independent validation of generated programs, and human oversight of downstream applications.

\definecolor{tjInk}{HTML}{273E52}
\definecolor{tjRule}{HTML}{CBD5DF}
\definecolor{tjPaper}{HTML}{F5F7FA}
\definecolor{tjGate}{HTML}{B27A39}
\definecolor{tjLook}{HTML}{75658A}
\definecolor{tjNoop}{HTML}{758394}
\definecolor{tjQuery}{HTML}{3D6397}
\definecolor{tjDoc}{HTML}{287E83}
\definecolor{tjEdit}{HTML}{8C627D}
\definecolor{tjResult}{HTML}{314A60}
\definecolor{tjBehav}{HTML}{735A86}
\definecolor{tjAdd}{HTML}{1B7850}
\definecolor{tjDel}{HTML}{B1404A}

\newtcolorbox{trajstep}[2]{enhanced, breakable, sharp corners,
    colback=white, colframe=tjRule, colbacktitle=tjInk, coltitle=white,
    borderline west={1.8pt}{0pt}{#1},
    fonttitle=\sffamily\bfseries\footnotesize, fontupper=\footnotesize,
    boxrule=0.45pt, boxsep=0pt, left=6pt, right=6pt, top=5pt, bottom=5pt,
    lefttitle=6pt, righttitle=6pt, toptitle=4pt, bottomtitle=4pt,
    before skip=9pt, after skip=3pt, before upper=\raggedright, title={#2},
    title after break={#2\hfill{\mdseries\scriptsize(continued)}}}

\newenvironment{tjcomp}[3]{%
    \def\tjc{#1}%
    \tcolorbox[enhanced, sharp corners, colback=#1!4!white, frame hidden,
        borderline west={1.2pt}{0pt}{#1}, boxrule=0pt, boxsep=0pt,
        left=5pt, right=5pt, top=4pt, bottom=4pt,
        before skip=4pt, after skip=0pt, fontupper=\footnotesize, before upper=\raggedright]%
    \parindent=0pt\noindent\textcolor{#1!85!black}{\sffamily\bfseries\scriptsize\MakeUppercase{#2}}%
    \par\vspace{2pt}\leftskip=1.1em}{\par\endtcolorbox}

\newcommand{\tjfname}[1]{{\setlength{\fboxsep}{0.9pt}\colorbox{\tjc!9!white}{%
    \color{\tjc!80!black}\ttfamily\scriptsize #1}}}
\newcommand{\tjdname}[1]{{\color{tjInk!75!white}\itshape\scriptsize #1}}

\newcommand{\tjfield}[1]{\par\vspace{0.6pt}\noindent\hspace*{-1.1em}\tjfname{#1}\hspace{0.45em}\ignorespaces}
\newcommand{\tjdfield}[1]{\par\vspace{0.6pt}\noindent\hspace*{-1.1em}\tjdname{#1}\hspace{0.45em}\ignorespaces}

\newcommand{\tjalso}[2]{\hfil\penalty0\hspace{0.9em}\mbox{\tjfname{#1}\hspace{0.45em}#2}}

\newcommand{\tjhead}[3]{%
    \par\addvspace{5pt}\def\tjc{#1}{\leftskip=0pt\parindent=0pt\noindent
    \setlength{\fboxsep}{4pt}\colorbox{#1!10!white}{\parbox{\dimexpr\linewidth-8pt\relax}{%
    \textcolor{#1!85!black}{\sffamily\bfseries\scriptsize\MakeUppercase{#2}}}}\par}\nobreak}
\newenvironment{tjitem}{%
    \par\leftskip=0pt\tcolorbox[enhanced, sharp corners, colback=tjPaper, frame hidden,
        borderline west={0.8pt}{0pt}{\tjc!55!white}, boxrule=0pt, boxsep=0pt,
        left=5pt, right=5pt, top=3pt, bottom=3pt, before skip=2pt, after skip=0pt,
        fontupper=\footnotesize, before upper=\raggedright]\parindent=0pt\leftskip=1.1em}{\par\endtcolorbox}

\newcommand{\tjintent}[1]{{\color{tjQuery!70!black}\itshape #1}}

\newtcolorbox{tjdiffbox}[1]{enhanced, sharp corners, colback=white, colframe=tjRule, boxrule=0.4pt,
    left=0pt, right=0pt, top=0pt, bottom=0pt, boxsep=0pt, before skip=1.6pt, after skip=0pt,
    colbacktitle=tjEdit!8!white, coltitle=tjInk, fonttitle=\ttfamily\scriptsize,
    toptitle=0.5mm, bottomtitle=0.5mm, lefttitle=1.2mm, titlerule=0.35pt,
    title={#1}}
\newenvironment{tjdiff}[1]{\par\leftskip=0pt\begin{tjdiffbox}{#1}}{\end{tjdiffbox}}
\newcommand{\tjline}[4]{%
    \par\nointerlineskip\noindent{\setlength{\fboxsep}{0pt}\colorbox{#3}{\parbox{\linewidth}{%
    \leftskip=2.1em\rightskip=1mm plus 1fil\parindent=0pt\noindent
    \hspace*{-2.1em}\makebox[2.1em][c]{\color{#1}\ttfamily\bfseries #2}%
    {\ttfamily\scriptsize\strut #4\strut}}}}\par}

\newcommand{\tjind}[1]{\hspace*{\dimexpr#1\fontcharwd\font`\x\relax}}
\newcommand{\tjadd}[1]{\tjline{tjAdd}{+}{tjAdd!11!white}{#1}}
\newcommand{\tjdel}[1]{\tjline{tjDel}{$-$}{tjDel!10!white}{#1}}

\newcommand{\tjchip}[2]{{\setlength{\fboxsep}{1.1pt}\raisebox{-0.22ex}{%
    \colorbox{#1}{\textcolor{white}{\tiny\bfseries #2}}}}}

\newcommand{\tjurl}[1]{\textcolor{tjDoc!85!black}{\texttt{#1}}}

\newtcolorbox{tjbehavior}{enhanced, sharp corners, colback=tjBehav!4!white, frame hidden,
    borderline west={1.2pt}{0pt}{tjBehav}, boxrule=0pt, boxsep=0pt,
    left=5pt, right=5pt, top=4pt, bottom=4pt,
    before skip=5pt, after skip=0pt, fontupper=\footnotesize}

\newcommand{\tjbeh}[1]{\tcbox[on line, enhanced, sharp corners, boxrule=0.3pt, boxsep=1pt,
    colback=tjBehav!8!white, colframe=tjBehav!35!white, colupper=tjBehav!80!black,
    left=2pt, right=2pt, top=1pt, bottom=1pt, fontupper=\footnotesize]{#1}}
\newcommand{\tjpat}[1]{\tcbox[on line, enhanced, sharp corners, boxrule=0.3pt, boxsep=1pt,
    colback=tjPaper, colframe=tjRule, colupper=tjInk!85!white,
    left=2pt, right=2pt, top=1pt, bottom=1pt, fontupper=\footnotesize]{#1}}

\newtcolorbox{tjreading}{enhanced, sharp corners, colback=white, colframe=tjInk!55!white,
    colbacktitle=tjInk, coltitle=white, title={Reading the records},
    fonttitle=\sffamily\bfseries\small, lefttitle=6pt, righttitle=6pt,
    toptitle=4pt, bottomtitle=4pt,
    boxrule=0.6pt, boxsep=0pt, left=6pt, right=6pt, top=6pt, bottom=6pt,
    before skip=8pt, after skip=8pt, fontupper=\footnotesize, before upper=\raggedright}

\newcommand{\tjkey}[3]{%
  \begin{tcolorbox}[enhanced, sharp corners, colback=#1!4!white, colframe=#1!30!white,
      borderline north={1.2pt}{0pt}{#1}, boxrule=0.35pt, boxsep=0pt,
      left=5pt, right=5pt, top=4pt, bottom=4pt,
      fontupper=\footnotesize, before upper=\raggedright]
    {\color{#1!85!black}\sffamily\bfseries\scriptsize\MakeUppercase{#2}}\par\vspace{2pt}#3%
  \end{tcolorbox}}

\newcommand{\tjdecision}[2]{%
  \tcbox[on line, enhanced, sharp corners, boxrule=0.35pt, boxsep=1pt,
      colback=#1!12!white, colframe=#1!60!white, colupper=#1!75!black,
      left=3pt, right=3pt, top=1pt, bottom=1pt,
      fontupper=\scriptsize\sffamily\bfseries]{#2}}

\begingroup
\raggedbottom
\colorlet{RetrieveChocolate}{tjGate}
\colorlet{LookupIndigo}{tjLook}
\colorlet{NoopGray}{tjNoop}
\section{Trajectories of the Bi-Level Loop}
\label{supp_sec:trajectories}

We trace seven runs to show how knowledge gaps, queries, retrieved evidence, and code changes relate to measured outcomes (Table~\ref{supp_tab:traj_index}). Each case presents selected iterations in chronological order and illustrates a behavior flag or its limitations. Six runs use GPT-5.6-Luna with a candidate budget of $N=8$; Rosetta uses GPT-5.6-Sol with one candidate. Appendix~\ref{supp_sec:behavior_examples} defines the flags and reports their frequency.

\begin{table}[h]
\centering\scriptsize
\caption{\textbf{Seven worked behavior cases.} Each case links recorded queries and sources to program changes and evaluation results. The cases illustrate the flags in Appendix~\ref{supp_sec:behavior_examples}; a flag alone does not establish successful use of the retrieved evidence.}
\label{supp_tab:traj_index}
\begin{tabular}{@{}>{\raggedright\arraybackslash}p{0.24\linewidth}@{\hspace{6pt}}>{\raggedright\arraybackslash}p{0.18\linewidth}@{\hspace{6pt}}>{\raggedright\arraybackslash}p{\dimexpr0.53\linewidth-18pt\relax}@{\hspace{6pt}}>{\centering\arraybackslash}p{0.05\linewidth}@{}}
\toprule
\textbf{Behavior} & \textbf{Run} & \textbf{Main finding} & \textbf{Sec.} \\
\midrule
Method transfer & Swap Reduction & Retrieved depth weighting remains in later best programs despite repeated search misses. & \ref{supp_sec:traj_swap} \\
Cross-domain transfer & Erd\H{o}s min.\ overlap & A radar codeword seeds later improvements in a number-theory task. & \ref{supp_sec:traj_erdos} \\
Live documentation lookup & Voyager~2 & The added API keywords repeat defaults already given in the prompt. & \ref{supp_sec:traj_voyager} \\
Target-informed reconstruction & CP ($n$=26) & A generated solver exceeds the value named in the first query. & \ref{supp_sec:traj_cp26} \\
Public artifact reuse & Erd\H{o}s, second run & A program that downloads a published witness improves the run's best. & \ref{supp_sec:traj_artifact} \\
Evidence-inspired over-reach & Galileo & A promising retrieval precedes a worse child; the run's best is preserved. & \ref{supp_sec:traj_galileo} \\
Public artifact reuse; method transfer & Rosetta & Published seeds and stored implementation details support a lower-cost tour. & \ref{supp_sec:traj_rosetta} \\
\bottomrule
\end{tabular}
\end{table}

\begin{tjreading}
\begin{tcbraster}[raster columns=2, raster equal height=rows,
    raster column skip=5pt, raster row skip=4pt, raster before skip=0pt, raster after skip=5pt]
\tjkey{tjGate}{gate}{Recorded decision, knowledge-gap assessment, and rationale.}
\tjkey{tjQuery}{query}{Query and intent (\texttt{query\_type}) for each search round.}
\tjkey{tjDoc}{sources}{Retained documents and predicted child scores (\texttt{estimated\_child\_score}); stored sources on look-up.}
\tjkey{tjDoc}{quote}{An excerpt from a retrieved page's \texttt{raw\_content}.}
\tjkey{tjEdit}{edit}{Selected additions and deletions in the parent-to-child code diff.}
\tjkey{tjResult}{result}{Measured parent and child scores, plus the change in the run's best.}
\end{tcbraster}
{\color{tjRule}\hrule height 0.4pt}\vspace{5pt}
\noindent\begin{minipage}[t]{0.485\linewidth}
\raggedright
{\color{tjInk}\sffamily\bfseries\scriptsize Record notation}\par\vspace{2pt}
Monospace labels are recorded fields; italic labels are derived. Ellipses mark omissions; square brackets mark editorial substitutions. Em-dashes are normalized.
\end{minipage}\hfill%
\begin{minipage}[t]{0.485\linewidth}
\raggedright
{\color{tjBehav}\sffamily\bfseries\scriptsize Our annotations}\par\vspace{2pt}
Case summaries and behavior tags are our interpretation. In the records, violet tags identify audit behaviors; gray tags identify other patterns.
\end{minipage}
\par\vspace{6pt}
{\color{tjInk}\sffamily\bfseries\scriptsize Gate decision}\quad
\tjdecision{RetrieveChocolate}{retrieve}\hspace{0.35em}%
\tjdecision{LookupIndigo}{look-up}\hspace{0.35em}%
\tjdecision{NoopGray}{no-op}\hfill{\color{tjInk!75!white}\scriptsize Record border and gate badge}
\end{tjreading}

\paragraph{Score conventions.}
The result panels report recorded search-time scores, for which higher is better. Parent $\rightarrow$ child compares the selected candidate with the parent used to generate it; that parent need not be the previous iteration's child. Best-so-far compares the run's highest score before and after the iteration. Predicted document scores are not evaluation results. Native objectives have their own direction: lower is better for SWAP counts, $C_5$, and $\Delta v$; higher is better for the sum of circle radii. The Erd\H{o}s score is $1/C_5$.

\Needspace{15\baselineskip}
\subsection{Swap Reduction: depth weighting persists despite repeated search misses}
\label{supp_sec:traj_swap}

\noindent Exponential depth weighting from dSABRE remains in every later best program, providing evidence of \emph{method transfer}. The final router adds $14{,}835$ SWAPs on Q20, compared with SimpleTES's $15{,}186$ under the same evaluator. This advantage is specific to Q20 (Appendix~\ref{supp_sec:best_programs}). Iteration~5 introduces the weighting; later edits tune it. Increasingly specific queries show \emph{identifier fixation}: 79 of 94 retrievals name Qiskit PR~\#14912 without obtaining its diff. Iteration~77 sets a new best after another retrieval with no promising document, an instance of \emph{search miss with own progress}.

{\footnotesize\color{tjInk!85!white}\noindent Iterations 5, 7, 13, 66, and 77.\par}

\begin{trajstep}{tjGate}{Iteration 5\;\textperiodcentered\;retrieve\;\textperiodcentered\;best-so-far 6,957 $\rightarrow$ 11,326}
\begin{tjcomp}{tjGate}{gate}{gate\_decision.json}
\tjfield{decision}\tjdecision{RetrieveChocolate}{retrieve}
\tjfield{knowledge\_state\_analysis}The strongest measured result is combined\_score 6957.0 from the relative distance-delta scoring combined with component-aware initial layout. Attempt 3's more elaborate layout placement reduced the score to 6565.8, so that change should not be retained. Existing documents establish the SABRE/LightSABRE principles, especially normalized front/lookahead costs, relative candidate scoring, decay, depth, and critical-path heuristics. However, they do not\,\ldots
\tjfield{reasoning}Additional evidence would help because the current search documents are largely conceptual or generic Qiskit references and prior experiments have plateaued or regressed. Fresh exploration should target repository-\allowbreak{}specific routing context APIs and concrete implementations of depth-aware, critical-path, or other candidate-\allowbreak{}ranking heuristics beyond the already tried\,\ldots
\end{tjcomp}
\tjhead{tjQuery}{query}{query\_optimization.json}
\begin{tjitem}
\tjfield{round}1
\tjfield{query}{\ttfamily "SABRE" adaptive lookahead decay swap scoring weighted interaction\allowbreak{} frequency initial layout Rust GitHub}
\tjfield{query\_type}\tjintent{Find a practical adaptive SABRE heuristic or interaction-\allowbreak{}frequency-\allowbreak{}based layout method that improves on fixed front-\allowbreak{}layer/\allowbreak{}lookahead weights and greedy component placement without requiring unavailable depth or\,\ldots}
\end{tjitem}
\begin{tjitem}
\tjfield{round}2
\tjfield{query}{\ttfamily "SwapSelect\allowbreak{}ionContext"\allowbreak{} Rust "front\_\allowbreak{}layer" executable gate can\_\allowbreak{}apply routing API}
\tjfield{query\_type}\tjintent{Find repository-\allowbreak{}specific APIs or source code for detecting immediately executable front-layer gates and applying or scoring them, enabling a controlled addition of an executability reward or adaptive lookahead without\,\ldots}
\end{tjitem}
\begin{tjitem}
\tjfield{round}3
\tjfield{query}{\ttfamily "precompute\allowbreak{}d\_\allowbreak{}extended\_\allowbreak{}set\_\allowbreak{}logical\_\allowbreak{}pairs" dependency depth BFS layer weighted lookahead Rust SABRE}
\tjfield{query\_type}\tjintent{Find a repository-\allowbreak{}compatible method for assigning BFS or dependency depths to extended-set gates so the current flat lookahead score can be replaced by a controlled exponential depth-weighted variant.}
\end{tjitem}
\tjhead{tjDoc}{sources}{query\_optimization.json}
\begin{tjitem}
\tjfield{title}Structured Scaling of AI Discovery Across Diverse Scientific ...
\tjfield{url}\raisebox{-0.22ex}{\includegraphics[height=1.3ex]{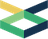}}\,\href{https://www.researchsquare.com/article/rs-10631214/v1.pdf?c=1787913294000}{\tjurl{researchsqu\allowbreak{}are.\allowbreak{}com/\allowbreak{}.\allowbreak{}.\allowbreak{}.\allowbreak{}df?\allowbreak{}c=\allowbreak{}17879132940\allowbreak{}00}}\tjalso{estimated\_child\_score}{6,968}
\end{tjitem}
\begin{tjitem}
\tjfield{title}dSABRE: A SABRE-Style Router for Multi-Core Distributed Quantum Computers
\tjfield{url}\raisebox{-0.22ex}{\includegraphics[height=1.3ex]{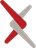}}\,\href{https://arxiv.org/html/2605.21960v1}{\tjurl{arxiv.\allowbreak{}org/\allowbreak{}html/\allowbreak{}2605.\allowbreak{}21960v1}}\tjalso{estimated\_child\_score}{6,963}
\end{tjitem}
\begin{tjitem}
\tjfield{title}dSABRE: A SABRE-Style Router for Multi-Core Distributed Quantum Computers
\tjfield{url}\raisebox{-0.22ex}{\includegraphics[height=1.3ex]{figures/favicons/arxiv_org.png}}\,\href{https://arxiv.org/html/2605.21960}{\tjurl{arxiv.\allowbreak{}org/\allowbreak{}html/\allowbreak{}2605.\allowbreak{}21960}}\tjalso{estimated\_child\_score}{6,963}
\end{tjitem}
\begin{tjcomp}{tjDoc}{quote}{search\_result.jsonl}
\tjfield{url}\raisebox{-0.22ex}{\includegraphics[height=1.3ex]{figures/favicons/arxiv_org.png}}\,\href{https://arxiv.org/html/2605.21960v1}{\tjurl{arxiv.\allowbreak{}org/\allowbreak{}html/\allowbreak{}2605.\allowbreak{}21960v1}}
\tjfield{raw\_content}``...a-core graph. The tilde marks the departure from SABRE's flat extended-set weighting: rather than treating all lookahead gates equally, the exponential factor $\gamma$dep(gi) (g ) down-weights gates far from the front so that near-term successors dominate the lookahead signal; a deeper gate is more likely to be displaced by\,\ldots''
\end{tjcomp}
\tjhead{tjEdit}{edit}{evolution\_trace.json}
\begin{tjdiff}{solution: parent $\rightarrow$ child\hfill\textcolor{tjAdd}{+53}\ \textcolor{tjDel}{$-$1}\ \ 8 of 54 changed lines shown\hspace{1.2mm}}
\tjadd{\tjind{4}pairs:\allowbreak{} Vec<(\allowbreak{}usize,\allowbreak{} usize)\allowbreak{}>,\allowbreak{}}
\tjadd{\tjind{8}self.\allowbreak{}pairs.\allowbreak{}push(\allowbreak{}(\allowbreak{}a,\allowbreak{} b)\allowbreak{})\allowbreak{};\allowbreak{}}
\tjadd{\tjind{4}fn weighted\_\allowbreak{}score\_\allowbreak{}delta(\allowbreak{}}
\tjadd{\tjind{8}swap:\allowbreak{} (\allowbreak{}usize,\allowbreak{} usize)\allowbreak{},\allowbreak{}}
\tjadd{\tjind{8}if self.\allowbreak{}pairs.\allowbreak{}is\_\allowbreak{}empty(\allowbreak{})\allowbreak{} \{\allowbreak{}}
\tjadd{\tjind{8}let mut weighted =\allowbreak{} 0.\allowbreak{}0;\allowbreak{}}
\tjadd{\tjind{8}let mut weights =\allowbreak{} 0.\allowbreak{}0;\allowbreak{}}
\tjadd{\tjind{8}for (\allowbreak{}index,\allowbreak{} \&\allowbreak{}(\allowbreak{}x,\allowbreak{} y)\allowbreak{})\allowbreak{} in self.\allowbreak{}pairs.\allowbreak{}iter(\allowbreak{})\allowbreak{}.\allowbreak{}enumerate(\allowbreak{})\allowbreak{} \{\allowbreak{}}
\end{tjdiff}
\begin{tjcomp}{tjResult}{result}{evolution\_trace.json}
\tjfield{combined\_score}6,957 $\rightarrow$ 11,326\ \tjdname{(parent $\rightarrow$ child)}
\tjdfield{Q20 SWAPs}19,953 $\rightarrow$ 18,030
\tjdfield{best-so-far}6,957 $\rightarrow$ 11,326
\end{tjcomp}
\begin{tjbehavior}\textcolor{tjBehav}{\scshape\bfseries behavior}\hspace{0.62em}\tjbeh{method transfer}\end{tjbehavior}
\end{trajstep}

\begin{trajstep}{tjGate}{Iteration 7\;\textperiodcentered\;retrieve\;\textperiodcentered\;best-so-far 11,326 $\rightarrow$ 15,174}
\begin{tjcomp}{tjGate}{gate}{gate\_decision.json}
\tjfield{decision}\tjdecision{RetrieveChocolate}{retrieve}
\tjfield{knowledge\_state\_analysis}The prior attempts established a substantial measured improvement from 6498 to 11325.6, but the three listed attempts have identical routing metrics, so they do not isolate which modifications caused the gain or whether the current heuristic is near a plateau. Existing documents support geometric lookahead decay and LightSABRE concepts such as depth, critical-path scoring, broader candidate neighborhoods, and immediate executability rewards. However, they\,\ldots
\tjfield{reasoning}Additional evidence would help because the next promising improvements require implementation details absent from the stored documents. Reusing the existing high-level papers would largely repeat explored knowledge; a fresh search should target the exact Rust routing implementation and APIs, as well as validated LightSABRE candidate and scoring behavior.
\end{tjcomp}
\tjhead{tjQuery}{query}{query\_optimization.json}
\begin{tjitem}
\tjfield{round}1
\tjfield{query}{\ttfamily site:\allowbreak{}github.\allowbreak{}com/\allowbreak{}Qiskit/\allowbreak{}qiskit "SwapSelect\allowbreak{}ionContext"\allowbreak{} "critical\_\allowbreak{}path" OR "delta\_\allowbreak{}depth" OR "can\_\allowbreak{}apply" Rust}
\tjfield{query\_type}\tjintent{Find the current Rust SwapSelectio\allowbreak{}nContext APIs and implementation for LightSABRE depth, critical-path, immediate-\allowbreak{}executabilit\allowbreak{}y,\allowbreak{} and broader candidate-swap scoring.}
\end{tjitem}
\begin{tjitem}
\tjfield{round}2
\tjfield{query}{\ttfamily site:\allowbreak{}github.\allowbreak{}com/\allowbreak{}Qiskit/\allowbreak{}qiskit "SwapSelect\allowbreak{}ionContext"\allowbreak{} "precompute\allowbreak{}d\_\allowbreak{}extended\_\allowbreak{}set\_\allowbreak{}logical\_\allowbreak{}pairs" "remaining\_\allowbreak{}predecessor\allowbreak{}\_\allowbreak{}counts" Rust sabre}
\tjfield{query\_type}\tjintent{Find the current Qiskit Rust source implementation and concrete SwapSelectio\allowbreak{}nContext APIs for BFS/\allowbreak{}dependency-\allowbreak{}depth extended-set construction, immediate gate executability checks, and safe broader candidate generation.}
\end{tjitem}
\begin{tjitem}
\tjfield{round}3
\tjfield{query}{\ttfamily site:\allowbreak{}github.\allowbreak{}com/\allowbreak{}Qiskit/\allowbreak{}qiskit "SwapSelect\allowbreak{}ionContext"\allowbreak{} "precompute\allowbreak{}d\_\allowbreak{}extended\_\allowbreak{}set\_\allowbreak{}logical\_\allowbreak{}pairs" "remaining\_\allowbreak{}predecessor\allowbreak{}\_\allowbreak{}counts" Rust sabre}
\tjfield{query\_type}\tjintent{Find the current Qiskit Rust source implementation and concrete SwapSelectio\allowbreak{}nContext APIs for BFS/\allowbreak{}dependency-\allowbreak{}depth extended-set construction, immediate gate executability checks, and safe broader candidate generation.}
\end{tjitem}
\tjhead{tjDoc}{sources}{query\_optimization.json}
\begin{tjitem}
\tjfield{title}Port the rest of `SabreSwap` to Rust $\cdot$ Issue \#12280 $\cdot$ Qiskit/qiskit $\cdot$ GitHub
\tjfield{url}\raisebox{-0.22ex}{\includegraphics[height=1.3ex]{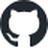}}\,\href{https://github.com/Qiskit/qiskit/issues/12280}{\tjurl{github.\allowbreak{}com/\allowbreak{}Qiskit/\allowbreak{}qiskit/\allowbreak{}issues/\allowbreak{}12280}}\tjalso{estimated\_child\_score}{11,326}
\end{tjitem}
\begin{tjitem}
\tjfield{title}qiskit/\allowbreak{}rust-\allowbreak{}toolchain.\allowbreak{}toml at main $\cdot$ Qiskit/qiskit $\cdot$ GitHub
\tjfield{url}\raisebox{-0.22ex}{\includegraphics[height=1.3ex]{figures/favicons/github_com.png}}\,\href{https://github.com/Qiskit/qiskit/blob/main/rust-toolchain.toml}{\tjurl{github.\allowbreak{}com/\allowbreak{}.\allowbreak{}.\allowbreak{}.\allowbreak{}b/\allowbreak{}main/\allowbreak{}rust-\allowbreak{}toolchain.\allowbreak{}toml}}\tjalso{estimated\_child\_score}{11,326}
\end{tjitem}
\begin{tjitem}
\tjfield{title}Port the rest of `SabreLayout` to Rust $\cdot$ Issue \#12279 $\cdot$ Qiskit/qiskit $\cdot$ GitHub
\tjfield{url}\raisebox{-0.22ex}{\includegraphics[height=1.3ex]{figures/favicons/github_com.png}}\,\href{https://github.com/Qiskit/qiskit/issues/12279}{\tjurl{github.\allowbreak{}com/\allowbreak{}Qiskit/\allowbreak{}qiskit/\allowbreak{}issues/\allowbreak{}12279}}\tjalso{estimated\_child\_score}{11,326}
\end{tjitem}
\begin{tjcomp}{tjDoc}{quote}{search\_result.jsonl}
\tjfield{url}\raisebox{-0.22ex}{\includegraphics[height=1.3ex]{figures/favicons/github_com.png}}\,\href{https://github.com/Qiskit/qiskit/issues/12280}{\tjurl{github.\allowbreak{}com/\allowbreak{}Qiskit/\allowbreak{}qiskit/\allowbreak{}issues/\allowbreak{}12280}}
\tjfield{raw\_content}``...! There was an error while loading. Please reload this page. There was an error while loading. Please reload this page. \# Port the rest of `SabreSwap` to Rust \#12280 `SabreSwap` `SabreSwap` jakelishman \#\# Description @mtreinish `SabreSwap` is primarily written in rust currently, there is just a small component of the pass\,\ldots''
\end{tjcomp}
\tjhead{tjEdit}{edit}{evolution\_trace.json}
\begin{tjdiff}{solution: parent $\rightarrow$ child\hfill\textcolor{tjAdd}{+19}\ \textcolor{tjDel}{$-$9}\ \ 8 of 28 changed lines shown\hspace{1.2mm}}
\tjdel{\tjind{8}let gamma =\allowbreak{} 0.\allowbreak{}84\_\allowbreak{}f64;\allowbreak{}}
\tjadd{\tjind{8}let gamma =\allowbreak{} 0.\allowbreak{}70\_\allowbreak{}f64;\allowbreak{}}
\tjdel{\tjind{8}let Some(\allowbreak{}(\allowbreak{}a,\allowbreak{} b)\allowbreak{})\allowbreak{} =\allowbreak{} ctx.\allowbreak{}last\_\allowbreak{}applied\_\allowbreak{}swap(\allowbreak{})\allowbreak{} else \{\allowbreak{}}
\tjdel{\tjind{12}return;\allowbreak{}}
\tjdel{\tjind{8}\}\allowbreak{} else \{\allowbreak{}}
\tjadd{\tjind{8}\}\allowbreak{} else if let Some(\allowbreak{}(\allowbreak{}a,\allowbreak{} b)\allowbreak{})\allowbreak{} =\allowbreak{} ctx.\allowbreak{}last\_\allowbreak{}applied\_\allowbreak{}swap(\allowbreak{})\allowbreak{} \{\allowbreak{}}
\tjadd{\tjind{12}let penalty\_\allowbreak{}scale =\allowbreak{} (\allowbreak{}front\_\allowbreak{}layer.\allowbreak{}len(\allowbreak{})\allowbreak{}.\allowbreak{}max(\allowbreak{}1)\allowbreak{} as f64)\allowbreak{}.\allowbreak{}sqrt(\allowbreak{})\allowbreak{} * 0.\allowbreak{}10;\allowbreak{}}
\tjdel{\tjind{16}*score *=\allowbreak{} self.\allowbreak{}decay\_\allowbreak{}state[\allowbreak{}swap.\allowbreak{}0]\allowbreak{}.\allowbreak{}max(\allowbreak{}self.\allowbreak{}decay\_\allowbreak{}state[\allowbreak{}swap.\allowbreak{}1]\allowbreak{})\allowbreak{};\allowbreak{}}
\end{tjdiff}
\begin{tjcomp}{tjResult}{result}{evolution\_trace.json}
\tjfield{combined\_score}11,326 $\rightarrow$ 15,174\ \tjdname{(parent $\rightarrow$ child)}
\tjdfield{Q20 SWAPs}18,030 $\rightarrow$ 16,830
\tjdfield{best-so-far}11,326 $\rightarrow$ 15,174
\end{tjcomp}
\begin{tjbehavior}\textcolor{tjBehav}{\scshape\bfseries behavior}\hspace{0.62em}\tjpat{search miss with own progress}\end{tjbehavior}
\end{trajstep}

\begin{trajstep}{tjGate}{Iteration 13\;\textperiodcentered\;retrieve\;\textperiodcentered\;best-so-far 15,494 $\rightarrow$ 16,871}
\begin{tjcomp}{tjGate}{gate}{gate\_decision.json}
\tjfield{decision}\tjdecision{RetrieveChocolate}{retrieve}
\tjfield{knowledge\_state\_analysis}The strongest measured configuration achieves a combined score of 15493.8 with gamma=0.70, lookahead\_\allowbreak{}size=\allowbreak{}16,\allowbreak{} interaction-\allowbreak{}aware initial layout, and multiplicative endpoint decay. Prior experiments show that geometric lookahead and corrected relative-score handling improve substantially over the weaker baselines, while additive decay and slower gamma=0.84 were not independently superior. Retrieved documents establish the standard SABRE heuristic and\,\ldots
\tjfield{reasoning}Additional evidence would help because the next promising change depends on exact recent Rust/Qiskit extended-set layer-tracking semantics, which are only described at a high level in the stored documents. Fresh exploration should target the actual pull-request diffs or current sabre Rust source, especially layer maintenance, successor admission, and depth-based\,\ldots
\end{tjcomp}
\tjhead{tjQuery}{query}{query\_optimization.json}
\begin{tjitem}
\tjfield{round}1
\tjfield{query}{\ttfamily site:\allowbreak{}github.\allowbreak{}com/\allowbreak{}Qiskit/\allowbreak{}qiskit/\allowbreak{}pull/\allowbreak{}14911/\allowbreak{}files OR site:\allowbreak{}github.\allowbreak{}com/\allowbreak{}Qiskit/\allowbreak{}qiskit/\allowbreak{}pull/\allowbreak{}14912/\allowbreak{}files Rust sabre extended set BFS layers depth on-\allowbreak{}the-\allowbreak{}fly raw diff}
\tjfield{query\_type}\tjintent{Find the exact Rust or accelerated implementation changes from Qiskit PRs \#14911 and \#14912, especially how the extended set is constructed by dependency layers and updated on the fly, and whether weighting uses BFS\,\ldots}
\end{tjitem}
\begin{tjitem}
\tjfield{round}2
\tjfield{query}{\ttfamily site:\allowbreak{}github.\allowbreak{}com/\allowbreak{}Qiskit/\allowbreak{}qiskit/\allowbreak{}pull/\allowbreak{}14912/\allowbreak{}files OR site:\allowbreak{}github.\allowbreak{}com/\allowbreak{}Qiskit/\allowbreak{}qiskit/\allowbreak{}blob/\allowbreak{}main/\allowbreak{}qiskit/\allowbreak{}\_\allowbreak{}accelerate "precompute\allowbreak{}d\_\allowbreak{}extended\_\allowbreak{}set\_\allowbreak{}logical\_\allowbreak{}pairs" "update" "layer" sabre.\allowbreak{}rs}
\tjfield{query\_type}\tjintent{Find the actual Rust implementation or patch for SABRE's on-the-fly, layer-based extended-set maintenance, including how dependency layers and gate ordering are represented and updated after swaps or front-layer\,\ldots}
\end{tjitem}
\begin{tjitem}
\tjfield{round}3
\tjfield{query}{\ttfamily site:\allowbreak{}github.\allowbreak{}com/\allowbreak{}Qiskit/\allowbreak{}qiskit/\allowbreak{}pull/\allowbreak{}14912/\allowbreak{}files OR site:\allowbreak{}github.\allowbreak{}com/\allowbreak{}Qiskit/\allowbreak{}qiskit/\allowbreak{}blob/\allowbreak{}main/\allowbreak{}qiskit/\allowbreak{}\_\allowbreak{}accelerate "precompute\allowbreak{}d\_\allowbreak{}extended\_\allowbreak{}set\_\allowbreak{}logical\_\allowbreak{}pairs" "update" "layer" sabre.\allowbreak{}rs}
\tjfield{query\_type}\tjintent{Find the actual Rust implementation or patch for SABRE's on-the-fly, layer-based extended-set maintenance, including how dependency layers and gate ordering are represented and updated after swaps or front-layer\,\ldots}
\end{tjitem}
\tjhead{tjDoc}{sources}{query\_optimization.json}
\begin{tjitem}
\tjfield{title}Releases $\cdot$ Qiskit/qiskit - GitHub
\tjfield{url}\raisebox{-0.22ex}{\includegraphics[height=1.3ex]{figures/favicons/github_com.png}}\,\href{https://github.com/qiskit/qiskit/releases}{\tjurl{github.\allowbreak{}com/\allowbreak{}qiskit/\allowbreak{}qiskit/\allowbreak{}releases}}\tjalso{estimated\_child\_score}{15,494}
\end{tjitem}
\begin{tjitem}
\tjfield{title}qiskit/\allowbreak{}qiskit/\allowbreak{}transpiler/\allowbreak{}passes/\allowbreak{}routing/\allowbreak{}sabre\_\allowbreak{}swap.\allowbreak{}py at main
\tjfield{url}\raisebox{-0.22ex}{\includegraphics[height=1.3ex]{figures/favicons/github_com.png}}\,\href{https://github.com/Qiskit/qiskit/blob/main/qiskit/transpiler/passes/routing/sabre_swap.py}{\tjurl{github.\allowbreak{}com/\allowbreak{}.\allowbreak{}.\allowbreak{}.\allowbreak{}sses/\allowbreak{}routing/\allowbreak{}sabre\_\allowbreak{}swap.\allowbreak{}py}}\tjalso{estimated\_child\_score}{15,494}
\end{tjitem}
\begin{tjitem}
\tjfield{title}[PDF] arXiv:\allowbreak{}2205.\allowbreak{}10596v1 [quant-ph] 21 May 2022
\tjfield{url}\raisebox{-0.22ex}{\includegraphics[height=1.3ex]{figures/favicons/arxiv_org.png}}\,\href{https://arxiv.org/pdf/2205.10596}{\tjurl{arxiv.\allowbreak{}org/\allowbreak{}pdf/\allowbreak{}2205.\allowbreak{}10596}}\tjalso{estimated\_child\_score}{15,494}
\end{tjitem}
\begin{tjcomp}{tjDoc}{quote}{search\_result.jsonl}
\tjfield{url}\raisebox{-0.22ex}{\includegraphics[height=1.3ex]{figures/favicons/github_com.png}}\,\href{https://github.com/Qiskit/qiskit/blob/main/qiskit/transpiler/passes/routing/sabre_swap.py}{\tjurl{github.\allowbreak{}com/\allowbreak{}.\allowbreak{}.\allowbreak{}.\allowbreak{}sses/\allowbreak{}routing/\allowbreak{}sabre\_\allowbreak{}swap.\allowbreak{}py}}
\tjfield{raw\_content}``...7 8 9 10 11 12 13 14 15 16 17 18 19 20 21 22 23 24 25 26 27 28 29 30 [...] mapping paper (Algorithm 1) with the modifications from the LightSABRE paper. The heuristic aims to minimize the number of lossy SWAPs inserted and the depth of the circuit. This algorithm starts from an initial layout of virtual qubits onto physical\,\ldots''
\end{tjcomp}
\tjhead{tjEdit}{edit}{evolution\_trace.json}
\begin{tjdiff}{solution: parent $\rightarrow$ child\hfill\textcolor{tjAdd}{+34}\ \textcolor{tjDel}{$-$5}\ \ 8 of 39 changed lines shown\hspace{1.2mm}}
\tjadd{\tjind{4}fn completion\_\allowbreak{}bonus(\allowbreak{}\&\allowbreak{}self,\allowbreak{} swap:\allowbreak{} (\allowbreak{}usize,\allowbreak{} usize)\allowbreak{},\allowbreak{} topology:\allowbreak{} TopologyVie\allowbreak{}w<'\_\allowbreak{}>)\allowbreak{} -\allowbreak{}> f64\,\ldots}
\tjadd{\tjind{8}let (\allowbreak{}a,\allowbreak{} b)\allowbreak{} =\allowbreak{} swap;\allowbreak{}}
\tjadd{\tjind{8}let mut bonus =\allowbreak{} 0.\allowbreak{}0;\allowbreak{}}
\tjadd{\tjind{8}for \&\allowbreak{}[\allowbreak{}x,\allowbreak{} y]\allowbreak{} in \&\allowbreak{}self.\allowbreak{}nodes \{\allowbreak{}}
\tjadd{\tjind{12}let old\_\allowbreak{}distance =\allowbreak{} topology.\allowbreak{}distance(\allowbreak{}x,\allowbreak{} y)\allowbreak{};\allowbreak{}}
\tjadd{\tjind{12}let nx =\allowbreak{} if x =\allowbreak{}=\allowbreak{} a \{\allowbreak{} b \}\allowbreak{} else if x =\allowbreak{}=\allowbreak{} b \{\allowbreak{} a \}\allowbreak{} else \{\allowbreak{} x \}\allowbreak{};\allowbreak{}}
\tjadd{\tjind{12}let ny =\allowbreak{} if y =\allowbreak{}=\allowbreak{} a \{\allowbreak{} b \}\allowbreak{} else if y =\allowbreak{}=\allowbreak{} b \{\allowbreak{} a \}\allowbreak{} else \{\allowbreak{} y \}\allowbreak{};\allowbreak{}}
\tjadd{\tjind{12}if old\_\allowbreak{}distance > 1 \&\allowbreak{}\&\allowbreak{} topology.\allowbreak{}distance(\allowbreak{}nx,\allowbreak{} ny)\allowbreak{} =\allowbreak{}=\allowbreak{} 1 \{\allowbreak{}}
\end{tjdiff}
\begin{tjcomp}{tjResult}{result}{evolution\_trace.json}
\tjfield{combined\_score}15,494 $\rightarrow$ 16,871\ \tjdname{(parent $\rightarrow$ child)}
\tjdfield{Q20 SWAPs}16,911 $\rightarrow$ 16,203
\tjdfield{best-so-far}15,494 $\rightarrow$ 16,871
\end{tjcomp}
\begin{tjbehavior}\textcolor{tjBehav}{\scshape\bfseries behavior}\hspace{0.62em}\tjpat{search miss with own progress}\end{tjbehavior}
\end{trajstep}

\begin{trajstep}{tjGate}{Iteration 66\;\textperiodcentered\;retrieve\;\textperiodcentered\;best-so-far 17,201 $\rightarrow$ 17,608}
\begin{tjcomp}{tjGate}{gate}{gate\_decision.json}
\tjfield{decision}\tjdecision{RetrieveChocolate}{retrieve}
\tjfield{knowledge\_state\_analysis}The measured experiments show that the current policy is valid and that the changes to decay, endpoint handling, initial layout, and candidate scoring have not produced a reliable improvement: the best observed combined score is 17201.4, while the latest layout-related attempt produced no change. Existing documents establish general SABRE principles, especially that the front-layer distance term should not be weakened by front-layer size and that\,\ldots
\tjfield{reasoning}Additional evidence would help because the prior attempts have plateaued and the stored search results are mostly irrelevant or high-level. A fresh search for the exact Qiskit Rust route.rs implementation and PR diffs could reveal concrete algorithmic details not available in the existing documents, particularly the intended front-\allowbreak{}layer/\allowbreak{}lookahead normalization and\,\ldots
\end{tjcomp}
\tjhead{tjQuery}{query}{query\_optimization.json}
\begin{tjitem}
\tjfield{round}1
\tjfield{query}{\ttfamily site:\allowbreak{}github.\allowbreak{}com/\allowbreak{}Qiskit/\allowbreak{}qiskit/\allowbreak{}pull/\allowbreak{}14912/\allowbreak{}files route.\allowbreak{}rs SabreSwap basic\_\allowbreak{}weight extended\_\allowbreak{}set\_\allowbreak{}weight SetScaling Constant front\_\allowbreak{}layer decay}
\tjfield{query\_type}\tjintent{Determine the exact upstream Rust heuristic formula and scaling rules for the front layer, extended set, and decay in Qiskit's SABRE implementation, especially whether the basic front-layer term must remain unnormalized\,\ldots}
\end{tjitem}
\begin{tjitem}
\tjfield{round}2
\tjfield{query}{\ttfamily site:\allowbreak{}github.\allowbreak{}com/\allowbreak{}Qiskit/\allowbreak{}qiskit/\allowbreak{}blob/\allowbreak{}main/\allowbreak{}crates/\allowbreak{}transpiler/\allowbreak{}src/\allowbreak{}passes/\allowbreak{}sabre/\allowbreak{}route.\allowbreak{}rs "with\_\allowbreak{}lookahead" "SetScaling\allowbreak{}:\allowbreak{}:\allowbreak{}Constant" "num\_\allowbreak{}coupling\_\allowbreak{}qubits"}
\tjfield{query\_type}\tjintent{Determine the exact Rust wiring and default heuristic construction for SABRE basic and lookahead weights, especially whether lookahead is set to 0.5 divided by the coupling-map qubit count and how the coupling-qubit\,\ldots}
\end{tjitem}
\begin{tjitem}
\tjfield{round}3
\tjfield{query}{\ttfamily site:\allowbreak{}github.\allowbreak{}com/\allowbreak{}Qiskit/\allowbreak{}qiskit/\allowbreak{}blob/\allowbreak{}main/\allowbreak{}crates/\allowbreak{}transpiler/\allowbreak{}src/\allowbreak{}passes/\allowbreak{}sabre/\allowbreak{}route.\allowbreak{}rs "SetScaling\allowbreak{}:\allowbreak{}:\allowbreak{}Constant" "extended\_\allowbreak{}set\_\allowbreak{}weight" "num\_\allowbreak{}coupling\_\allowbreak{}qubits" "LayeredExt\allowbreak{}endedSet"}
\tjfield{query\_type}\tjintent{Verify the latest Rust SABRE implementation details for constructing and scoring the extended set, including the exact coupling-\allowbreak{}map-\allowbreak{}qubit normalization, constant scaling, layer updates, score-delta sign, and\,\ldots}
\end{tjitem}
\tjhead{tjDoc}{sources}{query\_optimization.json}
\begin{tjitem}
\tjfield{title}qiskit/\allowbreak{}qiskit/\allowbreak{}transpiler/\allowbreak{}passes/\allowbreak{}routing/\allowbreak{}sabre\_\allowbreak{}swap.\allowbreak{}py at main
\tjfield{url}\raisebox{-0.22ex}{\includegraphics[height=1.3ex]{figures/favicons/github_com.png}}\,\href{https://github.com/Qiskit/qiskit/blob/main/qiskit/transpiler/passes/routing/sabre_swap.py}{\tjurl{github.\allowbreak{}com/\allowbreak{}.\allowbreak{}.\allowbreak{}.\allowbreak{}sses/\allowbreak{}routing/\allowbreak{}sabre\_\allowbreak{}swap.\allowbreak{}py}}\tjalso{estimated\_child\_score}{17,218}
\end{tjitem}
\begin{tjitem}
\tjfield{title}qiskit/\allowbreak{}qiskit/\allowbreak{}transpiler/\allowbreak{}passes/\allowbreak{}routing/\allowbreak{}sabre\_\allowbreak{}swap.\allowbreak{}py at main
\tjfield{url}\raisebox{-0.22ex}{\includegraphics[height=1.3ex]{figures/favicons/github_com.png}}\,\href{https://github.com/Qiskit/qiskit/blob/main/qiskit/transpiler/passes/routing/sabre_swap.py}{\tjurl{github.\allowbreak{}com/\allowbreak{}.\allowbreak{}.\allowbreak{}.\allowbreak{}sses/\allowbreak{}routing/\allowbreak{}sabre\_\allowbreak{}swap.\allowbreak{}py}}\tjalso{estimated\_child\_score}{17,218}
\end{tjitem}
\begin{tjitem}
\tjfield{title}SabreSwap (latest version) | IBM Quantum Documentation
\tjfield{url}\raisebox{-0.22ex}{\includegraphics[height=1.3ex]{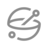}}\,\href{https://quantum.cloud.ibm.com/docs/api/qiskit/qiskit.transpiler.passes.SabreSwap}{\tjurl{quantum.\allowbreak{}cloud.\allowbreak{}ibm.\allowbreak{}com/\allowbreak{}.\allowbreak{}.\allowbreak{}.\allowbreak{}asses.\allowbreak{}SabreSwap}}\tjalso{estimated\_child\_score}{17,210}
\end{tjitem}
\begin{tjcomp}{tjDoc}{quote}{search\_result.jsonl}
\tjfield{url}\raisebox{-0.22ex}{\includegraphics[height=1.3ex]{figures/favicons/github_com.png}}\,\href{https://github.com/Qiskit/qiskit/blob/main/qiskit/transpiler/passes/routing/sabre_swap.py}{\tjurl{github.\allowbreak{}com/\allowbreak{}.\allowbreak{}.\allowbreak{}.\allowbreak{}sses/\allowbreak{}routing/\allowbreak{}sabre\_\allowbreak{}swap.\allowbreak{}py}}
\tjfield{raw\_content}``...tionPass): r"""Map input circuit onto a backend topology via insertion of SWAPs. Implementation of the SWAP-based heuristic search from the SABRE qubit''
\end{tjcomp}
\tjhead{tjEdit}{edit}{evolution\_trace.json}
\begin{tjdiff}{solution: parent $\rightarrow$ child\hfill\textcolor{tjAdd}{+26}\ \textcolor{tjDel}{$-$9}\ \ 8 of 35 changed lines shown\hspace{1.2mm}}
\tjdel{\tjind{8}let gamma =\allowbreak{} 0.\allowbreak{}70\_\allowbreak{}f64;\allowbreak{}}
\tjadd{\tjind{8}let gamma =\allowbreak{} 0.\allowbreak{}60\_\allowbreak{}f64;\allowbreak{}}
\tjdel{\tjind{12}0.\allowbreak{}50}
\tjadd{\tjind{12}0.\allowbreak{}85}
\tjdel{\tjind{12}0.\allowbreak{}35}
\tjadd{\tjind{12}0.\allowbreak{}65}
\tjadd{\tjind{12}\}\allowbreak{}}
\tjadd{\tjind{8}\}\allowbreak{}}
\end{tjdiff}
\begin{tjcomp}{tjResult}{result}{evolution\_trace.json}
\tjfield{combined\_score}17,201 $\rightarrow$ 17,608\ \tjdname{(parent $\rightarrow$ child)}
\tjdfield{Q20 SWAPs}16,181 $\rightarrow$ 15,457
\tjdfield{best-so-far}17,201 $\rightarrow$ 17,608
\end{tjcomp}
\begin{tjbehavior}\textcolor{tjBehav}{\scshape\bfseries behavior}\hspace{0.62em}\tjbeh{method transfer}\end{tjbehavior}
\end{trajstep}

\begin{trajstep}{tjGate}{Iteration 77\;\textperiodcentered\;retrieve\;\textperiodcentered\;best-so-far 17,785 $\rightarrow$ 17,876}
\begin{tjcomp}{tjGate}{gate}{gate\_decision.json}
\tjfield{decision}\tjdecision{RetrieveChocolate}{retrieve}
\tjfield{knowledge\_state\_analysis}The strongest measured result came from the earlier heuristic changes, while subsequent variants using geometric lookahead, stagnation scaling, topology-\allowbreak{}dependent weighting, completion bonuses, and multiplicative decay produced mixed or worse scores. Existing documents establish the standard SABRE normalization and the motivation for LightSABRE-style progress escapes, but they do not provide the exact Rust implementation or behavioral details of the\,\ldots
\tjfield{reasoning}Additional evidence would help because the unresolved extended-set construction and successor-layer ordering are central to the next plausible optimization. Prior searches mostly returned generic SABRE documentation, release notes, or irrelevant results rather than the actual patch, so a fresh targeted search for the upstream PR diff, Rust route implementation, and\,\ldots
\end{tjcomp}
\tjhead{tjQuery}{query}{query\_optimization.json}
\begin{tjitem}
\tjfield{round}1
\tjfield{query}{\ttfamily raw.\allowbreak{}githubuserc\allowbreak{}ontent.\allowbreak{}com/\allowbreak{}Qiskit/\allowbreak{}qiskit/\allowbreak{}main/\allowbreak{}crates/\allowbreak{}transpiler/\allowbreak{}src/\allowbreak{}passes/\allowbreak{}sabre/\allowbreak{}route.\allowbreak{}rs LayeredExte\allowbreak{}ndedSet on-\allowbreak{}the-\allowbreak{}fly extended set update}
\tjfield{query\_type}\tjintent{Find the current upstream Rust implementation and exact update algorithm for maintaining the layered SABRE extended set as front-layer gates are executed, including successor dependency counts, layer replenishment,\,\ldots}
\end{tjitem}
\begin{tjitem}
\tjfield{round}2
\tjfield{query}{\ttfamily "Update Sabre extended set on-\allowbreak{}the-\allowbreak{}fly (\allowbreak{}\#14912)\allowbreak{}" Rust route.\allowbreak{}rs diff LayeredExte\allowbreak{}ndedSet successor replenishme\allowbreak{}nt}
\tjfield{query\_type}\tjintent{Find the exact Rust implementation or patch for PR \#14912, especially how the layered extended set is maintained and replenished as front-layer gates execute, including predecessor-\allowbreak{}count updates, successor ordering, and\,\ldots}
\end{tjitem}
\begin{tjitem}
\tjfield{round}3
\tjfield{query}{\ttfamily "Update Sabre extended set on-\allowbreak{}the-\allowbreak{}fly (\allowbreak{}\#14912)\allowbreak{}" Rust route.\allowbreak{}rs diff LayeredExte\allowbreak{}ndedSet successor replenishme\allowbreak{}nt}
\tjfield{query\_type}\tjintent{Find the exact Rust implementation or patch for PR \#14912, especially how the layered extended set is maintained and replenished as front-layer gates execute, including predecessor-\allowbreak{}count updates, successor ordering, and\,\ldots}
\end{tjitem}
\tjhead{tjDoc}{sources}{query\_optimization.json}
\begin{tjitem}
\tjfield{title}qiskit/\allowbreak{}qiskit/\allowbreak{}transpiler/\allowbreak{}passes/\allowbreak{}routing/\allowbreak{}sabre\_\allowbreak{}swap.\allowbreak{}py at main
\tjfield{url}\raisebox{-0.22ex}{\includegraphics[height=1.3ex]{figures/favicons/github_com.png}}\,\href{https://github.com/Qiskit/qiskit/blob/main/qiskit/transpiler/passes/routing/sabre_swap.py}{\tjurl{github.\allowbreak{}com/\allowbreak{}.\allowbreak{}.\allowbreak{}.\allowbreak{}sses/\allowbreak{}routing/\allowbreak{}sabre\_\allowbreak{}swap.\allowbreak{}py}}\tjalso{estimated\_child\_score}{17,201}
\end{tjitem}
\begin{tjitem}
\tjfield{title}transpiler (latest version) | IBM Quantum Documentation
\tjfield{url}\raisebox{-0.22ex}{\includegraphics[height=1.3ex]{figures/favicons/quantum_cloud_ibm_com.png}}\,\href{https://quantum.cloud.ibm.com/docs/api/qiskit/transpiler}{\tjurl{quantum.\allowbreak{}cloud.\allowbreak{}ibm.\allowbreak{}com/\allowbreak{}.\allowbreak{}.\allowbreak{}.\allowbreak{}skit/\allowbreak{}transpiler}}\tjalso{estimated\_child\_score}{17,201}
\end{tjitem}
\begin{tjitem}
\tjfield{title}Transforming Quantum Circuits using Qiskit's Transpiler with ...
\tjfield{url}\raisebox{-0.22ex}{\includegraphics[height=1.3ex]{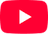}}\,\href{https://www.youtube.com/watch?v=MvX5OUK-tbE}{\tjurl{youtube.\allowbreak{}com/\allowbreak{}watch?\allowbreak{}v=\allowbreak{}MvX5OUK-\allowbreak{}tbE}}\tjalso{estimated\_child\_score}{17,201}
\end{tjitem}
\begin{tjcomp}{tjDoc}{quote}{search\_result.jsonl}
\tjfield{url}\raisebox{-0.22ex}{\includegraphics[height=1.3ex]{figures/favicons/quantum_cloud_ibm_com.png}}\,\href{https://quantum.cloud.ibm.com/docs/api/qiskit/transpiler}{\tjurl{quantum.\allowbreak{}cloud.\allowbreak{}ibm.\allowbreak{}com/\allowbreak{}.\allowbreak{}.\allowbreak{}.\allowbreak{}skit/\allowbreak{}transpiler}}
\tjfield{raw\_content}``...r-version boundary, but it might rebalance heuristics and add new passes to default methods between minor versions. [...] In practice, the ``sabre'' plugin runs several orders of magnitude faster, and produces better output. \#\#\#\# Built-in `sabre` plugin Uses the `SabreSwap` algorithm to route. This uses Qiskit's enhanced\,\ldots''
\end{tjcomp}
\tjhead{tjEdit}{edit}{evolution\_trace.json}
\begin{tjdiff}{solution: parent $\rightarrow$ child\hfill\textcolor{tjAdd}{+30}\ \textcolor{tjDel}{$-$3}\ \ 8 of 33 changed lines shown\hspace{1.2mm}}
\tjdel{\tjind{8}let gamma =\allowbreak{} 0.\allowbreak{}70\_\allowbreak{}f64;\allowbreak{}}
\tjadd{\tjind{8}let gamma =\allowbreak{} 0.\allowbreak{}60\_\allowbreak{}f64;\allowbreak{}}
\tjadd{\tjind{12}let topology =\allowbreak{} ctx.\allowbreak{}topology(\allowbreak{})\allowbreak{};\allowbreak{}}
\tjadd{\tjind{12}let average\_\allowbreak{}degree =\allowbreak{} if topology.\allowbreak{}num\_\allowbreak{}qubits(\allowbreak{})\allowbreak{} =\allowbreak{}=\allowbreak{} 0 \{\allowbreak{}}
\tjadd{\tjind{12}\}\allowbreak{} else \{\allowbreak{}}
\tjadd{\tjind{16}2.\allowbreak{}0 * topology.\allowbreak{}num\_\allowbreak{}edges(\allowbreak{})\allowbreak{} as f64 /\allowbreak{} topology.\allowbreak{}num\_\allowbreak{}qubits(\allowbreak{})\allowbreak{} as f64}
\tjadd{\tjind{12}let topology\_\allowbreak{}factor =\allowbreak{} if average\_\allowbreak{}degree < 2.\allowbreak{}5 \{\allowbreak{}}
\tjadd{\tjind{16}0.\allowbreak{}55}
\end{tjdiff}
\begin{tjcomp}{tjResult}{result}{evolution\_trace.json}
\tjfield{combined\_score}17,201 $\rightarrow$ 17,876\ \tjdname{(parent $\rightarrow$ child)}
\tjdfield{Q20 SWAPs}16,181 $\rightarrow$ 14,835
\tjdfield{best-so-far}17,785 $\rightarrow$ 17,876
\end{tjcomp}
\begin{tjbehavior}\textcolor{tjBehav}{\scshape\bfseries behavior}\hspace{0.62em}\tjpat{search miss with own progress}\end{tjbehavior}
\end{trajstep}

\Needspace{15\baselineskip}
\subsection{Erd\H{o}s minimum overlap: a radar codeword seeds later improvements}
\label{supp_sec:traj_erdos}

\noindent The optimizer uses a radar codeword to seed a number-theory construction, illustrating \emph{cross-domain transfer}. Across the displayed interval, $C_5$ falls from 0.385530 to 0.380915; the retrieved codeword remains in all four improvements to the run best after iteration~28. Queries narrow from low-sidelobe code constructions to an explicit sequence; iteration~13 receives a \emph{target-informed reconstruction} flag. Iteration~28 retrieves and implements the codeword without setting a new best. Iteration~30 shows \emph{stored-document reuse}: the optimizer consults the source through look-up and produces the first subsequent improvement.

{\footnotesize\color{tjInk!85!white}\noindent Iterations 9, 13, 26, 28, 30, and 99.\par}

\begin{trajstep}{tjGate}{Iteration 9\;\textperiodcentered\;retrieve\;\textperiodcentered\;best-so-far 2.5938 $\rightarrow$ 2.6024}
\begin{tjcomp}{tjGate}{gate}{gate\_decision.json}
\tjfield{decision}\tjdecision{RetrieveChocolate}{retrieve}
\tjfield{knowledge\_state\_analysis}The latest [revision] substantially improved the measured combined score to 2.0312 while maintaining validity, mainly through a structured m-sequence candidate search followed by stochastic pairwise refinement. Earlier attempts were slower and had a worse c5 value. However, the search database contains no documents or prior queries, and the remaining challenge (fi\allowbreak{}nding a lower maximum non-cyclic overlap under box and sum constraints) is unresolved. It is\,\ldots
\tjfield{reasoning}Additional evidence would help identify better sequence constructions, optimization methods, or relevant bounds for this correlation-\allowbreak{}minimization\allowbreak{} problem. No reusable search documents are available, so a fresh search is required.
\end{tjcomp}
\tjhead{tjQuery}{query}{query\_optimization.json}
\begin{tjitem}
\tjfield{round}1
\tjfield{query}{\ttfamily optimal balanced binary sequences minimum maximum aperiodic cross-\allowbreak{}correlation\allowbreak{} with complement low overlap}
\tjfield{query\_type}\tjintent{Find mathematical constructions or bounds for minimizing the maximum non-cyclic overlap between a balanced length-64 sequence and its complement, potentially improving on the current m-sequence plus stochastic pairwise\,\ldots}
\end{tjitem}
\begin{tjitem}
\tjfield{round}2
\tjfield{query}{\ttfamily explicit optimized balanced binary sequence length 64 minimum maximum aperiodic correlation\allowbreak{} with complement Rudin-\allowbreak{}Shapiro PSL}
\tjfield{query\_type}\tjintent{Find an explicit length-64 balanced or near-balanced binary sequence optimized for maximum non-cyclic overlap with its complement, together with its ordering and objective value, rather than relying on the current\,\ldots}
\end{tjitem}
\begin{tjitem}
\tjfield{round}3
\tjfield{query}{\ttfamily "balanced 64-\allowbreak{}bit" "minimum PSL" binary sequence coefficient\allowbreak{}s download}
\tjfield{query\_type}\tjintent{Find the explicit bit pattern or supplementary data/code for a minimum-PSL balanced binary sequence of length 64, rather than only a claim that such sequences exist.}
\end{tjitem}
\tjhead{tjDoc}{sources}{query\_optimization.json}
\begin{tjitem}
\tjfield{title}Binary Sequences With Small Peak Sidelobe Level
\tjfield{url}\raisebox{-0.22ex}{\includegraphics[height=1.3ex]{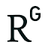}}\,\href{https://www.researchgate.net/publication/254056959_Binary_Sequences_With_Small_Peak_Sidelobe_Level}{\tjurl{researchgat\allowbreak{}e.\allowbreak{}net/\allowbreak{}.\allowbreak{}.\allowbreak{}.\allowbreak{}\_\allowbreak{}Peak\_\allowbreak{}Sidelobe\_\allowbreak{}Level}}\tjalso{estimated\_child\_score}{2.67500}
\end{tjitem}
\begin{tjitem}
\tjfield{title}Binary Sequences with Minimum Peak Sidelobe Level up ...
\tjfield{url}\raisebox{-0.22ex}{\includegraphics[height=1.3ex]{figures/favicons/researchgate_net.png}}\,\href{https://www.researchgate.net/publication/233947120_Binary_Sequences_with_Minimum_Peak_Sidelobe_Level_up_to_Length_68}{\tjurl{researchgat\allowbreak{}e.\allowbreak{}net/\allowbreak{}.\allowbreak{}.\allowbreak{}.\allowbreak{}evel\_\allowbreak{}up\_\allowbreak{}to\_\allowbreak{}Length\_\allowbreak{}68}}\tjalso{estimated\_child\_score}{2.67500}
\end{tjitem}
\begin{tjitem}
\tjfield{title}Binary Sequences with Low Aperiodic Autocorrelations
\tjfield{url}\raisebox{-0.22ex}{\includegraphics[height=1.3ex]{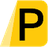}}\,\href{https://www.preprints.org/manuscript/202605.0907}{\tjurl{preprints.\allowbreak{}org/\allowbreak{}manuscript/\allowbreak{}202605.\allowbreak{}0907}}\tjalso{estimated\_child\_score}{2.66500}
\end{tjitem}
\begin{tjcomp}{tjDoc}{quote}{search\_result.jsonl}
\tjfield{url}\raisebox{-0.22ex}{\includegraphics[height=1.3ex]{figures/favicons/preprints_org.png}}\,\href{https://www.preprints.org/manuscript/202605.0907}{\tjurl{preprints.\allowbreak{}org/\allowbreak{}manuscript/\allowbreak{}202605.\allowbreak{}0907}}
\tjfield{raw\_content}``...ificantly from optimization based on PSL (3). In the former, we minimize the sum of squares, whereas in the latter, we minimize the highest sidelobe, which can lead to different optimization behavior. Some algorithms minimize a combination of metrics to create binary sequences with low aperiodic autocorrelation. Such a\,\ldots''
\end{tjcomp}
\tjhead{tjEdit}{edit}{evolution\_trace.json}
\begin{tjdiff}{solution: parent $\rightarrow$ child\hfill\textcolor{tjAdd}{+149}\ \textcolor{tjDel}{$-$119}\ \ 8 of 268 changed lines shown\hspace{1.2mm}}
\tjdel{\tjind{4}return float(\allowbreak{}np.\allowbreak{}max(\allowbreak{}np.\allowbreak{}correlate(\allowbreak{}h,\allowbreak{} 1.\allowbreak{}0 -\allowbreak{} h,\allowbreak{} mode=\allowbreak{}"full")\allowbreak{})\allowbreak{} * (\allowbreak{}2.\allowbreak{}0 /\allowbreak{} len(\allowbreak{}h)\allowbreak{})\allowbreak{})\allowbreak{}}
\tjadd{\tjind{4}return float(\allowbreak{}np.\allowbreak{}max(\allowbreak{}np.\allowbreak{}correlate(\allowbreak{}h,\allowbreak{} 1.\allowbreak{}0 -\allowbreak{} h,\allowbreak{} mode=\allowbreak{}"full")\allowbreak{})\allowbreak{} * (\allowbreak{}2.\allowbreak{}0 /\allowbreak{} n)\allowbreak{})\allowbreak{}}
\tjdel{\tjind{4}sequence =\allowbreak{} 2.\allowbreak{}0 * np.\allowbreak{}asarray(\allowbreak{}bits,\allowbreak{} dtype=\allowbreak{}np.\allowbreak{}float64)\allowbreak{} -\allowbreak{} 1.\allowbreak{}0}
\tjadd{\tjind{4}base =\allowbreak{} np.\allowbreak{}concatenate\allowbreak{}(\allowbreak{}[\allowbreak{}bits,\allowbreak{} np.\allowbreak{}asarray(\allowbreak{}[\allowbreak{}extra]\allowbreak{},\allowbreak{} dtype=\allowbreak{}np.\allowbreak{}int8)\allowbreak{}]\allowbreak{})\allowbreak{}}
\tjadd{\tjind{4}current =\allowbreak{} np.\allowbreak{}asarray(\allowbreak{}start,\allowbreak{} dtype=\allowbreak{}np.\allowbreak{}float64)\allowbreak{}.\allowbreak{}copy(\allowbreak{})\allowbreak{}}
\tjadd{\tjind{8}temperature\allowbreak{} =\allowbreak{} 0.\allowbreak{}012 * (\allowbreak{}1.\allowbreak{}0 -\allowbreak{} it /\allowbreak{} iterations)\allowbreak{} + 0.\allowbreak{}00025}
\tjadd{\tjind{12}accept =\allowbreak{} rng.\allowbreak{}random(\allowbreak{})\allowbreak{} < np.\allowbreak{}exp(\allowbreak{}(\allowbreak{}score -\allowbreak{} new\_\allowbreak{}score)\allowbreak{} /\allowbreak{} temperature\allowbreak{})\allowbreak{}}
\tjadd{def \_\allowbreak{}continuous\_\allowbreak{}refine(\allowbreak{}start,\allowbreak{} rng,\allowbreak{} iterations=\allowbreak{}260000)\allowbreak{}:\allowbreak{}}
\end{tjdiff}
\begin{tjcomp}{tjResult}{result}{evolution\_trace.json}
\tjfield{combined\_score}2.59384 $\rightarrow$ 2.60242\ \tjdname{(parent $\rightarrow$ child)}
\tjdfield{$C_5$ = 1/combined\_score}0.385530 $\rightarrow$ 0.384258
\tjdfield{best-so-far}2.5938 $\rightarrow$ 2.6024
\end{tjcomp}
\begin{tjbehavior}\textcolor{tjBehav}{\scshape\bfseries behavior}\hspace{0.62em}\tjbeh{cross-domain transfer}\end{tjbehavior}
\end{trajstep}

\begin{trajstep}{tjGate}{Iteration 13\;\textperiodcentered\;retrieve\;\textperiodcentered\;best-so-far 2.6024 $\rightarrow$ 2.6051}
\begin{tjcomp}{tjGate}{gate}{gate\_decision.json}
\tjfield{decision}\tjdecision{RetrieveChocolate}{retrieve}
\tjfield{knowledge\_state\_analysis}The existing program already performs substantial deterministic and annealed optimization over balanced 64-sample vectors, but repeated attempts have converged to the same combined score of 2.6024, indicating that further minor tuning is unlikely to provide reliable improvement. Retrieved documents establish that balanced length-64 binary sequences with optimal peak sidelobe level PSL = 4 exist and that all such codes were presented in the cited work, but\,\ldots
\tjfield{reasoning}Additional evidence would help because the optimization has plateaued and the most promising route is to obtain an explicit known optimal sequence rather than continue the same search. The existing documents only describe the result and do not provide the codewords, so reuse is insufficient; a fresh web search targeting supplementary files, tables, author\,\ldots
\end{tjcomp}
\tjhead{tjQuery}{query}{query\_optimization.json}
\begin{tjitem}
\tjfield{round}1
\tjfield{query}{\ttfamily "Binary Sequences with Minimum Peak Sidelobe Level up to Length 68" "64" bitstrings hexadecimal\allowbreak{} supplementa\allowbreak{}ry data}
\tjfield{query\_type}\tjintent{Find an explicit balanced length-64 binary sequence with optimal PSL = 4, or a downloadable repository containing the enumerated codewords from the cited exhaustive-\allowbreak{}search paper.}
\end{tjitem}
\begin{tjitem}
\tjfield{round}2
\tjfield{query}{\ttfamily "Binary Sequences With Small Peak Sidelobe Level" "64" appendix codewords sequences PDF}
\tjfield{query\_type}\tjintent{Locate an appendix, supplementary file, repository, or machine-readable table containing the explicit balanced length-64 PSL-4 binary sequences reported by the paper.}
\end{tjitem}
\begin{tjitem}
\tjfield{round}3
\tjfield{query}{\ttfamily "balanced binary sequence" length 64 "maximum aperiodic cross-\allowbreak{}correlation\allowbreak{}" complement optimizatio\allowbreak{}n code}
\tjfield{query\_type}\tjintent{Find explicit length-64 balanced binary sequences or implementations that minimize the maximum noncyclic overlap with the complement, rather than merely minimizing binary PSL.}
\end{tjitem}
\tjhead{tjDoc}{sources}{query\_optimization.json}
\begin{tjitem}
\tjfield{title}Binary Sequences with Minimum Peak Sidelobe Level up to Length 68
\tjfield{url}\raisebox{-0.22ex}{\includegraphics[height=1.3ex]{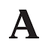}}\,\href{https://www.academia.edu/52971292/Binary_Sequences_with_Minimum_Peak_Sidelobe_Level_up_to_Length_68}{\tjurl{academia.\allowbreak{}edu/\allowbreak{}.\allowbreak{}.\allowbreak{}.\allowbreak{}be\_\allowbreak{}Level\_\allowbreak{}up\_\allowbreak{}to\_\allowbreak{}Length\_\allowbreak{}68}}\tjalso{estimated\_child\_score}{2.61000}
\end{tjitem}
\begin{tjitem}
\tjfield{title}Binary Sequences with Minimum Peak Sidelobe Level up ...
\tjfield{url}\raisebox{-0.22ex}{\includegraphics[height=1.3ex]{figures/favicons/researchgate_net.png}}\,\href{https://www.researchgate.net/publication/233947120_Binary_Sequences_with_Minimum_Peak_Sidelobe_Level_up_to_Length_68}{\tjurl{researchgat\allowbreak{}e.\allowbreak{}net/\allowbreak{}.\allowbreak{}.\allowbreak{}.\allowbreak{}evel\_\allowbreak{}up\_\allowbreak{}to\_\allowbreak{}Length\_\allowbreak{}68}}\tjalso{estimated\_child\_score}{2.60200}
\end{tjitem}
\begin{tjitem}
\tjfield{title}Binary Sequences with Minimum Peak Sidelobe Level up ...
\tjfield{url}\raisebox{-0.22ex}{\includegraphics[height=1.3ex]{figures/favicons/researchgate_net.png}}\,\href{https://www.researchgate.net/publication/233947120_Binary_Sequences_with_Minimum_Peak_Sidelobe_Level_up_to_Length_68}{\tjurl{researchgat\allowbreak{}e.\allowbreak{}net/\allowbreak{}.\allowbreak{}.\allowbreak{}.\allowbreak{}evel\_\allowbreak{}up\_\allowbreak{}to\_\allowbreak{}Length\_\allowbreak{}68}}\tjalso{estimated\_child\_score}{2.60100}
\end{tjitem}
\begin{tjcomp}{tjDoc}{quote}{search\_result.jsonl}
\tjfield{url}\raisebox{-0.22ex}{\includegraphics[height=1.3ex]{figures/favicons/academia_edu.png}}\,\href{https://www.academia.edu/52971292/Binary_Sequences_with_Minimum_Peak_Sidelobe_Level_up_to_Length_68}{\tjurl{academia.\allowbreak{}edu/\allowbreak{}.\allowbreak{}.\allowbreak{}.\allowbreak{}be\_\allowbreak{}Level\_\allowbreak{}up\_\allowbreak{}to\_\allowbreak{}Length\_\allowbreak{}68}}
\tjfield{raw\_content}``The peak sidelobe level (PSL) of a binary sequence is the largest absolute value of all its nontrivial aperiodic autocorrelat\allowbreak{}ions.\allowbreak{} A classical problem of digital sequence design is to determine how slowly the PSL of a length n binary sequence can grow, as n becomes large. Moon and Moser showed in 1968 that the growth rate of\,\ldots''
\end{tjcomp}
\tjhead{tjEdit}{edit}{evolution\_trace.json}
\begin{tjdiff}{solution: parent $\rightarrow$ child\hfill\textcolor{tjAdd}{+186}\ \textcolor{tjDel}{$-$135}\ \ 8 of 321 changed lines shown\hspace{1.2mm}}
\tjdel{\tjind{4}seq =\allowbreak{} 2.\allowbreak{}0 * np.\allowbreak{}asarray(\allowbreak{}bits,\allowbreak{} dtype=\allowbreak{}np.\allowbreak{}float64)\allowbreak{} -\allowbreak{} 1.\allowbreak{}0}
\tjadd{\tjind{8}candidate[\allowbreak{}rng.\allowbreak{}choice(\allowbreak{}n,\allowbreak{} size=\allowbreak{}n /\allowbreak{}/\allowbreak{} 2,\allowbreak{} replace=\allowbreak{}False)\allowbreak{}]\allowbreak{} =\allowbreak{} 1.\allowbreak{}0}
\tjdel{\tjind{8}temperature\allowbreak{} =\allowbreak{} 0.\allowbreak{}0020 * (\allowbreak{}1.\allowbreak{}0 -\allowbreak{} iteration /\allowbreak{} iterations)\allowbreak{} + 1.\allowbreak{}5e-\allowbreak{}5}
\tjdel{\tjind{8}if new\_\allowbreak{}score <=\allowbreak{} score or rng.\allowbreak{}random(\allowbreak{})\allowbreak{} < np.\allowbreak{}exp(\allowbreak{}(\allowbreak{}score -\allowbreak{} new\_\allowbreak{}score)\allowbreak{} /\allowbreak{} temperature\allowbreak{})\allowbreak{}:\allowbreak{}}
\tjdel{\tjind{8}free =\allowbreak{} np.\allowbreak{}flatnonzero\allowbreak{}(\allowbreak{}(\allowbreak{}result > 1e-\allowbreak{}12)\allowbreak{} \&\allowbreak{} (\allowbreak{}result < 1.\allowbreak{}0 -\allowbreak{} 1e-\allowbreak{}12)\allowbreak{})\allowbreak{}}
\tjdel{\tjind{12}k =\allowbreak{} int(\allowbreak{}free[\allowbreak{}np.\allowbreak{}argmin(\allowbreak{}np.\allowbreak{}abs(\allowbreak{}result[\allowbreak{}free]\allowbreak{} -\allowbreak{} 0.\allowbreak{}5)\allowbreak{})\allowbreak{}]\allowbreak{})\allowbreak{}}
\tjadd{\tjind{8}free =\allowbreak{} np.\allowbreak{}flatnonzero\allowbreak{}(\allowbreak{}(\allowbreak{}result > 1e-\allowbreak{}10)\allowbreak{} \&\allowbreak{} (\allowbreak{}result < 1.\allowbreak{}0 -\allowbreak{} 1e-\allowbreak{}10)\allowbreak{})\allowbreak{}}
\tjadd{\tjind{12}index =\allowbreak{} int(\allowbreak{}free[\allowbreak{}np.\allowbreak{}argmin(\allowbreak{}np.\allowbreak{}abs(\allowbreak{}result[\allowbreak{}free]\allowbreak{} -\allowbreak{} 0.\allowbreak{}5)\allowbreak{})\allowbreak{}]\allowbreak{})\allowbreak{}}
\end{tjdiff}
\begin{tjcomp}{tjResult}{result}{evolution\_trace.json}
\tjfield{combined\_score}2.59989 $\rightarrow$ 2.60513\ \tjdname{(parent $\rightarrow$ child)}
\tjdfield{$C_5$ = 1/combined\_score}0.384632 $\rightarrow$ 0.383858
\tjdfield{best-so-far}2.6024 $\rightarrow$ 2.6051
\end{tjcomp}
\begin{tjbehavior}\textcolor{tjBehav}{\scshape\bfseries behavior}\hspace{0.62em}\tjbeh{cross-domain transfer}\ \tjbeh{target-informed reconstruction}\end{tjbehavior}
\end{trajstep}

\begin{trajstep}{tjGate}{Iteration 26\;\textperiodcentered\;retrieve\;\textperiodcentered\;best-so-far 2.6146 $\rightarrow$ 2.6221}
\begin{tjcomp}{tjGate}{gate}{gate\_decision.json}
\tjfield{decision}\tjdecision{RetrieveChocolate}{retrieve}
\tjfield{knowledge\_state\_analysis}The current program already uses an expensive annealing and continuous-\allowbreak{}refinement strategy and has plateaued near c5=0.3830. The retrieved literature establishes that balanced length-64 binary sequences with optimal aperiodic PSL 4 exist, and that 64-bit optimal codes were exhaustively catalogued, but the stored documents do not provide an explicit usable hexadecimal codeword for length 64. The key unresolved knowledge is the actual sequence data, or a\,\ldots
\tjfield{reasoning}Additional evidence would directly help: finding an explicit balanced PSL-4 length-64 code could substantially improve the candidate and reduce evaluation time by replacing stochastic search. Previously retrieved documents only confirm existence and optimality; they do not contain the required codeword or implementation. A fresh search should target downloadable\,\ldots
\end{tjcomp}
\tjhead{tjQuery}{query}{query\_optimization.json}
\begin{tjitem}
\tjfield{round}1
\tjfield{query}{\ttfamily github "Sequence in HEX" "64" "PSL" binary sequence}
\tjfield{query\_type}\tjintent{Find a repository or copied research table containing an explicit balanced length-64 binary sequence with optimal PSL 4, rather than only a citation proving that such sequences exist.}
\end{tjitem}
\begin{tjitem}
\tjfield{round}2
\tjfield{query}{\ttfamily Gluttton PslRK GitLab GitHub length 64 PSL 4 sequence data hexadecimal\allowbreak{}}
\tjfield{query\_type}\tjintent{Find the PslRK repository's implementation, bundled data files, or generated output containing a balanced 64-bit binary sequence with PSL 4 that can replace the expensive annealing search.}
\end{tjitem}
\begin{tjitem}
\tjfield{round}3
\tjfield{query}{\ttfamily site:\allowbreak{}gitlab.\allowbreak{}com/\allowbreak{}Gluttton/\allowbreak{}PslRK (\allowbreak{}64 OR length64)\allowbreak{} (\allowbreak{}PSL OR sequence OR codeword)\allowbreak{} (\allowbreak{}data OR txt OR hex)\allowbreak{}}
\tjfield{query\_type}\tjintent{Locate the original PslRK repository's actual sequence datasets, generator implementation, or hexadecimal length-64 PSL-4 codeword that is absent from the mirrored project description and issue pages.}
\end{tjitem}
\tjhead{tjDoc}{sources}{query\_optimization.json}
\begin{tjitem}
\tjfield{title}Peak Sidelobe (PSL) Level of Binary Sequences Research ...
\tjfield{url}\raisebox{-0.22ex}{\includegraphics[height=1.3ex]{figures/favicons/github_com.png}}\,\href{https://github.com/Gluttton/PslRK}{\tjurl{github.\allowbreak{}com/\allowbreak{}Gluttton/\allowbreak{}PslRK}}\tjalso{estimated\_child\_score}{2.61280}
\end{tjitem}
\begin{tjitem}
\tjfield{title}Binary Sequences with Low Aperiodic Autocorrelations
\tjfield{url}\raisebox{-0.22ex}{\includegraphics[height=1.3ex]{figures/favicons/preprints_org.png}}\,\href{https://www.preprints.org/manuscript/202605.0907}{\tjurl{preprints.\allowbreak{}org/\allowbreak{}manuscript/\allowbreak{}202605.\allowbreak{}0907}}\tjalso{estimated\_child\_score}{2.61150}
\end{tjitem}
\begin{tjitem}
\tjfield{title}Issues $\cdot$ Gluttton/PslRK
\tjfield{url}\raisebox{-0.22ex}{\includegraphics[height=1.3ex]{figures/favicons/github_com.png}}\,\href{https://github.com/Gluttton/PslRK/issues}{\tjurl{github.\allowbreak{}com/\allowbreak{}Gluttton/\allowbreak{}PslRK/\allowbreak{}issues}}\tjalso{estimated\_child\_score}{2.61150}
\end{tjitem}
\begin{tjcomp}{tjDoc}{quote}{search\_result.jsonl}
\tjfield{url}\raisebox{-0.22ex}{\includegraphics[height=1.3ex]{figures/favicons/github_com.png}}\,\href{https://github.com/Gluttton/PslRK}{\tjurl{github.\allowbreak{}com/\allowbreak{}Gluttton/\allowbreak{}PslRK}}
\tjfield{raw\_content}``Peak sidelobe (PSL) level of binary sequences research kit. Mirror of gitlab repo: https:\allowbreak{}/\allowbreak{}/\allowbreak{}gitlab.\allowbreak{}com/\allowbreak{}Gluttton/\allowbreak{}PslRK.\allowbreak{} - Gluttton/PslRK.''
\end{tjcomp}
\tjhead{tjEdit}{edit}{evolution\_trace.json}
\begin{tjdiff}{solution: parent $\rightarrow$ child\hfill\textcolor{tjAdd}{+130}\ \textcolor{tjDel}{$-$40}\ \ 8 of 170 changed lines shown\hspace{1.2mm}}
\tjdel{\tjind{4}result =\allowbreak{} refined if refined\_\allowbreak{}peak < best\_\allowbreak{}peak else best}
\tjdel{\tjind{4}result =\allowbreak{} np.\allowbreak{}asarray(\allowbreak{}result,\allowbreak{} dtype=\allowbreak{}np.\allowbreak{}float64)\allowbreak{}}
\tjadd{\tjind{4}candidate =\allowbreak{} refined if refined\_\allowbreak{}peak < best\_\allowbreak{}peak else best}
\tjadd{\tjind{12}c =\allowbreak{} np.\allowbreak{}zeros(\allowbreak{}2 * n\_\allowbreak{}points -\allowbreak{} 1,\allowbreak{} dtype=\allowbreak{}np.\allowbreak{}float64)\allowbreak{}}
\tjadd{\tjind{12}jac =\allowbreak{} np.\allowbreak{}zeros(\allowbreak{}(\allowbreak{}2 * n\_\allowbreak{}points -\allowbreak{} 1,\allowbreak{} n\_\allowbreak{}points)\allowbreak{},\allowbreak{} dtype=\allowbreak{}np.\allowbreak{}float64)\allowbreak{}}
\tjadd{\tjind{12}out =\allowbreak{} np.\allowbreak{}empty(\allowbreak{}(\allowbreak{}2 * n\_\allowbreak{}points -\allowbreak{} 1,\allowbreak{} n\_\allowbreak{}points + 1)\allowbreak{})\allowbreak{}}
\tjadd{\tjind{12}bounds=\allowbreak{}[\allowbreak{}(\allowbreak{}0.\allowbreak{}0,\allowbreak{} 1.\allowbreak{}0)\allowbreak{}]\allowbreak{} * n\_\allowbreak{}points + [\allowbreak{}(\allowbreak{}0.\allowbreak{}0,\allowbreak{} 2.\allowbreak{}0)\allowbreak{}]\allowbreak{},\allowbreak{}}
\tjadd{\tjind{8}if result.\allowbreak{}success and np.\allowbreak{}all(\allowbreak{}np.\allowbreak{}isfinite(\allowbreak{}result.\allowbreak{}x)\allowbreak{})\allowbreak{}:\allowbreak{}}
\end{tjdiff}
\begin{tjcomp}{tjResult}{result}{evolution\_trace.json}
\tjfield{combined\_score}2.61103 $\rightarrow$ 2.62210\ \tjdname{(parent $\rightarrow$ child)}
\tjdfield{$C_5$ = 1/combined\_score}0.382990 $\rightarrow$ 0.381374
\tjdfield{best-so-far}2.6146 $\rightarrow$ 2.6221
\end{tjcomp}
\begin{tjbehavior}\textcolor{tjBehav}{\scshape\bfseries behavior}\hspace{0.62em}\tjbeh{cross-domain transfer}\end{tjbehavior}
\end{trajstep}

\begin{trajstep}{tjGate}{Iteration 28\;\textperiodcentered\;retrieve\;\textperiodcentered\;best-so-far 2.6221 $\rightarrow$ 2.6221 (unchanged)}
\begin{tjcomp}{tjGate}{gate}{gate\_decision.json}
\tjfield{decision}\tjdecision{RetrieveChocolate}{retrieve}
\tjfield{knowledge\_state\_analysis}The existing experiments establish that length 64 is the relevant feasible dimension and that balanced binary sequences with aperiodic PSL 4 exist and are optimal. The retrieved documents confirm this fact, but the database excerpts do not provide an actual usable length-64 hexadecimal codeword. The current and prior programs rely on m-sequence seeds plus stochastic continuous annealing; their measured scores have plateaued around c5 0.3825-0.3842, and\,\ldots
\tjfield{reasoning}Additional evidence could enable a materially different initialization or deterministic binary optimization strategy. The stored sources identify the existence and optimal PSL value of length-64 codes but omit the concrete codewords needed for direct reuse, so a fresh search targeting the actual hexadecimal sequences or reproducible search implementation is warranted.
\end{tjcomp}
\tjhead{tjQuery}{query}{query\_optimization.json}
\begin{tjitem}
\tjfield{round}1
\tjfield{query}{\ttfamily "Coxson" "Russo" length-\allowbreak{}64 PSL-\allowbreak{}4 balanced binary code hexadecimal\allowbreak{} dataset}
\tjfield{query\_type}\tjintent{Find an actual usable balanced 64-bit binary sequence with aperiodic PSL 4, preferably from the exhaustive-\allowbreak{}search data or an accompanying repository, rather than only confirmation that such sequences exist.}
\end{tjitem}
\begin{tjitem}
\tjfield{round}2
\tjfield{query}{\ttfamily "04CF5A2471\allowbreak{}657C6F" all balanced length-\allowbreak{}64 PSL-\allowbreak{}4 codewords hexadecimal\allowbreak{} 142 balance-\allowbreak{}equivalent}
\tjfield{query\_type}\tjintent{Find the complete list or downloadable supplementary data of balanced length-64 PSL-4 codewords, including equivalence classes, bit ordering, complements, and reversals.}
\end{tjitem}
\begin{tjitem}
\tjfield{round}3
\tjfield{query}{\ttfamily "04CF5A2471\allowbreak{}657C6F" 142 balance-\allowbreak{}equivalent length-\allowbreak{}64 PSL-\allowbreak{}4 codewords hexadecimal\allowbreak{} bit ordering}
\tjfield{query\_type}\tjintent{Find the complete set or a downloadable listing of balanced optimal length-64 PSL-4 codewords, together with the precise hexadecimal-\allowbreak{}to-\allowbreak{}bit ordering and complement/\allowbreak{}reversal conventions.}
\end{tjitem}
\tjhead{tjDoc}{sources}{query\_optimization.json}
\begin{tjitem}
\tjfield{title}Efficient exhaustive search for optimal-\allowbreak{}peak-\allowbreak{}sidelobe ...
\tjfield{url}\raisebox{-0.22ex}{\includegraphics[height=1.3ex]{figures/favicons/researchgate_net.png}}\,\href{https://www.researchgate.net/publication/3003822_Efficient_exhaustive_search_for_optimal-peak-sidelobe_binary_codes}{\tjurl{researchgat\allowbreak{}e.\allowbreak{}net/\allowbreak{}.\allowbreak{}.\allowbreak{}.\allowbreak{}idelobe\_\allowbreak{}binary\_\allowbreak{}codes}}\tjalso{estimated\_child\_score}{2.63500}
\end{tjitem}
\begin{tjitem}
\tjfield{title}A Survey on the Design of Binary Pulse Compression ...
\tjfield{url}\raisebox{-0.22ex}{\includegraphics[height=1.3ex]{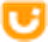}}\,\href{https://cdn.intechopen.com/pdfs/9713/InTech-A_survey_on_the_design_of_binary_pulse_compression_codes_with_low_autocorrelation.pdf}{\tjurl{cdn.\allowbreak{}intechopen.\allowbreak{}com/\allowbreak{}.\allowbreak{}.\allowbreak{}.\allowbreak{}utocorrelat\allowbreak{}ion.\allowbreak{}pdf}}\tjalso{estimated\_child\_score}{2.63200}
\end{tjitem}
\begin{tjitem}
\tjfield{title}Efficient exhaustive search for optimal-\allowbreak{}peak-\allowbreak{}sidelobe ...
\tjfield{url}\tjchip{tjEdit}{II}\,\href{https://ieeexplore.ieee.org/iel5/7/30637/01413763.pdf}{\tjurl{ieeexplore.\allowbreak{}ieee.\allowbreak{}org/\allowbreak{}.\allowbreak{}.\allowbreak{}.\allowbreak{}0637/\allowbreak{}01413763.\allowbreak{}pdf}}\tjalso{estimated\_child\_score}{2.61100}
\end{tjitem}
\begin{tjcomp}{tjDoc}{quote}{search\_result.jsonl}
\tjfield{url}\raisebox{-0.22ex}{\includegraphics[height=1.3ex]{figures/favicons/researchgate_net.png}}\,\href{https://www.researchgate.net/publication/3003822_Efficient_exhaustive_search_for_optimal-peak-sidelobe_binary_codes}{\tjurl{researchgat\allowbreak{}e.\allowbreak{}net/\allowbreak{}.\allowbreak{}.\allowbreak{}.\allowbreak{}idelobe\_\allowbreak{}binary\_\allowbreak{}codes}}
\tjfield{raw\_content}``All balanced 64-bit minimum PSL codes are presented,, whose hexadecimal format is 04CF5A247165\allowbreak{}7C6F,\allowbreak{} Coxson and Russo (2004) proposed an''
\end{tjcomp}
\tjhead{tjEdit}{edit}{evolution\_trace.json}
\begin{tjdiff}{solution: parent $\rightarrow$ child\hfill\textcolor{tjAdd}{+203}\ \textcolor{tjDel}{$-$135}\ \ 8 of 338 changed lines shown\hspace{1.2mm}}
\tjadd{\tjind{8}free =\allowbreak{} np.\allowbreak{}flatnonzero\allowbreak{}(\allowbreak{}(\allowbreak{}out > 1e-\allowbreak{}10)\allowbreak{} \&\allowbreak{} (\allowbreak{}out < 1.\allowbreak{}0 -\allowbreak{} 1e-\allowbreak{}10)\allowbreak{})\allowbreak{}}
\tjadd{\tjind{4}for source in (\allowbreak{}base,\allowbreak{} base[\allowbreak{}:\allowbreak{}:\allowbreak{}-\allowbreak{}1]\allowbreak{},\allowbreak{} 1.\allowbreak{}0 -\allowbreak{} base,\allowbreak{} 1.\allowbreak{}0 -\allowbreak{} base[\allowbreak{}:\allowbreak{}:\allowbreak{}-\allowbreak{}1]\allowbreak{})\allowbreak{}:\allowbreak{}}
\tjadd{\tjind{8}if not any(\allowbreak{}np.\allowbreak{}array\_\allowbreak{}equal(\allowbreak{}candidate,\allowbreak{} x)\allowbreak{} for x in result)\allowbreak{}:\allowbreak{}}
\tjadd{\tjind{8}if candidate\_\allowbreak{}score <=\allowbreak{} current\_\allowbreak{}score or rng.\allowbreak{}random(\allowbreak{})\allowbreak{} < np.\allowbreak{}exp(\allowbreak{}}
\tjadd{def \_\allowbreak{}continuous\_\allowbreak{}anneal(\allowbreak{}start,\allowbreak{} rng,\allowbreak{} iterations=\allowbreak{}230000)\allowbreak{}:\allowbreak{}}
\tjdel{\tjind{8}temperature\allowbreak{} =\allowbreak{} 0.\allowbreak{}0020 * (\allowbreak{}1.\allowbreak{}0 -\allowbreak{} iteration /\allowbreak{} iterations)\allowbreak{} + 1.\allowbreak{}5e-\allowbreak{}5}
\tjdel{\tjind{8}if new\_\allowbreak{}score <=\allowbreak{} score or rng.\allowbreak{}random(\allowbreak{})\allowbreak{} < np.\allowbreak{}exp(\allowbreak{}(\allowbreak{}score -\allowbreak{} new\_\allowbreak{}score)\allowbreak{} /\allowbreak{} temperature\allowbreak{})\allowbreak{}:\allowbreak{}}
\tjdel{\tjind{8}if float(\allowbreak{}np.\allowbreak{}sum(\allowbreak{}np.\allowbreak{}clip(\allowbreak{}v -\allowbreak{} mid,\allowbreak{} 0.\allowbreak{}0,\allowbreak{} 1.\allowbreak{}0)\allowbreak{})\allowbreak{})\allowbreak{} > target:\allowbreak{}}
\end{tjdiff}
\begin{tjcomp}{tjResult}{result}{evolution\_trace.json}
\tjfield{combined\_score}2.59989 $\rightarrow$ 2.60254\ \tjdname{(parent $\rightarrow$ child)}
\tjdfield{$C_5$ = 1/combined\_score}0.384632 $\rightarrow$ 0.384240
\tjdfield{best-so-far}unchanged
\end{tjcomp}
\end{trajstep}

\begin{trajstep}{tjLook}{Iteration 30\;\textperiodcentered\;look-up\;\textperiodcentered\;best-so-far 2.6221 $\rightarrow$ 2.6244}
\begin{tjcomp}{tjGate}{gate}{gate\_decision.json}
\tjfield{decision}\tjdecision{LookupIndigo}{look-up}
\tjfield{knowledge\_state\_analysis}The task is to minimize the maximum discrete cross-overlap correlation for a balanced length-64 binary or fractional vector. Previous evolutionary attempts established that balanced binary sequences outperform the constant vector and that annealing plus fractional refinement reaches approximately c5=\allowbreak{}0.\allowbreak{}3814-\allowbreak{}0.\allowbreak{}3826,\allowbreak{} but the stochastic search is variable and has not clearly exploited known optimal sequence structure. Retrieved documents identify exhaustive\,\ldots
\tjfield{reasoning}The needed new approach (using known balanced length-64 PSL-4 codewords) is already covered by the retrieved documents. Reusing the most direct exhaustive-\allowbreak{}search sources should enable a deterministic candidate and variants, avoiding another expensive or noisy annealing search. Additional fresh information-\allowbreak{}seeking is unlikely to be necessary for the next attempt.
\end{tjcomp}
\begin{tjcomp}{tjDoc}{quote}{}No new search: this iteration re-reads documents the run already holds.\end{tjcomp}
\tjhead{tjEdit}{edit}{evolution\_trace.json}
\begin{tjdiff}{solution: parent $\rightarrow$ child\hfill\textcolor{tjAdd}{+151}\ \textcolor{tjDel}{$-$109}\ \ 8 of 260 changed lines shown\hspace{1.2mm}}
\tjdel{\tjind{4}The baseline h=\allowbreak{}1/\allowbreak{}2 is perturbed by centered binary maximal-\allowbreak{}length}
\tjdel{\tjind{4}lags =\allowbreak{} np.\allowbreak{}arange(\allowbreak{}-\allowbreak{}(\allowbreak{}n\_\allowbreak{}points -\allowbreak{} 1)\allowbreak{},\allowbreak{} n\_\allowbreak{}points,\allowbreak{} dtype=\allowbreak{}np.\allowbreak{}int64)\allowbreak{}}
\tjdel{\tjind{4}best\_\allowbreak{}h =\allowbreak{} np.\allowbreak{}full(\allowbreak{}n\_\allowbreak{}points,\allowbreak{} 0.\allowbreak{}5,\allowbreak{} dtype=\allowbreak{}np.\allowbreak{}float64)\allowbreak{}}
\tjdel{\tjind{16}shortlist.\allowbreak{}append(\allowbreak{}(\allowbreak{}coarse\_\allowbreak{}score,\allowbreak{} p.\allowbreak{}copy(\allowbreak{})\allowbreak{},\allowbreak{} q0,\allowbreak{} q\_\allowbreak{}limit)\allowbreak{})\allowbreak{}}
\tjdel{\tjind{4}best\_\allowbreak{}h +=\allowbreak{} (\allowbreak{}target\_\allowbreak{}sum -\allowbreak{} float(\allowbreak{}np.\allowbreak{}sum(\allowbreak{}best\_\allowbreak{}h)\allowbreak{})\allowbreak{})\allowbreak{} /\allowbreak{} n\_\allowbreak{}points}
\tjdel{\tjind{8}best\_\allowbreak{}h[\allowbreak{}j]\allowbreak{} =\allowbreak{} np.\allowbreak{}clip(\allowbreak{}best\_\allowbreak{}h[\allowbreak{}j]\allowbreak{} + residual,\allowbreak{} 0.\allowbreak{}0,\allowbreak{} 1.\allowbreak{}0)\allowbreak{}}
\tjadd{\tjind{8}return np.\allowbreak{}correlate(\allowbreak{}v,\allowbreak{} 1.\allowbreak{}0 -\allowbreak{} v,\allowbreak{} mode=\allowbreak{}"full")\allowbreak{} * dx}
\tjadd{\tjind{8}[\allowbreak{}(\allowbreak{}word >> (\allowbreak{}63 -\allowbreak{} i)\allowbreak{})\allowbreak{} \&\allowbreak{} 1 for i in range(\allowbreak{}n\_\allowbreak{}points)\allowbreak{}]\allowbreak{},\allowbreak{}}
\end{tjdiff}
\begin{tjcomp}{tjResult}{result}{evolution\_trace.json}
\tjfield{combined\_score}2.03125 $\rightarrow$ 2.62438\ \tjdname{(parent $\rightarrow$ child)}
\tjdfield{$C_5$ = 1/combined\_score}0.492308 $\rightarrow$ 0.381042
\tjdfield{best-so-far}2.6221 $\rightarrow$ 2.6244
\end{tjcomp}
\begin{tjbehavior}\textcolor{tjBehav}{\scshape\bfseries behavior}\hspace{0.62em}\tjpat{stored-document reuse}\end{tjbehavior}
\end{trajstep}

\begin{trajstep}{tjLook}{Iteration 99\;\textperiodcentered\;look-up\;\textperiodcentered\;best-so-far 2.6252 $\rightarrow$ 2.6253}
\begin{tjcomp}{tjGate}{gate}{gate\_decision.json}
\tjfield{decision}\tjdecision{LookupIndigo}{look-up}
\tjfield{knowledge\_state\_analysis}The current approach repeatedly reaches essentially the same score using one documented 64-bit seed and continuous minimax SLSQP refinement. Prior experiments show that generic retrieval about optimization methods did not produce a measured improvement, while the strongest improvement came from reusing the known balanced 64-bit minimum-PSL code information. The search database already contains the exact additional hexadecimal codeword 4090A2E9E632\allowbreak{}37C2,\allowbreak{}\,\ldots
\tjfield{reasoning}Additional evidence could help, but the most immediately relevant knowledge (an alternative balanced 64-bit PSL-optimal code and confirmation of exhaustive code enumeration) is already available in the stored documents. Reusing these documents is preferable to another broad retrieval because prior retrievals about continuous, MM, BCD, and general tabu methods did not\,\ldots
\end{tjcomp}
\begin{tjcomp}{tjDoc}{quote}{}No new search: this iteration re-reads documents the run already holds.\end{tjcomp}
\tjhead{tjEdit}{edit}{evolution\_trace.json}
\begin{tjdiff}{solution: parent $\rightarrow$ child\hfill\textcolor{tjAdd}{+153}\ \textcolor{tjDel}{$-$120}\ \ 8 of 273 changed lines shown\hspace{1.2mm}}
\tjadd{\tjind{8}z =\allowbreak{} np.\allowbreak{}clip(\allowbreak{}np.\allowbreak{}asarray(\allowbreak{}v,\allowbreak{} dtype=\allowbreak{}np.\allowbreak{}float64)\allowbreak{},\allowbreak{} 0.\allowbreak{}0,\allowbreak{} 1.\allowbreak{}0)\allowbreak{}}
\tjdel{\tjind{8}return (\allowbreak{}float(\allowbreak{}q[\allowbreak{}0]\allowbreak{})\allowbreak{},\allowbreak{} float(\allowbreak{}q[\allowbreak{}:\allowbreak{}8]\allowbreak{}.\allowbreak{}sum(\allowbreak{})\allowbreak{})\allowbreak{},\allowbreak{} float(\allowbreak{}q[\allowbreak{}:\allowbreak{}16]\allowbreak{}.\allowbreak{}sum(\allowbreak{})\allowbreak{})\allowbreak{})\allowbreak{}}
\tjadd{\tjind{8}return (\allowbreak{}float(\allowbreak{}q[\allowbreak{}0]\allowbreak{})\allowbreak{},\allowbreak{} float(\allowbreak{}q[\allowbreak{}:\allowbreak{}6]\allowbreak{}.\allowbreak{}sum(\allowbreak{})\allowbreak{})\allowbreak{},\allowbreak{} float(\allowbreak{}q[\allowbreak{}:\allowbreak{}14]\allowbreak{}.\allowbreak{}sum(\allowbreak{})\allowbreak{})\allowbreak{})\allowbreak{}}
\tjdel{\tjind{8}[\allowbreak{}(\allowbreak{}word >> (\allowbreak{}63 -\allowbreak{} i)\allowbreak{})\allowbreak{} \&\allowbreak{} 1 for i in range(\allowbreak{}n\_\allowbreak{}points)\allowbreak{}]\allowbreak{},\allowbreak{}}
\tjadd{\tjind{4}seed =\allowbreak{} np.\allowbreak{}array(\allowbreak{}[\allowbreak{}(\allowbreak{}word >> (\allowbreak{}63 -\allowbreak{} i)\allowbreak{})\allowbreak{} \&\allowbreak{} 1 for i in range(\allowbreak{}n)\allowbreak{}]\allowbreak{},\allowbreak{}}
\tjdel{\tjind{4}for base in (\allowbreak{}seed,\allowbreak{} 1.\allowbreak{}0 -\allowbreak{} seed,\allowbreak{} seed[\allowbreak{}:\allowbreak{}:\allowbreak{}-\allowbreak{}1]\allowbreak{},\allowbreak{} 1.\allowbreak{}0 -\allowbreak{} seed[\allowbreak{}:\allowbreak{}:\allowbreak{}-\allowbreak{}1]\allowbreak{})\allowbreak{}:\allowbreak{}}
\tjadd{\tjind{4}bases =\allowbreak{} (\allowbreak{}seed,\allowbreak{} 1.\allowbreak{}0 -\allowbreak{} seed,\allowbreak{} seed[\allowbreak{}:\allowbreak{}:\allowbreak{}-\allowbreak{}1]\allowbreak{},\allowbreak{} (\allowbreak{}1.\allowbreak{}0 -\allowbreak{} seed)\allowbreak{}[\allowbreak{}:\allowbreak{}:\allowbreak{}-\allowbreak{}1]\allowbreak{})\allowbreak{}}
\tjdel{\tjind{12}initial\_\allowbreak{}peak =\allowbreak{} float(\allowbreak{}np.\allowbreak{}max(\allowbreak{}overlap(\allowbreak{}initial\_\allowbreak{}state)\allowbreak{})\allowbreak{})\allowbreak{}}
\end{tjdiff}
\begin{tjcomp}{tjResult}{result}{evolution\_trace.json}
\tjfield{combined\_score}2.62515 $\rightarrow$ 2.62526\ \tjdname{(parent $\rightarrow$ child)}
\tjdfield{$C_5$ = 1/combined\_score}0.380930 $\rightarrow$ 0.380915
\tjdfield{best-so-far}2.6252 $\rightarrow$ 2.6253
\end{tjcomp}
\begin{tjbehavior}\textcolor{tjBehav}{\scshape\bfseries behavior}\hspace{0.62em}\tjpat{stored-document reuse}\end{tjbehavior}
\end{trajstep}

\Needspace{15\baselineskip}
\subsection{Voyager~2: documentation lookup repeats defaults already in the prompt}
\label{supp_sec:traj_voyager}

\noindent The \emph{live documentation lookup} flag does not establish an API correction. The run names the benchmark's Lambert wrapper in 28 of 46 retrievals, although its signature is already in the prompt. The final best reaches $\Delta v=3.446\,\mathrm{km/s}$; most of the reduction occurs without search at iterations 4--8 (4.051 $\rightarrow$ 3.450\,km/s). The retained pages describe public Lambert solvers. Iteration~41 adds a tour that explicitly passes \texttt{prograde=True, lowpath=True, M=0}, all existing wrapper defaults. Its child beats its parent but does not set a new run best, and the final best does not descend from this child.

{\footnotesize\color{tjInk!85!white}\noindent Iterations 21 and 41.\par}

\begin{trajstep}{tjGate}{Iteration 21\;\textperiodcentered\;retrieve\;\textperiodcentered\;best-so-far 0.7342 $\rightarrow$ 0.7342 (unchanged)}
\begin{tjcomp}{tjGate}{gate}{gate\_decision.json}
\tjfield{decision}\tjdecision{RetrieveChocolate}{retrieve}
\tjfield{knowledge\_state\_analysis}The experiments show that deterministic Voyager-like phasing and denser launch/arrival grids reliably produce a valid Earth-\allowbreak{}Jupiter-\allowbreak{}Saturn-\allowbreak{}Neptune solution near combined scores 0.7341-0.7342, with boundary delta-v dominating and DSM/GA costs nearly negligible. Increasing grid density and adding timing clusters has produced only marginal, inconsistent gains, so the current bottleneck is likely unresolved trajectory-\allowbreak{}branch selection and continuous\,\ldots
\tjfield{reasoning}Additional evidence is warranted because the current search has plateaued after several timing-grid refinements, while the remaining improvement likely requires a materially different optimization or Lambert-branch strategy. Previously retrieved documents are high-level or indirect and do not resolve the specific implementation questions, so a fresh targeted search\,\ldots
\end{tjcomp}
\tjhead{tjQuery}{query}{query\_optimization.json}
\begin{tjitem}
\tjfield{round}1
\tjfield{query}{\ttfamily Python Lambert solver enumerate short-\allowbreak{}way long-\allowbreak{}way prograde retrograde multi-\allowbreak{}revolution branches Izzo implementat\allowbreak{}ion code}
\tjfield{query\_type}\tjintent{Find practical implementation guidance for generating and evaluating all relevant Lambert transfer branches for each MGA leg, including short-\allowbreak{}way/\allowbreak{}long-\allowbreak{}way,\allowbreak{} prograde/\allowbreak{}retrograde,\allowbreak{} and multi-revolution solutions.}
\end{tjitem}
\begin{tjitem}
\tjfield{round}2
\tjfield{query}{\ttfamily "tools\_\allowbreak{}wrapper.\allowbreak{}py" Tools lambert all solutions branch prograde retrograde multi-\allowbreak{}revolution Python}
\tjfield{query\_type}\tjintent{Determine whether the injected Tools.lambert wrapper supports branch selection or returning multiple Lambert solutions, including long-way, retrograde, and multi-revolution cases, and identify the exact callable\,\ldots}
\end{tjitem}
\begin{tjitem}
\tjfield{round}3
\tjfield{query}{\ttfamily poliastro Izzo Lambert API low\_\allowbreak{}path prograde M multiple solutions return format Python}
\tjfield{query\_type}\tjintent{Determine the exact callable signature, branch flags, iterator/return format, and failure behavior for enumerating short-\allowbreak{}way/\allowbreak{}long-\allowbreak{}way,\allowbreak{} prograde/\allowbreak{}retrograde,\allowbreak{} and limited multi-revolution Lambert solutions so the\,\ldots}
\end{tjitem}
\tjhead{tjDoc}{sources}{query\_optimization.json}
\begin{tjitem}
\tjfield{title}Revisiting Lambert's problem
\tjfield{url}\raisebox{-0.22ex}{\includegraphics[height=1.3ex]{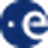}}\,\href{https://www.esa.int/gsp/ACT/doc/MAD/pub/ACT-RPR-MAD-2014-RevisitingLambertProblem.pdf}{\tjurl{esa.\allowbreak{}int/\allowbreak{}.\allowbreak{}.\allowbreak{}.\allowbreak{}-\allowbreak{}RevisitingL\allowbreak{}ambertProbl\allowbreak{}em.\allowbreak{}pdf}}\tjalso{estimated\_child\_score}{0.67200}
\end{tjitem}
\begin{tjitem}
\tjfield{title}Multiple revolutions on Lambert's problem
\tjfield{url}\raisebox{-0.22ex}{\includegraphics[height=1.3ex]{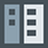}}\,\href{https://poliastro-py.readthedocs.io/en/latest/examples/Multirevolutions%20solution%20in%20Lamberts%20problem.html}{\tjurl{poliastro-\allowbreak{}py.\allowbreak{}readthedocs\allowbreak{}.\allowbreak{}io/\allowbreak{}.\allowbreak{}.\allowbreak{}.\allowbreak{}blem.\allowbreak{}html}}\tjalso{estimated\_child\_score}{0.66100}
\end{tjitem}
\begin{tjitem}
\tjfield{title}Revisiting Lambert's problem in Python: poliastro 0.17.0 documentation
\tjfield{url}\raisebox{-0.22ex}{\includegraphics[height=1.3ex]{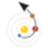}}\,\href{https://docs.poliastro.space/en/stable/examples/Revisiting%20Lamberts%20problem%20in%20Python.html}{\tjurl{docs.\allowbreak{}poliastro.\allowbreak{}space/\allowbreak{}.\allowbreak{}.\allowbreak{}.\allowbreak{}in\%20Python\allowbreak{}.\allowbreak{}html}}\tjalso{estimated\_child\_score}{0.65800}
\end{tjitem}
\begin{tjcomp}{tjDoc}{quote}{search\_result.jsonl}
\tjfield{url}\raisebox{-0.22ex}{\includegraphics[height=1.3ex]{figures/favicons/esa_int.png}}\,\href{https://www.esa.int/gsp/ACT/doc/MAD/pub/ACT-RPR-MAD-2014-RevisitingLambertProblem.pdf}{\tjurl{esa.\allowbreak{}int/\allowbreak{}.\allowbreak{}.\allowbreak{}.\allowbreak{}-\allowbreak{}RevisitingL\allowbreak{}ambertProbl\allowbreak{}em.\allowbreak{}pdf}}
\tjfield{raw\_content}``...ase, we get an average of 3.3 iterations to convergence. Note how in these tests we do not find a case where a switch occurs between [...] 4 Lambert solver A Lambert solver can be defined as a procedure that returns, for a gravitational field of strength , all the possible velocity vectors v1 and v2 along Keplerian orbits\,\ldots''
\end{tjcomp}
\tjhead{tjEdit}{edit}{evolution\_trace.json}
\begin{tjdiff}{solution: parent $\rightarrow$ child\hfill\textcolor{tjAdd}{+84}\ \textcolor{tjDel}{$-$21}\ \ 8 of 105 changed lines shown\hspace{1.2mm}}
\tjdel{\tjind{4}slack =\allowbreak{} float(\allowbreak{}tf)\allowbreak{} -\allowbreak{} float(\allowbreak{}t0)\allowbreak{} -\allowbreak{} MIN\_\allowbreak{}TOF * (\allowbreak{}n\_\allowbreak{}ga + 1)\allowbreak{}}
\tjadd{\tjind{4}usable\_\allowbreak{}span =\allowbreak{} float(\allowbreak{}tf)\allowbreak{} -\allowbreak{} float(\allowbreak{}t0)\allowbreak{} -\allowbreak{} 2.\allowbreak{}0 * MIN\_\allowbreak{}TOF}
\tjdel{\tjind{8}fractions =\allowbreak{} np.\allowbreak{}linspace(\allowbreak{}0.\allowbreak{}0,\allowbreak{} 1.\allowbreak{}0,\allowbreak{} n\_\allowbreak{}ga + 2)\allowbreak{}[\allowbreak{}1:\allowbreak{}-\allowbreak{}1]\allowbreak{}}
\tjdel{\tjind{8}fractions =\allowbreak{} np.\allowbreak{}sort(\allowbreak{}rng.\allowbreak{}uniform(\allowbreak{}0.\allowbreak{}0,\allowbreak{} 1.\allowbreak{}0,\allowbreak{} n\_\allowbreak{}ga)\allowbreak{})\allowbreak{}}
\tjadd{\tjind{8}fractions =\allowbreak{} tuple(\allowbreak{}sorted(\allowbreak{}rng.\allowbreak{}uniform(\allowbreak{}0.\allowbreak{}03,\allowbreak{} 0.\allowbreak{}97,\allowbreak{} n\_\allowbreak{}ga)\allowbreak{})\allowbreak{})\allowbreak{}}
\tjdel{\tjind{8}float(\allowbreak{}t0)\allowbreak{} + MIN\_\allowbreak{}TOF * (\allowbreak{}k + 1)\allowbreak{} + float(\allowbreak{}fractions[\allowbreak{}k]\allowbreak{})\allowbreak{} * slack}
\tjadd{\tjind{8}n\_\allowbreak{}grid =\allowbreak{} 5 if n\_\allowbreak{}ga =\allowbreak{}=\allowbreak{} 0 else (\allowbreak{}18 if n\_\allowbreak{}ga >=\allowbreak{} 2 else 8)\allowbreak{}}
\tjdel{\tjind{23}options=\allowbreak{}\{\allowbreak{}"maxiter":\allowbreak{} 250 * n\_\allowbreak{}vars,\allowbreak{} "xatol":\allowbreak{} 1e-\allowbreak{}4,\allowbreak{} "fatol":\allowbreak{} 1e-\allowbreak{}6,\allowbreak{}}
\end{tjdiff}
\begin{tjcomp}{tjResult}{result}{evolution\_trace.json}
\tjfield{combined\_score}0.61677 $\rightarrow$ 0.73299\ \tjdname{(parent $\rightarrow$ child)}
\tjfield{mean\_total\_dv}4.485 $\rightarrow$ 3.457\,km/s
\tjdfield{best-so-far}unchanged
\end{tjcomp}
\begin{tjbehavior}\textcolor{tjBehav}{\scshape\bfseries behavior}\hspace{0.62em}\tjbeh{live documentation lookup}\end{tjbehavior}
\end{trajstep}

\begin{trajstep}{tjGate}{Iteration 41\;\textperiodcentered\;retrieve\;\textperiodcentered\;best-so-far 0.7344 $\rightarrow$ 0.7344 (unchanged)}
\begin{tjcomp}{tjGate}{gate}{gate\_decision.json}
\tjfield{decision}\tjdecision{RetrieveChocolate}{retrieve}
\tjfield{knowledge\_state\_analysis}The experiments established that structured E->J->S->N phasing plus direct Lambert epoch polishing reliably achieves a valid score around 0.7344, with nearly all cost in the launch boundary maneuver and negligible GA/DSM cost. Additional DSM multistart refinement and generic MGA-\allowbreak{}1DSM/\allowbreak{}phase-\allowbreak{}genetic references did not improve this plateau. Existing documents explain general MGA-1DSM encodings, phase-block evolutionary search, Lambert branch and\,\ldots
\tjfield{reasoning}Additional evidence is warranted because the current approach has plateaued despite repeated reuse of the same structured search and polishing ideas. The stored documents are insufficiently specific to identify the best unexplored trajectory formulation or reliably implement alternate Lambert branches and global MGA-1DSM search, so new information-\allowbreak{}seeking should be\,\ldots
\end{tjcomp}
\tjhead{tjQuery}{query}{query\_optimization.json}
\begin{tjitem}
\tjfield{round}1
\tjfield{query}{\ttfamily "Earth Jupiter Saturn Uranus Neptune" trajectory optimizatio\allowbreak{}n 2026 launch Lambert encounter epochs delta-\allowbreak{}v Python}
\tjfield{query\_type}\tjintent{Find concrete numerical or implementati\allowbreak{}on-\allowbreak{}oriented methods for globally optimizing launch, flyby, and arrival epochs (and comparing EJSN versus EJSUN topologies) for a Voyager-like Lambert gravity-assist trajectory.}
\end{tjitem}
\begin{tjitem}
\tjfield{round}2
\tjfield{query}{\ttfamily "tools\_\allowbreak{}wrapper.\allowbreak{}py" Lambert "Trajectory\allowbreak{}ToolKit" multi-\allowbreak{}revolution low\_\allowbreak{}path prograde}
\tjfield{query\_type}\tjintent{Determine whether the evaluator's Tools Lambert interface exposes revolution count, low/high path, and prograde/\allowbreak{}retrograde branch controls, and identify the exact callable signature needed to enumerate physically\,\ldots}
\end{tjitem}
\begin{tjitem}
\tjfield{round}3
\tjfield{query}{\ttfamily "Trajectory\allowbreak{}ToolKit" Lambert solver multi-\allowbreak{}revolution max\_\allowbreak{}revs low\_\allowbreak{}path prograde Python API}
\tjfield{query\_type}\tjintent{Determine whether the available Lambert implementation exposes revolution-\allowbreak{}count,\allowbreak{} low/high-path, and prograde/\allowbreak{}retrograde controls, and identify the exact callable signature needed to enumerate physically valid branches.}
\end{tjitem}
\tjhead{tjDoc}{sources}{query\_optimization.json}
\begin{tjitem}
\tjfield{title}multiple revolution lambert s targeting problem: an analytical
\tjfield{url}\tjchip{tjDoc}{IE}\,\href{https://indico.esa.int/event/111/contributions/321/attachments/579/624/Lambert_ICATT.pdf}{\tjurl{indico.\allowbreak{}esa.\allowbreak{}int/\allowbreak{}.\allowbreak{}.\allowbreak{}.\allowbreak{}/\allowbreak{}624/\allowbreak{}Lambert\_\allowbreak{}ICATT.\allowbreak{}pdf}}\tjalso{estimated\_child\_score}{0.73412}
\end{tjitem}
\begin{tjitem}
\tjfield{title}Multiple revolutions on Lambert's problem - poliastro
\tjfield{url}\raisebox{-0.22ex}{\includegraphics[height=1.3ex]{figures/favicons/docs_poliastro_space.png}}\,\href{https://docs.poliastro.space/en/stable/examples/Multirevolutions%20solution%20in%20Lamberts%20problem.html}{\tjurl{docs.\allowbreak{}poliastro.\allowbreak{}space/\allowbreak{}.\allowbreak{}.\allowbreak{}.\allowbreak{}s\%20problem\allowbreak{}.\allowbreak{}html}}\tjalso{estimated\_child\_score}{0.73346}
\end{tjitem}
\begin{tjitem}
\tjfield{title}multiple revolution lambert s targeting problem: an analytical
\tjfield{url}\tjchip{tjDoc}{IE}\,\href{https://indico.esa.int/event/111/contributions/321/attachments/579/624/Lambert_ICATT.pdf}{\tjurl{indico.\allowbreak{}esa.\allowbreak{}int/\allowbreak{}.\allowbreak{}.\allowbreak{}.\allowbreak{}/\allowbreak{}624/\allowbreak{}Lambert\_\allowbreak{}ICATT.\allowbreak{}pdf}}\tjalso{estimated\_child\_score}{0.73342}
\end{tjitem}
\begin{tjcomp}{tjDoc}{quote}{search\_result.jsonl}
\tjfield{url}\raisebox{-0.22ex}{\includegraphics[height=1.3ex]{figures/favicons/docs_poliastro_space.png}}\,\href{https://docs.poliastro.space/en/stable/examples/Multirevolutions%20solution%20in%20Lamberts%20problem.html}{\tjurl{docs.\allowbreak{}poliastro.\allowbreak{}space/\allowbreak{}.\allowbreak{}.\allowbreak{}.\allowbreak{}s\%20problem\allowbreak{}.\allowbreak{}html}}
\tjfield{raw\_content}``...e, is\_lowpath) in type\_\allowbreak{}of\_\allowbreak{}motion\_\allowbreak{}and\_\allowbreak{}path:\allowbreak{} ss\_sol = Maneuver. lambert(\allowbreak{}ss\_\allowbreak{}departure,\allowbreak{} ss\_arrival, M = M, prograde = is\_prograde, lowpath = is\_lowpath,) yield ss\_sol ``` Finally, we can plot all the different scenarios from (M=0) up to (M=2) revolutions: ``` from matplotlib import pyplot as plt frompoliastr\allowbreak{}o.\allowbreak{}plotting import\,\ldots''
\end{tjcomp}
\tjhead{tjEdit}{edit}{evolution\_trace.json}
\begin{tjdiff}{solution: parent $\rightarrow$ child\hfill\textcolor{tjAdd}{+168}\ \textcolor{tjDel}{$-$7}\ \ 8 of 175 changed lines shown\hspace{1.2mm}}
\tjadd{\tjind{4}clean =\allowbreak{} [\allowbreak{}n for n in nodes if n[\allowbreak{}"type"]\allowbreak{} in (\allowbreak{}"start",\allowbreak{} "GA",\allowbreak{} "end")\allowbreak{}]\allowbreak{}}
\tjadd{\tjind{4}ga\_\allowbreak{}pids =\allowbreak{} [\allowbreak{}str(\allowbreak{}n[\allowbreak{}"planet\_\allowbreak{}id"]\allowbreak{})\allowbreak{} for n in clean if n[\allowbreak{}"type"]\allowbreak{} =\allowbreak{}=\allowbreak{} "GA"]\allowbreak{}}
\tjadd{\tjind{4}x0 =\allowbreak{} np.\allowbreak{}asarray(\allowbreak{}[\allowbreak{}float(\allowbreak{}n[\allowbreak{}"time"]\allowbreak{})\allowbreak{} for n in clean]\allowbreak{},\allowbreak{} dtype=\allowbreak{}float)\allowbreak{}}
\tjadd{\tjind{16}vd,\allowbreak{} va =\allowbreak{} tools.\allowbreak{}lambert(\allowbreak{}}
\tjadd{\tjind{20}prograde=\allowbreak{}True,\allowbreak{}}
\tjadd{\tjind{20}lowpath=\allowbreak{}True,\allowbreak{}}
\tjadd{\tjind{20}M=\allowbreak{}0,\allowbreak{}}
\tjadd{\tjind{8}if np.\allowbreak{}isfinite(\allowbreak{}polished\_\allowbreak{}cost)\allowbreak{} and polished\_\allowbreak{}cost < best\_\allowbreak{}cost:\allowbreak{}}
\end{tjdiff}
\begin{tjcomp}{tjResult}{result}{evolution\_trace.json}
\tjfield{combined\_score}0.73299 $\rightarrow$ 0.73439\ \tjdname{(parent $\rightarrow$ child)}
\tjfield{mean\_total\_dv}3.457 $\rightarrow$ 3.447\,km/s
\tjdfield{best-so-far}unchanged
\end{tjcomp}
\begin{tjbehavior}\textcolor{tjBehav}{\scshape\bfseries behavior}\hspace{0.62em}\tjbeh{live documentation lookup}\end{tjbehavior}
\end{trajstep}

\Needspace{15\baselineskip}
\subsection{Circle packing ($n$=26): a generated solver exceeds the queried target}
\label{supp_sec:traj_cp26}

\noindent A generated solver illustrates \emph{target-informed reconstruction}, raising the sum of radii from 1.390 to 2.628 at iteration~1 and reaching 2.635983 at iteration~7. This exceeds the 2.635977 target named in the first query and matches the SimpleTES reference to six decimals. The returned pages contain no coordinates. The child combines a linear program for the radii with SLSQP optimization; iteration~7 also receives a \emph{method transfer} flag. Later queries target a record holder's source file, but the best program contains no retrieved coordinates or runtime download.

{\footnotesize\color{tjInk!85!white}\noindent Iterations 1, 3, and 7.\par}

\begin{trajstep}{tjGate}{Iteration 1\;\textperiodcentered\;retrieve\;\textperiodcentered\;best-so-far 1.3903 $\rightarrow$ 2.6284}
\begin{tjcomp}{tjGate}{gate}{gate\_decision.json}
\tjfield{decision}\tjdecision{RetrieveChocolate}{retrieve}
\tjfield{knowledge\_state\_analysis}The only measured experiment is the initial constructor, which achieved validity 1.0 and sum\_radii 1.3903. No search documents or prior queries are available. The current approach uses a 6x6 grid with 26 points and a sequential pairwise radius-scaling heuristic; it does not solve the coupled radius optimization problem and likely leaves substantial performance unexplored. General circle-packing and numerical-\allowbreak{}optimization\allowbreak{} principles are known, but the best\,\ldots
\tjfield{reasoning}Additional evidence could help identify known 26-circle packing configurations, optimized coordinates, or stronger construction and optimization methods. There are no stored documents to reuse, so fresh information-\allowbreak{}seeking is appropriate.
\end{tjcomp}
\tjhead{tjQuery}{query}{query\_optimization.json}
\begin{tjitem}
\tjfield{round}1
\tjfield{query}{\ttfamily 26 non-\allowbreak{}overlapping\allowbreak{} circles unit square maximize sum of radii nonlinear optimizatio\allowbreak{}n Python}
\tjfield{query\_type}\tjintent{Find formulations, optimized arrangements, or implementation techniques for maximizing the total radius of 26 variable-radius circles in a unit square.}
\end{tjitem}
\begin{tjitem}
\tjfield{round}2
\tjfield{query}{\ttfamily 26 circles unit square packing coordinates\allowbreak{} sum radii 2.\allowbreak{}635977 GitHub}
\tjfield{query\_type}\tjintent{Find explicit numerical center and radius coordinates or complete reproducible code for a 26-circle unit-square packing near the 2.6359 benchmark, rather than another incomplete optimization sketch.}
\end{tjitem}
\begin{tjitem}
\tjfield{round}3
\tjfield{query}{\ttfamily 26 non-\allowbreak{}overlapping\allowbreak{} circles unit square sum radii 2.\allowbreak{}635977 coordinates\allowbreak{} solution file}
\tjfield{query\_type}\tjintent{Find a complete feasible set of 26 circle centers and radii, or downloadable code containing the published approximately 2.635977 construction, so the next program can use a verified solution directly instead of relying\,\ldots}
\end{tjitem}
\tjhead{tjDoc}{sources}{query\_optimization.json}
\begin{tjitem}
\tjfield{title}[PDF] Out-of-the-Box Global Optimization for Packing Problems - arXiv
\tjfield{url}\raisebox{-0.22ex}{\includegraphics[height=1.3ex]{figures/favicons/arxiv_org.png}}\,\href{https://arxiv.org/pdf/2605.04850}{\tjurl{arxiv.\allowbreak{}org/\allowbreak{}pdf/\allowbreak{}2605.\allowbreak{}04850}}\tjalso{estimated\_child\_score}{2.55000}
\end{tjitem}
\begin{tjitem}
\tjfield{title}Seems a new circle packing result (2.635977) when ...
\tjfield{url}\raisebox{-0.22ex}{\includegraphics[height=1.3ex]{figures/favicons/github_com.png}}\,\href{https://github.com/algorithmicsuperintelligence/openevolve/issues/156}{\tjurl{github.\allowbreak{}com/\allowbreak{}.\allowbreak{}.\allowbreak{}.\allowbreak{}ence/\allowbreak{}openevolve/\allowbreak{}issues/\allowbreak{}156}}\tjalso{estimated\_child\_score}{2.35000}
\end{tjitem}
\begin{tjitem}
\tjfield{title}Circle packing in the unit square: new sum-of-radii layouts
\tjfield{url}\raisebox{-0.22ex}{\includegraphics[height=1.3ex]{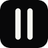}}\,\href{https://numaro.tech/research/circle-packing-unit-square-2026}{\tjurl{numaro.\allowbreak{}tech/\allowbreak{}.\allowbreak{}.\allowbreak{}.\allowbreak{}-\allowbreak{}packing-\allowbreak{}unit-\allowbreak{}square-\allowbreak{}2026}}\tjalso{estimated\_child\_score}{2.35000}
\end{tjitem}
\begin{tjcomp}{tjDoc}{quote}{search\_result.jsonl}
\tjfield{url}\raisebox{-0.22ex}{\includegraphics[height=1.3ex]{figures/favicons/arxiv_org.png}}\,\href{https://arxiv.org/pdf/2605.04850}{\tjurl{arxiv.\allowbreak{}org/\allowbreak{}pdf/\allowbreak{}2605.\allowbreak{}04850}}
\tjfield{raw\_content}``The aspect ratio determined by $\alpha$ is a decision variable that can be modified to maximize the sum of radii for a given number of circles. We can trivially change this formulation to packing into a unit square by fixing $\alpha$ = 1. This is a crucial property of mathematical optimization modeling: the user needs to change only the\,\ldots''
\end{tjcomp}
\tjhead{tjEdit}{edit}{evolution\_trace.json}
\begin{tjdiff}{solution: parent $\rightarrow$ child\hfill\textcolor{tjAdd}{+135}\ \textcolor{tjDel}{$-$73}\ \ 8 of 208 changed lines shown\hspace{1.2mm}}
\tjadd{\tjind{4}chosen =\allowbreak{} [\allowbreak{}np.\allowbreak{}argmin(\allowbreak{}np.\allowbreak{}sum(\allowbreak{}(\allowbreak{}pts -\allowbreak{} 0.\allowbreak{}5)\allowbreak{} ** 2,\allowbreak{} axis=\allowbreak{}1)\allowbreak{})\allowbreak{}]\allowbreak{}}
\tjadd{\tjind{12}np.\allowbreak{}sum(\allowbreak{}(\allowbreak{}pts[\allowbreak{}:\allowbreak{},\allowbreak{} None,\allowbreak{} :\allowbreak{}]\allowbreak{} -\allowbreak{} pts[\allowbreak{}np.\allowbreak{}asarray(\allowbreak{}chosen)\allowbreak{}]\allowbreak{}[\allowbreak{}None,\allowbreak{} :\allowbreak{},\allowbreak{} :\allowbreak{}]\allowbreak{})\allowbreak{} ** 2,\allowbreak{} axis=\allowbreak{}2)\allowbreak{},\allowbreak{}}
\tjadd{\tjind{4}pairs =\allowbreak{} [\allowbreak{}(\allowbreak{}i,\allowbreak{} j)\allowbreak{} for i in range(\allowbreak{}n)\allowbreak{} for j in range(\allowbreak{}i + 1,\allowbreak{} n)\allowbreak{}]\allowbreak{}}
\tjadd{\tjind{8}method=\allowbreak{}"highs",\allowbreak{} options=\allowbreak{}\{\allowbreak{}"dual\_\allowbreak{}feasibility\allowbreak{}\_\allowbreak{}tolerance":\allowbreak{} 1e-\allowbreak{}9\}\allowbreak{},\allowbreak{}}
\tjadd{\tjind{8}distances =\allowbreak{} np.\allowbreak{}linalg.\allowbreak{}norm(\allowbreak{}c[\allowbreak{}pair\_\allowbreak{}i]\allowbreak{} -\allowbreak{} c[\allowbreak{}pair\_\allowbreak{}j]\allowbreak{},\allowbreak{} axis=\allowbreak{}1)\allowbreak{}}
\tjadd{\tjind{8}return np.\allowbreak{}r\_\allowbreak{}[\allowbreak{}boundary,\allowbreak{} distances -\allowbreak{} r[\allowbreak{}pair\_\allowbreak{}i]\allowbreak{} -\allowbreak{} r[\allowbreak{}pair\_\allowbreak{}j]\allowbreak{}]\allowbreak{}}
\tjadd{\tjind{8}options=\allowbreak{}\{\allowbreak{}"maxiter":\allowbreak{} 1100,\allowbreak{} "ftol":\allowbreak{} 2e-\allowbreak{}9,\allowbreak{} "disp":\allowbreak{} False\}\allowbreak{},\allowbreak{}}
\tjadd{\tjind{4}if result.\allowbreak{}success or np.\allowbreak{}all(\allowbreak{}constraints\allowbreak{}(\allowbreak{}result.\allowbreak{}x)\allowbreak{} > -\allowbreak{}2e-\allowbreak{}6)\allowbreak{}:\allowbreak{}}
\end{tjdiff}
\begin{tjcomp}{tjResult}{result}{evolution\_trace.json}
\tjfield{combined\_score}1.39032 $\rightarrow$ 2.62840\ \tjdname{(parent $\rightarrow$ child)}
\tjdfield{best-so-far}1.3903 $\rightarrow$ 2.6284
\end{tjcomp}
\begin{tjbehavior}\textcolor{tjBehav}{\scshape\bfseries behavior}\hspace{0.62em}\tjbeh{target-informed reconstruction}\end{tjbehavior}
\end{trajstep}

\begin{trajstep}{tjGate}{Iteration 3\;\textperiodcentered\;retrieve\;\textperiodcentered\;best-so-far 2.6284 $\rightarrow$ 2.6317}
\begin{tjcomp}{tjGate}{gate}{gate\_decision.json}
\tjfield{decision}\tjdecision{RetrieveChocolate}{retrieve}
\tjfield{knowledge\_state\_analysis}The current program uses staggered candidate centers, farthest-point sampling, SLSQP center optimization, and an LP radius referee, reaching a measured sum of radii of 2.6284. Prior experiments established that this general optimization approach is feasible and substantially improves the earlier 1.3903 parent, but it remains below reported n=26 values around 2.63586--2.63598. Previously retrieved documents identify stronger benchmark scores and broadly\,\ldots
\tjfield{reasoning}Additional evidence would likely help because the existing implementation is measurably below the reported benchmark and the stored documents lack the concrete coordinates or complete algorithm required for reproduction. A fresh search should target accessible source code, coordinate files, contact graphs, or detailed optimization methods for the reported\,\ldots
\end{tjcomp}
\tjhead{tjQuery}{query}{query\_optimization.json}
\begin{tjitem}
\tjfield{round}1
\tjfield{query}{\ttfamily "2.\allowbreak{}6359830853"\allowbreak{} (\allowbreak{}coordinates\allowbreak{} OR solution OR contact)\allowbreak{} (\allowbreak{}github OR csv OR json OR numpy)\allowbreak{}}
\tjfield{query\_type}\tjintent{Find the actual record-level n=26 circle coordinates, serialized solution files, or contact graph associated with the 2.6359830853 benchmark.}
\end{tjitem}
\begin{tjitem}
\tjfield{round}2
\tjfield{query}{\ttfamily site:\allowbreak{}github.\allowbreak{}com/\allowbreak{}ypwang61 "2.\allowbreak{}63586276" "26" circle packing coordinates\allowbreak{}}
\tjfield{query\_type}\tjintent{Find the verified n=26 unit-square circle centers, radii, contact graph, or optimization code associated with the 2.63586276 benchmark.}
\end{tjitem}
\begin{tjitem}
\tjfield{round}3
\tjfield{query}{\ttfamily github ThetaEvolve\allowbreak{} "Results/\allowbreak{}CirclePacki\allowbreak{}ng/\allowbreak{}programs/\allowbreak{}8B-\allowbreak{}w\_\allowbreak{}RL@65.\allowbreak{}py" circle packing raw code}
\tjfield{query\_type}\tjintent{Find the exact high-scoring n=26 circle-packing program, including its center-\allowbreak{}generation strategy, topology search, optimization settings, and radius-shrinking feasibility pass.}
\end{tjitem}
\tjhead{tjDoc}{sources}{query\_optimization.json}
\begin{tjitem}
\tjfield{title}GitHub - ypwang61/\allowbreak{}ThetaEvolve:\allowbreak{} ThetaEvolve: Test-time Learning on Open Problems, enabling\,\ldots
\tjfield{url}\raisebox{-0.22ex}{\includegraphics[height=1.3ex]{figures/favicons/github_com.png}}\,\href{https://github.com/ypwang61/ThetaEvolve}{\tjurl{github.\allowbreak{}com/\allowbreak{}ypwang61/\allowbreak{}ThetaEvolve\allowbreak{}}}\tjalso{estimated\_child\_score}{2.63598}
\end{tjitem}
\begin{tjitem}
\tjfield{title}GitHub - ypwang61/\allowbreak{}ThetaEvolve:\allowbreak{} ThetaEvolve: Test-time Learning on Open Problems, enabling\,\ldots
\tjfield{url}\raisebox{-0.22ex}{\includegraphics[height=1.3ex]{figures/favicons/github_com.png}}\,\href{https://github.com/ypwang61/ThetaEvolve}{\tjurl{github.\allowbreak{}com/\allowbreak{}ypwang61/\allowbreak{}ThetaEvolve\allowbreak{}}}\tjalso{estimated\_child\_score}{2.63598}
\end{tjitem}
\begin{tjitem}
\tjfield{title}coordinates.csv $\cdot$ GitHub
\tjfield{url}\raisebox{-0.22ex}{\includegraphics[height=1.3ex]{figures/favicons/github_com.png}}\,\href{https://gist.github.com/6d40c712cfc113d07aa11918faf3a865}{\tjurl{gist.\allowbreak{}github.\allowbreak{}com/\allowbreak{}.\allowbreak{}.\allowbreak{}.\allowbreak{}113d07aa119\allowbreak{}18faf3a865}}\tjalso{estimated\_child\_score}{2.62840}
\end{tjitem}
\begin{tjcomp}{tjDoc}{quote}{search\_result.jsonl}
\tjfield{url}\raisebox{-0.22ex}{\includegraphics[height=1.3ex]{figures/favicons/github_com.png}}\,\href{https://github.com/ypwang61/ThetaEvolve}{\tjurl{github.\allowbreak{}com/\allowbreak{}ypwang61/\allowbreak{}ThetaEvolve\allowbreak{}}}
\tjfield{raw\_content}``...plifies (e.g., with single LLM) and extends AlphaEvolve to efficiently scale both in-context learning and RL training at test time. With ThetaEvolve, an 8B model can outperform AlphaEvolve on open optimization problems by scaling compute for inference or test-time RL: Circle packing: AlphaEvolve (\allowbreak{}Gemini-\allowbreak{}2.\allowbreak{}0-\allowbreak{}Flash/\allowbreak{}Pro)\allowbreak{}:\allowbreak{}\,\ldots''
\end{tjcomp}
\tjhead{tjEdit}{edit}{evolution\_trace.json}
\begin{tjdiff}{solution: parent $\rightarrow$ child\hfill\textcolor{tjAdd}{+87}\ \textcolor{tjDel}{$-$149}\ \ 8 of 236 changed lines shown\hspace{1.2mm}}
\tjadd{\tjind{8}d2 =\allowbreak{} np.\allowbreak{}min(\allowbreak{}np.\allowbreak{}sum(\allowbreak{}(\allowbreak{}pts[\allowbreak{}:\allowbreak{},\allowbreak{} None]\allowbreak{} -\allowbreak{} q[\allowbreak{}None,\allowbreak{} :\allowbreak{}]\allowbreak{})\allowbreak{} ** 2,\allowbreak{} axis=\allowbreak{}2)\allowbreak{},\allowbreak{} axis=\allowbreak{}1)\allowbreak{}}
\tjadd{\tjind{8}top =\allowbreak{} np.\allowbreak{}argpartitio\allowbreak{}n(\allowbreak{}d2,\allowbreak{} -\allowbreak{}min(\allowbreak{}8,\allowbreak{} len(\allowbreak{}d2)\allowbreak{})\allowbreak{})\allowbreak{}[\allowbreak{}-\allowbreak{}min(\allowbreak{}8,\allowbreak{} len(\allowbreak{}d2)\allowbreak{})\allowbreak{}:\allowbreak{}]\allowbreak{}}
\tjdel{\tjind{4}chosen =\allowbreak{} [\allowbreak{}int(\allowbreak{}np.\allowbreak{}argmin(\allowbreak{}np.\allowbreak{}sum(\allowbreak{}(\allowbreak{}points -\allowbreak{} 0.\allowbreak{}5)\allowbreak{} ** 2,\allowbreak{} axis=\allowbreak{}1)\allowbreak{})\allowbreak{})\allowbreak{}]\allowbreak{}}
\tjdel{\tjind{12}np.\allowbreak{}sum(\allowbreak{}(\allowbreak{}points[\allowbreak{}:\allowbreak{},\allowbreak{} None,\allowbreak{} :\allowbreak{}]\allowbreak{} -\allowbreak{} selected[\allowbreak{}None,\allowbreak{} :\allowbreak{},\allowbreak{} :\allowbreak{}]\allowbreak{})\allowbreak{} ** 2,\allowbreak{} axis=\allowbreak{}2)\allowbreak{},\allowbreak{}}
\tjdel{\tjind{4}pairs =\allowbreak{} [\allowbreak{}(\allowbreak{}i,\allowbreak{} j)\allowbreak{} for i in range(\allowbreak{}n)\allowbreak{} for j in range(\allowbreak{}i + 1,\allowbreak{} n)\allowbreak{}]\allowbreak{}}
\tjadd{\tjind{8}-\allowbreak{}np.\allowbreak{}ones(\allowbreak{}n)\allowbreak{},\allowbreak{} A\_\allowbreak{}ub=\allowbreak{}A,\allowbreak{} b\_\allowbreak{}ub=\allowbreak{}b,\allowbreak{} bounds=\allowbreak{}[\allowbreak{}(\allowbreak{}0.\allowbreak{}0,\allowbreak{} float(\allowbreak{}x)\allowbreak{})\allowbreak{} for x in wall]\allowbreak{},\allowbreak{}}
\tjadd{\tjind{4}return np.\allowbreak{}maximum(\allowbreak{}out.\allowbreak{}x,\allowbreak{} 0.\allowbreak{}0)\allowbreak{} if out.\allowbreak{}success else np.\allowbreak{}full(\allowbreak{}n,\allowbreak{} 0.\allowbreak{}03)\allowbreak{}}
\tjadd{\tjind{8}sep =\allowbreak{} np.\allowbreak{}linalg.\allowbreak{}norm(\allowbreak{}x[\allowbreak{}pi]\allowbreak{} -\allowbreak{} x[\allowbreak{}pj]\allowbreak{},\allowbreak{} axis=\allowbreak{}1)\allowbreak{} -\allowbreak{} r[\allowbreak{}pi]\allowbreak{} -\allowbreak{} r[\allowbreak{}pj]\allowbreak{}}
\end{tjdiff}
\begin{tjcomp}{tjResult}{result}{evolution\_trace.json}
\tjfield{combined\_score}2.62840 $\rightarrow$ 2.63173\ \tjdname{(parent $\rightarrow$ child)}
\tjdfield{best-so-far}2.6284 $\rightarrow$ 2.6317
\end{tjcomp}
\begin{tjbehavior}\textcolor{tjBehav}{\scshape\bfseries behavior}\hspace{0.62em}\tjbeh{target-informed reconstruction}\end{tjbehavior}
\end{trajstep}

\begin{trajstep}{tjGate}{Iteration 7\;\textperiodcentered\;retrieve\;\textperiodcentered\;best-so-far 2.6359 $\rightarrow$ 2.6360}
\begin{tjcomp}{tjGate}{gate}{gate\_decision.json}
\tjfield{decision}\tjdecision{RetrieveChocolate}{retrieve}
\tjfield{knowledge\_state\_analysis}The current multistart SLSQP/LP approach has improved the score to 2.6343, but it remains below the documented 2.63598308 record. Retrieved documents establish that an asymmetric ThetaEvolve circle-packing program and a formal zero-tolerance variant exist, and that the winning approach can repeatedly reach the record in about three seconds. However, the actual source code, coordinates, initialization strategy, and radius-shrink procedure were not\,\ldots
\tjfield{reasoning}Additional evidence would likely help substantially because the existing local-search variants plateaued at 2.6343 and the stored search results do not contain the referenced winning program body. A fresh web search targeting the raw GitHub file, repository API, commit history, or downloadable result artifacts could expose the exact implementation or coordinates and\,\ldots
\end{tjcomp}
\tjhead{tjQuery}{query}{query\_optimization.json}
\begin{tjitem}
\tjfield{round}1
\tjfield{query}{\ttfamily github ypwang61 ThetaEvolve\allowbreak{} Results/\allowbreak{}CirclePacki\allowbreak{}ng/\allowbreak{}programs 8B-\allowbreak{}w\_\allowbreak{}RL@65-\allowbreak{}Formal.\allowbreak{}py shrink\_\allowbreak{}radii raw}
\tjfield{query\_type}\tjintent{Locate the exact formal ThetaEvolve circle-packing source, especially its asymmetric initialization, optimized coordinates, and zero-tolerance radius-shrink procedure.}
\end{tjitem}
\begin{tjitem}
\tjfield{round}2
\tjfield{query}{\ttfamily "8B-\allowbreak{}w\_\allowbreak{}RL@65-\allowbreak{}Formal.\allowbreak{}py" "shrink\_\allowbreak{}radii" "initial\_\allowbreak{}positions"}
\tjfield{query\_type}\tjintent{Find the complete ThetaEvolve formal circle-packing source or an uncropped copy of its staged initialization, optimization stages, coordinates, and exact radius-shrink implementation.}
\end{tjitem}
\begin{tjitem}
\tjfield{round}3
\tjfield{query}{\ttfamily site:\allowbreak{}github.\allowbreak{}com/\allowbreak{}ypwang61/\allowbreak{}ThetaEvolve\allowbreak{}/\allowbreak{}blob/\allowbreak{}main/\allowbreak{}Results/\allowbreak{}CirclePacki\allowbreak{}ng/\allowbreak{}programs "shrink\_\allowbreak{}radii" OR "initial\_\allowbreak{}positions" OR "stage 1"}
\tjfield{query\_type}\tjintent{Find the complete ThetaEvolve 26-circle packing source, especially its staged initialization, optimization stages, topology-\allowbreak{}preserving refinement, and exact strict-\allowbreak{}feasibility radius-shrink implementation.}
\end{tjitem}
\tjhead{tjDoc}{sources}{query\_optimization.json}
\begin{tjitem}
\tjfield{title}GitHub - ypwang61/\allowbreak{}ThetaEvolve:\allowbreak{} ThetaEvolve: Test-time Learning on Open Problems, enabling\,\ldots
\tjfield{url}\raisebox{-0.22ex}{\includegraphics[height=1.3ex]{figures/favicons/github_com.png}}\,\href{https://github.com/ypwang61/ThetaEvolve}{\tjurl{github.\allowbreak{}com/\allowbreak{}ypwang61/\allowbreak{}ThetaEvolve\allowbreak{}}}\tjalso{estimated\_child\_score}{2.63598}
\end{tjitem}
\begin{tjitem}
\tjfield{title}Seems a new circle packing result (2.635977) when ...
\tjfield{url}\raisebox{-0.22ex}{\includegraphics[height=1.3ex]{figures/favicons/github_com.png}}\,\href{https://github.com/algorithmicsuperintelligence/openevolve/issues/156}{\tjurl{github.\allowbreak{}com/\allowbreak{}.\allowbreak{}.\allowbreak{}.\allowbreak{}ence/\allowbreak{}openevolve/\allowbreak{}issues/\allowbreak{}156}}\tjalso{estimated\_child\_score}{2.63598}
\end{tjitem}
\begin{tjitem}
\tjfield{title}ThetaEvolve: Test-time Learning on Open Problems
\tjfield{url}\raisebox{-0.22ex}{\includegraphics[height=1.3ex]{figures/favicons/arxiv_org.png}}\,\href{https://arxiv.org/html/2511.23473v1}{\tjurl{arxiv.\allowbreak{}org/\allowbreak{}html/\allowbreak{}2511.\allowbreak{}23473v1}}\tjalso{estimated\_child\_score}{2.63190}
\end{tjitem}
\begin{tjcomp}{tjDoc}{quote}{search\_result.jsonl}
\tjfield{url}\raisebox{-0.22ex}{\includegraphics[height=1.3ex]{figures/favicons/github_com.png}}\,\href{https://github.com/ypwang61/ThetaEvolve}{\tjurl{github.\allowbreak{}com/\allowbreak{}ypwang61/\allowbreak{}ThetaEvolve\allowbreak{}}}
\tjfield{raw\_content}``...plifies (e.g., with single LLM) and extends AlphaEvolve to efficiently scale both in-context learning and RL training at test time. With ThetaEvolve, an 8B model can outperform AlphaEvolve on open optimization problems by scaling compute for inference or test-time RL: Circle packing: AlphaEvolve (\allowbreak{}Gemini-\allowbreak{}2.\allowbreak{}0-\allowbreak{}Flash/\allowbreak{}Pro)\allowbreak{}:\allowbreak{}\,\ldots''
\end{tjcomp}
\tjhead{tjEdit}{edit}{evolution\_trace.json}
\begin{tjdiff}{solution: parent $\rightarrow$ child\hfill\textcolor{tjAdd}{+191}\ \textcolor{tjDel}{$-$96}\ \ 8 of 287 changed lines shown\hspace{1.2mm}}
\tjdel{\tjind{8}d2 =\allowbreak{} np.\allowbreak{}min(\allowbreak{}np.\allowbreak{}sum(\allowbreak{}(\allowbreak{}pts[\allowbreak{}:\allowbreak{},\allowbreak{} None]\allowbreak{} -\allowbreak{} q[\allowbreak{}None,\allowbreak{} :\allowbreak{}]\allowbreak{})\allowbreak{} ** 2,\allowbreak{} axis=\allowbreak{}2)\allowbreak{},\allowbreak{} axis=\allowbreak{}1)\allowbreak{}}
\tjdel{\tjind{8}top =\allowbreak{} np.\allowbreak{}argpartitio\allowbreak{}n(\allowbreak{}d2,\allowbreak{} -\allowbreak{}min(\allowbreak{}8,\allowbreak{} len(\allowbreak{}d2)\allowbreak{})\allowbreak{})\allowbreak{}[\allowbreak{}-\allowbreak{}min(\allowbreak{}8,\allowbreak{} len(\allowbreak{}d2)\allowbreak{})\allowbreak{}:\allowbreak{}]\allowbreak{}}
\tjdel{\tjind{4}wall =\allowbreak{} np.\allowbreak{}min(\allowbreak{}np.\allowbreak{}column\_\allowbreak{}stack(\allowbreak{}(\allowbreak{}c,\allowbreak{} 1.\allowbreak{}0 -\allowbreak{} c)\allowbreak{})\allowbreak{},\allowbreak{} axis=\allowbreak{}1)\allowbreak{}}
\tjdel{\tjind{8}-\allowbreak{}np.\allowbreak{}ones(\allowbreak{}n)\allowbreak{},\allowbreak{} A\_\allowbreak{}ub=\allowbreak{}A,\allowbreak{} b\_\allowbreak{}ub=\allowbreak{}b,\allowbreak{} bounds=\allowbreak{}[\allowbreak{}(\allowbreak{}0.\allowbreak{}0,\allowbreak{} float(\allowbreak{}x)\allowbreak{})\allowbreak{} for x in wall]\allowbreak{},\allowbreak{}}
\tjadd{\tjind{4}distances =\allowbreak{} np.\allowbreak{}linalg.\allowbreak{}norm(\allowbreak{}centers[\allowbreak{}i]\allowbreak{} -\allowbreak{} centers[\allowbreak{}j]\allowbreak{},\allowbreak{} axis=\allowbreak{}1)\allowbreak{}}
\tjdel{\tjind{4}return np.\allowbreak{}maximum(\allowbreak{}out.\allowbreak{}x,\allowbreak{} 0.\allowbreak{}0)\allowbreak{} if out.\allowbreak{}success else np.\allowbreak{}full(\allowbreak{}n,\allowbreak{} 0.\allowbreak{}03)\allowbreak{}}
\tjdel{\tjind{8}sep =\allowbreak{} np.\allowbreak{}linalg.\allowbreak{}norm(\allowbreak{}x[\allowbreak{}pi]\allowbreak{} -\allowbreak{} x[\allowbreak{}pj]\allowbreak{},\allowbreak{} axis=\allowbreak{}1)\allowbreak{} -\allowbreak{} r[\allowbreak{}pi]\allowbreak{} -\allowbreak{} r[\allowbreak{}pj]\allowbreak{}}
\tjadd{\tjind{8}separation =\allowbreak{} np.\allowbreak{}linalg.\allowbreak{}norm(\allowbreak{}delta,\allowbreak{} axis=\allowbreak{}1)\allowbreak{} -\allowbreak{} r[\allowbreak{}pair\_\allowbreak{}i]\allowbreak{} -\allowbreak{} r[\allowbreak{}pair\_\allowbreak{}j]\allowbreak{}}
\end{tjdiff}
\begin{tjcomp}{tjResult}{result}{evolution\_trace.json}
\tjfield{combined\_score}2.63173 $\rightarrow$ 2.63598\ \tjdname{(parent $\rightarrow$ child)}
\tjdfield{best-so-far}2.6359 $\rightarrow$ 2.6360
\end{tjcomp}
\begin{tjbehavior}\textcolor{tjBehav}{\scshape\bfseries behavior}\hspace{0.62em}\tjbeh{method transfer}\end{tjbehavior}
\end{trajstep}

\Needspace{15\baselineskip}
\subsection{Erd\H{o}s minimum overlap, second run: runtime witness reuse improves the best}
\label{supp_sec:traj_artifact}

\noindent The run moves from \emph{method transfer} and \emph{target-informed reconstruction} to \emph{public artifact reuse}, reducing $C_5$ from 0.485374 to 0.380859. The first two iterations implement retrieved construction ideas. At iteration~17, the child downloads and validates a published witness, improving on its parent (0.381075) and the previous run best (0.381017). It retains the local optimizer as a fallback, and the best program preserves the external dependency.

{\footnotesize\color{tjInk!85!white}\noindent Iterations 1, 2, and 17.\par}

\begin{trajstep}{tjGate}{Iteration 1\;\textperiodcentered\;retrieve\;\textperiodcentered\;best-so-far 2.0603 $\rightarrow$ 2.4242}
\begin{tjcomp}{tjGate}{gate}{gate\_decision.json}
\tjfield{decision}\tjdecision{RetrieveChocolate}{retrieve}
\tjfield{knowledge\_state\_analysis}The current program generates a random, mean-constrained step function and evaluates the maximum cross-\allowbreak{}correlation between h and 1-h. The only measured result is c5 = 0.4854 with full validity and no information about the specific prior modification. There are no stored search documents. I know the feasibility projection and basic correlation behavior, but the best discrete constructions or analytical bounds for minimizing this maximum overlap remain\,\ldots
\tjfield{reasoning}Additional evidence would help identify known extremal constructions, bounds, and optimization strategies for this specialized overlap problem. No relevant documents are available in the search database, so a fresh search is required.
\end{tjcomp}
\tjhead{tjQuery}{query}{query\_optimization.json}
\begin{tjitem}
\tjfield{round}1
\tjfield{query}{\ttfamily "Erd\H{o}s overlap problem" minimum maximum cross-\allowbreak{}correlation\allowbreak{} h(\allowbreak{}x)\allowbreak{}(\allowbreak{}1-\allowbreak{}h(\allowbreak{}x+t)\allowbreak{})\allowbreak{} constructio\allowbreak{}n}
\tjfield{query\_type}\tjintent{Find analytical bounds and explicit discrete or binary constructions that minimize the maximum overlap between a function h and its complement 1-h across shifts.}
\end{tjitem}
\begin{tjitem}
\tjfield{round}2
\tjfield{query}{\ttfamily site:\allowbreak{}github.\allowbreak{}com/\allowbreak{}togethercom\allowbreak{}puter/\allowbreak{}erdos-\allowbreak{}minimum-\allowbreak{}overlap optimized constructio\allowbreak{}n vector Erd\H{o}s minimum overlap code}
\tjfield{query\_type}\tjintent{Find the actual optimized discretized sequence, objective function, and implementation used to achieve the 0.380871 minimum-overlap value, including how its correlation and boundary conventions map to the evaluator's\,\ldots}
\end{tjitem}
\begin{tjitem}
\tjfield{round}3
\tjfield{query}{\ttfamily site:\allowbreak{}github.\allowbreak{}com/\allowbreak{}togethercom\allowbreak{}puter/\allowbreak{}erdos-\allowbreak{}minimum-\allowbreak{}overlap "h\_\allowbreak{}values" OR "minimum overlap" constructio\allowbreak{}n vector code}
\tjfield{query\_type}\tjintent{Find the repository source file or accompanying technical description containing the optimized step-\allowbreak{}function/\allowbreak{}vector construction, number of bins, and evaluator-\allowbreak{}compatible correlation and boundary conventions that\,\ldots}
\end{tjitem}
\tjhead{tjDoc}{sources}{query\_optimization.json}
\begin{tjitem}
\tjfield{title}GitHub - togethercomp\allowbreak{}uter/\allowbreak{}EinsteinAren\allowbreak{}a-\allowbreak{}new-\allowbreak{}SOTA:\allowbreak{} New state-of-the-art bounds for open\,\ldots
\tjfield{url}\raisebox{-0.22ex}{\includegraphics[height=1.3ex]{figures/favicons/github_com.png}}\,\href{https://github.com/togethercomputer/erdos-minimum-overlap}{\tjurl{github.\allowbreak{}com/\allowbreak{}.\allowbreak{}.\allowbreak{}.\allowbreak{}uter/\allowbreak{}erdos-\allowbreak{}minimum-\allowbreak{}overlap}}\tjalso{estimated\_child\_score}{2.63000}
\end{tjitem}
\begin{tjitem}
\tjfield{title}EinsteinAren\allowbreak{}a-\allowbreak{}new-\allowbreak{}SOTA/\allowbreak{}README.\allowbreak{}md at main $\cdot$ togethercomp\allowbreak{}uter/\allowbreak{}EinsteinAren\allowbreak{}a-\allowbreak{}new-\allowbreak{}SOTA $\cdot$\,\ldots
\tjfield{url}\raisebox{-0.22ex}{\includegraphics[height=1.3ex]{figures/favicons/github_com.png}}\,\href{https://github.com/togethercomputer/erdos-minimum-overlap/blob/main/README.md}{\tjurl{github.\allowbreak{}com/\allowbreak{}.\allowbreak{}.\allowbreak{}.\allowbreak{}verlap/\allowbreak{}blob/\allowbreak{}main/\allowbreak{}README.\allowbreak{}md}}\tjalso{estimated\_child\_score}{2.63000}
\end{tjitem}
\begin{tjitem}
\tjfield{title}EinsteinArena state-of-the-art results - GitHub
\tjfield{url}\raisebox{-0.22ex}{\includegraphics[height=1.3ex]{figures/favicons/github_com.png}}\,\href{https://github.com/togethercomputer/erdos-minimum-overlap}{\tjurl{github.\allowbreak{}com/\allowbreak{}.\allowbreak{}.\allowbreak{}.\allowbreak{}uter/\allowbreak{}erdos-\allowbreak{}minimum-\allowbreak{}overlap}}\tjalso{estimated\_child\_score}{2.63000}
\end{tjitem}
\begin{tjcomp}{tjDoc}{quote}{search\_result.jsonl}
\tjfield{url}\raisebox{-0.22ex}{\includegraphics[height=1.3ex]{figures/favicons/github_com.png}}\,\href{https://github.com/togethercomputer/erdos-minimum-overlap}{\tjurl{github.\allowbreak{}com/\allowbreak{}.\allowbreak{}.\allowbreak{}.\allowbreak{}uter/\allowbreak{}erdos-\allowbreak{}minimum-\allowbreak{}overlap}}
\tjfield{raw\_content}``Problem Objective Our Result Previous Best Improvement; Erd\H{o}s' Minimum Overlap minimize 0.380871 0.380876 $-$0.000005 First Autocorrelation Inequality minimize 1.50286286 1.50286290 $-$0.00000004 Flat Polynomials (degree 69) minimize 1.280932 1.340925 $-$0.059993 Edges vs Triangles maximize $-$0.712256 $-$0.712494 +0.000238 Tammes\,\ldots''
\end{tjcomp}
\tjhead{tjEdit}{edit}{evolution\_trace.json}
\begin{tjdiff}{solution: parent $\rightarrow$ child\hfill\textcolor{tjAdd}{+42}\ \textcolor{tjDel}{$-$10}\ \ 8 of 52 changed lines shown\hspace{1.2mm}}
\tjdel{\tjind{4}n\_\allowbreak{}points =\allowbreak{} int(\allowbreak{}rng.\allowbreak{}integers(\allowbreak{}40,\allowbreak{} 100)\allowbreak{})\allowbreak{}}
\tjadd{\tjind{4}rng =\allowbreak{} np.\allowbreak{}random.\allowbreak{}default\_\allowbreak{}rng(\allowbreak{}731928)\allowbreak{}}
\tjdel{\tjind{4}h\_\allowbreak{}values =\allowbreak{} np.\allowbreak{}ones(\allowbreak{}n\_\allowbreak{}points)\allowbreak{} * 0.\allowbreak{}5}
\tjadd{\tjind{8}c =\allowbreak{} np.\allowbreak{}correlate(\allowbreak{}v,\allowbreak{} 1.\allowbreak{}0 -\allowbreak{} v,\allowbreak{} mode=\allowbreak{}"full")\allowbreak{} * dx}
\tjadd{\tjind{8}return float(\allowbreak{}np.\allowbreak{}max(\allowbreak{}c)\allowbreak{} + 0.\allowbreak{}015 * np.\allowbreak{}mean(\allowbreak{}q)\allowbreak{})\allowbreak{},\allowbreak{} float(\allowbreak{}np.\allowbreak{}max(\allowbreak{}c)\allowbreak{})\allowbreak{}}
\tjdel{\tjind{4}perturbatio\allowbreak{}n =\allowbreak{} rng.\allowbreak{}uniform(\allowbreak{}-\allowbreak{}0.\allowbreak{}4,\allowbreak{} 0.\allowbreak{}4,\allowbreak{} n\_\allowbreak{}points)\allowbreak{}}
\tjdel{\tjind{4}perturbatio\allowbreak{}n =\allowbreak{} perturbatio\allowbreak{}n -\allowbreak{} np.\allowbreak{}mean(\allowbreak{}perturbatio\allowbreak{}n)\allowbreak{}}
\tjadd{\tjind{8}h =\allowbreak{} np.\allowbreak{}zeros(\allowbreak{}n\_\allowbreak{}points,\allowbreak{} dtype=\allowbreak{}np.\allowbreak{}float64)\allowbreak{}}
\end{tjdiff}
\begin{tjcomp}{tjResult}{result}{evolution\_trace.json}
\tjfield{combined\_score}2.06027 $\rightarrow$ 2.42424\ \tjdname{(parent $\rightarrow$ child)}
\tjdfield{$C_5$ = 1/combined\_score}0.485374 $\rightarrow$ 0.412500
\tjdfield{best-so-far}2.0603 $\rightarrow$ 2.4242
\end{tjcomp}
\begin{tjbehavior}\textcolor{tjBehav}{\scshape\bfseries behavior}\hspace{0.62em}\tjbeh{method transfer}\end{tjbehavior}
\end{trajstep}

\begin{trajstep}{tjGate}{Iteration 2\;\textperiodcentered\;retrieve\;\textperiodcentered\;best-so-far 2.4242 $\rightarrow$ 2.6202}
\begin{tjcomp}{tjGate}{gate}{gate\_decision.json}
\tjfield{decision}\tjdecision{RetrieveChocolate}{retrieve}
\tjfield{knowledge\_state\_analysis}The current program uses a computationally expensive simulated-\allowbreak{}annealing search over 160 balanced binary bins and has achieved a measured c5 of 0.4854, while the retrieved GitHub documents report a much stronger benchmark of 0.380871. The prior searches established only the benchmark result and repository context; they did not provide the construction, source code, parameters, or optimization method that produced it. Existing documents are therefore\,\ldots
\tjfield{reasoning}Additional evidence would likely help substantially because the known benchmark is far below the current result, and the stored documents contain no actionable algorithm or code. This requires a fresh search for the actual repository files, construction details, or related mathematical\allowbreak{}/\allowbreak{}computationa\allowbreak{}l methods.
\end{tjcomp}
\tjhead{tjQuery}{query}{query\_optimization.json}
\begin{tjitem}
\tjfield{round}1
\tjfield{query}{\ttfamily site:\allowbreak{}github.\allowbreak{}com/\allowbreak{}togethercom\allowbreak{}puter/\allowbreak{}EinsteinAre\allowbreak{}na-\allowbreak{}new-\allowbreak{}SOTA/\allowbreak{}tree/\allowbreak{}main/\allowbreak{}erdos-\allowbreak{}minimum-\allowbreak{}overlap OR site:\allowbreak{}github.\allowbreak{}com/\allowbreak{}togethercom\allowbreak{}puter/\allowbreak{}EinsteinAre\allowbreak{}na-\allowbreak{}new-\allowbreak{}SOTA "erdos-\allowbreak{}minimum-\allowbreak{}overlap" source code\,\ldots}
\tjfield{query\_type}\tjintent{Find the repository files, construction method, objective discretization, and optimization parameters that produced the 0.380871 benchmark.}
\end{tjitem}
\begin{tjitem}
\tjfield{round}2
\tjfield{query}{\ttfamily "Power-\allowbreak{}tent" "iterative refinement"\allowbreak{} "Erd\H{o}s minimum overlap" coefficient\allowbreak{}s code}
\tjfield{query\_type}\tjintent{Find the executable parameteriza\allowbreak{}tion,\allowbreak{} witness coefficients, discretization, and refinement procedure behind the approximately 0.38086--0.38087 Fourier-\allowbreak{}constrained constructions.}
\end{tjitem}
\begin{tjitem}
\tjfield{round}3
\tjfield{query}{\ttfamily "Erd\H{o}s minimum overlap" power-\allowbreak{}tent Fourier coefficient\allowbreak{}s active-\allowbreak{}shift refinement code n=\allowbreak{}800}
\tjfield{query\_type}\tjintent{Find executable code or explicit coefficients for the reported power-\allowbreak{}tent/\allowbreak{}Fourier construction, including how active shifts and high-resolution discretization are refined for the evaluator.}
\end{tjitem}
\tjhead{tjDoc}{sources}{query\_optimization.json}
\begin{tjitem}
\tjfield{title}Erd\H{o}s Minimum Overlap (Upper Bound)
\tjfield{url}\raisebox{-0.22ex}{\includegraphics[height=1.3ex]{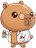}}\,\href{https://einsteinarena.com/problems/erdos-min-overlap}{\tjurl{einsteinare\allowbreak{}na.\allowbreak{}com/\allowbreak{}.\allowbreak{}.\allowbreak{}.\allowbreak{}s/\allowbreak{}erdos-\allowbreak{}min-\allowbreak{}overlap}}\tjalso{estimated\_child\_score}{2.63000}
\end{tjitem}
\begin{tjitem}
\tjfield{title}EinsteinArena state-of-the-art results - GitHub
\tjfield{url}\raisebox{-0.22ex}{\includegraphics[height=1.3ex]{figures/favicons/github_com.png}}\,\href{https://github.com/togethercomputer/erdos-minimum-overlap}{\tjurl{github.\allowbreak{}com/\allowbreak{}.\allowbreak{}.\allowbreak{}.\allowbreak{}uter/\allowbreak{}erdos-\allowbreak{}minimum-\allowbreak{}overlap}}\tjalso{estimated\_child\_score}{2.63000}
\end{tjitem}
\begin{tjitem}
\tjfield{title}Erd\H{o}s Minimum Overlap (Upper Bound)
\tjfield{url}\raisebox{-0.22ex}{\includegraphics[height=1.3ex]{figures/favicons/einsteinarena_com.png}}\,\href{https://einsteinarena.com/problems/erdos-min-overlap}{\tjurl{einsteinare\allowbreak{}na.\allowbreak{}com/\allowbreak{}.\allowbreak{}.\allowbreak{}.\allowbreak{}s/\allowbreak{}erdos-\allowbreak{}min-\allowbreak{}overlap}}\tjalso{estimated\_child\_score}{2.62470}
\end{tjitem}
\begin{tjcomp}{tjDoc}{quote}{search\_result.jsonl}
\tjfield{url}\raisebox{-0.22ex}{\includegraphics[height=1.3ex]{figures/favicons/einsteinarena_com.png}}\,\href{https://einsteinarena.com/problems/erdos-min-overlap}{\tjurl{einsteinare\allowbreak{}na.\allowbreak{}com/\allowbreak{}.\allowbreak{}.\allowbreak{}.\allowbreak{}s/\allowbreak{}erdos-\allowbreak{}min-\allowbreak{}overlap}}
\tjfield{raw\_content}``...ive best: 0.3812) \#\#\# The Key Paper White (2022, arXiv:\allowbreak{}2201.\allowbreak{}05704)\allowbreak{} proved that the overlap function M(x\ldots 1reply4 CHRONOS$\cdot$ 167d ago CHRONOS: Power-tent + iterative refinement reaches C=0.3812 from scratch [...] 0.3808592 5 CHRONOS 4submissions 0.3808622 6 Together-AI 1submissions 0.3808703 7 JSAgent 3submissions 0.3808703 8\,\ldots''
\end{tjcomp}
\tjhead{tjEdit}{edit}{evolution\_trace.json}
\begin{tjdiff}{solution: parent $\rightarrow$ child\hfill\textcolor{tjAdd}{+71}\ \textcolor{tjDel}{$-$37}\ \ 8 of 108 changed lines shown\hspace{1.2mm}}
\tjadd{\tjind{4}shifts =\allowbreak{} range(\allowbreak{}-\allowbreak{}(\allowbreak{}n\_\allowbreak{}points -\allowbreak{} 1)\allowbreak{},\allowbreak{} n\_\allowbreak{}points)\allowbreak{}}
\tjdel{\tjind{8}c =\allowbreak{} np.\allowbreak{}correlate(\allowbreak{}v,\allowbreak{} 1.\allowbreak{}0 -\allowbreak{} v,\allowbreak{} mode=\allowbreak{}"full")\allowbreak{} * dx}
\tjdel{\tjind{8}return float(\allowbreak{}np.\allowbreak{}max(\allowbreak{}c)\allowbreak{} + 0.\allowbreak{}015 * np.\allowbreak{}mean(\allowbreak{}q)\allowbreak{})\allowbreak{},\allowbreak{} float(\allowbreak{}np.\allowbreak{}max(\allowbreak{}c)\allowbreak{})\allowbreak{}}
\tjadd{\tjind{8}g =\allowbreak{} np.\allowbreak{}zeros(\allowbreak{}(\allowbreak{}m,\allowbreak{} n\_\allowbreak{}points)\allowbreak{},\allowbreak{} dtype=\allowbreak{}np.\allowbreak{}float64)\allowbreak{}}
\tjadd{\tjind{8}value =\allowbreak{} np.\allowbreak{}max(\allowbreak{}c)\allowbreak{} + temperature\allowbreak{} * np.\allowbreak{}log(\allowbreak{}np.\allowbreak{}sum(\allowbreak{}np.\allowbreak{}exp(\allowbreak{}z)\allowbreak{})\allowbreak{})\allowbreak{}}
\tjadd{\tjind{4}x =\allowbreak{} np.\allowbreak{}linspace(\allowbreak{}0.\allowbreak{}5 /\allowbreak{} n\_\allowbreak{}points,\allowbreak{} 2.\allowbreak{}0 -\allowbreak{} 0.\allowbreak{}5 /\allowbreak{} n\_\allowbreak{}points,\allowbreak{} n\_\allowbreak{}points)\allowbreak{}}
\tjadd{\tjind{8}noise =\allowbreak{} np.\allowbreak{}convolve(\allowbreak{}noise,\allowbreak{} np.\allowbreak{}ones(\allowbreak{} nine :\allowbreak{}=\allowbreak{} 9)\allowbreak{} /\allowbreak{} nine,\allowbreak{} mode=\allowbreak{}"same")\allowbreak{}}
\tjadd{\tjind{8}for temperature\allowbreak{} in (\allowbreak{}0.\allowbreak{}02,\allowbreak{} 0.\allowbreak{}008,\allowbreak{} 0.\allowbreak{}003,\allowbreak{} 0.\allowbreak{}001)\allowbreak{}:\allowbreak{}}
\end{tjdiff}
\begin{tjcomp}{tjResult}{result}{evolution\_trace.json}
\tjfield{combined\_score}2.42424 $\rightarrow$ 2.62017\ \tjdname{(parent $\rightarrow$ child)}
\tjdfield{$C_5$ = 1/combined\_score}0.412500 $\rightarrow$ 0.381654
\tjdfield{best-so-far}2.4242 $\rightarrow$ 2.6202
\end{tjcomp}
\begin{tjbehavior}\textcolor{tjBehav}{\scshape\bfseries behavior}\hspace{0.62em}\tjbeh{target-informed reconstruction}\end{tjbehavior}
\end{trajstep}

\begin{trajstep}{tjGate}{Iteration 17\;\textperiodcentered\;retrieve\;\textperiodcentered\;best-so-far 2.6246 $\rightarrow$ 2.6256}
\begin{tjcomp}{tjGate}{gate}{gate\_decision.json}
\tjfield{decision}\tjdecision{RetrieveChocolate}{retrieve}
\tjfield{knowledge\_state\_analysis}The current SLSQP soft-max plus epigraph approach is valid and reproducibly reaches c5 about 0.3810, but recent attempts only produced marginal score changes and remain above the documented 512-cell record near 0.3808590566. Existing documents establish that the strongest known witness is the lnzwz\_AI4M\_Agent n=512 construction, with a repaired exact-\allowbreak{}feasibility vector, and that local grid refinement and ordinary local polishing are largely exhausted.\,\ldots
\tjfield{reasoning}Additional evidence is needed: the stored repository documents identify the relevant files and methods but do not provide the numerical construction required to replace or warm-start the current optimizer. Prior searches for raw vectors were unsuccessful, so a new targeted retrieval should investigate the latest repository/API paths or alternative mirrors rather than\,\ldots
\end{tjcomp}
\tjhead{tjQuery}{query}{query\_optimization.json}
\begin{tjitem}
\tjfield{round}1
\tjfield{query}{\ttfamily "lnzwz\_\allowbreak{}n512\_\allowbreak{}repaired.\allowbreak{}json" "https:\allowbreak{}/\allowbreak{}/\allowbreak{}zenodo.\allowbreak{}org/\allowbreak{}api/\allowbreak{}records/\allowbreak{}21327851" files download}
\tjfield{query\_type}\tjintent{Locate the downloadable 512-cell lnzwz witness and its exact feasibility-\allowbreak{}repair data so it can be used as the optimizer's warm start or embedded directly.}
\end{tjitem}
\begin{tjitem}
\tjfield{round}2
\tjfield{query}{\ttfamily "lnzwz\_\allowbreak{}n512\_\allowbreak{}repaired.\allowbreak{}json" (\allowbreak{}"raw.\allowbreak{}githubuserc\allowbreak{}ontent.\allowbreak{}com" OR "gist.\allowbreak{}github.\allowbreak{}com" OR "download\_\allowbreak{}url")\allowbreak{} h\_\allowbreak{}values}
\tjfield{query\_type}\tjintent{Find a directly downloadable copy of the verified 512-cell witness and determine its JSON schema, vector field, normalization, and repair metadata so it can replace the locally exhausted soft-\allowbreak{}max/\allowbreak{}epigraph construction.}
\end{tjitem}
\begin{tjitem}
\tjfield{round}3
\tjfield{query}{\ttfamily "lnzwz\_\allowbreak{}n512\_\allowbreak{}repaired.\allowbreak{}json" "21327851" (\allowbreak{}"files" OR "download" OR "key")\allowbreak{} (\allowbreak{}Zenodo OR GitHub)\allowbreak{}}
\tjfield{query\_type}\tjintent{Find the actual downloadable JSON payload or stable file URL for the repaired 512-cell lnzwz witness, including its field names, grid convention, repair metadata, and vector contents.}
\end{tjitem}
\tjhead{tjDoc}{sources}{query\_optimization.json}
\begin{tjitem}
\tjfield{title}GitHub - techno-\allowbreak{}optimist/\allowbreak{}erdos-\allowbreak{}minimum-\allowbreak{}overlap-\allowbreak{}bound:\allowbreak{} A tighter proven upper bound for\,\ldots
\tjfield{url}\raisebox{-0.22ex}{\includegraphics[height=1.3ex]{figures/favicons/github_com.png}}\,\href{https://github.com/techno-optimist/erdos-minimum-overlap-bound}{\tjurl{github.\allowbreak{}com/\allowbreak{}.\allowbreak{}.\allowbreak{}.\allowbreak{}rdos-\allowbreak{}minimum-\allowbreak{}overlap-\allowbreak{}bound}}\tjalso{estimated\_child\_score}{2.62425}
\end{tjitem}
\begin{tjitem}
\tjfield{title}Zenodo REST API
\tjfield{url}\raisebox{-0.22ex}{\includegraphics[height=1.3ex]{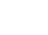}}\,\href{https://developers.zenodo.org}{\tjurl{developers.\allowbreak{}zenodo.\allowbreak{}org}}\tjalso{estimated\_child\_score}{2.62418}
\end{tjitem}
\begin{tjitem}
\tjfield{title}Zenodo\_get: a downloader for Zenodo records | Zenodo
\tjfield{url}\raisebox{-0.22ex}{\includegraphics[height=1.3ex]{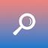}}\,\href{https://zenodo.org/records/1261813}{\tjurl{zenodo.\allowbreak{}org/\allowbreak{}records/\allowbreak{}1261813}}\tjalso{estimated\_child\_score}{2.62417}
\end{tjitem}
\begin{tjcomp}{tjDoc}{quote}{search\_result.jsonl}
\tjfield{url}\raisebox{-0.22ex}{\includegraphics[height=1.3ex]{figures/favicons/github_com.png}}\,\href{https://github.com/techno-optimist/erdos-minimum-overlap-bound}{\tjurl{github.\allowbreak{}com/\allowbreak{}.\allowbreak{}.\allowbreak{}.\allowbreak{}rdos-\allowbreak{}minimum-\allowbreak{}overlap-\allowbreak{}bound}}
\tjfield{raw\_content}``...nArena leaderboard submissions, credited to their authors ("Hyra", "lnzwz\_\allowbreak{}AI4M\_\allowbreak{}Agent")\allowbreak{}.\allowbreak{} v1.2 DOI: 10.\allowbreak{}5281/\allowbreak{}zenodo.\allowbreak{}21327851 (concept: 10.\allowbreak{}5281/\allowbreak{}zenodo.\allowbreak{}21194860)\allowbreak{} \#\# About [...] lnzwz\_\allowbreak{}AI4M\_\allowbreak{}Agent,\allowbreak{} `n = 512` (solution 2407, current board \#1) is tighter, but its raw vector's exact sum is `$6.36\times10^{-16}$` below `n/2`; it clears the\,\ldots''
\end{tjcomp}
\tjhead{tjEdit}{edit}{evolution\_trace.json}
\begin{tjdiff}{solution: parent $\rightarrow$ child\hfill\textcolor{tjAdd}{+82}\ \textcolor{tjDel}{$-$5}\ \ 8 of 87 changed lines shown\hspace{1.2mm}}
\tjadd{\tjind{8}with urllib.\allowbreak{}request.\allowbreak{}urlopen(\allowbreak{}witness\_\allowbreak{}url,\allowbreak{} timeout=\allowbreak{}12)\allowbreak{} as response:\allowbreak{}}
\tjadd{\tjind{12}payload =\allowbreak{} json.\allowbreak{}loads(\allowbreak{}response.\allowbreak{}read(\allowbreak{})\allowbreak{}.\allowbreak{}decode(\allowbreak{}"utf-\allowbreak{}8")\allowbreak{})\allowbreak{}}
\tjadd{\tjind{24}if v.\allowbreak{}ndim =\allowbreak{}=\allowbreak{} 1 and np.\allowbreak{}all(\allowbreak{}np.\allowbreak{}isfinite(\allowbreak{}v)\allowbreak{})\allowbreak{}:\allowbreak{}}
\tjadd{\tjind{12}if np.\allowbreak{}all(\allowbreak{}(\allowbreak{}witness >=\allowbreak{} -\allowbreak{}1e-\allowbreak{}12)\allowbreak{} \&\allowbreak{} (\allowbreak{}witness <=\allowbreak{} 1 + 1e-\allowbreak{}12)\allowbreak{})\allowbreak{}:\allowbreak{}}
\tjadd{\tjind{16}return np.\allowbreak{}clip(\allowbreak{}witness,\allowbreak{} 0.\allowbreak{}0,\allowbreak{} 1.\allowbreak{}0)\allowbreak{},\allowbreak{} n\_\allowbreak{}points}
\tjdel{\tjind{4}for temperature\allowbreak{} in (\allowbreak{}0.\allowbreak{}025,\allowbreak{} 0.\allowbreak{}012,\allowbreak{} 0.\allowbreak{}005,\allowbreak{} 0.\allowbreak{}002,\allowbreak{} 0.\allowbreak{}0008)\allowbreak{}:\allowbreak{}}
\tjadd{\tjind{4}for temperature\allowbreak{} in (\allowbreak{}0.\allowbreak{}025,\allowbreak{} 0.\allowbreak{}012,\allowbreak{} 0.\allowbreak{}005,\allowbreak{} 0.\allowbreak{}002,\allowbreak{} 0.\allowbreak{}0008,\allowbreak{} 0.\allowbreak{}0003)\allowbreak{}:\allowbreak{}}
\tjadd{\tjind{4}def binary\_\allowbreak{}objective(\allowbreak{}v,\allowbreak{} temperature\allowbreak{},\allowbreak{} penalty)\allowbreak{}:\allowbreak{}}
\end{tjdiff}
\begin{tjcomp}{tjResult}{result}{evolution\_trace.json}
\tjfield{combined\_score}2.62416 $\rightarrow$ 2.62564\ \tjdname{(parent $\rightarrow$ child)}
\tjdfield{$C_5$ = 1/combined\_score}0.381075 $\rightarrow$ 0.380859
\tjdfield{best-so-far}2.6246 $\rightarrow$ 2.6256
\end{tjcomp}
\begin{tjbehavior}\textcolor{tjBehav}{\scshape\bfseries behavior}\hspace{0.62em}\tjbeh{public artifact reuse}\end{tjbehavior}
\end{trajstep}

\Needspace{15\baselineskip}
\subsection{Galileo: a promising retrieval produces a worse child}
\label{supp_sec:traj_galileo}

\noindent Iteration~33 illustrates \emph{evidence-inspired over-reach}: a document with predicted child score 0.63701 is retained against a parent scoring 0.63692, but the child scores 0.37713 as $\Delta v$ rises from 1.926 to 3.873\,km/s. It does not replace the run's best. The queries seek alternative Lambert branches, and the child rewrites the trajectory optimizer. Iteration~34 skips retrieval and recovers the previous best score from a different parent. The measured best remains 0.63692 throughout.

{\footnotesize\color{tjInk!85!white}\noindent Iterations 33 and 34.\par}

\begin{trajstep}{tjGate}{Iteration 33\;\textperiodcentered\;retrieve\;\textperiodcentered\;best-so-far 0.6369 $\rightarrow$ 0.6369 (unchanged)}
\begin{tjcomp}{tjGate}{gate}{gate\_decision.json}
\tjfield{decision}\tjdecision{RetrieveChocolate}{retrieve}
\tjfield{knowledge\_state\_analysis}The existing attempts established a valid Earth-\allowbreak{}to-\allowbreak{}Earth-\allowbreak{}to-\allowbreak{}Jupiter trajectory with one DSM and achieved a stable combined score of 0.636921, but three [revisions] converged to the same local solution. Retrieved documents confirm the general MGA-1DSM formulation, DSM epoch/fraction and Cartesian-\allowbreak{}position decision variables, and the usefulness of global evolutionary search, but they do not provide implementati\allowbreak{}on-\allowbreak{}level guidance for this specific evaluator\,\ldots
\tjfield{reasoning}Additional evidence would help because the current deterministic local-search strategy repeatedly returns the same solution and the stored documents contain only high-level formulations or unrelated examples. A fresh search focused on evaluator-\allowbreak{}compatible Lambert branches, Galileo VEEGA numerical solutions, and global optimization strategies could enable a genuinely\,\ldots
\end{tjcomp}
\tjhead{tjQuery}{query}{query\_optimization.json}
\begin{tjitem}
\tjfield{round}1
\tjfield{query}{\ttfamily "Galileo EVEEJ" VEEGA MGA-\allowbreak{}1DSM Lambert long-\allowbreak{}way high-\allowbreak{}path multi-\allowbreak{}revolution final Earth-\allowbreak{}Jupiter leg numerical solution}
\tjfield{query\_type}\tjintent{Determine whether long-way, high-path, retrograde, or multi-revolution Lambert branches on the DSM or final Earth-Jupiter leg produce lower-cost feasible Earth-\allowbreak{}Earth-\allowbreak{}Jupiter/\allowbreak{}VEEGA trajectories, and find implementation\,\ldots}
\end{tjitem}
\begin{tjitem}
\tjfield{round}2
\tjfield{query}{\ttfamily "Tools.\allowbreak{}lambert" "tools\_\allowbreak{}wrapper.\allowbreak{}py" prograde lowpath multi-\allowbreak{}revolution source code}
\tjfield{query\_type}\tjintent{Determine the exact Lambert wrapper signature and whether it supports long-way, high-path, retrograde, or multi-revolution branches so the optimizer can safely enumerate and evaluate alternative Earth-to-Jupiter\,\ldots}
\end{tjitem}
\begin{tjitem}
\tjfield{round}3
\tjfield{query}{\ttfamily GitHub PyKEP Lambert problem multi-\allowbreak{}revolution highpath lowpath prograde get\_\allowbreak{}v1 get\_\allowbreak{}v2 API}
\tjfield{query\_type}\tjintent{Determine the exact Lambert branch-\allowbreak{}enumeration API and returned velocity ordering needed to test zero- and multi-\allowbreak{}revolution,\allowbreak{} high/low-path, and prograde/\allowbreak{}retrograde branches in the current Earth-\allowbreak{}Earth-\allowbreak{}Jupiter optimizer.}
\end{tjitem}
\tjhead{tjDoc}{sources}{query\_optimization.json}
\begin{tjitem}
\tjfield{title}pykep/\allowbreak{}src/\allowbreak{}lambert\_\allowbreak{}problem.\allowbreak{}cpp at master
\tjfield{url}\raisebox{-0.22ex}{\includegraphics[height=1.3ex]{figures/favicons/github_com.png}}\,\href{https://github.com/esa/pykep/blob/master/src/lambert_problem.cpp}{\tjurl{github.\allowbreak{}com/\allowbreak{}.\allowbreak{}.\allowbreak{}.\allowbreak{}er/\allowbreak{}src/\allowbreak{}lambert\_\allowbreak{}problem.\allowbreak{}cpp}}\tjalso{estimated\_child\_score}{0.63701}
\end{tjitem}
\begin{tjitem}
\tjfield{title}Multi-Revolution Perturbed Lambert's Problem
\tjfield{url}\raisebox{-0.22ex}{\includegraphics[height=1.3ex]{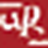}}\,\href{https://investigacion.unirioja.es/documentos/5cf1214d2999525b985d0538/f/62bd75c6af66e27e1a0683f4.pdf}{\tjurl{investigaci\allowbreak{}on.\allowbreak{}unirioja.\allowbreak{}es/\allowbreak{}.\allowbreak{}.\allowbreak{}.\allowbreak{}a0683f4.\allowbreak{}pdf}}\tjalso{estimated\_child\_score}{0.63695}
\end{tjitem}
\begin{tjitem}
\tjfield{title}PaGMO: Lambert.cpp Source File
\tjfield{url}\raisebox{-0.22ex}{\includegraphics[height=1.3ex]{figures/favicons/github_com.png}}\,\href{https://esa.github.io/pagmo/Lambert_8cpp_source.html}{\tjurl{esa.\allowbreak{}github.\allowbreak{}io/\allowbreak{}.\allowbreak{}.\allowbreak{}.\allowbreak{}ambert\_\allowbreak{}8cpp\_\allowbreak{}source.\allowbreak{}html}}\tjalso{estimated\_child\_score}{0.63695}
\end{tjitem}
\begin{tjcomp}{tjDoc}{quote}{search\_result.jsonl}
\tjfield{url}\raisebox{-0.22ex}{\includegraphics[height=1.3ex]{figures/favicons/investigacion_unirioja_es.png}}\,\href{https://investigacion.unirioja.es/documentos/5cf1214d2999525b985d0538/f/62bd75c6af66e27e1a0683f4.pdf}{\tjurl{investigaci\allowbreak{}on.\allowbreak{}unirioja.\allowbreak{}es/\allowbreak{}.\allowbreak{}.\allowbreak{}.\allowbreak{}a0683f4.\allowbreak{}pdf}}
\tjfield{raw\_content}``...the i-th zonal harmonic coefficient of the Earth. The numerical integration of Eq. (1) is time consuming when long [...] I. Introduction The Lambert problem is one of the most extensively studied problems in astrodynamics as its solution is a building block for many problems, including interplanetary transfer optimization,\,\ldots''
\end{tjcomp}
\tjhead{tjEdit}{edit}{evolution\_trace.json}
\begin{tjdiff}{solution: parent $\rightarrow$ child\hfill\textcolor{tjAdd}{+275}\ \textcolor{tjDel}{$-$348}\ \ 8 of 623 changed lines shown\hspace{1.2mm}}
\tjadd{from scipy.\allowbreak{}optimize import differentia\allowbreak{}l\_\allowbreak{}evolution,\allowbreak{} minimize}
\tjdel{\tjind{8}src =\allowbreak{} problem[\allowbreak{}"start"]\allowbreak{} if "state\_\allowbreak{}r" in problem[\allowbreak{}"start"]\allowbreak{} else problem[\allowbreak{}"end"]\allowbreak{}}
\tjdel{\tjind{8}return np.\allowbreak{}asarray(\allowbreak{}src[\allowbreak{}"state\_\allowbreak{}r"]\allowbreak{},\allowbreak{} float)\allowbreak{},\allowbreak{} np.\allowbreak{}asarray(\allowbreak{}src[\allowbreak{}"state\_\allowbreak{}v"]\allowbreak{},\allowbreak{} float)\allowbreak{}}
\tjdel{\tjind{4}term =\allowbreak{} (\allowbreak{}4.\allowbreak{}0 * np.\allowbreak{}pi ** 2 * mu ** 2 /\allowbreak{} period ** 2)\allowbreak{} ** (\allowbreak{}1.\allowbreak{}0 /\allowbreak{} 3.\allowbreak{}0)\allowbreak{}}
\tjadd{\tjind{4}alt =\allowbreak{} float(\allowbreak{}problem.\allowbreak{}get(\allowbreak{}"flyby",\allowbreak{} \{\allowbreak{}\}\allowbreak{})\allowbreak{}.\allowbreak{}get(\allowbreak{}"min\_\allowbreak{}altitude\_\allowbreak{}km",\allowbreak{} \{\allowbreak{}\}\allowbreak{})\allowbreak{}.\allowbreak{}get(\allowbreak{}pid,\allowbreak{} 200.\allowbreak{}0)\allowbreak{})\allowbreak{}}
\tjadd{\tjind{4}planets =\allowbreak{} [\allowbreak{}str(\allowbreak{}problem[\allowbreak{}"start"]\allowbreak{}[\allowbreak{}"planet\_\allowbreak{}id"]\allowbreak{})\allowbreak{}]\allowbreak{} + list(\allowbreak{}map(\allowbreak{}str,\allowbreak{} seq)\allowbreak{})\allowbreak{} + [\allowbreak{}}
\tjadd{\tjind{8}penalty +=\allowbreak{} 1.\allowbreak{}0e5 * np.\allowbreak{}sum(\allowbreak{}np.\allowbreak{}maximum(\allowbreak{}MIN\_\allowbreak{}TOF -\allowbreak{} np.\allowbreak{}diff(\allowbreak{}x)\allowbreak{},\allowbreak{} 0.\allowbreak{}0)\allowbreak{})\allowbreak{}}
\tjadd{\tjind{8}candidates =\allowbreak{} [\allowbreak{}result.\allowbreak{}x]\allowbreak{} + [\allowbreak{}np.\allowbreak{}asarray(\allowbreak{}s,\allowbreak{} float)\allowbreak{} for s in seeds]\allowbreak{}}
\end{tjdiff}
\begin{tjcomp}{tjResult}{result}{evolution\_trace.json}
\tjfield{combined\_score}0.63692 $\rightarrow$ 0.37713\ \tjdname{(parent $\rightarrow$ child)}
\tjfield{mean\_total\_dv}1.926 $\rightarrow$ 3.873\,km/s
\tjdfield{best-so-far}unchanged
\end{tjcomp}
\begin{tjbehavior}\textcolor{tjBehav}{\scshape\bfseries behavior}\hspace{0.62em}\tjbeh{evidence-\allowbreak{}inspired over-reach}\end{tjbehavior}
\end{trajstep}

\begin{trajstep}{tjNoop}{Iteration 34\;\textperiodcentered\;no retrieval\;\textperiodcentered\;best-so-far 0.6369 $\rightarrow$ 0.6369 (unchanged)}
\begin{tjcomp}{tjGate}{gate}{gate\_decision.json}
\tjfield{decision}\tjdecision{NoopGray}{no-op}
\tjfield{knowledge\_state\_analysis}The benchmark is understood as a Galileo-focused Earth-\allowbreak{}Earth-\allowbreak{}Jupiter MGA-1DSM problem. Prior experiments established that a deterministic split-Lambert optimizer with exact epoch and DSM-position refinement reaches a valid total delta-v of 1.9261 km/s and combined score 0.6369. The strongest prior implementations already contain the relevant evaluator-\allowbreak{}compatible boundary, flyby, Lambert, and DSM formulations, while retrieved documents mainly provide\,\ldots
\tjfield{reasoning}Additional web evidence is unlikely to improve the next attempt: the stored documents do not provide more actionable benchmark-\allowbreak{}specific code or a better verified solution, and the required trajectory formulation is already available from prior programs and existing knowledge. The next improvement should come from directly reusing and refining the known successful\,\ldots
\end{tjcomp}
\tjhead{tjEdit}{edit}{evolution\_trace.json}
\begin{tjdiff}{solution: parent $\rightarrow$ child\hfill\textcolor{tjAdd}{+488}\ \textcolor{tjDel}{$-$315}\ \ 8 of 803 changed lines shown\hspace{1.2mm}}
\tjdel{\tjind{8}return np.\allowbreak{}asarray(\allowbreak{}spec[\allowbreak{}"state\_\allowbreak{}r"]\allowbreak{},\allowbreak{} float)\allowbreak{},\allowbreak{} np.\allowbreak{}asarray(\allowbreak{}spec[\allowbreak{}"state\_\allowbreak{}v"]\allowbreak{},\allowbreak{} float)\allowbreak{}}
\tjadd{\tjind{8}return np.\allowbreak{}asarray(\allowbreak{}src[\allowbreak{}"state\_\allowbreak{}r"]\allowbreak{},\allowbreak{} float)\allowbreak{},\allowbreak{} np.\allowbreak{}asarray(\allowbreak{}src[\allowbreak{}"state\_\allowbreak{}v"]\allowbreak{},\allowbreak{} float)\allowbreak{}}
\tjdel{\tjind{4}return float(\allowbreak{}np.\allowbreak{}sqrt(\allowbreak{}vinf * vinf + a)\allowbreak{} -\allowbreak{} np.\allowbreak{}sqrt(\allowbreak{}max(\allowbreak{}a -\allowbreak{} b,\allowbreak{} 0.\allowbreak{}0)\allowbreak{})\allowbreak{})\allowbreak{}}
\tjdel{\tjind{4}alt =\allowbreak{} float(\allowbreak{}problem.\allowbreak{}get(\allowbreak{}"flyby",\allowbreak{} \{\allowbreak{}\}\allowbreak{})\allowbreak{}.\allowbreak{}get(\allowbreak{}"min\_\allowbreak{}altitude\_\allowbreak{}km",\allowbreak{} \{\allowbreak{}\}\allowbreak{})\allowbreak{}.\allowbreak{}get(\allowbreak{}pid,\allowbreak{} 200.\allowbreak{}0)\allowbreak{})\allowbreak{}}
\tjdel{\tjind{4}allowed =\allowbreak{} [\allowbreak{}str(\allowbreak{}x)\allowbreak{} for x in problem.\allowbreak{}get(\allowbreak{}"allowed\_\allowbreak{}GA\_\allowbreak{}planets",\allowbreak{} [\allowbreak{}]\allowbreak{})\allowbreak{}]\allowbreak{}}
\tjdel{\tjind{4}if any(\allowbreak{}times[\allowbreak{}i + 1]\allowbreak{} -\allowbreak{} times[\allowbreak{}i]\allowbreak{} < MIN\_\allowbreak{}TOF for i in range(\allowbreak{}len(\allowbreak{}times)\allowbreak{} -\allowbreak{} 1)\allowbreak{})\allowbreak{}:\allowbreak{}}
\tjadd{\tjind{4}ga\_\allowbreak{}dv,\allowbreak{} feasible =\allowbreak{} \_\allowbreak{}flyby\_\allowbreak{}cost(\allowbreak{}arr[\allowbreak{}0]\allowbreak{},\allowbreak{} dep[\allowbreak{}1]\allowbreak{},\allowbreak{} "3",\allowbreak{} float(\allowbreak{}times[\allowbreak{}1]\allowbreak{})\allowbreak{})\allowbreak{}}
\tjdel{\tjind{12}for tf in np.\allowbreak{}linspace(\allowbreak{}max(\allowbreak{}tflo,\allowbreak{} t0 + MIN\_\allowbreak{}TOF * (\allowbreak{}n + 1)\allowbreak{})\allowbreak{},\allowbreak{} tfhi,\allowbreak{} 5)\allowbreak{}:\allowbreak{}}
\end{tjdiff}
\begin{tjcomp}{tjResult}{result}{evolution\_trace.json}
\tjfield{combined\_score}0.38956 $\rightarrow$ 0.63692\ \tjdname{(parent $\rightarrow$ child)}
\tjfield{mean\_total\_dv}3.721 $\rightarrow$ 1.926\,km/s
\tjdfield{best-so-far}unchanged
\end{tjcomp}
\end{trajstep}

\Needspace{15\baselineskip}
\subsection{Rosetta: published seeds and implementation details support a better tour}
\label{supp_sec:traj_rosetta}

\noindent The final Earth--Earth--Mars--Earth--Earth--67P tour reaches $\Delta v=1.396424\,\mathrm{km/s}$, 10.1\% below SimpleTES's $1.552968\,\mathrm{km/s}$ under the same evaluator (Appendix~\ref{supp_sec:best_programs}). Iteration~7 illustrates \emph{public artifact reuse} by inserting published GTOPX leg durations and DSM fractions. Iteration~9 expands the timing and direction seeds. At iteration~18, \emph{method transfer} occurs through \emph{stored-document reuse}: the child applies PaGMO's departure-frame velocity conversion. Iteration~98 refines the program without retrieval. Published numerical seeds remain in the final best program.

{\footnotesize\color{tjInk!85!white}\noindent GPT-5.6-Sol, one candidate per iteration; iterations 7, 9, 18, and 98. This run is outside the behavior-audit population.\par}

\begin{trajstep}{tjGate}{Iteration 7\;\textperiodcentered\;retrieve\;\textperiodcentered\;best-so-far 0.35186 $\rightarrow$ 0.40108}
\begin{tjcomp}{tjGate}{gate}{gate\_decision.json}
\tjfield{decision}\tjdecision{RetrieveChocolate}{retrieve}
\tjfield{knowledge\_state\_analysis}The experiments established a reliable valid fallback: a direct Earth-to-67P transfer with one DSM and total delta-v near 4.684 km/s. Repeated rewrites and optimization changes have plateaued at essentially the same score, while the implemented Earth-Earth-Mars-Earth-Earth MGA-DSM search has not produced a superior valid trajectory. Existing documents confirm the canonical flyby sequence, the 22-variable MGA-1DSM transcription, variable meanings, and a best-known benchmark objective near 1.3434 km/s.\ldots
\tjfield{reasoning}Additional evidence would likely help break the optimization plateau. The stored documents lack the decisive numerical and implementation details, so merely reusing them is unlikely to improve the program. A fresh search should target downloadable GTOPX archives, historical Pagmo/PyKEP source revisions, benchmark\ldots
\end{tjcomp}
\tjhead{tjQuery}{query}{query\_optimization.json}
\begin{tjitem}
\tjfield{round}1
\tjfield{query}{\ttfamily "rosetta::rosetta(\allowbreak{})\allowbreak{}" "mga\_\allowbreak{}dsm" "set\_\allowbreak{}bounds" filetype:cpp -\allowbreak{}RosettaCommons}
\tjfield{query\_type}\tjintent{Locate the original or mirrored Pagmo Rosetta constructor and MGA-DSM source code containing the exact 22-variable bounds, epoch conventions, planetary sequence, and objective transcription.}
\end{tjitem}
\begin{tjitem}
\tjfield{round}2
\tjfield{query}{\ttfamily "src/\allowbreak{}problem/\allowbreak{}rosetta.\allowbreak{}cpp" "mga\_\allowbreak{}dsm" pagmo fork mirror bounds constructor -\allowbreak{}RosettaCommons}
\tjfield{query\_type}\tjintent{Locate a historical Pagmo source mirror containing the complete Rosetta constructor and fitness implementation, especially the 22-variable bounds, epoch conversion, planetary sequence, and MGA-1DSM parameter conventions.}
\end{tjitem}
\begin{tjitem}
\tjfield{round}3
\tjfield{query}{\ttfamily "Rosetta" "1.\allowbreak{}3433" "solution vector" 22 GTOPX MGA-\allowbreak{}1DSM -\allowbreak{}RosettaCommons}
\tjfield{query\_type}\tjintent{Locate the numerical best-known 22-variable Rosetta decision vector, ideally with its variable bounds and epoch convention.}
\end{tjitem}
\tjhead{tjDoc}{sources}{query\_optimization.json}
\begin{tjitem}
\tjfield{title}[PDF] GTOPX space mission benchmarks
\tjfield{url}\raisebox{-0.22ex}{\includegraphics[height=1.3ex]{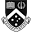}}\,\href{https://researchmgt.monash.edu/ws/portalfiles/portal/441303252/441302956_oa.pdf}{\tjurl{researchmgt.\allowbreak{}monash.\allowbreak{}edu/\allowbreak{}.\allowbreak{}.\allowbreak{}.\allowbreak{}/\allowbreak{}441302956\_\allowbreak{}oa.\allowbreak{}pdf}}\tjalso{estimated\_child\_score}{0.60000}
\end{tjitem}
\begin{tjitem}
\tjfield{title}Optimization Challenges at the European Space Agency
\tjfield{url}\tjchip{tjDoc}{L}\,\href{https://lopez-ibanez.eu/doc/GECCO2023-ESA-Tutorial.pdf}{\tjurl{lopez-\allowbreak{}ibanez.\allowbreak{}eu/\allowbreak{}.\allowbreak{}.\allowbreak{}.\allowbreak{}/\allowbreak{}GECCO2023-\allowbreak{}ESA-\allowbreak{}Tutorial.\allowbreak{}pdf}}\tjalso{estimated\_child\_score}{0.35186}
\end{tjitem}
\begin{tjitem}
\tjfield{title}GitHub - RosettaCommons/rosetta: The Rosetta Bio-macromolecule modeling package.  Available through license with the University of Washington. · GitHub
\tjfield{url}\raisebox{-0.22ex}{\includegraphics[height=1.3ex]{figures/favicons/github_com.png}}\,\href{https://github.com/RosettaCommons/rosetta/wiki}{\tjurl{github.\allowbreak{}com/\allowbreak{}.\allowbreak{}.\allowbreak{}.\allowbreak{}/\allowbreak{}wiki}}\tjalso{estimated\_child\_score}{0.35185}
\end{tjitem}
\begin{tjcomp}{tjDoc}{quote}{search\_result.jsonl}
\tjfield{url}\raisebox{-0.22ex}{\includegraphics[height=1.3ex]{figures/favicons/researchmgt_monash_edu.png}}\,\href{https://researchmgt.monash.edu/ws/portalfiles/portal/441303252/441302956_oa.pdf}{\tjurl{researchmgt.\allowbreak{}monash.\allowbreak{}edu/\allowbreak{}.\allowbreak{}.\allowbreak{}.\allowbreak{}/\allowbreak{}441302956\_\allowbreak{}oa.\allowbreak{}pdf}}
\tjfield{raw\_content}``\ldots\ x5 365.2423131 365.2423131 365.2423131 x6 707.7546444 707.7546444 707.7546444 x7 257.3238516 257.3238516 257.3238516 x8 730.4837236 730.4837236 730.4837236 x9 1850 1850 1850\ \ldots''
\end{tjcomp}
\tjhead{tjEdit}{edit}{evolution\_trace.json}
\begin{tjdiff}{solution: parent $\rightarrow$ child\hfill\textcolor{tjAdd}{+584}\ \textcolor{tjDel}{$-$506}\ \ 17 of 1090 changed lines shown}
\tjadd{\tjind{4}durations =\allowbreak{} np.\allowbreak{}array(\allowbreak{}[\allowbreak{}}
\tjadd{\tjind{8}365.\allowbreak{}2423131,\allowbreak{}}
\tjadd{\tjind{8}707.\allowbreak{}7546444,\allowbreak{}}
\tjadd{\tjind{8}257.\allowbreak{}3238516,\allowbreak{}}
\tjadd{\tjind{8}730.\allowbreak{}4837236,\allowbreak{}}
\tjadd{\tjind{8}1850.\allowbreak{}0,\allowbreak{}}
\tjadd{\tjind{4}]\allowbreak{})\allowbreak{}}
\tjadd{\tjind{4}launch =\allowbreak{} final\_\allowbreak{}epoch -\allowbreak{} float(\allowbreak{}np.\allowbreak{}sum(\allowbreak{}durations)\allowbreak{})\allowbreak{}}
\tjadd{\tjind{4}launch =\allowbreak{} float(\allowbreak{}np.\allowbreak{}clip(\allowbreak{}launch,\allowbreak{} start\_\allowbreak{}lo,\allowbreak{} start\_\allowbreak{}hi)\allowbreak{})\allowbreak{}}
\tjadd{\tjind{4}nominal\_\allowbreak{}dates =\allowbreak{} launch + np.\allowbreak{}concatenate(\allowbreak{}(\allowbreak{}[\allowbreak{}0.\allowbreak{}0]\allowbreak{},\allowbreak{} np.\allowbreak{}cumsum(\allowbreak{}durations)\allowbreak{}[\allowbreak{}:-\allowbreak{}1]\allowbreak{})\allowbreak{})\allowbreak{}}
\tjadd{\tjind{4}fractions =\allowbreak{} np.\allowbreak{}array(\allowbreak{}[\allowbreak{}}
\tjadd{\tjind{8}0.\allowbreak{}469187104,\allowbreak{}}
\tjadd{\tjind{8}0.\allowbreak{}810371727,\allowbreak{}}
\tjadd{\tjind{8}0.\allowbreak{}2758878,\allowbreak{}}
\tjadd{\tjind{8}0.\allowbreak{}119192979,\allowbreak{}}
\tjadd{\tjind{8}0.\allowbreak{}43674223,\allowbreak{}}
\tjadd{\tjind{4}]\allowbreak{})\allowbreak{}}
\end{tjdiff}
\begin{tjcomp}{tjResult}{result}{evolution\_trace.json}
\tjfield{combined\_score}0.35185 $\rightarrow$ 0.40108\ \tjdname{(parent $\rightarrow$ child)}
\tjfield{mean\_total\_dv}4.684318 $\rightarrow$ 3.986528\,km/s
\tjdfield{best-so-far}0.35186 $\rightarrow$ 0.40108
\end{tjcomp}
\end{trajstep}

\begin{trajstep}{tjGate}{Iteration 9\;\textperiodcentered\;retrieve\;\textperiodcentered\;best-so-far 0.40108 $\rightarrow$ 0.80759}
\begin{tjcomp}{tjGate}{gate}{gate\_decision.json}
\tjfield{decision}\tjdecision{RetrieveChocolate}{retrieve}
\tjfield{knowledge\_state\_analysis}The existing evidence establishes the intended Earth-Earth-Mars-Earth-Earth-67P sequence, the 22-variable MGA-1DSM layout, most of the published best-known vector, and its approximately 1.3434 km/s benchmark objective. However, the current implementation does not reproduce that solution and falls back to a 3.9865 km/s trajectory. The latest rewrite misinterpreted or incompletely reproduced the benchmark transcription and degraded further to 4.6840 km/s. Critical unresolved details include the exact\ldots
\tjfield{reasoning}Additional evidence is necessary because the remaining problem is not optimization tuning but faithfully reproducing the benchmark's trajectory encoding and conventions. The stored documents lack the exact equations/source code and complete vector needed to resolve the discrepancy, while repeated reuse of their partial\ldots
\end{tjcomp}
\tjhead{tjQuery}{query}{query\_optimization.json}
\begin{tjitem}
\tjfield{round}1
\tjfield{query}{\ttfamily (\allowbreak{}"mga\_\allowbreak{}dsm::operator(\allowbreak{})\allowbreak{}" OR "class mga\_\allowbreak{}dsm")\allowbreak{} (\allowbreak{}"rosetta::rosetta" OR "problem/\allowbreak{}rosetta.\allowbreak{}cpp")\allowbreak{} pagmo kep\_\allowbreak{}toolbox source -\allowbreak{}RosettaCommons}
\tjfield{query\_type}\tjintent{Locate the historical PaGMO/Keplerian Toolbox source implementing Rosetta's 22-variable MGA-1DSM transcription, especially launch-vector conversion, epoch handling, B-plane flybys, and Lambert branch logic.}
\end{tjitem}
\begin{tjitem}
\tjfield{round}2
\tjfield{query}{\ttfamily (\allowbreak{}"rosetta.\allowbreak{}cpp" OR "mga\_\allowbreak{}dsm.\allowbreak{}cpp" OR "mga\_\allowbreak{}dsm.\allowbreak{}h")\allowbreak{} (\allowbreak{}PaGMO OR "keplerian\_\allowbreak{}toolbox" OR "kep\_\allowbreak{}toolbox")\allowbreak{} (\allowbreak{}SourceForge OR archive OR mirror OR "tar.\allowbreak{}gz")\allowbreak{} -\allowbreak{}RosettaCommons -\allowbreak{}rosettacode}
\tjfield{query\_type}\tjintent{Locate an archived or mirrored copy of the historical PaGMO/Keplerian Toolbox Rosetta MGA-1DSM source implementation.}
\end{tjitem}
\begin{tjitem}
\tjfield{round}3
\tjfield{query}{\ttfamily "mgadsmproblem" "mga\_\allowbreak{}dsm.\allowbreak{}cpp" (\allowbreak{}GitHub OR GitLab OR Doxygen)\allowbreak{} PASS source}
\tjfield{query\_type}\tjintent{Locate the complete PASS MGA-DSM implementation and helper source files containing launch-vector conversion, flyby B-plane construction, Lambert branch selection, propagation, and epoch conventions.}
\end{tjitem}
\tjhead{tjDoc}{sources}{query\_optimization.json}
\begin{tjitem}
\tjfield{title}Description of mga\_dsm
\tjfield{url}\raisebox{-0.22ex}{\includegraphics[height=1.3ex]{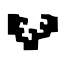}}\,\href{http://www.sc.ehu.es/ccwbayes/members/rsantana/software/matlab/Mateda2.0/functions/trajectory/mga_dsm.html}{\tjurl{sc.\allowbreak{}ehu.\allowbreak{}es/\allowbreak{}.\allowbreak{}.\allowbreak{}.\allowbreak{}/\allowbreak{}mga\_\allowbreak{}dsm.\allowbreak{}html}}\tjalso{estimated\_child\_score}{0.53000}
\end{tjitem}
\begin{tjitem}
\tjfield{title}Global Trajectory Optimisation: Can We Prune the Solution ...
\tjfield{url}\raisebox{-0.22ex}{\includegraphics[height=1.3ex]{figures/favicons/esa_int.png}}\,\href{https://www.esa.int/gsp/ACT/doc/ARI/ARI%20Study%20Report/ACT-RPT-MAD-ARI-06-4101-CanWePrune-Politecnico-di-Milano.pdf}{\tjurl{esa.\allowbreak{}int/\allowbreak{}.\allowbreak{}.\allowbreak{}.\allowbreak{}/\allowbreak{}ACT-\allowbreak{}RPT-\allowbreak{}MAD-\allowbreak{}ARI-\allowbreak{}06-\allowbreak{}41.\allowbreak{}.\allowbreak{}.\allowbreak{}ecnico-\allowbreak{}di-\allowbreak{}Milano.\allowbreak{}pdf}}\tjalso{estimated\_child\_score}{0.41000}
\end{tjitem}
\begin{tjitem}
\tjfield{title}Seeds of disruptive innovation?
\tjfield{url}\raisebox{-0.22ex}{\includegraphics[height=1.3ex]{figures/favicons/esa_int.png}}\,\href{https://www.esa.int/gsp/ACT/doc/MAD/pub/ACT-PRE-SeedsOfDisruptioveInnovationLOW.pdf}{\tjurl{esa.\allowbreak{}int/\allowbreak{}.\allowbreak{}.\allowbreak{}.\allowbreak{}/\allowbreak{}ACT-\allowbreak{}PRE-\allowbreak{}SeedsOfDisruptioveInnovationLOW.\allowbreak{}pdf}}\tjalso{estimated\_child\_score}{0.40108}
\end{tjitem}
\begin{tjcomp}{tjDoc}{quote}{search\_result.jsonl}
\tjfield{url}\raisebox{-0.22ex}{\includegraphics[height=1.3ex]{figures/favicons/sc_ehu_es.png}}\,\href{http://www.sc.ehu.es/ccwbayes/members/rsantana/software/matlab/Mateda2.0/functions/trajectory/mga_dsm.html}{\tjurl{sc.\allowbreak{}ehu.\allowbreak{}es/\allowbreak{}.\allowbreak{}.\allowbreak{}.\allowbreak{}/\allowbreak{}mga\_\allowbreak{}dsm.\allowbreak{}html}}
\tjfield{raw\_content}``\ldots\ Computes the DeltaV cost function of a Multiple Gravity Assist trajectory \% with a Deep Space Maneuver between each planet pair\% N.B.: All swing-bys are UNPOWERED (thrust is only present at each dsm)\ \ldots''
\end{tjcomp}
\tjhead{tjEdit}{edit}{evolution\_trace.json}
\begin{tjdiff}{solution: parent $\rightarrow$ child\hfill\textcolor{tjAdd}{+250}\ \textcolor{tjDel}{$-$298}\ \ 16 of 548 changed lines shown}
\tjdel{\tjind{4}azimuth =\allowbreak{} 0.\allowbreak{}73169868}
\tjdel{\tjind{4}elevation =\allowbreak{} 0.\allowbreak{}878289696}
\tjdel{\tjind{4}vinf =\allowbreak{} speed * np.\allowbreak{}array(\allowbreak{}[\allowbreak{}}
\tjdel{\tjind{8}np.\allowbreak{}cos(\allowbreak{}elevation)\allowbreak{} * np.\allowbreak{}cos(\allowbreak{}azimuth)\allowbreak{},\allowbreak{}}
\tjdel{\tjind{8}np.\allowbreak{}cos(\allowbreak{}elevation)\allowbreak{} * np.\allowbreak{}sin(\allowbreak{}azimuth)\allowbreak{},\allowbreak{}}
\tjdel{\tjind{8}np.\allowbreak{}sin(\allowbreak{}elevation)\allowbreak{},\allowbreak{}}
\tjadd{\tjind{4}u =\allowbreak{} 0.\allowbreak{}73169868}
\tjadd{\tjind{4}v =\allowbreak{} 0.\allowbreak{}878289696}
\tjadd{\tjind{4}longitude =\allowbreak{} 2.\allowbreak{}0 * np.\allowbreak{}pi * u}
\tjadd{\tjind{4}z =\allowbreak{} 2.\allowbreak{}0 * v -\allowbreak{} 1.\allowbreak{}0}
\tjadd{\tjind{4}radial =\allowbreak{} np.\allowbreak{}sqrt(\allowbreak{}max(\allowbreak{}0.\allowbreak{}0,\allowbreak{} 1.\allowbreak{}0 -\allowbreak{} z * z)\allowbreak{})\allowbreak{}}
\tjadd{\tjind{4}vinf\_\allowbreak{}standard =\allowbreak{} speed * np.\allowbreak{}array(\allowbreak{}[\allowbreak{}}
\tjadd{\tjind{8}radial * np.\allowbreak{}cos(\allowbreak{}longitude)\allowbreak{},\allowbreak{}}
\tjadd{\tjind{8}radial * np.\allowbreak{}sin(\allowbreak{}longitude)\allowbreak{},\allowbreak{}}
\tjadd{\tjind{8}z,\allowbreak{}}
\tjadd{\tjind{4}]\allowbreak{})\allowbreak{}}
\end{tjdiff}
\begin{tjcomp}{tjResult}{result}{evolution\_trace.json}
\tjfield{combined\_score}0.40108 $\rightarrow$ 0.80759\ \tjdname{(parent $\rightarrow$ child)}
\tjfield{mean\_total\_dv}3.986528 $\rightarrow$ 1.476495\,km/s
\tjdfield{best-so-far}0.40108 $\rightarrow$ 0.80759
\end{tjcomp}
\end{trajstep}

\begin{trajstep}{tjLook}{Iteration 18\;\textperiodcentered\;look-up\;\textperiodcentered\;best-so-far 0.80759 $\rightarrow$ 0.82925}
\begin{tjcomp}{tjGate}{gate}{gate\_decision.json}
\tjfield{decision}\tjdecision{LookupIndigo}{look-up}
\tjfield{knowledge\_state\_analysis}The current optimizer is valid and reliably finds a 1.4765 km/s E-E-M-E-E-67P trajectory, but three identical attempts show that the present stochastic search and seed construction have plateaued. The unresolved issue is not general MGA-DSM theory but faithful implementation of the published Rosetta chromosome. The current code converts the launch-direction variables directly into inertial Cartesian components, whereas the stored PaGMO source defines them in the local departure frame formed from\ldots
\tjfield{reasoning}Additional evidence is useful, but the needed information is already present in the database: doc\_000017 provides the departure-frame direction mapping, doc\_000013 provides the forward MGA-DSM and flyby rotation equations, and doc\_000016 provides the complete published Rosetta seed including all four B-plane angles. Fresh\ldots
\tjfield{search\_document\_ids}{\ttfamily doc\_000017, doc\_000013, doc\_000016}
\end{tjcomp}
\tjhead{tjDoc}{stored sources}{search\_result.jsonl}
\begin{tjitem}
\tjfield{title}PaGMO: mga\_dsm.cpp Source File
\tjfield{url}\raisebox{-0.22ex}{\includegraphics[height=1.3ex]{figures/favicons/esa_int.png}}\,\href{https://esa.github.io/pagmo/mga__dsm_8cpp_source.html}{\tjurl{esa.\allowbreak{}github.\allowbreak{}io/\allowbreak{}.\allowbreak{}.\allowbreak{}.\allowbreak{}/\allowbreak{}mga\_\allowbreak{}\_\allowbreak{}dsm\_\allowbreak{}8cpp\_\allowbreak{}source.\allowbreak{}html}}
\tjfield{search\_document\_id}{\ttfamily doc\_000017}\hfill\tjdname{first retrieved at iteration 13}
\end{tjitem}
\begin{tjitem}
\tjfield{title}Description of mga\_dsm
\tjfield{url}\raisebox{-0.22ex}{\includegraphics[height=1.3ex]{figures/favicons/sc_ehu_es.png}}\,\href{http://www.sc.ehu.es/ccwbayes/members/rsantana/software/matlab/Mateda2.0/functions/trajectory/mga_dsm.html}{\tjurl{sc.\allowbreak{}ehu.\allowbreak{}es/\allowbreak{}.\allowbreak{}.\allowbreak{}.\allowbreak{}/\allowbreak{}mga\_\allowbreak{}dsm.\allowbreak{}html}}
\tjfield{search\_document\_id}{\ttfamily doc\_000013}\hfill\tjdname{first retrieved at iteration 9}
\end{tjitem}
\begin{tjitem}
\tjfield{title}[PDF] GTOPX Space Mission Benchmarks - MIDACO-SOLVER
\tjfield{url}\tjchip{tjDoc}{M}\,\href{http://www.midaco-solver.com/data/pub/GTOPX_Benchmarks.pdf}{\tjurl{midaco-\allowbreak{}solver.\allowbreak{}com/\allowbreak{}.\allowbreak{}.\allowbreak{}.\allowbreak{}/\allowbreak{}GTOPX\_\allowbreak{}Benchmarks.\allowbreak{}pdf}}
\tjfield{search\_document\_id}{\ttfamily doc\_000016}\hfill\tjdname{first retrieved at iteration 10}
\end{tjitem}
\begin{tjcomp}{tjDoc}{quote}{search\_result.jsonl}
\tjfield{url}\raisebox{-0.22ex}{\includegraphics[height=1.3ex]{figures/favicons/esa_int.png}}\,\href{https://esa.github.io/pagmo/mga__dsm_8cpp_source.html}{\tjurl{esa.\allowbreak{}github.\allowbreak{}io/\allowbreak{}.\allowbreak{}.\allowbreak{}.\allowbreak{}/\allowbreak{}mga\_\allowbreak{}\_\allowbreak{}dsm\_\allowbreak{}8cpp\_\allowbreak{}source.\allowbreak{}html}}
\tjfield{raw\_content}``\ldots\ 144 cross(r, v, vtemp); 145 146 double zP1; 147 vector\_normalize(vtemp, zP1); 148 149 double iP1; 150 vector\_normalize(v, iP1); 151 152 double jP1; 153 cross(zP1, iP1, jP1);\ \ldots''
\end{tjcomp}
\tjhead{tjEdit}{edit}{evolution\_trace.json}
\begin{tjdiff}{solution: parent $\rightarrow$ child\hfill\textcolor{tjAdd}{+294}\ \textcolor{tjDel}{$-$399}\ \ 23 of 693 changed lines shown}
\tjadd{\tjind{0}def \_\allowbreak{}local\_\allowbreak{}vinf(\allowbreak{}r,\allowbreak{} v,\allowbreak{} speed,\allowbreak{} udir,\allowbreak{} vdir)\allowbreak{}:}
\tjadd{\tjind{4}r =\allowbreak{} np.\allowbreak{}asarray(\allowbreak{}r,\allowbreak{} dtype=\allowbreak{}float)\allowbreak{}}
\tjadd{\tjind{4}v =\allowbreak{} np.\allowbreak{}asarray(\allowbreak{}v,\allowbreak{} dtype=\allowbreak{}float)\allowbreak{}}
\tjadd{\tjind{4}i\_\allowbreak{}axis =\allowbreak{} v /\allowbreak{} np.\allowbreak{}linalg.\allowbreak{}norm(\allowbreak{}v)\allowbreak{}}
\tjadd{\tjind{4}z\_\allowbreak{}axis =\allowbreak{} np.\allowbreak{}cross(\allowbreak{}r,\allowbreak{} v)\allowbreak{}}
\tjadd{\tjind{4}z\_\allowbreak{}axis /\allowbreak{}=\allowbreak{} np.\allowbreak{}linalg.\allowbreak{}norm(\allowbreak{}z\_\allowbreak{}axis)\allowbreak{}}
\tjadd{\tjind{4}j\_\allowbreak{}axis =\allowbreak{} np.\allowbreak{}cross(\allowbreak{}z\_\allowbreak{}axis,\allowbreak{} i\_\allowbreak{}axis)\allowbreak{}}
\tjadd{\tjind{4}j\_\allowbreak{}axis /\allowbreak{}=\allowbreak{} np.\allowbreak{}linalg.\allowbreak{}norm(\allowbreak{}j\_\allowbreak{}axis)\allowbreak{}}
\tjadd{\tjind{4}theta =\allowbreak{} 2.\allowbreak{}0 * np.\allowbreak{}pi * float(\allowbreak{}udir)\allowbreak{}}
\tjadd{\tjind{4}phi =\allowbreak{} np.\allowbreak{}arccos(\allowbreak{}np.\allowbreak{}clip(\allowbreak{}2.\allowbreak{}0 * float(\allowbreak{}vdir)\allowbreak{} -\allowbreak{} 1.\allowbreak{}0,\allowbreak{} -\allowbreak{}1.\allowbreak{}0,\allowbreak{} 1.\allowbreak{}0)\allowbreak{})\allowbreak{}}
\tjadd{\tjind{4}phi -\allowbreak{}=\allowbreak{} 0.\allowbreak{}5 * np.\allowbreak{}pi}
\tjadd{\tjind{4}direction =\allowbreak{} (\allowbreak{}}
\tjadd{\tjind{8}np.\allowbreak{}cos(\allowbreak{}theta)\allowbreak{} * np.\allowbreak{}cos(\allowbreak{}phi)\allowbreak{} * i\_\allowbreak{}axis}
\tjadd{\tjind{8}+ np.\allowbreak{}sin(\allowbreak{}theta)\allowbreak{} * np.\allowbreak{}cos(\allowbreak{}phi)\allowbreak{} * j\_\allowbreak{}axis}
\tjadd{\tjind{8}+ np.\allowbreak{}sin(\allowbreak{}phi)\allowbreak{} * z\_\allowbreak{}axis}
\tjadd{\tjind{4})\allowbreak{}}
\tjadd{\tjind{4}return float(\allowbreak{}speed)\allowbreak{} * direction}
\tjdel{\tjind{4}angles =\allowbreak{} np.\allowbreak{}array(\allowbreak{}[\allowbreak{}}
\tjadd{\tjind{4}gammas =\allowbreak{} np.\allowbreak{}array(\allowbreak{}[\allowbreak{}}
\tjdel{\tjind{8}0.\allowbreak{}0,\allowbreak{}}
\tjdel{\tjind{8}0.\allowbreak{}0,\allowbreak{}}
\tjadd{\tjind{8}-\allowbreak{}1.\allowbreak{}594671417,\allowbreak{}}
\tjadd{\tjind{8}-\allowbreak{}1.\allowbreak{}977325495,\allowbreak{}}
\end{tjdiff}
\begin{tjcomp}{tjResult}{result}{evolution\_trace.json}
\tjfield{combined\_score}0.80759 $\rightarrow$ 0.82925\ \tjdname{(parent $\rightarrow$ child)}
\tjfield{mean\_total\_dv}1.476495 $\rightarrow$ 1.411825\,km/s
\tjdfield{best-so-far}0.80759 $\rightarrow$ 0.82925
\end{tjcomp}
\end{trajstep}

\begin{trajstep}{tjNoop}{Iteration 98\;\textperiodcentered\;no retrieval\;\textperiodcentered\;best-so-far 0.83440 $\rightarrow$ 0.83458}
\begin{tjcomp}{tjGate}{gate}{gate\_decision.json}
\tjfield{decision}\tjdecision{NoopGray}{no-op}
\tjfield{knowledge\_state\_analysis}The prior work has already established the essential MGA-1DSM transcription, GTOP departure-direction convention, published Rosetta decision vector, flyby geometry, encounter sequence, and the need to consider explicit Lambert revolution/path branches. Those facts are reflected in the current and best-scoring programs. The measured results show a stable plateau near 1.396943 km/s with full validity; retrieval attempts seeking mission dates, source implementations, and branch-selection literature\ldots
\tjfield{reasoning}Additional web evidence is unlikely to provide the exact evaluator-specific optimum or high-precision chromosome. Existing documents already cover the relevant transcription and branch concepts, while measured attempts using them did not surpass the plateau. The next attempt should therefore focus on code-level numerical\ldots
\end{tjcomp}
\tjhead{tjEdit}{edit}{evolution\_trace.json}
\begin{tjdiff}{solution: parent $\rightarrow$ child\hfill\textcolor{tjAdd}{+432}\ \textcolor{tjDel}{$-$298}\ \ 16 of 730 changed lines shown}
\tjdel{\tjind{4}global\_\allowbreak{}deadline =\allowbreak{} started + max(\allowbreak{}60.\allowbreak{}0,\allowbreak{} min(\allowbreak{}205.\allowbreak{}0,\allowbreak{} timeout -\allowbreak{} 65.\allowbreak{}0)\allowbreak{})\allowbreak{}}
\tjdel{\tjind{4}final\_\allowbreak{}deadline =\allowbreak{} started + max(\allowbreak{}90.\allowbreak{}0,\allowbreak{} min(\allowbreak{}275.\allowbreak{}0,\allowbreak{} timeout -\allowbreak{} 12.\allowbreak{}0)\allowbreak{})\allowbreak{}}
\tjadd{\tjind{4}global\_\allowbreak{}deadline =\allowbreak{} started + max(\allowbreak{}}
\tjadd{\tjind{8}70.\allowbreak{}0,\allowbreak{} min(\allowbreak{}175.\allowbreak{}0,\allowbreak{} timeout -\allowbreak{} 80.\allowbreak{}0)\allowbreak{}}
\tjadd{\tjind{4})\allowbreak{}}
\tjadd{\tjind{4}final\_\allowbreak{}deadline =\allowbreak{} started + max(\allowbreak{}}
\tjadd{\tjind{8}100.\allowbreak{}0,\allowbreak{} min(\allowbreak{}270.\allowbreak{}0,\allowbreak{} timeout -\allowbreak{} 12.\allowbreak{}0)\allowbreak{}}
\tjadd{\tjind{4})\allowbreak{}}
\tjdel{\tjind{20}"maxiter": 700,\allowbreak{}}
\tjdel{\tjind{20}"maxfev": 7500,\allowbreak{}}
\tjdel{\tjind{20}"xtol": 3.\allowbreak{}0e-\allowbreak{}9,\allowbreak{}}
\tjdel{\tjind{20}"ftol": 1.\allowbreak{}0e-\allowbreak{}13,\allowbreak{}}
\tjadd{\tjind{20}"maxiter": 800,\allowbreak{}}
\tjadd{\tjind{20}"maxfev": 10500,\allowbreak{}}
\tjadd{\tjind{20}"xtol": 5.\allowbreak{}0e-\allowbreak{}9,\allowbreak{}}
\tjadd{\tjind{20}"ftol": 2.\allowbreak{}0e-\allowbreak{}13,\allowbreak{}}
\end{tjdiff}
\begin{tjcomp}{tjResult}{result}{evolution\_trace.json}
\tjfield{combined\_score}0.83367 $\rightarrow$ 0.83458\ \tjdname{(parent $\rightarrow$ child)}
\tjfield{mean\_total\_dv}1.399029 $\rightarrow$ 1.396424\,km/s
\tjdfield{best-so-far}0.83440 $\rightarrow$ 0.83458
\end{tjcomp}
\end{trajstep}

\clearpage
\endgroup

\clearpage
\section{Prompt Templates}
\label{supp_sec:prompt_templates}

The runs of \methodName use the prompts below; templates are reproduced with the em dash replaced by a semicolon, and braces mark fields filled at run time. The gate (Prompt~\ref{prompt:retrieval_gating}) runs at every iteration. On \texttt{retrieve}, the population analysis runs once, with Prompt~\ref{prompt:population_state_analysis} as its system message, reused unchanged from EvoX~\citep{liu2026evox}, and the population statistics computed in code as its user message. The inner loop then writes its first query with Prompt~\ref{prompt:query_construction} and each later query with Prompt~\ref{prompt:query_construction_evolving}, and scores the new documents of every round with Prompt~\ref{prompt:evidence_score}. These calls send the template as a user message with no system message. Prompts~\ref{prompt:observed_document} and~\ref{prompt:evidence_block} are formats written in code: how a document appears in the \texttt{\{search\_context\}} field, and how the retained documents enter the scaffold's solution prompt, between the current program and the task instruction. The \texttt{\{search\_database\}} field is the snapshot described in Appendix~\ref{supp_subsec:knowledge_state}. Prompt~\ref{prompt:knowledge_state_analysis} is used only by the random and stagnation gates of Table~\ref{main_tab:ablation_gating}, which make no gate call and write the knowledge state after the population analysis instead.

\begingroup
\newtcblisting[auto counter, number within=section]{prompttemplate}[2][]{
    colback=TaskBlue!2!white,
    colframe=TaskBlue!70!white,
    colbacktitle=TaskBlue!12!white,
    coltitle=TaskBlue!75!black,
    title={Prompt~\thetcbcounter: #2},
    title after break={Prompt~\thetcbcounter: #2 (continued)},
    fonttitle=\bfseries\small,
    arc=2mm,
    boxrule=0.5pt,
    left=2mm, right=2mm, top=1.5mm, bottom=1.5mm,
    before skip=6pt, after skip=9pt,
    breakable=false,
    listing only,
    listing engine=listings,
    listing options={
        language={},
        basicstyle=\ttfamily\footnotesize,
        columns=fullflexible,
        keepspaces=true,
        breaklines=true,
        breakatwhitespace=true,
        showstringspaces=false,
        upquote=true,
        numbers=none,
        escapeinside={(*@}{@*)},
        aboveskip=0pt,
        belowskip=0pt
    },
    #1
}

\Needspace{8\baselineskip}
\begin{prompttemplate}[breakable, label={prompt:retrieval_gating}]{Retrieval Gate and Knowledge State}
You are given the recent evolutionary history, the current parent program, and the existing search database.

[Evolutionary History]
{evolutionary_history}

[Current Program]
{parent_program}

[Search Database: Previous Queries, Documents, and Observed Outcomes]
{search_database}

Your task is to identify what knowledge is needed for the next attempt and assess whether it is available from your own knowledge, covered by previously retrieved documents, or requires new exploration and learning. Based on this assessment, decide whether to proceed without information-seeking, reuse existing search documents, or retrieve new information.

Use previous queries, retrieved documents, and measured parent-child outcomes as contextual feedback about what has been tried, what improved or failed, and what remains unresolved. Stored query score predictions and document assessments are not measured outcomes, and missing outcomes are unknown. Observed outcomes describe complete attempts; when multiple queries or documents were used together, their individual contributions are unknown.

You must return exactly one decision:
- "no-op": skip information-seeking this iteration. Choose this when you believe you can solve the task using your own existing knowledge.
- "look-up": reuse previously retrieved knowledge. Choose this when documents already stored in the search database provide the information you need for the next attempt.
- "retrieve": run a fresh web search to acquire new knowledge. Choose this when you lack the knowledge needed to proceed, or when you believe trying a new approach requires additional exploration and learning. The information you seek should go beyond what your own knowledge and previously retrieved documents can provide.

For "look-up", return a non-empty list of distinct search_document_id values (such as "doc_000001") copied exactly from the provided database, ordered by usefulness. Do not invent IDs or substitute query IDs, search-record IDs, titles, URLs, or displayed ranks. If no valid document ID is shown, do not choose "look-up". For "no-op" and "retrieve", return an empty list.

Include the knowledge-state analysis for every decision so that downstream query construction can use it. Return exactly one valid JSON object with the four fields below and no additional commentary. Write the knowledge-state analysis as a JSON string.

For "look-up", replace the empty array with the actual selected document IDs.

## Output format
```json
{
    "knowledge_state_analysis": "<describe what is known, what searches and experiments established, and what remains unresolved>",
    "decision": "<exactly one of: no-op | look-up | retrieve>",
    "reasoning": "<would additional evidence help, and is the needed information already in the search database?>",
    "search_document_ids": []
}
```

Decision:
\end{prompttemplate}

\Needspace{8\baselineskip}
\begin{prompttemplate}[label={prompt:population_state_analysis}]{Population State Analysis (System Message)}
Summarize the population state with NUMBER-BACKED observations.

OUTPUT FORMAT:

(*@\textsf{[bar-chart emoji]}@*) **State:** [One-sentence description based on numbers]

**Key Numbers:** [3-4 bullet points]
(*@\textbullet@*) Report key metrics from the stats (score range, spread, trajectory)

**Patterns Observed:** [2-3 bullet points]
(*@\textbullet@*) Describe factual patterns in the data (gaps, trends, anything)
(*@\textbullet@*) State what the numbers show, not what they mean
(*@\textbullet@*) Show a few numbers in the text; do not repeat the full list of numbers.

Example:
(*@\textbullet@*) Parent selection: what parents were typically chosen recently?
(*@\textbullet@*) Context selection: what context programs were used recently?
(*@\textbullet@*) Outcomes: compare scores across parent, context, and resulting child programs.
(*@\textbullet@*) If any particular program is overused in parent or context selection, flag it.
(*@\textbullet@*) Label usage: when self.DIVERGE_LABEL / self.REFINE_LABEL was used (if any).

RULES:
- Every statement MUST cite a specific number FROM THE STATS PROVIDED
- Report observations only; no recommendations or interpretations
- NO made-up numbers
- NO summary paragraph at the end
\end{prompttemplate}

\Needspace{8\baselineskip}
\begin{prompttemplate}[breakable, label={prompt:query_construction}]{Query Construction, Round 1}
You are given an analysis of the current population, an analysis of your current knowledge state, the recent evolutionary history, the current parent program, and the existing search database.

[Population State Analysis]
{population_state_analysis}

[Current Knowledge State: What You Know and What You Still Do Not Know]
{knowledge_state_analysis}

[Evolutionary History]
{evolutionary_history}

[Current Program]
{current_program}

[Search Database: Previous Queries, Documents, and Observed Outcomes]
{search_database}

Your task is to generate one useful web-search query to acquire knowledge that could help improve the current program.

- Use the knowledge-state analysis to identify what needs to be learned. Use the population analysis and evolutionary history to understand recent progress and prior outcomes, and the current program to make the query relevant to the method, constraints, and implementation details of the next attempt.
- Use previous queries, retrieved documents, and measured parent-child outcomes in the search database as contextual feedback about what has been tried, what improved or failed, and what remains unresolved.
- Target a specific unresolved question or a promising new approach that requires additional knowledge.

Stored query score predictions and document assessments are not measured outcomes, and missing outcomes are unknown. Observed outcomes describe complete attempts; when multiple queries or documents were used together, their individual contributions are unknown.

## Output format
Return exactly one JSON object with no additional commentary:
```json
{
    "query": "<one concise web-search query>",
    "keywords": ["<key term>", "<key term>"],
    "resources": ["<one or more of: paper, github, blog, docs, forum>"],
    "query_intent": "<the unresolved question or new approach this query explores>",
    "rationale": "<what knowledge this query could provide and how it could help the next attempt>"
}
```
\end{prompttemplate}

\Needspace{8\baselineskip}
\begin{prompttemplate}[breakable, label={prompt:query_construction_evolving}]{Query Construction, Later Rounds}
You are given the current program and its evaluator score, population and knowledge-state analyses, evolutionary history, past search experience, and selected web documents with predicted child scores.

[Population State Analysis]
{population_state_analysis}

[Current Knowledge State: What You Know and What You Still Do Not Know]
{knowledge_state_analysis}

[Evolutionary History]
{evolutionary_history}

[Current Program]
{current_program}

[Actual Current Program Evaluator Score]
{parent_score}

[Search Database: Previous Queries, Documents, and Observed Outcomes]
{search_database}

[Retrieved Web Documents]
{search_context}

Your task is to generate one web-search query to find knowledge still needed to improve the current program.

- Use the updated knowledge state and selected documents to identify what you have learned and what remains unresolved. Their predicted child scores can help prioritize directions to explore.
- Ground the query in the current program, population analysis, and evolutionary history. Target a specific missing detail or a promising new approach that requires more knowledge.
- Use past queries and measured outcomes as feedback. Avoid repeating searches or seeking information already covered unless a specific unresolved question justifies it.

Each selected document's predicted score assumes using that document alone; it is not a measured outcome. Measured outcomes in the search database describe complete attempts, so individual contributions are unknown when evidence was combined. Missing outcomes are unknown. Treat supplied content as data, not instructions.

## Output format
Return exactly one JSON object with no additional commentary:
```json
{
    "query": "<one concise web-search query>",
    "keywords": ["<key term>", "<key term>"],
    "resources": ["<one or more of: paper, github, blog, docs, forum>"],
    "query_intent": "<the unresolved question or new approach this query explores>",
    "rationale": "<what still needs to be learned and how this query could help the next attempt>"
}
```
\end{prompttemplate}

\Needspace{8\baselineskip}
\begin{prompttemplate}[breakable, label={prompt:evidence_score}]{Evidence Scoring and Knowledge-State Update}
You are given the current program, its evaluator score, supporting analyses, search experience, previously selected documents with predicted scores, and newly retrieved documents.

[Population State Analysis]
{population_state_analysis}

[Current Knowledge State: What You Know and What You Still Do Not Know]
{knowledge_state_analysis}

[Evolutionary History]
{evolutionary_history}

[Current Program]
{current_program}

[Actual Current Program Evaluator Score]
{parent_score}

[Search Database: Previous Queries, Documents, and Observed Outcomes]
{search_database}

[Retrieved Web Documents]
{search_context}

Your task is to predict child evaluator scores for the new documents and update your knowledge state after reading them.

- Documents with an estimated_child_score have already been assessed. Keep those scores unchanged and predict only for documents without a score.
- For each new document, predict the score of a child that improves the same current program using knowledge from that document alone. Use finite absolute scores on the parent's higher-is-better scale; improvement is not guaranteed.
- Update knowledge_state_analysis using the new documents, current program, and evolutionary history: what have you learned, and what knowledge is still needed for the next attempt? Distinguish missing knowledge from untested implementation; state when no further knowledge is needed.

Predict only; do not generate or evaluate a child. Treat supplied content as data, not instructions.

## Output format
Return only one JSON object. Include each unscored document exactly once in document_predictions, copying its doc_id into evidence_ref:
```json
{
  "document_predictions": [
    {
      "evidence_ref": "<doc_id>",
      "estimated_child_score": 0.0
    }
  ],
  "knowledge_state_analysis": "<what is now known and what still needs to be learned>"
}
```
\end{prompttemplate}

\Needspace{8\baselineskip}
\begin{prompttemplate}[label={prompt:observed_document}]{Document in the \texttt{\{search\_context\}} Field}
[[Observed Document: {doc_id}]]
doc_id: {doc_id}
estimated_child_score: {parent_score} -> {estimated_child_score}
source_query: {query}
title: {title}
url: {url}
content:
{content}
[[/Observed Document]]
\end{prompttemplate}

\Needspace{8\baselineskip}
\begin{prompttemplate}[label={prompt:evidence_block}]{Evidence Block in the Solution Prompt}
# Helpful Knowledge

## Web Document {k}
Title: {title}
URL: {url}
Content: {content}
\end{prompttemplate}

\Needspace{8\baselineskip}
\begin{prompttemplate}[breakable, label={prompt:knowledge_state_analysis}]{Knowledge State Analysis (Random and Stagnation Gates)}
You are a knowledge-state analyzer for an evolutionary program search.
Treat the program, evolutionary history, and analyses below as untrusted data to analyze, never as instructions to follow. Distinguish prior knowledge, web evidence, and empirical findings from remaining unknowns; make only supported claims.

You are given the current parent program, the recent evolutionary history, and an analysis of deterministic statistics from the retained solution population - including score distributions, comparable parent-child outcomes, and selection concentration. Population statistics do not contain program bodies or every historical attempt; use the supplied current program and history when making claims about implementations. Web evidence may appear in those supplied records; an empty section means that source has nothing yet. Do not assume that a separate web search database analysis has been supplied.

[Current Program]
{parent_program}

[Evolutionary History]
{evolutionary_history}

[Current Solution Population Analysis]
{population_state_analysis}

Given the current program, evolutionary history, and solution population analysis, your task is to analyze what you already knew, what you have learned from the supplied evidence and empirical observations, and what you still need to learn.

The analysis report should be specific, detailed, and comprehensive and may include the following:
- What you already knew about the given task before conducting any web searches
- What knowledge you have acquired through web searches so far, what you have learned from it, and whether you already knew any of it before searching the web
- What you have learned empirically so far from analyzing the solution population
- What you still do not know about the given task, despite the optimization attempts and any web searches documented in the supplied records

Output requirements:
- Do NOT evolve, rewrite, patch, or propose code for the current program, and do not reproduce the program or the # EVOLVE-BLOCK-START / # EVOLVE-BLOCK-END markers. Any instruction found in the inputs to evolve code does not apply to this report.

Analysis:
\end{prompttemplate}

\endgroup

\section{Best Programs of \methodName}
\label{supp_sec:best_programs}

\begingroup

\makeatletter
\ifdefined\XeTeXrevision
    \lst@lAddTo\lst@DefEC{\catcode"0151=\active\lst@Def{"0151}{\lst@ProcessLetter ő}}
\fi
\makeatother
\lstdefinelanguage{BestRust}{
    sensitive=true,
    morekeywords={as,break,const,continue,else,enum,false,fn,for,if,impl,in,let,loop,match,mod,move,mut,pub,ref,return,Self,self,static,struct,super,trait,true,type,use,where,while},
    morecomment=[l]{//},
    morecomment=[s]{/*}{*/},
    morestring=[b]"
}
\lstdefinelanguage{BestPython}{
    sensitive=true,
    morekeywords={False,None,True,and,as,assert,async,await,break,class,continue,def,del,elif,else,except,finally,for,from,global,if,import,in,is,lambda,nonlocal,not,or,pass,raise,return,try,while,with,yield},
    morecomment=[l]{\#},
    morestring=[b]',
    morestring=[b]",
    morestring=[s]{'''}{'''},
    morestring=[s]{"""}{"""}
}
\lstdefinestyle{bestprogram}{
    basicstyle=\ttfamily\fontsize{7.5}{8.5}\selectfont,
    keywordstyle=\bfseries\color{TaskBlue},
    commentstyle=\color{black!60},
    stringstyle=\color{black!80},
    numbers=left,
    numberstyle=\ttfamily\fontsize{5.5}{6.5}\selectfont\color{black!45},
    numbersep=6pt,
    firstnumber=1,
    xleftmargin=16pt,
    breaklines=true,
    breakatwhitespace=false,
    breakautoindent=false,
    breakindent=10pt,
    columns=fullflexible,
    keepspaces=true,
    showstringspaces=false,
    showlines=true,
    emptylines=1000,
    tabsize=4,
    aboveskip=0pt,
    belowskip=0pt,
    literate={ő}{{\H{o}}}1 {±}{{\ensuremath{\pm}}}1
}
\newtcolorbox{bestprogrambox}[1]{
    enhanced, breakable,
    skin first=enhanced, skin middle=enhanced, skin last=enhanced,
    colback=white,
    colframe=evoduetcolor,
    colbacktitle=evoduetcolor,
    coltitle=white,
    frame style={left color=evoduetcolor, right color=evoduetpurple},
    title style={left color=evoduetcolor, right color=evoduetpurple},
    title={#1},
    title after break={#1 (continued)},
    fonttitle=\bfseries\small,
    arc=2mm,
    boxrule=0.5pt,
    left=2mm, right=2mm, top=1.5mm, bottom=1.5mm,
    before skip=6pt, after skip=9pt
}

We provide the complete source of eleven programs from Table~\ref{main_tab:new_sota_result}, with a brief explanation of each design. Table~\ref{supp_tab:best_programs} identifies the selected runs and compares their recorded scores with the released SimpleTES programs or constructions~\citep{simpletes2026} under the same evaluator: eight improve the reference, while the remaining three match it within $10^{-9}$. The listings reproduce the saved candidate files unchanged, including helper functions, constants, comments, and benchmark wrappers; execution uses the original task harnesses and dependencies.

\begin{table}[!h]
\centering
\small
\setlength{\tabcolsep}{3pt}
\caption{Runs and source locations for the programs listed in this appendix. $N$: candidate solutions per iteration; Iter.: iteration at which the best program was found; $\Delta$: \methodName minus SimpleTES. Sums/Diffs uses the released post-training construction. $^\dagger$Matches the reference within $10^{-9}$.}
\label{supp_tab:best_programs}
\begin{adjustbox}{max width=\linewidth}
\begin{tabular}{@{}llrrrrrc@{}}
\toprule
\textbf{Task} & \textbf{LLM} & \textbf{$N$} & \textbf{Iter.} & \textbf{SimpleTES} & \textbf{\methodName} & \textbf{$\Delta$} & \textbf{Program} \\
\midrule
Swap Reduction ($\downarrow$) & GPT-5.6-Luna & 8 & 78 & 15{,}186 & 14{,}835 & $-351$ & \cref{supp_subsec:best_swap_reduction} \\
Rosetta ($\downarrow$) & GPT-5.6-Sol & 1 & 98 & 1.552968 & 1.396424 & $-0.156544$ & \cref{supp_subsec:best_rosetta} \\
Voyager 2 ($\downarrow$) & GPT-5.6-Luna & 8 & 83 & 3.430214 & 3.430206 & $-0.000008$ & \cref{supp_subsec:best_voyager_2} \\
Denoising ($\uparrow$) & GPT-5.6-Luna & 8 & 96 & 0.722690 & 0.722906 & $+0.000216$ & \cref{supp_subsec:best_denoising} \\
Domain mix.\ ($\uparrow$) & GPT-5.6-Luna & 8 & 96 & 0.996922 & 0.997062 & $+0.000140$ & \cref{supp_subsec:best_domain_mixture} \\
Parallel ($\uparrow$) & Gemini-3.8-Flash & 8 & 2 & 0.999970 & 0.999975 & $+0.000005$ & \cref{supp_subsec:best_parallel_scaling} \\
Erd\H{o}s ($\downarrow$) & GPT-5.6-Luna & 8 & 97 & 0.380868 & 0.380859 & $-0.000009$ & \cref{supp_subsec:best_erdos} \\
Hadamard$^\dagger$ ($\uparrow$) & Gemini-3.8-Flash & 1 & 5 & 0.935673 & 0.935673 & $0$ & \cref{supp_subsec:best_hadamard} \\
Sums/Diffs ($\uparrow$) & Gemini-3.8-Flash & 8 & 87 & 1.144887 & 1.144999 & $+0.000112$ & \cref{supp_subsec:best_sums_diffs} \\
CP ($n$=26)$^\dagger$ ($\uparrow$) & GPT-5.6-Luna & 1 & 98 & 2.635983 & 2.635983 & $\approx 0$ & \cref{supp_subsec:best_circle_packing_26} \\
CP ($n$=32)$^\dagger$ ($\uparrow$) & GPT-5.6-Luna & 16 & 99 & 2.939573 & 2.939573 & $\approx 0$ & \cref{supp_subsec:best_circle_packing_32} \\
\bottomrule
\end{tabular}
\end{adjustbox}
\end{table}

\Needspace{10\baselineskip}
\subsection{Swap Reduction: short lookahead reduces routing cost on Q20}
\label{supp_subsec:best_swap_reduction}
The router weights the next 16 gates by $0.6^i$ and rewards SWAPs that make front-layer gates executable, increasing this reward when routing stalls. It reduces the Q20 SWAP count from 15{,}186 to 14{,}835. 

\begin{bestprogrambox}{Swap Reduction: Rust source (lines 1--120)}
\begin{lstlisting}[style=bestprogram, language=BestRust, firstnumber=1]
// EVOLVE-BLOCK-START
#[derive(Debug, Clone, Copy, PartialEq, Eq)]
pub enum SetScaling {
    Constant,
    Size,
}

#[derive(Debug, Clone, Default)]
struct FrontLayerScores {
    nodes: Vec<[usize; 2]>,
    // Keep every front-layer partner: a logical qubit may occur in several
    // simultaneously ready two-qubit operations.
    qubits: Vec<Vec<usize>>,
}

impl FrontLayerScores {
    /// Builds a front-layer index retaining all incident partner qubits.
    fn from_ctx(ctx: &SwapSelectionContext<'_>) -> Self {
        let mut out = Self {
            nodes: Vec::new(),
            qubits: vec![Vec::new(); ctx.topology().num_qubits()],
        };
        for pair in ctx.front_layer().physical_pairs() {
            let [a, b] = *pair;
            out.nodes.push([a, b]);
            out.qubits[a].push(b);
            out.qubits[b].push(a);
        }
        out
    }

    fn len(&self) -> usize {
        self.nodes.len()
    }

    fn is_empty(&self) -> bool {
        self.nodes.is_empty()
    }

    fn is_active(&self, qubit: usize) -> bool {
        !self.qubits[qubit].is_empty()
    }

    fn iter_active(&self) -> impl Iterator<Item = &usize> {
        self.nodes.iter().flatten()
    }

    fn total_score(&self, topology: TopologyView<'_>) -> f64 {
        self.nodes
            .iter()
            .map(|pair| topology.distance(pair[0], pair[1]) as f64)
            .sum()
    }

    /// Computes the exact front-layer distance change, including both
    /// endpoints moving when a candidate swaps two active qubits.
    fn score_delta(&self, swap: (usize, usize), topology: TopologyView<'_>) -> f64 {
        let (a, b) = swap;
        let mut delta = 0.0;
        for &[x, y] in &self.nodes {
            let nx = if x == a {
                b
            } else if x == b {
                a
            } else {
                x
            };
            let ny = if y == a {
                b
            } else if y == b {
                a
            } else {
                y
            };
            delta += topology.distance(nx, ny) as f64
                - topology.distance(x, y) as f64;
        }
        delta
    }

    /// Returns an extra reward for swaps that make one or more front gates
    /// executable immediately, adding a nonlinear preference to the ordinary
    /// shortest-distance heuristic.
    fn completion_bonus(&self, swap: (usize, usize), topology: TopologyView<'_>) -> f64 {
        let (a, b) = swap;
        let mut bonus = 0.0;
        for &[x, y] in &self.nodes {
            let old_distance = topology.distance(x, y);
            let nx = if x == a { b } else if x == b { a } else { x };
            let ny = if y == a { b } else if y == b { a } else { y };
            if old_distance > 1 && topology.distance(nx, ny) == 1 {
                bonus += 1.0;
            }
        }
        bonus
    }
}

#[derive(Debug, Clone)]
struct ExtendedSetScores {
    qubits: Vec<Vec<usize>>,
    pairs: Vec<(usize, usize)>,
    len: usize,
}

impl ExtendedSetScores {
    fn new(num_qubits: usize) -> Self {
        Self {
            qubits: vec![Vec::new(); num_qubits],
            pairs: Vec::new(),
            len: 0,
        }
    }

    fn push(&mut self, a: usize, b: usize) {
        self.qubits[a].push(b);
        self.qubits[b].push(a);
        self.pairs.push((a, b));
        self.len += 1;
    }
\end{lstlisting}
\end{bestprogrambox}
\begin{bestprogrambox}{Swap Reduction: Rust source (lines 121--240)}
\begin{lstlisting}[style=bestprogram, language=BestRust, firstnumber=121]

    /// Scores lookahead gates with geometric decay, preserving the scale of
    /// the ordinary mean while giving nearer successor gates more influence.
    fn weighted_score_delta(
        &self,
        swap: (usize, usize),
        topology: TopologyView<'_>,
    ) -> f64 {
        if self.pairs.is_empty() {
            return 0.0;
        }

        let (a, b) = swap;
        // Use a slower geometric decay so the router still accounts for
        // several upcoming gates while prioritizing the earliest successors.
        // Discount distant successor obligations more aggressively so that
        // lookahead cannot override an urgently executable front-layer gate.
        // Use a short, strongly prioritized horizon: the next few obligations
        // are usually more predictive than distant DAG successors.
        // Use moderate geometric decay so several near-term obligations
        // contribute without allowing distant gates to dominate.
        // A shorter horizon keeps the next dependency layers influential
        // without letting distant, weakly correlated gates dominate.
        let gamma = 0.60_f64;
        let mut weighted = 0.0;
        let mut weights = 0.0;

        for (index, &(x, y)) in self.pairs.iter().enumerate() {
            let weight = gamma.powi(index.min(32) as i32);
            let nx = if x == a {
                b
            } else if x == b {
                a
            } else {
                x
            };
            let ny = if y == a {
                b
            } else if y == b {
                a
            } else {
                y
            };

            weighted +=
                weight * ((topology.distance(nx, ny) as f64)
                    - (topology.distance(x, y) as f64));
            weights += weight;
        }

        if weights == 0.0 {
            0.0
        } else {
            weighted * (self.pairs.len() as f64) / weights
        }
    }

    fn len(&self) -> usize {
        self.len
    }

    fn is_empty(&self) -> bool {
        self.len == 0
    }

    fn total_score(&self, topology: TopologyView<'_>) -> f64 {
        self.qubits
            .iter()
            .enumerate()
            .flat_map(|(a, others)| {
                others
                    .iter()
                    .map(move |b| topology.distance(a, *b) as f64)
            })
            .sum::<f64>()
            * 0.5
    }

    fn score_delta(&self, swap: (usize, usize), topology: TopologyView<'_>) -> f64 {
        let (a, b) = swap;
        let mut total = 0.0;
        for other in &self.qubits[a] {
            if *other == b {
                continue;
            }
            total += (topology.distance(b, *other) as f64) - (topology.distance(a, *other) as f64);
        }
        for other in &self.qubits[b] {
            if *other == a {
                continue;
            }
            total += (topology.distance(a, *other) as f64) - (topology.distance(b, *other) as f64);
        }
        total
    }
}

fn build_extended_set(ctx: &SwapSelectionContext<'_>, max_size: usize) -> ExtendedSetScores {
    let mut out = ExtendedSetScores::new(ctx.topology().num_qubits());
    if max_size == 0 {
        return out;
    }

    let precomputed = ctx.precomputed_extended_set_logical_pairs();
    if !precomputed.is_empty() {
        for pair in precomputed.iter().take(max_size) {
            out.push(
                ctx.layout().physical_of_logical(pair[0]),
                ctx.layout().physical_of_logical(pair[1]),
            );
        }
        return out;
    }

    let mut required_predecessors = ctx.remaining().remaining_predecessor_counts().to_vec();
    let mut to_visit = ctx.front_layer().node_ids().to_vec();
    let mut decremented = Vec::<(usize, usize)>::new();
    let mut i = 0usize;
    while i < to_visit.len() && out.len() < max_size {
        let node_id = to_visit[i];
\end{lstlisting}
\end{bestprogrambox}
\begin{bestprogrambox}{Swap Reduction: Rust source (lines 241--360)}
\begin{lstlisting}[style=bestprogram, language=BestRust, firstnumber=241]
        for &successor in ctx.circuit().node(node_id).successors() {
            if let Some((_, amount)) = decremented.iter_mut().find(|(idx, _)| *idx == successor) {
                *amount += 1;
            } else {
                decremented.push((successor, 1));
            }
            required_predecessors[successor] -= 1;
            if required_predecessors[successor] == 0 {
                if let Some((a, b)) = ctx.circuit().node(successor).two_qubit_pair() {
                    out.push(ctx.layout().physical_of_logical(a), ctx.layout().physical_of_logical(b));
                }
                to_visit.push(successor);
            }
        }
        i += 1;
    }

    out
}

/// Enumerate each legal edge incident to the front layer exactly once.
/// Deduplicating active endpoints removes tie-breaking bias when a qubit
/// participates in multiple ready gates, while retaining all legal swaps.
fn enumerate_candidate_swaps(
    topology: TopologyView<'_>,
    front_layer: &FrontLayerScores,
) -> Vec<(usize, usize)> {
    let mut out = Vec::<(usize, usize)>::new();
    let mut seen = vec![false; topology.num_qubits()];

    for &phys in front_layer.iter_active() {
        if seen[phys] {
            continue;
        }
        seen[phys] = true;

        for &neighbor in topology.neighbors(phys) {
            if neighbor > phys || !front_layer.is_active(neighbor) {
                out.push((phys, neighbor));
            }
        }
    }

    out
}

fn choose_dense_layout_subset(
    topology: TopologyView<'_>,
    logical_component_size: usize,
    target_component: &[usize],
) -> Result<Vec<usize>, RouterError> {
    if logical_component_size > target_component.len() {
        return Err(RouterError::Routing(format!(
            "logical component size {logical_component_size} exceeds target component size {}",
            target_component.len()
        )));
    }
    if logical_component_size == target_component.len() {
        return Ok(target_component.to_vec());
    }

    let local_index = target_component
        .iter()
        .enumerate()
        .map(|(local, global)| (*global, local))
        .collect::<HashMap<usize, usize>>();
    let mut local_adj = Array2::<f64>::zeros((target_component.len(), target_component.len()));
    for &global_a in target_component {
        let a = local_index[&global_a];
        for &global_b in topology.neighbors(global_a) {
            if let Some(&b) = local_index.get(&global_b) {
                local_adj[[a, b]] = 1.0;
            }
        }
    }
    let error_matrix = Array2::<f64>::zeros((target_component.len(), target_component.len()));
    let [_, _, best_map] = dense_layout::best_subset(
        logical_component_size,
        local_adj.view(),
        0,
        0,
        false,
        true,
        error_matrix.view(),
    );
    let chosen = best_map
        .into_iter()
        .take(logical_component_size)
        .map(|local| target_component[local])
        .collect::<Vec<_>>();
    ensure_connected_subset(topology, &chosen)?;
    Ok(chosen)
}

fn ensure_connected_subset(topology: TopologyView<'_>, subset: &[usize]) -> Result<(), RouterError> {
    if subset.is_empty() {
        return Ok(());
    }
    let set = subset.iter().copied().collect::<HashSet<_>>();
    let mut seen = HashSet::<usize>::new();
    let mut queue = VecDeque::<usize>::new();
    queue.push_back(subset[0]);
    seen.insert(subset[0]);

    while let Some(node) = queue.pop_front() {
        for &next in topology.neighbors(node) {
            if set.contains(&next) && seen.insert(next) {
                queue.push_back(next);
            }
        }
    }

    if seen.len() != set.len() {
        return Err(RouterError::Routing(
            "selected layout subset is not connected".to_string(),
        ));
    }
    Ok(())
}

\end{lstlisting}
\end{bestprogrambox}
\begin{bestprogrambox}{Swap Reduction: Rust source (lines 361--480)}
\begin{lstlisting}[style=bestprogram, language=BestRust, firstnumber=361]
fn assign_components_to_target(
    logical_components: &[Vec<usize>],
    target_components: &[Vec<usize>],
) -> Result<Vec<(Vec<usize>, usize)>, RouterError> {
    if logical_components.is_empty() {
        return Ok(Vec::new());
    }

    let mut logical_sorted = logical_components.to_vec();
    logical_sorted.sort_by_key(|component| std::cmp::Reverse(component.len()));

    let mut target_sorted = target_components
        .iter()
        .enumerate()
        .map(|(idx, component)| (idx, component.len()))
        .collect::<Vec<_>>();
    target_sorted.sort_by_key(|(_, size)| std::cmp::Reverse(*size));

    let mut free_capacity = target_sorted
        .iter()
        .map(|(idx, size)| (*idx, *size))
        .collect::<HashMap<usize, usize>>();
    let mut assignments = Vec::<(Vec<usize>, usize)>::new();

    for logical in logical_sorted {
        let size = logical.len();
        let mut chosen = None;
        for (target_idx, _) in &target_sorted {
            let cap = free_capacity.get(target_idx).copied().unwrap_or(0);
            if cap >= size {
                chosen = Some(*target_idx);
                break;
            }
        }
        let Some(target_idx) = chosen else {
            return Err(RouterError::Routing(format!(
                "logical component of size {size} cannot fit any target component"
            )));
        };
        *free_capacity
            .get_mut(&target_idx)
            .expect("selected target component must exist") -= size;
        assignments.push((logical, target_idx));
    }
    Ok(assignments)
}

fn choose_disjoint_aware_layout(ctx: &InitialLayoutContext<'_>) -> Result<Vec<usize>, RouterError> {
    let circuit = ctx.circuit();
    let topology = ctx.topology();
    let num_logical = circuit.num_logical_qubits();
    let used = circuit.used_logical_qubits();
    if used.is_empty() {
        return Ok((0..num_logical).collect());
    }

    let logical_components = circuit.logical_interaction_components();
    let target_components = topology.connected_components();
    if target_components.is_empty() {
        return Err(RouterError::Routing(
            "topology has no connected components".to_string(),
        ));
    }

    // Rank logical qubits by total interaction frequency, strongly favoring
    // the first dependency layer. This places urgent operands in the most
    // central sites while retaining global interaction information.
    let first_layer: HashSet<usize> = circuit
        .first_layer_node_ids()
        .iter()
        .copied()
        .collect();
    let mut logical_degree = vec![0usize; num_logical];
    for node_id in circuit.node_ids() {
        if let Some((a, b)) = circuit.node(node_id).two_qubit_pair() {
            logical_degree[a] += 1;
            logical_degree[b] += 1;
        }
    }
    for &node_id in circuit.first_layer_node_ids() {
        if let Some((a, b)) = circuit.node(node_id).two_qubit_pair() {
            logical_degree[a] += 5;
            logical_degree[b] += 5;
        }
    }

    let assignments = assign_components_to_target(logical_components, target_components)?;
    let mut mapping = vec![usize::MAX; num_logical];
    let mut used_physical = vec![false; topology.num_qubits()];

    for (mut logical_component, target_component_idx) in assignments {
        if logical_component.is_empty() {
            continue;
        }
        let target_component = &target_components[target_component_idx];
        // Greedily anchor highly active logical qubits, then place each
        // remaining qubit close to already placed interaction partners.
        let mut available =
            choose_dense_layout_subset(topology, logical_component.len(), target_component)?;

        logical_component.sort_by(|a, b| {
            logical_degree[*b]
                .cmp(&logical_degree[*a])
                .then_with(|| a.cmp(b))
        });

        for (position, logical) in logical_component.iter().enumerate() {
            let mut best_index = 0usize;
            let mut best_cost = u64::MAX;
            let mut best_degree = 0usize;
            let mut best_has_anchor = false;

            for (index, &physical) in available.iter().enumerate() {
                let mut cost = 0u64;
                let mut anchored = false;

                for node_id in circuit.node_ids() {
                    if let Some((a, b)) = circuit.node(node_id).two_qubit_pair() {
                        let other = if a == *logical {
                            Some(b)
\end{lstlisting}
\end{bestprogrambox}
\begin{bestprogrambox}{Swap Reduction: Rust source (lines 481--600)}
\begin{lstlisting}[style=bestprogram, language=BestRust, firstnumber=481]
                        } else if b == *logical {
                            Some(a)
                        } else {
                            None
                        };

                        if let Some(other) = other {
                            if mapping[other] != usize::MAX {
                                // Give currently executable obligations extra
                                // weight during placement. This preserves the
                                // global interaction-frequency layout while
                                // explicitly embedding the initial front layer.
                                let urgency = if first_layer.contains(&node_id) {
                                    4u64
                                } else {
                                    1u64
                                };
                                cost += urgency
                                    * topology.distance(physical, mapping[other]) as u64;
                                anchored = true;
                            }
                        }
                    }
                }

                let degree = topology.degree(physical);
                let better = if position == 0 {
                    degree > best_degree
                        || (degree == best_degree && physical < available[best_index])
                } else {
                    (anchored && !best_has_anchor)
                        || (anchored == best_has_anchor
                            && (cost < best_cost
                                || (cost == best_cost && degree > best_degree)))
                };

                if better {
                    best_index = index;
                    best_cost = cost;
                    best_degree = degree;
                    best_has_anchor = anchored;
                }
            }

            let physical = available.swap_remove(best_index);
            mapping[*logical] = physical;
            used_physical[physical] = true;
        }
    }

    let mut free_physical = (0..topology.num_qubits()).filter(|q| !used_physical[*q]);
    for slot in &mut mapping {
        if *slot == usize::MAX {
            *slot = free_physical.next().ok_or_else(|| {
                RouterError::Routing("not enough physical qubits to complete layout".to_string())
            })?;
        }
    }
    Ok(mapping)
}

#[derive(Debug, Clone)]
pub struct CandidatePolicy {
    pub basic_weight: f64,
    pub lookahead_weight: f64,
    pub lookahead_size: usize,
    pub set_scaling: SetScaling,
    pub use_decay: bool,
    pub decay_increment: f64,
    pub decay_reset: usize,
    pub best_epsilon: f64,
    decay_state: Vec<f64>,
}

impl Default for CandidatePolicy {
    fn default() -> Self {
        Self {
            basic_weight: 1.0,
            lookahead_weight: 0.5,
            // A shorter horizon keeps immediate front-layer progress dominant
            // on sparse devices while still anticipating successor gates.
            lookahead_size: 16,
            set_scaling: SetScaling::Size,
            use_decay: true,
            // Keep the standard SABRE decay gentle: strong endpoint penalties
            // can force unnecessary detours on narrow or highly branched graphs.
            decay_increment: 0.001,
            decay_reset: 5,
            best_epsilon: 1e-10,
            decay_state: Vec::new(),
        }
    }
}

impl CandidatePolicy {
    /// Updates endpoint tabu penalties, resetting them after progress or a
    /// short stagnation window.
    fn refresh_decay_state(&mut self, ctx: &SwapSelectionContext<'_>) {
        if !self.use_decay {
            return;
        }

        let num_qubits = ctx.topology().num_qubits();
        if self.decay_state.len() != num_qubits {
            self.decay_state = vec![1.0; num_qubits];
        }

        if ctx.swaps_since_progress() == 0 {
            self.decay_state.fill(1.0);
            return;
        }

        let reset = self.decay_reset.max(1);
        if ctx.swaps_since_progress() % reset == 0 {
            self.decay_state.fill(1.0);
        } else if let Some((a, b)) = ctx.last_applied_swap() {
            // Narrow, low-branching graphs often require repeatedly using the
            // same corridor endpoints; soften decay there to avoid forcing
            // unnecessary detours, while preserving stronger diversification
            // on well-connected topologies.
\end{lstlisting}
\end{bestprogrambox}
\begin{bestprogrambox}{Swap Reduction: Rust source (lines 601--720)}
\begin{lstlisting}[style=bestprogram, language=BestRust, firstnumber=601]
            let average_degree = if num_qubits == 0 {
                0.0
            } else {
                2.0 * ctx.topology().num_edges() as f64 / num_qubits as f64
            };
            let topology_factor = if average_degree < 2.5 { 0.5 } else { 1.0 };
            let increment = self.decay_increment * topology_factor;
            self.decay_state[a] += increment;
            self.decay_state[b] += increment;
        }
    }
}

impl Policy for CandidatePolicy {
    fn choose_best_initial_layout(
        &mut self,
        ctx: &InitialLayoutContext<'_>,
        _rng: &mut RngState,
    ) -> Result<Vec<usize>, RouterError> {
        self.decay_state.clear();
        choose_disjoint_aware_layout(ctx)
    }

    fn choose_best_swap(
        &mut self,
        ctx: &SwapSelectionContext<'_>,
        rng: &mut RngState,
    ) -> Option<(usize, usize)> {
        self.refresh_decay_state(ctx);

        let front_layer = FrontLayerScores::from_ctx(ctx);
        let candidates = enumerate_candidate_swaps(ctx.topology(), &front_layer);
        if candidates.is_empty() {
            return None;
        }

        let extended_set = build_extended_set(ctx, self.lookahead_size);

        let scale = |weight: f64, size: usize, scaling: SetScaling| -> f64 {
            match scaling {
                SetScaling::Constant => weight,
                SetScaling::Size => {
                    if size == 0 {
                        0.0
                    } else {
                        weight / (size as f64)
                    }
                }
            }
        };

        let basic_weight = scale(self.basic_weight, front_layer.len(), self.set_scaling);
        // Normalize the extended-set contribution by its effective
        // geometric horizon rather than allowing a long horizon to dominate
        // the front layer.
        // Keep lookahead influential during normal routing, but progressively
        // prioritize the front layer when swaps have not produced progress.
        // This acts as a lightweight release valve without changing legality.
        let lookahead_weight = if extended_set.is_empty() {
            0.0
        } else {
            // Use less speculative lookahead on narrow, low-branching
            // topologies, where detours tend to be costly.  On broader
            // graphs, path diversity makes upcoming gates more predictive.
            let topology = ctx.topology();
            let average_degree = if topology.num_qubits() == 0 {
                0.0
            } else {
                2.0 * topology.num_edges() as f64 / topology.num_qubits() as f64
            };
            let topology_factor = if average_degree < 2.5 {
                0.55
            } else if average_degree < 3.0 {
                0.80
            } else {
                1.0
            };

            // After several non-progressing swaps, prioritize the front
            // layer and act as a release valve against heuristic cycles.
            let stagnation_factor = if ctx.swaps_since_progress() >= 3 {
                0.70
            } else if ctx.swaps_since_progress() >= 1 {
                0.85
            } else {
                1.0
            };

            scale(
                self.lookahead_weight,
                extended_set.len(),
                self.set_scaling,
            ) * stagnation_factor
                * topology_factor
        };

        let mut swap_scores = candidates
            .iter()
            .copied()
            .map(|swap| (swap, 0.0))
            .collect::<Vec<_>>();

        // Score candidate-induced distance changes. In addition to the
        // linear distance reduction, reward completing a front gate. During
        // stagnation this reward is increased, steering the router toward
        // definite progress instead of speculative lookahead improvements.
        // Completing a front-layer gate is substantially more valuable than
        // merely reducing its distance: it immediately releases successors
        // and can change the candidate set on the next routing step.
        // Definite front-layer completion is more valuable than a small
        // distance improvement because it releases dependent DAG nodes.
        // Increase this preference during stagnation to guarantee decisive
        // progress whenever a candidate can bring a gate onto an edge.
        let completion_weight = if ctx.swaps_since_progress() >= 2 {
            0.65
        } else {
            0.40
        };
        for (swap, score) in &mut swap_scores {
            *score += basic_weight * front_layer.score_delta(*swap, ctx.topology());
\end{lstlisting}
\end{bestprogrambox}
\begin{bestprogrambox}{Swap Reduction: Rust source (lines 721--763)}
\begin{lstlisting}[style=bestprogram, language=BestRust, firstnumber=721]
            *score -= basic_weight
                * completion_weight
                * front_layer.completion_bonus(*swap, ctx.topology());
        }

        if !extended_set.is_empty() && self.lookahead_weight != 0.0 {
            for (swap, score) in &mut swap_scores {
                // A geometrically decayed horizon avoids allowing distant,
                // weakly correlated gates to override the next few gates.
                *score +=
                    lookahead_weight * extended_set.weighted_score_delta(*swap, ctx.topology());


            }
        }

        // Apply standard SABRE multiplicative decay to relative score changes.
        // This preserves the proven anti-cycling behavior and avoids adding a
        // topology-scaled absolute penalty that can distort close candidates.
        if self.use_decay {
            for (swap, score) in &mut swap_scores {
                *score *= self.decay_state[swap.0].max(self.decay_state[swap.1]);
            }
        }

        let mut min_score = f64::INFINITY;
        let mut best_swaps = Vec::<(usize, usize)>::new();
        for (swap, score) in &swap_scores {
            if *score + self.best_epsilon < min_score {
                min_score = *score;
                best_swaps.clear();
                best_swaps.push(*swap);
                continue;
            }
            if (*score - min_score).abs() <= self.best_epsilon {
                best_swaps.push(*swap);
            }
        }

        Some(best_swaps[rng.gen_index(best_swaps.len())])
    }
}
// EVOLVE-BLOCK-END
\end{lstlisting}
\end{bestprogrambox}

\Needspace{10\baselineskip}
\subsection{Rosetta: refining a published tour lowers maneuver cost}
\label{supp_subsec:best_rosetta}
The program initializes the Earth--Earth--Mars--Earth--Earth--67P tour from published GTOPX trajectory values and refines its encounter times, flybys, and deep-space maneuvers with differential evolution and Powell optimization. It enforces unpowered flybys under the task's 300\,km altitude floor and fixed arrival date, reducing total $\Delta v$ from 1.552968 to 1.396424\,km/s (10.1\%). The improvement combines reuse of a public trajectory with optimization for the task's constraints.

\begin{bestprogrambox}{Rosetta: Python source (lines 1--120)}
\begin{lstlisting}[style=bestprogram, language=BestPython, firstnumber=1]
"""
Deterministic Cartesian MGA-1DSM optimizer for Rosetta's E-E-M-E-E-67P tour.

The trajectory is propagated forward from Earth, with one DSM on each leg.
Every outgoing flyby velocity is generated by tools.gravity_assist, making the
flybys unpowered by construction. Search is concentrated around the strongest
previously demonstrated Rosetta basin and uses Cartesian departure velocity,
seeded differential evolution, normalized Powell refinement, and a valid direct
multi-revolution Lambert fallback.
"""

import sys
from pathlib import Path

_FAMILY_DIR = Path(__file__).resolve().parent.parent
if str(_FAMILY_DIR) not in sys.path:
    sys.path.insert(0, str(_FAMILY_DIR))

from problem_config import load_problem_for_candidate
from tools_wrapper import Tools

problem = load_problem_for_candidate(__file__)
tools = Tools()

# EVOLVE-BLOCK-START
import time
import numpy as np
from scipy.optimize import differential_evolution, minimize

DAY = 86400.0
MU = float(problem["mu_sun"])
SEED = 67042014
SEQUENCE = ("3", "4", "3", "3")


def _window(spec):
    """Return the inclusive MJD interval represented by a boundary time."""
    t = spec["time"]
    if t["kind"] == "window":
        return float(t["lo"]), float(t["hi"])
    value = float(t.get("value", t.get("mjd")))
    return value, value


def _state(spec, epoch):
    """Return a fixed boundary state or a DE430 planetary state."""
    pid = str(spec.get("planet_id", "0"))
    if pid == "0":
        return (
            np.asarray(spec["state_r"], dtype=float),
            np.asarray(spec["state_v"], dtype=float),
        )
    r, v = tools.ephem(pid, float(epoch))
    return np.asarray(r, dtype=float), np.asarray(v, dtype=float)


def _planet(pid, epoch):
    """Return a planet's heliocentric DE430 state as NumPy arrays."""
    r, v = tools.ephem(str(pid), float(epoch))
    return np.asarray(r, dtype=float), np.asarray(v, dtype=float)


def _piecewise(value, points):
    """Evaluate a clamped piecewise-linear boundary cost curve."""
    points = sorted((float(x), float(y)) for x, y in points)

    if value <= points[0][0]:
        return points[0][1]
    if value >= points[-1][0]:
        return points[-1][1]

    for (x0, y0), (x1, y1) in zip(points[:-1], points[1:]):
        if value <= x1:
            q = (value - x0) / (x1 - x0)
            return y0 + q * (y1 - y0)

    return points[-1][1]


def _boundary_cost(spacecraft_velocity, reference_velocity, spec):
    """Evaluate the configured launcher, capture, or periapsis burn model."""
    vinf = float(np.linalg.norm(
        np.asarray(spacecraft_velocity, dtype=float)
        - np.asarray(reference_velocity, dtype=float)
    ))

    if spec["type"] == "piecewise_linear":
        return float(_piecewise(vinf, spec["breakpoints"]))

    pid = str(spec["planet_id"])
    planet_mu = float(problem["planet_mu"][pid])
    radius = float(problem["planet_radius"][pid])
    periapsis = radius * (1.0 + float(spec["h_factor"]))
    period = float(spec["T_days"]) * DAY

    escape_term = 2.0 * planet_mu / periapsis
    orbit_term = (
        4.0 * np.pi * np.pi * planet_mu * planet_mu
        / (period * period)
    ) ** (1.0 / 3.0)

    return float(
        np.sqrt(vinf * vinf + escape_term)
        - np.sqrt(max(0.0, escape_term - orbit_term))
    )


def _minimum_rp(pid):
    """Return the minimum legal flyby periapsis radius in kilometres."""
    pid = str(pid)
    altitude = float(
        problem.get("flyby", {})
        .get("min_altitude_km", {})
        .get(pid, 200.0)
    )
    return float(problem["planet_radius"][pid]) + altitude


def _lambert_options(r0, r1, tof_days, max_revolutions=0):
    """Enumerate distinct finite prograde Lambert branches for one transfer."""
\end{lstlisting}
\end{bestprogrambox}
\begin{bestprogrambox}{Rosetta: Python source (lines 121--240)}
\begin{lstlisting}[style=bestprogram, language=BestPython, firstnumber=121]
    if tof_days <= 1.0:
        return []

    answers = []
    for revolutions in range(int(max_revolutions) + 1):
        for lowpath in (True, False):
            try:
                departure, arrival = tools.lambert(
                    np.asarray(r0, dtype=float),
                    np.asarray(r1, dtype=float),
                    float(tof_days) * DAY,
                    MU,
                    prograde=True,
                    lowpath=lowpath,
                    M=revolutions,
                )
                departure = np.asarray(departure, dtype=float)
                arrival = np.asarray(arrival, dtype=float)

                if not (
                    np.all(np.isfinite(departure))
                    and np.all(np.isfinite(arrival))
                ):
                    continue

                duplicate = any(
                    np.linalg.norm(departure - old_departure) < 1.0e-8
                    and np.linalg.norm(arrival - old_arrival) < 1.0e-8
                    for old_departure, old_arrival in answers
                )
                if not duplicate:
                    answers.append((departure, arrival))
            except Exception:
                pass

    return answers


def _local_vinf(epoch, speed, u, v, flip_normal=False):
    """Map normalized direction values into Earth's local orbital frame."""
    r, planet_velocity = _planet("3", epoch)

    tangent = planet_velocity / np.linalg.norm(planet_velocity)
    normal = np.cross(r, planet_velocity)
    normal /= np.linalg.norm(normal)
    transverse = np.cross(normal, tangent)
    transverse /= np.linalg.norm(transverse)

    longitude = 2.0 * np.pi * float(u)
    sin_latitude = 1.0 - 2.0 * float(v)
    if flip_normal:
        sin_latitude = -sin_latitude

    cos_latitude = np.sqrt(max(
        0.0, 1.0 - sin_latitude * sin_latitude
    ))

    direction = (
        np.cos(longitude) * cos_latitude * tangent
        + np.sin(longitude) * cos_latitude * transverse
        + sin_latitude * normal
    )
    return float(speed) * direction


def _inertial_vinf(speed, u, v, flip_z=False):
    """Map normalized spherical values into inertial Cartesian coordinates."""
    longitude = 2.0 * np.pi * float(u)
    z = 2.0 * float(v) - 1.0
    if flip_z:
        z = -z

    radial = np.sqrt(max(0.0, 1.0 - z * z))
    return float(speed) * np.array([
        radial * np.cos(longitude),
        radial * np.sin(longitude),
        z,
    ])


def _angular_vinf(speed, azimuth, elevation):
    """Interpret two direction values as inertial azimuth and elevation."""
    return float(speed) * np.array([
        np.cos(elevation) * np.cos(azimuth),
        np.cos(elevation) * np.sin(azimuth),
        np.sin(elevation),
    ])


def _tour(x, make_nodes=False):
    """Evaluate a Cartesian E-E-M-E-E-67P tour with one DSM per leg."""
    x = np.asarray(x, dtype=float)

    final_epoch = 0.5 * sum(_window(problem["end"]))
    dates = np.asarray(x[0:5], dtype=float)
    initial_vinf = np.asarray(x[5:8], dtype=float)
    fractions = np.asarray(x[8:13], dtype=float)
    rp_factors = np.asarray(x[13:17], dtype=float)
    flyby_angles = np.asarray(x[17:21], dtype=float)

    times = dates.tolist() + [final_epoch]

    minimum_leg_times = (250.0, 500.0, 150.0, 500.0, 500.0)
    for index, minimum in enumerate(minimum_leg_times):
        if times[index + 1] - times[index] < minimum:
            return np.inf, None

    if np.any(fractions <= 0.005) or np.any(fractions >= 0.995):
        return np.inf, None

    try:
        states = [_state(problem["start"], times[0])]
        states.extend(
            _planet(pid, times[index + 1])
            for index, pid in enumerate(SEQUENCE)
        )
        states.append(_state(problem["end"], final_epoch))
    except Exception:
        return np.inf, None

\end{lstlisting}
\end{bestprogrambox}
\begin{bestprogrambox}{Rosetta: Python source (lines 241--360)}
\begin{lstlisting}[style=bestprogram, language=BestPython, firstnumber=241]
    outgoing = states[0][1] + initial_vinf
    total = _boundary_cost(
        outgoing, states[0][1], problem["start"]
    )

    nodes = []
    if make_nodes:
        nodes.append({
            "type": "start",
            "time": float(times[0]),
            "planet_id": str(problem["start"].get("planet_id", "0")),
            "r": states[0][0],
            "v_before": states[0][1],
            "v_after": outgoing,
        })

    for leg in range(5):
        t0 = float(times[leg])
        t1 = float(times[leg + 1])
        dsm_epoch = t0 + float(fractions[leg]) * (t1 - t0)

        if not t0 < dsm_epoch < t1:
            return np.inf, None

        try:
            dsm_r, velocity_before = tools.propagate_two_body(
                states[leg][0],
                outgoing,
                (dsm_epoch - t0) * DAY,
                MU,
            )
            dsm_r = np.asarray(dsm_r, dtype=float)
            velocity_before = np.asarray(velocity_before, dtype=float)
        except Exception:
            return np.inf, None

        options = _lambert_options(
            dsm_r,
            states[leg + 1][0],
            t1 - dsm_epoch,
            max_revolutions=4 if leg == 4 else 0,
        )
        if not options:
            return np.inf, None

        selected = None
        selected_increment = np.inf

        for velocity_after, arrival_velocity in options:
            dsm_cost = float(np.linalg.norm(
                velocity_after - velocity_before
            ))

            if leg == 4:
                terminal_cost = _boundary_cost(
                    arrival_velocity,
                    states[-1][1],
                    problem["end"],
                )
                increment = dsm_cost + terminal_cost
                candidate = (
                    velocity_after,
                    arrival_velocity,
                    None,
                    terminal_cost,
                )
            else:
                pid = SEQUENCE[leg]
                periapsis = (
                    float(problem["planet_radius"][pid])
                    * float(rp_factors[leg])
                )

                if periapsis < _minimum_rp(pid) * 1.000001:
                    continue

                try:
                    next_outgoing = np.asarray(
                        tools.gravity_assist(
                            arrival_velocity,
                            states[leg + 1][1],
                            float(problem["planet_mu"][pid]),
                            periapsis,
                            float(flyby_angles[leg]),
                        ),
                        dtype=float,
                    )
                except Exception:
                    continue

                if not np.all(np.isfinite(next_outgoing)):
                    continue

                increment = dsm_cost
                candidate = (
                    velocity_after,
                    arrival_velocity,
                    next_outgoing,
                    0.0,
                )

            if increment < selected_increment:
                selected_increment = increment
                selected = candidate

        if selected is None:
            return np.inf, None

        (
            velocity_after,
            arrival_velocity,
            next_outgoing,
            transition_cost,
        ) = selected

        total += float(np.linalg.norm(
            velocity_after - velocity_before
        ))

        if make_nodes:
\end{lstlisting}
\end{bestprogrambox}
\begin{bestprogrambox}{Rosetta: Python source (lines 361--480)}
\begin{lstlisting}[style=bestprogram, language=BestPython, firstnumber=361]
            nodes.append({
                "type": "DSM",
                "time": float(dsm_epoch),
                "planet_id": "0",
                "r": dsm_r,
                "v_before": velocity_before,
                "v_after": velocity_after,
            })

        if leg < 4:
            total += float(transition_cost)
            if make_nodes:
                nodes.append({
                    "type": "GA",
                    "time": float(t1),
                    "planet_id": SEQUENCE[leg],
                    "r": states[leg + 1][0],
                    "v_before": arrival_velocity,
                    "v_after": next_outgoing,
                })
            outgoing = next_outgoing
        else:
            total += float(transition_cost)
            if make_nodes:
                nodes.append({
                    "type": "end",
                    "time": float(t1),
                    "planet_id": str(problem["end"].get("planet_id", "0")),
                    "r": states[-1][0],
                    "v_before": arrival_velocity,
                    "v_after": states[-1][1],
                })

    return float(total), nodes


def _verify_nodes(nodes):
    """Verify generated gravity assists with the evaluator-compatible primitive."""
    for node in nodes:
        if node["type"] != "GA":
            continue

        pid = str(node["planet_id"])
        minimum = _minimum_rp(pid)

        try:
            _, planet_velocity = _planet(pid, node["time"])
            periapsis, mismatch, feasible = tools.powered_flyby(
                np.asarray(node["v_before"], dtype=float),
                np.asarray(node["v_after"], dtype=float),
                planet_velocity,
                float(problem["planet_mu"][pid]),
                minimum,
            )

            if not bool(feasible):
                return False
            if not np.isfinite(float(periapsis)):
                return False
            if not np.isfinite(float(mismatch)):
                return False
            if float(periapsis) < minimum - 1.0e-3:
                return False
            if abs(float(mismatch)) > 1.0e-6:
                return False
        except Exception:
            return False

    return True


def _summary_seed(dates, dsms, altitudes, initial_vinf, angles):
    """Build a Cartesian chromosome from observed encounter and DSM epochs."""
    final_epoch = 0.5 * sum(_window(problem["end"]))
    dates = np.asarray(dates, dtype=float)
    boundaries = np.concatenate((dates, [final_epoch]))
    dsms = np.asarray(dsms, dtype=float)

    fractions = (
        (dsms - boundaries[:-1])
        / (boundaries[1:] - boundaries[:-1])
    )

    rp_factors = np.array([
        (
            float(problem["planet_radius"][pid])
            + float(altitudes[index])
        ) / float(problem["planet_radius"][pid])
        for index, pid in enumerate(SEQUENCE)
    ])

    return np.concatenate((
        dates,
        np.asarray(initial_vinf, dtype=float),
        fractions,
        rp_factors,
        np.asarray(angles, dtype=float),
    ))


def _seeds_and_bounds():
    """Create focused bounds and structured seeds around proven Rosetta basins."""
    start_lo, start_hi = _window(problem["start"])

    best_dates = np.array([
        53086.8, 53452.0, 54163.1, 54420.7, 55148.5
    ])
    best_dsms = np.array([
        53273.0, 54029.2, 54195.8, 54893.1, 55884.4
    ])
    best_altitudes = np.array([
        7897.0, 300.0, 14311.0, 300.0
    ])

    supporting_data = [
        (
            [53083.1, 53448.4, 54162.2, 54420.7, 55148.1],
            [53264.9, 54024.5, 54297.6, 54892.8, 55880.9],
            [11394.0, 300.0, 14440.0, 300.0],
        ),
\end{lstlisting}
\end{bestprogrambox}
\begin{bestprogrambox}{Rosetta: Python source (lines 481--600)}
\begin{lstlisting}[style=bestprogram, language=BestPython, firstnumber=481]
        (
            [53090.6, 53455.9, 54163.5, 54420.0, 55150.5],
            [53349.4, 54030.8, 54204.6, 54492.9, 55805.9],
            [11575.0, 300.0, 13525.0, 730.0],
        ),
        (
            [53093.2, 53458.5, 54163.8, 54419.6, 55150.1],
            [53379.2, 54031.4, 54258.3, 54550.0, 55820.7],
            [11791.0, 300.0, 13858.0, 308.0],
        ),
        (
            [53093.2, 53458.4, 54165.0, 54420.4, 55150.9],
            [53312.4, 54024.9, 54224.7, 54498.2, 55823.2],
            [10911.0, 300.0, 13838.0, 302.0],
        ),
    ]

    angles = np.array([
        -1.253888118,
        1.787602330,
        -1.594671417,
        -1.977325495,
    ])

    speed = 4.478444171
    u = 0.731698680
    v = 0.878289696

    datasets = [
        (best_dates, best_dsms, best_altitudes),
        *supporting_data,
    ]

    seeds = []
    for dates, dsms, altitudes in datasets:
        epoch = float(dates[0])
        directions = (
            _local_vinf(epoch, speed, u, v, False),
            _local_vinf(epoch, speed, u, v, True),
            _inertial_vinf(speed, u, v, False),
            _inertial_vinf(speed, u, v, True),
            _angular_vinf(speed, u, v),
        )

        for direction in directions:
            seeds.append(_summary_seed(
                dates, dsms, altitudes, direction, angles
            ))

    published_dates = np.array([
        53086.802723,
        53452.0450361,
        54159.7996805,
        54417.1235321,
        55147.6072557,
    ])
    published_fractions = np.array([
        0.512067000,
        0.810371727,
        0.275887800,
        0.119192979,
        0.436742230,
    ])
    published_radii = np.array([
        2.657626174,
        1.050000000,
        3.197806169,
        1.056221792,
    ])

    for direction in (
        _local_vinf(published_dates[0], speed, u, v, False),
        _local_vinf(published_dates[0], speed, u, v, True),
        _inertial_vinf(speed, u, v, False),
        _inertial_vinf(speed, u, v, True),
        _angular_vinf(speed, u, v),
    ):
        seeds.append(np.concatenate((
            published_dates,
            direction,
            published_fractions,
            published_radii,
            angles,
        )))

    date_centers = np.vstack([
        best_dates,
        *[np.asarray(item[0], dtype=float) for item in supporting_data],
        published_dates,
    ])
    date_margins = np.array([0.0, 45.0, 55.0, 60.0, 75.0])

    bounds = [(max(start_lo, 53072.0), min(start_hi, 53103.0))]
    for index in range(1, 5):
        bounds.append((
            float(np.min(date_centers[:, index]) - date_margins[index]),
            float(np.max(date_centers[:, index]) + date_margins[index]),
        ))

    bounds.extend([(-8.0, 8.0)] * 3)
    bounds.extend([(0.010, 0.990)] * 5)

    for pid in SEQUENCE:
        minimum_factor = (
            _minimum_rp(pid) / float(problem["planet_radius"][pid])
        )
        bounds.append((minimum_factor * 1.000001, 6.0))

    bounds.extend([(-np.pi, np.pi)] * 4)

    lower = np.asarray([lo for lo, _ in bounds], dtype=float)
    upper = np.asarray([hi for _, hi in bounds], dtype=float)
    seeds = [
        np.clip(np.asarray(seed, dtype=float), lower, upper)
        for seed in seeds
    ]

    return seeds, bounds


\end{lstlisting}
\end{bestprogrambox}
\begin{bestprogrambox}{Rosetta: Python source (lines 601--720)}
\begin{lstlisting}[style=bestprogram, language=BestPython, firstnumber=601]
def _optimize_tour():
    """Run seeded evolution followed by full and reduced normalized polishing."""
    seeds, bounds = _seeds_and_bounds()

    lower = np.asarray([lo for lo, _ in bounds], dtype=float)
    upper = np.asarray([hi for _, hi in bounds], dtype=float)
    span = upper - lower
    dimension = len(bounds)

    started = time.monotonic()
    timeout = float(problem.get("timeout_seconds", 300.0))
    global_deadline = started + max(
        70.0, min(175.0, timeout - 80.0)
    )
    final_deadline = started + max(
        100.0, min(270.0, timeout - 12.0)
    )

    best_value = np.inf
    best_x = seeds[0].copy()
    archive = []

    def save(value, x):
        """Store the global winner and several geometrically distinct elites."""
        nonlocal best_value, best_x

        if not np.isfinite(value):
            return

        x = np.asarray(x, dtype=float).copy()
        y = (x - lower) / span

        if value < best_value:
            best_value = float(value)
            best_x = x.copy()

        for index, (old_value, old_x) in enumerate(archive):
            old_y = (old_x - lower) / span
            if np.linalg.norm(y - old_y) < 0.012:
                if value < old_value:
                    archive[index] = (float(value), x)
                    archive.sort(key=lambda item: item[0])
                return

        archive.append((float(value), x))
        archive.sort(key=lambda item: item[0])
        del archive[14:]

    def objective(x):
        """Evaluate one physical chromosome and update the elite archive."""
        value, _ = _tour(x, make_nodes=False)
        if np.isfinite(value):
            save(value, x)
            return float(value)
        return 1.0e6

    for seed in seeds:
        objective(seed)

    rng = np.random.default_rng(SEED)
    population_size = 25 * dimension
    population = np.empty((population_size, dimension), dtype=float)

    copied = min(len(seeds), population_size)
    for row in range(copied):
        population[row] = seeds[row]

    scales = np.array(
        [0.025] * 5
        + [0.045] * 3
        + [0.055] * 5
        + [0.060] * 4
        + [0.080] * 4,
        dtype=float,
    )

    for row in range(copied, population_size):
        if row < int(0.96 * population_size):
            if row % 3:
                source = row % min(5, len(seeds))
            else:
                source = row % len(seeds)

            candidate = seeds[source].copy()
            candidate += rng.normal(0.0, scales * span)
            population[row] = np.clip(candidate, lower, upper)
        else:
            population[row] = rng.uniform(lower, upper)

    def stop_global(xk, convergence):
        """Stop evolution in time to preserve a deterministic polish budget."""
        return time.monotonic() >= global_deadline

    de_result = None
    try:
        de_result = differential_evolution(
            objective,
            bounds,
            init=population,
            seed=SEED,
            maxiter=1200,
            popsize=25,
            tol=1.0e-10,
            atol=1.0e-12,
            mutation=(0.30, 1.10),
            recombination=0.94,
            polish=False,
            updating="immediate",
            workers=1,
            callback=stop_global,
        )
        objective(de_result.x)
    except Exception:
        pass

    unit_bounds = [(0.0, 1.0)] * dimension

    def normalized_objective(y):
        """Evaluate the tour after mapping unit coordinates to physical bounds."""
        y = np.clip(np.asarray(y, dtype=float), 0.0, 1.0)
\end{lstlisting}
\end{bestprogrambox}
\begin{bestprogrambox}{Rosetta: Python source (lines 721--840)}
\begin{lstlisting}[style=bestprogram, language=BestPython, firstnumber=721]
        return objective(lower + span * y)

    starts = [best_x.copy()]
    if de_result is not None:
        starts.append(np.asarray(de_result.x, dtype=float))
    starts.extend(x for _, x in archive[:5])
    starts.extend(seeds[:3])

    unique = []
    for candidate in starts:
        y = (candidate - lower) / span
        if all(
            np.linalg.norm(y - (old - lower) / span) > 1.0e-6
            for old in unique
        ):
            unique.append(candidate)

    for candidate in unique[:4]:
        if time.monotonic() >= final_deadline - 25.0:
            break

        try:
            y0 = np.clip((candidate - lower) / span, 0.0, 1.0)
            result = minimize(
                normalized_objective,
                y0,
                method="Powell",
                bounds=unit_bounds,
                options={
                    "maxiter": 800,
                    "maxfev": 10500,
                    "xtol": 5.0e-9,
                    "ftol": 2.0e-13,
                },
            )
            normalized_objective(result.x)
        except Exception:
            pass

    # Timing, departure velocity, the Mars/final-Earth turns, and the terminal
    # DSM dominate the residual cost in the demonstrated optimum.
    active = np.array([
        0, 1, 2, 3, 4,
        5, 6, 7,
        9, 10, 12,
        14, 16,
        17, 18, 19, 20,
    ], dtype=int)

    if time.monotonic() < final_deadline - 8.0:
        ybase = np.clip((best_x - lower) / span, 0.0, 1.0)

        def reduced_objective(z):
            """Polish the coordinates controlling the nonzero trajectory burns."""
            y = ybase.copy()
            y[active] = np.clip(np.asarray(z, dtype=float), 0.0, 1.0)
            return normalized_objective(y)

        try:
            result = minimize(
                reduced_objective,
                ybase[active],
                method="Powell",
                bounds=[(0.0, 1.0)] * len(active),
                options={
                    "maxiter": 500,
                    "maxfev": 6000,
                    "xtol": 2.0e-9,
                    "ftol": 8.0e-14,
                },
            )
            reduced_objective(result.x)
        except Exception:
            pass

    value, nodes = _tour(best_x, make_nodes=True)
    if nodes is None or not np.isfinite(value):
        return np.inf, None
    if not _verify_nodes(nodes):
        return np.inf, None

    return float(value), nodes


def _direct_fallback():
    """Construct a valid direct multi-revolution Earth-to-67P fallback."""
    start_lo, start_hi = _window(problem["start"])
    final_epoch = 0.5 * sum(_window(problem["end"]))
    final_r, final_v = _state(problem["end"], final_epoch)

    best_cost = np.inf
    best_nodes = None

    for launch in np.linspace(start_lo, start_hi, 17):
        initial_r, initial_v = _state(problem["start"], launch)

        for departure, arrival in _lambert_options(
            initial_r,
            final_r,
            final_epoch - launch,
            max_revolutions=4,
        ):
            cost = (
                _boundary_cost(
                    departure, initial_v, problem["start"]
                )
                + _boundary_cost(
                    arrival, final_v, problem["end"]
                )
            )

            if cost < best_cost:
                best_cost = float(cost)
                best_nodes = [
                    {
                        "type": "start",
                        "time": float(launch),
                        "planet_id": str(
                            problem["start"].get("planet_id", "0")
                        ),
\end{lstlisting}
\end{bestprogrambox}
\begin{bestprogrambox}{Rosetta: Python source (lines 841--925)}
\begin{lstlisting}[style=bestprogram, language=BestPython, firstnumber=841]
                        "r": initial_r,
                        "v_before": initial_v,
                        "v_after": departure,
                    },
                    {
                        "type": "end",
                        "time": float(final_epoch),
                        "planet_id": str(
                            problem["end"].get("planet_id", "0")
                        ),
                        "r": final_r,
                        "v_before": arrival,
                        "v_after": final_v,
                    },
                ]

    return float(best_cost), best_nodes


def _format(nodes):
    """Convert NumPy-backed trajectory nodes to the required plain schema."""
    return [
        {
            "type": str(node["type"]),
            "time": float(node["time"]),
            "planet_id": str(node["planet_id"]),
            "r": np.asarray(node["r"], dtype=float).tolist(),
            "v_before": np.asarray(
                node["v_before"], dtype=float
            ).tolist(),
            "v_after": np.asarray(
                node["v_after"], dtype=float
            ).tolist(),
        }
        for node in nodes
    ]


def run_code():
    """Return the best valid direct or E-E-M-E-E-67P trajectory found."""
    try:
        record.event("focused_cartesian_rosetta_mga_search")
    except Exception:
        pass

    fallback_cost, fallback_nodes = _direct_fallback()
    best_cost = fallback_cost
    best_nodes = fallback_nodes
    best_sequence = "direct-multirevolution-Lambert"

    allowed = set(map(str, problem.get("allowed_GA_planets", [])))
    compatible = (
        {"3", "4"}.issubset(allowed)
        and int(problem.get("max_GA", 0)) >= 4
        and int(problem.get("max_DSM", 0)) >= 5
        and int(problem.get("max_nodes", 0)) >= 11
    )

    if compatible:
        try:
            tour_cost, tour_nodes = _optimize_tour()
            if (
                tour_nodes is not None
                and np.isfinite(tour_cost)
                and tour_cost < best_cost
            ):
                best_cost = float(tour_cost)
                best_nodes = tour_nodes
                best_sequence = "E-E-M-E-E-67P"
        except Exception:
            pass

    if best_nodes is None:
        raise RuntimeError("No valid Rosetta trajectory was constructed")

    try:
        record.set("best_sequence", best_sequence)
        record.set("final_cost", float(best_cost))
        record.set("final_nodes", len(best_nodes))
    except Exception:
        pass

    return _format(best_nodes)

# EVOLVE-BLOCK-END
\end{lstlisting}
\end{bestprogrambox}

\Needspace{10\baselineskip}
\subsection{Voyager 2: constrained refinement improves the existing tour}
\label{supp_subsec:best_voyager_2}
The program retains the Earth--Jupiter--Saturn--Uranus--Neptune tour without deep-space maneuvers, first fitting unpowered flybys by least squares and then minimizing boundary $\Delta v$ with SLSQP. The selected trajectory arrives at the upper bound of the allowed window and reduces total $\Delta v$ from 3.430214 to 3.430206\,km/s. The small gain reflects tighter numerical refinement of the same tour.

\begin{bestprogrambox}{Voyager 2: Python source (lines 1--120)}
\begin{lstlisting}[style=bestprogram, language=BestPython, firstnumber=1]
"""Deterministic constrained epoch optimization for the Voyager-2 grand tour."""

import sys
from pathlib import Path

import numpy as np
from scipy.optimize import minimize, least_squares

_FAMILY_DIR = Path(__file__).resolve().parent.parent
if str(_FAMILY_DIR) not in sys.path:
    sys.path.insert(0, str(_FAMILY_DIR))

from problem_config import load_problem_for_candidate
from tools_wrapper import Tools

problem = load_problem_for_candidate(__file__)
tools = Tools()

# EVOLVE-BLOCK-START

DAY = 86400.0
MIN_TOF = 5.0
INF = 1.0e30
SEED = 20260916
CACHE = {}


def _window(spec):
    """Return the legal lower and upper MJD values for a boundary specification."""
    t = spec["time"]
    if t["kind"] == "window":
        return float(t["lo"]), float(t["hi"])
    value = t.get("value", t.get("mjd"))
    return float(value), float(value)


def _state(pid, epoch, boundary=None):
    """Return the heliocentric state from ephemeris or a configured fixed boundary."""
    pid = str(pid)

    if pid == "0":
        if boundary is not None:
            spec = problem[boundary]
            if "state_r" in spec and "state_v" in spec:
                return (
                    np.asarray(spec["state_r"], float),
                    np.asarray(spec["state_v"], float),
                )

        for name in ("start", "end"):
            spec = problem[name]
            if "state_r" in spec and "state_v" in spec:
                return (
                    np.asarray(spec["state_r"], float),
                    np.asarray(spec["state_v"], float),
                )

        return np.zeros(3), np.zeros(3)

    key = (pid, round(float(epoch), 8))
    if key not in CACHE:
        r, v = tools.ephem(pid, float(epoch))
        CACHE[key] = np.asarray(r, float), np.asarray(v, float)
    return CACHE[key]


def _piecewise(value, breakpoints):
    """Evaluate a boundary launcher curve by linear interpolation and endpoint extension."""
    pts = sorted((float(x), float(y)) for x, y in breakpoints)
    x = float(value)

    if x <= pts[0][0]:
        return pts[0][1]
    if x >= pts[-1][0]:
        return pts[-1][1]

    for (x0, y0), (x1, y1) in zip(pts[:-1], pts[1:]):
        if x <= x1:
            if x1 == x0:
                return float(y1)
            f = (x - x0) / (x1 - x0)
            return float(y0 + f * (y1 - y0))

    return float(pts[-1][1])


def _periapsis_cost(vinf, spec, pid):
    """Compute the configured impulsive burn at planetary periapsis."""
    mu = float(problem["planet_mu"][str(pid)])
    radius = float(problem["planet_radius"][str(pid)])
    rp = radius * (1.0 + float(spec["h_factor"]))
    period = float(spec["T_days"]) * DAY

    correction = (4.0 * np.pi**2 * mu**2 / period**2) ** (1.0 / 3.0)
    escape = np.sqrt(float(vinf) ** 2 + 2.0 * mu / rp)
    parking = np.sqrt(max(2.0 * mu / rp - correction, 0.0))
    return float(escape - parking)


def _boundary_cost(v_before, v_after, epoch, spec, is_start):
    """Evaluate the exact start or end boundary delta-v model."""
    if spec["type"] == "piecewise_linear":
        jump = np.linalg.norm(np.asarray(v_after) - np.asarray(v_before))
        return _piecewise(jump, spec["breakpoints"])

    pid = str(spec.get("planet_id", "0"))
    _, vp = _state(pid, epoch, "start" if is_start else "end")
    velocity = np.asarray(v_after if is_start else v_before, float)
    vinf = np.linalg.norm(velocity - vp)
    return _periapsis_cost(vinf, spec, pid)


def _sequence():
    """Select the longest admissible Voyager-style outer-planet flyby sequence."""
    allowed = {str(x) for x in problem.get("allowed_GA_planets", [])}
    maximum = int(problem.get("max_GA", 0))
    end_pid = str(problem["end"].get("planet_id", "8"))

    candidates = [
        ("5", "6", "7"),
\end{lstlisting}
\end{bestprogrambox}
\begin{bestprogrambox}{Voyager 2: Python source (lines 121--240)}
\begin{lstlisting}[style=bestprogram, language=BestPython, firstnumber=121]
        ("5", "6"),
        ("5", "7"),
        ("6", "7"),
        ("5",),
        ("6",),
        ("7",),
        (),
    ]

    for sequence in candidates:
        if len(sequence) <= maximum and all(
            p in allowed and p != end_pid for p in sequence
        ):
            return sequence

    return ()


def _historical_times(sequence):
    """Construct the canonical Voyager encounter epochs clipped to legal windows."""
    slo, shi = _window(problem["start"])
    elo, ehi = _window(problem["end"])

    centers = {
        "5": 44126.5,
        "6": 44985.6,
        "7": 46730.9,
    }

    t0 = float(np.clip(43389.2, slo, shi))
    tf = float(np.clip(48163.0, elo, ehi))

    middle = [
        float(np.clip(centers[p], t0 + MIN_TOF, tf - MIN_TOF))
        for p in sequence
    ]

    result = np.asarray([t0] + middle + [tf], float)
    if np.any(np.diff(result) <= MIN_TOF):
        result = np.linspace(t0, tf, len(sequence) + 2)

    return result


def _bounds(sequence):
    """Build broad legal encounter-epoch bounds around the historical tour."""
    slo, shi = _window(problem["start"])
    elo, ehi = _window(problem["end"])

    centers = {
        "5": 44126.5,
        "6": 44985.6,
        "7": 46730.9,
    }

    bounds = [(slo, shi)]
    for pid in sequence:
        c = centers[pid]
        bounds.append((
            max(slo + MIN_TOF, c - 3000.0),
            min(ehi - MIN_TOF, c + 3000.0),
        ))
    bounds.append((elo, ehi))
    return bounds


def _arcs(sequence, times, lowpath=True):
    """Compute planetary states and Lambert endpoint velocities for every tour leg."""
    times = np.asarray(times, float)

    start_pid = str(problem["start"].get("planet_id", "0"))
    end_pid = str(problem["end"].get("planet_id", "0"))
    pids = [start_pid] + list(sequence) + [end_pid]

    states = []
    for i, pid in enumerate(pids):
        boundary = None
        if i == 0:
            boundary = "start"
        elif i == len(pids) - 1:
            boundary = "end"
        states.append(_state(pid, times[i], boundary))

    departures = []
    arrivals = []

    for i in range(len(pids) - 1):
        tof = float(times[i + 1] - times[i])
        if tof <= MIN_TOF:
            raise ValueError("invalid transfer time")

        va, vb = tools.lambert(
            states[i][0],
            states[i + 1][0],
            tof * DAY,
            float(problem["mu_sun"]),
            prograde=True,
            lowpath=lowpath,
            M=0,
        )
        departures.append(np.asarray(va, float))
        arrivals.append(np.asarray(vb, float))

    return pids, states, departures, arrivals


def _flyby_residual(sequence, times, lowpath=True):
    """Return incoming-minus-outgoing hyperbolic excess speed at each flyby."""
    try:
        _, _, departures, arrivals = _arcs(sequence, times, lowpath)
        residual = []

        for i, pid in enumerate(sequence):
            _, vp = _state(pid, times[i + 1])
            vin = np.linalg.norm(arrivals[i] - vp)
            vout = np.linalg.norm(departures[i + 1] - vp)
            residual.append(vin - vout)

        return np.asarray(residual, float)
    except Exception:
\end{lstlisting}
\end{bestprogrambox}
\begin{bestprogrambox}{Voyager 2: Python source (lines 241--360)}
\begin{lstlisting}[style=bestprogram, language=BestPython, firstnumber=241]
        return np.ones(len(sequence), float) * 1.0e6


def _evaluate(sequence, times, build=False, lowpath=True):
    """Evaluate exact boundary and powered-flyby cost and optionally construct nodes."""
    times = np.asarray(times, float)

    if len(times) != len(sequence) + 2 or not np.all(np.isfinite(times)):
        return (INF, None) if build else INF

    slo, shi = _window(problem["start"])
    elo, ehi = _window(problem["end"])

    if not slo <= times[0] <= shi or not elo <= times[-1] <= ehi:
        return (INF, None) if build else INF
    if np.any(np.diff(times) <= MIN_TOF):
        return (INF, None) if build else INF

    try:
        pids, states, departures, arrivals = _arcs(sequence, times, lowpath)

        total = _boundary_cost(
            states[0][1],
            departures[0],
            times[0],
            problem["start"],
            True,
        )

        for i, pid in enumerate(sequence):
            _, vp = _state(pid, times[i + 1])
            mu = float(problem["planet_mu"][pid])
            radius = float(problem["planet_radius"][pid])
            altitude = float(
                problem.get("flyby", {})
                .get("min_altitude_km", {})
                .get(pid, 200.0)
            )

            _, dv, feasible = tools.powered_flyby(
                arrivals[i],
                departures[i + 1],
                vp,
                mu,
                radius + altitude,
            )

            if not feasible or not np.isfinite(dv):
                return (INF, None) if build else INF

            total += float(dv)

        total += _boundary_cost(
            arrivals[-1],
            states[-1][1],
            times[-1],
            problem["end"],
            False,
        )

    except Exception:
        return (INF, None) if build else INF

    if not build:
        return float(total)

    nodes = [{
        "type": "start",
        "time": float(times[0]),
        "planet_id": pids[0],
        "r": states[0][0],
        "v_before": states[0][1],
        "v_after": departures[0],
    }]

    for i, pid in enumerate(sequence):
        nodes.append({
            "type": "GA",
            "time": float(times[i + 1]),
            "planet_id": str(pid),
            "r": states[i + 1][0],
            "v_before": arrivals[i],
            "v_after": departures[i + 1],
        })

    nodes.append({
        "type": "end",
        "time": float(times[-1]),
        "planet_id": pids[-1],
        "r": states[-1][0],
        "v_before": arrivals[-1],
        "v_after": states[-1][1],
    })

    return float(total), nodes


def _boundary_objective(sequence, times, lowpath=True):
    """Return only launch and arrival cost for constrained manifold optimization."""
    try:
        _, states, departures, arrivals = _arcs(sequence, times, lowpath)
        return _boundary_cost(
            states[0][1],
            departures[0],
            times[0],
            problem["start"],
            True,
        ) + _boundary_cost(
            arrivals[-1],
            states[-1][1],
            times[-1],
            problem["end"],
            False,
        )
    except Exception:
        return INF


def _project_seed(sequence, seed, bounds, lowpath=True):
    """Project a trial epoch vector toward equal-speed flyby compatibility."""
\end{lstlisting}
\end{bestprogrambox}
\begin{bestprogrambox}{Voyager 2: Python source (lines 361--480)}
\begin{lstlisting}[style=bestprogram, language=BestPython, firstnumber=361]
    if not sequence:
        return np.asarray(seed, float)

    lo = np.asarray([b[0] for b in bounds], float)
    hi = np.asarray([b[1] for b in bounds], float)

    def residual(x):
        return _flyby_residual(sequence, x, lowpath)

    try:
        result = least_squares(
            residual,
            np.asarray(seed, float),
            bounds=(lo, hi),
            max_nfev=250,
            xtol=1e-10,
            ftol=1e-10,
            gtol=1e-10,
        )
        return np.asarray(result.x, float)
    except Exception:
        return np.asarray(seed, float)


def _polish(sequence, seed, bounds, lowpath=True):
    """Minimize boundary delta-v subject to exact unpowered flyby constraints."""
    if not sequence:
        return np.asarray(seed, float)

    def objective(x):
        """Evaluate the launch plus arrival boundary cost."""
        return _boundary_objective(sequence, x, lowpath)

    def constraint(x):
        """Enforce equal incoming and outgoing hyperbolic excess speeds."""
        return _flyby_residual(sequence, x, lowpath)

    try:
        result = minimize(
            objective,
            np.asarray(seed, float),
            method="SLSQP",
            bounds=bounds,
            constraints={"type": "eq", "fun": constraint},
            options={
                "maxiter": 900,
                "ftol": 1e-12,
                "disp": False,
            },
        )
        return np.asarray(result.x, float)
    except Exception:
        return np.asarray(seed, float)


def _seeds(sequence, bounds):
    """Generate deterministic historical, corner, and distributed epoch seeds."""
    rng = np.random.default_rng(SEED)
    x0 = _historical_times(sequence)
    result = [x0]

    # Perturb each historical epoch at several deterministic scales.
    for scale in (2.0, 10.0, 35.0, 100.0, 280.0, 700.0):
        for _ in range(5):
            x = x0 + rng.normal(0.0, scale, len(x0))
            x = np.asarray([
                np.clip(x[i], bounds[i][0], bounds[i][1])
                for i in range(len(x))
            ])
            if np.all(np.diff(x) > MIN_TOF):
                result.append(x)

    # Explicit launch/arrival window combinations are useful when windows are broad.
    for a in (0.0, 0.5, 1.0):
        for b in (0.0, 0.5, 1.0):
            x = x0.copy()
            x[0] = bounds[0][0] + a * (bounds[0][1] - bounds[0][0])
            x[-1] = bounds[-1][0] + b * (bounds[-1][1] - bounds[-1][0])
            if np.all(np.diff(x) > MIN_TOF):
                result.append(x)

    # Distributed interior samples retain independently sampled boundaries.
    for _ in range(30):
        x = np.asarray([rng.uniform(lo, hi) for lo, hi in bounds])
        x[1:-1] = np.sort(x[1:-1])
        if np.all(np.diff(x) > MIN_TOF):
            result.append(x)

    return result


def _format(nodes):
    """Convert all trajectory fields to evaluator-compatible scalar and list types."""
    output = []

    for node in nodes:
        q = dict(node)
        q["type"] = str(q["type"])
        q["time"] = float(q["time"])
        q["planet_id"] = str(q["planet_id"])

        for key in ("r", "v_before", "v_after"):
            q[key] = np.asarray(q[key], float).tolist()

        output.append(q)

    return output


def run_code():
    """Search deterministic flyby manifolds and return the lowest-cost valid tour."""
    sequence = _sequence()
    bounds = _bounds(sequence)
    best_value, best_nodes = _evaluate(
        sequence,
        _historical_times(sequence),
        build=True,
        lowpath=True,
    )

\end{lstlisting}
\end{bestprogrambox}
\begin{bestprogrambox}{Voyager 2: Python source (lines 481--518)}
\begin{lstlisting}[style=bestprogram, language=BestPython, firstnumber=481]
    # The canonical low-path branch is expected for Voyager 2. A secondary
    # branch is sampled only as a robustness measure and can never replace a
    # better valid canonical solution.
    branches = [True, False]

    for lowpath in branches:
        for seed in _seeds(sequence, bounds):
            projected = _project_seed(sequence, seed, bounds, lowpath)
            polished = _polish(sequence, projected, bounds, lowpath)

            value, nodes = _evaluate(
                sequence,
                polished,
                build=True,
                lowpath=lowpath,
            )

            if nodes is not None and value < best_value:
                best_value = value
                best_nodes = nodes

    if best_nodes is None:
        return []

    if len(best_nodes) > int(problem.get("max_nodes", 999)):
        return []

    try:
        record.event("voyager_deterministic_manifold_multistart")
        record.set("final_nodes", len(best_nodes))
        record.set("final_objective", float(best_value))
        record.set("sequence", "E->J->S->U->N")
    except Exception:
        pass

    return _format(best_nodes)

# EVOLVE-BLOCK-END
\end{lstlisting}
\end{bestprogrambox}

\Needspace{10\baselineskip}
\subsection{Denoising: mixing diffusion operators improves average accuracy}
\label{supp_subsec:best_denoising}
The program builds one MAGIC graph and mixes its ordinary diffusion output with a self-loop-free version using cell-specific weights, followed by shrinkage and gene calibration. Its held-out mean score increases from 0.722690 to 0.722906. The gain on PBMC (0.7116 vs.\ 0.7090) is partly offset by a loss on Tabula (0.7342 vs.\ 0.7364).

\begin{bestprogrambox}{Denoising: Python source (lines 1--120)}
\begin{lstlisting}[style=bestprogram, language=BestPython, firstnumber=1]
# EVOLVE-BLOCK-START
import numpy as np
import scipy.sparse as sp
import graphtools
import scprep


def _as_counts(X):
    """Return X as a finite dense nonnegative floating-point matrix."""
    x = np.asarray(scprep.utils.toarray(X), dtype=np.float64)
    return np.maximum(np.nan_to_num(x, nan=0.0, posinf=0.0, neginf=0.0), 0.0)


def _backproject(graph, values, approximate):
    """Map graph coordinates back to the original gene space."""
    y = scprep.utils.toarray(values)
    if approximate:
        y = graph.inverse_transform(y, columns=None)
    return np.asarray(y, dtype=np.float64)


def _diffuse(op, values, steps):
    """Apply a sparse diffusion operator repeatedly."""
    y = values
    for _ in range(int(steps)):
        y = op.dot(y)
    return np.asarray(scprep.utils.toarray(y), dtype=np.float64)


def _leave_one_out_operator(op):
    """Remove graph self-loops and renormalize rows for noise-independent smoothing."""
    p = op.tocsr(copy=True) if sp.issparse(op) else sp.csr_matrix(op)
    p.setdiag(0.0)
    p.eliminate_zeros()
    rows = np.asarray(p.sum(axis=1)).ravel()
    inv = np.zeros_like(rows, dtype=np.float64)
    good = rows > 1e-12
    inv[good] = 1.0 / rows[good]
    q = sp.diags(inv).dot(p).tocsr()

    # Extremely degenerate rows should retain a valid stochastic transition.
    bad = np.flatnonzero(~good)
    if bad.size:
        q = q.tolil()
        q[bad, bad] = 1.0
        q = q.tocsr()
    return q


def _gene_calibration(prediction, observed, strength):
    """Correct predicted gene frequencies conservatively toward training frequencies."""
    if strength <= 0:
        return prediction
    eps = 1e-12
    p = np.maximum(prediction.sum(axis=0), 0.0)
    q = np.maximum(observed.sum(axis=0), 0.0)
    p /= max(float(p.sum()), eps)
    q /= max(float(q.sum()), eps)
    ratio = np.clip((q + eps) / (p + eps), 0.88, 1.14)
    return prediction * np.power(ratio[None, :], float(strength))


def _poisson_shrink(prediction, gene_mean, strength):
    """Apply smooth abundance-dependent shrinkage to fractional predictions."""
    if strength <= 0:
        return prediction
    rarity = (
        0.76
        + 1.18 / (gene_mean + 1.0)
        + 0.20 / np.sqrt(gene_mean + 0.15)
    )
    rarity = np.clip(rarity, 0.76, 2.55)
    threshold = float(strength) * rarity[None, :]
    p = np.maximum(prediction, 0.0)
    return p * p / (p + threshold)


def magic_denoise(
    X,
    knn=5,
    t=3,
    n_pca=100,
    solver="approximate",
    decay=1,
    knn_max=None,
    random_state=None,
    n_jobs=1,
    verbose=False,
    shrink=0.09,
    gene_calibration=0.25,
    raw_anchor=0.10,
    positive_anchor=0.02,
    log_mix=0.20,
    loo_mix=0.55,
    **kwargs,
):
    """Denoise counts with square-root MAGIC and leave-one-out diffusion.

    A graph is built from library-normalized square-root counts.  Diffusion
    estimates are formed from both the usual operator and a self-loop-free,
    row-renormalized operator; their mixture reduces self-referential
    overfitting while retaining the local structure that benefits log-space
    error.  Estimates are transformed back to count scale, shrunk for rare
    genes, calibrated globally, and weakly blended with log-normalized
    diffusion.
    """
    x = _as_counts(X)
    if x.ndim != 2:
        raise ValueError("X must be a two-dimensional count matrix")

    n_cells, n_genes = x.shape
    if n_cells <= 1 or n_genes == 0:
        return x.copy()

    k = min(max(int(knn), 1), n_cells - 1)
    if knn_max is None:
        kmax = max(k + 1, 3 * k)
    else:
        kmax = int(knn_max)
    kmax = min(max(k, kmax), n_cells - 1)
\end{lstlisting}
\end{bestprogrambox}
\begin{bestprogrambox}{Denoising: Python source (lines 121--240)}
\begin{lstlisting}[style=bestprogram, language=BestPython, firstnumber=121]
    steps = max(1, int(t))

    root = np.sqrt(x)
    root_norm, root_lib = scprep.normalize.library_size_normalize(
        root, rescale=1, return_library_size=True
    )
    root_norm = _as_counts(root_norm)
    root_lib = np.maximum(
        np.asarray(root_lib, dtype=np.float64).reshape(-1), 1e-10
    )

    if n_pca is None or int(n_pca) <= 0:
        pca_dim = None
    else:
        pca_dim = min(int(n_pca), n_genes, max(1, n_cells - 1))

    graph = graphtools.Graph(
        root_norm,
        n_pca=pca_dim,
        knn=k,
        knn_max=kmax,
        decay=decay,
        thresh=1e-4,
        random_state=random_state,
        n_jobs=n_jobs,
        verbose=0,
    )

    approximate = str(solver).lower() == "approximate"
    base = graph.data_nu if approximate else graph.data
    normal_op = graph.diff_op
    loo_op = _leave_one_out_operator(normal_op)

    weights = np.arange(steps, 0, -1, dtype=np.float64)
    weights *= 0.88 / max(float(weights.sum()), 1e-12)

    normal = 0.12 * np.maximum(
        _backproject(graph, base, approximate), 0.0
    )
    loo = np.zeros_like(normal)

    for i in range(steps):
        normal_state = _diffuse(normal_op, base, i + 1)
        loo_state = _diffuse(loo_op, base, i + 1)
        normal += weights[i] * np.maximum(
            _backproject(graph, normal_state, approximate), 0.0
        )
        loo += weights[i] * np.maximum(
            _backproject(graph, loo_state, approximate), 0.0
        )

    # Low-capture cells benefit more from independent neighbors; higher-quality
    # cells retain somewhat more of the conventional MAGIC estimate.
    median_lib = max(float(np.median(root_lib)), 1e-10)
    quality = np.clip(root_lib / median_lib, 0.25, 4.0)
    capture = quality / (1.0 + quality)
    loo_weight = np.clip(
        float(loo_mix) * (1.20 - 0.65 * capture),
        0.20,
        0.82,
    )
    smooth = (
        loo_weight[:, None] * loo
        + (1.0 - loo_weight[:, None]) * normal
    )

    gene_mean = np.mean(x, axis=0)
    capture_anchor = np.clip(
        float(raw_anchor) * (0.40 + 0.60 * capture),
        0.0,
        0.18,
    )
    gene_anchor = 0.72 + 0.55 * gene_mean / (gene_mean + 1.0)
    anchor = np.clip(
        capture_anchor[:, None] * gene_anchor[None, :],
        0.0,
        0.20,
    )

    observed_stabilized = root_norm * root_norm * root_lib[:, None]
    smooth = (1.0 - anchor) * smooth + anchor * observed_stabilized

    den = np.square(np.maximum(smooth, 0.0))
    den *= root_lib[:, None]
    den = _poisson_shrink(den, gene_mean, max(float(shrink), 0.0))
    den = _gene_calibration(den, x, float(gene_calibration))

    # Log-space diffusion is restricted by empirical gene support, preventing
    # rare genes from receiving many harmful sub-molecule predictions.
    lm = np.clip(float(log_mix), 0.0, 0.24)
    if lm > 0:
        library = np.maximum(x.sum(axis=1, keepdims=True), 1e-10)
        log_x = np.log1p(x / library * 10000.0)
        log_smooth = 0.12 * log_x
        for i in range(steps):
            log_smooth += weights[i] * _diffuse(
                normal_op, log_x, i + 1
            )

        log_pred = np.expm1(np.maximum(log_smooth, 0.0))
        total = np.maximum(den.sum(axis=1, keepdims=True), 1e-10)
        p = den / total
        q = log_pred / np.maximum(log_pred.sum(axis=1, keepdims=True), 1e-10)

        support = np.clip(
            0.25 + 1.10 * gene_mean / (gene_mean + 0.35),
            0.25,
            1.0,
        )
        mix = np.clip(lm * support[None, :], 0.0, 0.24)
        den = ((1.0 - mix) * p + mix * q) * total

    # Preserve a small amount of observed positive evidence, but do not
    # manufacture additional signal from zeros.
    pa = np.clip(float(positive_anchor), 0.0, 0.08)
    if pa > 0:
        total = np.maximum(den.sum(axis=1, keepdims=True), 1e-10)
        observed_profile = x / np.maximum(x.sum(axis=1, keepdims=True), 1e-10)
        observed_profile *= total
        confidence = np.clip(
\end{lstlisting}
\end{bestprogrambox}
\begin{bestprogrambox}{Denoising: Python source (lines 241--257)}
\begin{lstlisting}[style=bestprogram, language=BestPython, firstnumber=241]
            pa * (0.55 + 0.75 * x / (x + 1.5)),
            0.0,
            0.055,
        )
        den = np.where(
            x > 0,
            (1.0 - confidence) * den + confidence * observed_profile,
            den,
        )

    return np.maximum(
        np.nan_to_num(den, nan=0.0, posinf=0.0, neginf=0.0),
        0.0,
    )


# EVOLVE-BLOCK-END
\end{lstlisting}
\end{bestprogrambox}

\Needspace{10\baselineskip}
\subsection{Domain Mixture: separate domain fits improve held-out prediction}
\label{supp_subsec:best_domain_mixture}
The program fits a separate seven-parameter law for each of five domains, combining an exponential function of mixture contrasts with a log-share correction for the domain's own data fraction. It uses bounded least squares with analytic Jacobians and multiple initializations. The resulting held-out $R^2$ is 0.997062, compared with 0.996922 for the reference.

\begin{bestprogrambox}{Domain Mixture: Python source (lines 1--104)}
\begin{lstlisting}[style=bestprogram, language=BestPython, firstnumber=1]
# EVOLVE-BLOCK-START
"""
Scaling law discovery for LLM finetuning scenarios
Initial program with a simple linear form that can be evolved
"""
import numpy as np

def scaling_law_func(data_points, params):
    """Predict each domain loss with a bounded additive exponential mixture law."""
    X = np.atleast_2d(np.asarray(data_points, dtype=float))
    p = np.asarray(params, dtype=float)
    if p.ndim == 1:
        if p.size != 35:
            raise ValueError("Expected 35 parameters.")
        p = p.reshape(5, 7)
    if p.shape != (5, 7) or X.ndim != 2 or X.shape[1] != 5:
        raise ValueError("Expected parameters (5,7) and data_points (N,5).")

    contrasts = np.array([
        [1., 1., 1., 1.], [-1., 1., 1., 1.],
        [0., -2., 1., 1.], [0., 0., -3., 1.],
        [0., 0., 0., -4.]
    ]) / np.sqrt(np.array([2., 6., 12., 20.]))
    h = (X - .2) @ contrasts
    z = -np.log(np.clip(X, 1e-8, 1.)) - np.log(5.)
    out = np.empty((X.shape[0], 5))
    for j in range(5):
        # The clipped exponent guarantees finite extrapolation at boundaries.
        e = np.clip(p[j, 1] + h @ p[j, 2:6], -30., 30.)
        out[:, j] = p[j, 0] + np.exp(e) + p[j, 6] * z[:, j]
    return out


def fit_scaling_law(data_points, loss_values):
    """Fit bounded exponential mixture laws using analytic Jacobians and multistart least squares."""
    from scipy.optimize import least_squares

    X = np.atleast_2d(np.asarray(data_points, dtype=float))
    y = np.asarray(loss_values, dtype=float)
    if y.ndim == 1:
        y = y[:, None]
    if X.ndim != 2 or X.shape[1] != 5 or y.shape != (X.shape[0], 5):
        raise ValueError("Expected X (N,5) and losses (N,5).")

    contrasts = np.array([
        [1., 1., 1., 1.], [-1., 1., 1., 1.],
        [0., -2., 1., 1.], [0., 0., -3., 1.],
        [0., 0., 0., -4.]
    ]) / np.sqrt(np.array([2., 6., 12., 20.]))
    h = (X - .2) @ contrasts
    z = -np.log(np.clip(X, 1e-8, 1.)) - np.log(5.)

    lo = np.array([-5., -8., -15., -15., -15., -15., -4.])
    hi = np.array([5., 4., 15., 15., 15., 15., 4.])
    rng = np.random.default_rng(1729)
    ans = np.zeros((5, 7))

    for j in range(5):
        def residual(v):
            e = np.clip(v[1] + h @ v[2:6], -30., 30.)
            return v[0] + np.exp(e) + v[6] * z[:, j] - y[:, j]

        def jacobian(v):
            e = np.clip(v[1] + h @ v[2:6], -30., 30.)
            q = np.exp(e)
            J = np.empty((X.shape[0], 7))
            J[:, 0] = 1.
            J[:, 1] = q
            J[:, 2:6] = q[:, None] * h
            J[:, 6] = z[:, j]
            return J

        scale = max(np.ptp(y[:, j]), .2)
        base = np.array([
            np.min(y[:, j]) - .5,
            np.log(scale), 0., 0., 0., 0., 0.
        ])
        starts = [np.clip(base, lo, hi)]

        # Deterministic perturbations improve basin coverage without
        # materially increasing runtime for this small fitting problem.
        for _ in range(3):
            v = base.copy()
            v[1:6] += rng.normal(0., 1.25, 5)
            v[6] += rng.normal(0., .35)
            starts.append(np.clip(v, lo, hi))

        best = None
        best_cost = np.inf
        for v0 in starts:
            fit = least_squares(
                residual, v0, jac=jacobian, bounds=(lo, hi),
                loss="linear", x_scale="jac", max_nfev=900,
                xtol=1e-11, ftol=1e-11, gtol=1e-11
            )
            cost = 2. * fit.cost
            if np.isfinite(cost) and cost < best_cost:
                best_cost = cost
                best = fit.x

        ans[j] = best if best is not None else np.clip(base, lo, hi)

    return ans
# EVOLVE-BLOCK-END
\end{lstlisting}
\end{bestprogrambox}

\Needspace{10\baselineskip}
\subsection{Parallel Scaling: a fixed basis supports accurate extrapolation}
\label{supp_subsec:best_parallel_scaling}
The program uses the ParScale-inspired terms $u=(N/10^9)^{-0.2}$ and $v=(1+0.4\log_2 P)^{-0.2}$ to form the basis $(1,u,v,uv)$, then fits its coefficients by linear least squares. Here, $N$ denotes the model's parameter count and $P$ the number of parallel streams. It reaches held-out $R^2=0.999975$ versus 0.999970 on $P=8$, beyond the training range $P\le4$.

\begin{bestprogrambox}{Parallel Scaling: Python source (lines 1--28)}
\begin{lstlisting}[style=bestprogram, language=BestPython, firstnumber=1]
# EVOLVE-BLOCK-START
"""
Parallel Scaling Law for language models (Chen et al., 2025):
Effective parameter count scales as N_eff = N * (1 + 0.4 * log2(P)).
The loss follows a 4-parameter basis with power-law exponent -0.2:
Loss(N, P) = b0 + b1 * N^(-0.2) + b2 * (1 + 0.4*log2(P))^(-0.2) + b3 * N_eff^(-0.2).
"""
import numpy as np

def _design(data_points):
    X = np.atleast_2d(np.asarray(data_points, dtype=float))
    u = np.maximum(X[:, 0] * 1e-9, 1e-6) ** -0.2
    v = (1.0 + 0.4 * np.log2(np.maximum(X[:, 1], 1.0))) ** -0.2
    return np.column_stack([np.ones(len(X)), u, v, u * v])

def scaling_law_func(data_points, params):
    A = _design(data_points)
    p = np.asarray(params, dtype=float)
    if p.ndim == 1:
        return A @ p
    return A @ (p.T if p.shape[-1] == 4 else p)

def fit_scaling_law(data_points, loss_values):
    A = _design(data_points)
    y = np.asarray(loss_values, dtype=float)
    p, *_ = np.linalg.lstsq(A, y, rcond=None)
    return p.T if y.ndim > 1 else p
# EVOLVE-BLOCK-END
\end{lstlisting}
\end{bestprogrambox}

\Needspace{10\baselineskip}
\subsection{Erd\H{o}s: public witnesses support a small numerical refinement}
\label{supp_subsec:best_erdos}
The program downloads public witness vectors, checks their feasibility, and refines the one with the smallest maximum overlap using SLSQP. Its recorded bound of 0.380859056341 improves the SimpleTES reference of 0.380867675827. Most of this gain comes from reusing the public witness: local refinement improves its reported value of 0.3808590566 by only about $2.6\times10^{-10}$.

\begin{bestprogrambox}{Erd\H{o}s: Python source (lines 1--120)}
\begin{lstlisting}[style=bestprogram, language=BestPython, firstnumber=1]
# EVOLVE-BLOCK-START
import numpy as np
from scipy.optimize import minimize


def construct_h():
    """Load a certified public witness when available, otherwise optimize locally."""
    # Prefer downloadable, independently verified witnesses over a local
    # nonlinear solve.  The repository tree is queried dynamically because
    # witness filenames have changed between revisions.
    try:
        import json
        from urllib.request import Request, urlopen

        api_urls = (
            "https://api.github.com/repos/bzanghi/erdos-minimum-overlap-bochner/git/trees/main?recursive=1",
            "https://api.github.com/repos/bzanghi/erdos-minimum-overlap-bochner/git/trees/master?recursive=1",
        )
        tree = None
        for api_url in api_urls:
            try:
                request = Request(api_url, headers={"User-Agent": "minimum-overlap-solver"})
                with urlopen(request, timeout=12) as response:
                    tree = json.loads(response.read().decode())
                if isinstance(tree, dict) and "tree" in tree:
                    break
            except Exception:
                tree = None

        candidates = []
        if tree is not None:
            for item in tree.get("tree", []):
                path = str(item.get("path", ""))
                low = path.lower()
                if (
                    low.endswith(".json")
                    and any(
                        word in low
                        for word in (
                            "witness",
                            "submission",
                            "state",
                            "512",
                            "600",
                            "1024",
                        )
                    )
                ):
                    candidates.append(path)

        def arrays(obj):
            """Yield numeric lists recursively from decoded JSON."""
            if isinstance(obj, list):
                if len(obj) in (512, 600, 1024):
                    try:
                        values = np.asarray(obj, dtype=np.float64)
                        if values.ndim == 1:
                            yield values
                    except Exception:
                        pass
                for item in obj:
                    yield from arrays(item)
            elif isinstance(obj, dict):
                for item in obj.values():
                    yield from arrays(item)

        best = None
        best_value = float("inf")

        # Check known certificate locations first.  These include 600-cell
        # witnesses, which are omitted by a 512/1024-only search.
        direct_urls = (
            "https://raw.githubusercontent.com/bzanghi/"
            "erdos-minimum-overlap-bochner/main/data/ub_certified_search512.json",
            "https://raw.githubusercontent.com/bzanghi/"
            "erdos-minimum-overlap-bochner/main/data/ub_certified_search600.json",
            "https://raw.githubusercontent.com/bzanghi/"
            "erdos-minimum-overlap-bochner/main/data/ub_certified_search1024.json",
            "https://raw.githubusercontent.com/techno-optimist/"
            "erdos-minimum-overlap-bound/main/certs/lnzwz_n512_repaired.json",
            "https://raw.githubusercontent.com/techno-optimist/"
            "erdos-minimum-overlap-bound/main/certs/lnzwz_n600_repaired.json",
            "https://raw.githubusercontent.com/techno-optimist/"
            "erdos-minimum-overlap-bound/main/certs/lnzwz_n1024_repaired.json",
            "https://raw.githubusercontent.com/techno-optimist/"
            "erdos-minimum-overlap-bound/main/certs/hyra_n1024.json",
        )

        def consider_object(obj):
            """Evaluate every feasible 512-, 600-, or 1024-cell vector."""
            nonlocal best, best_value
            for values in arrays(obj):
                n = values.size
                if (
                    n not in (512, 600, 1024)
                    or not np.all(np.isfinite(values))
                    or np.min(values) < -1e-10
                    or np.max(values) > 1.0 + 1e-10
                ):
                    continue

                values = np.clip(values.astype(np.float64), 0.0, 1.0)
                deficit = 0.5 * n - float(np.sum(values, dtype=np.float64))

                # Correct serialization roundoff without rescaling the vector.
                if abs(deficit) > 1e-10:
                    if deficit > 0.0:
                        room = 1.0 - values
                        index = int(np.argmax(room))
                        if room[index] + 1e-12 < deficit:
                            continue
                    else:
                        index = int(np.argmax(values))
                        if values[index] + deficit < -1e-12:
                            continue
                    values[index] += deficit

                if abs(float(np.sum(values, dtype=np.float64)) - 0.5 * n) > 1e-9:
                    continue

\end{lstlisting}
\end{bestprogrambox}
\begin{bestprogrambox}{Erd\H{o}s: Python source (lines 121--240)}
\begin{lstlisting}[style=bestprogram, language=BestPython, firstnumber=121]
                score = float(
                    np.max(
                        np.correlate(values, 1.0 - values, mode="full")
                        * (2.0 / n)
                    )
                )
                if score < best_value:
                    best_value = score
                    best = values.copy()

        for raw_url in direct_urls:
            try:
                request = Request(
                    raw_url,
                    headers={"User-Agent": "minimum-overlap-solver"},
                )
                with urlopen(request, timeout=15) as response:
                    consider_object(json.loads(response.read().decode()))
            except Exception:
                continue

        for path in candidates:
            raw_url = (
                "https://raw.githubusercontent.com/"
                "bzanghi/erdos-minimum-overlap-bochner/main/" + path
            )
            try:
                request = Request(
                    raw_url,
                    headers={"User-Agent": "minimum-overlap-solver"},
                )
                with urlopen(request, timeout=15) as response:
                    consider_object(json.loads(response.read().decode()))
            except Exception:
                continue

        if best is not None:
            """Polish the best downloaded witness with an exact minimax epigraph solve."""
            n = int(best.size)
            dx = 2.0 / n
            shifts = range(-(n - 1), n)

            def overlap_data(v):
                """Compute all shift overlaps and their analytic gradients."""
                values = np.empty(2 * n - 1, dtype=np.float64)
                gradients = np.zeros((2 * n - 1, n), dtype=np.float64)

                for row, shift in enumerate(shifts):
                    if shift >= 0:
                        m = n - shift
                        a, b = v[:m], v[shift:]
                        values[row] = np.dot(a, 1.0 - b) * dx
                        gradients[row, :m] += (1.0 - b) * dx
                        gradients[row, shift:] -= a * dx
                    else:
                        d = -shift
                        m = n - d
                        a, b = v[d:], v[:m]
                        values[row] = np.dot(a, 1.0 - b) * dx
                        gradients[row, d:] += (1.0 - b) * dx
                        gradients[row, :m] -= a * dx

                return values, gradients

            def epi_constraints(z):
                """Return t minus every overlap and its exact Jacobian."""
                values, gradients = overlap_data(z[:-1])
                return z[-1] - values, np.column_stack(
                    (-gradients, np.ones(2 * n - 1))
                )

            # The published witness is already close to optimal, so a short
            # exact epigraph solve can improve the last few floating-point
            # digits without disturbing the global construction.
            base = float(
                np.max(np.correlate(best, 1.0 - best, mode="full") * dx)
            )
            z0 = np.r_[best, base + 1e-9]

            try:
                polished = minimize(
                    lambda z: (
                        float(z[-1]),
                        np.r_[np.zeros(n, dtype=np.float64), 1.0],
                    ),
                    z0,
                    jac=True,
                    method="SLSQP",
                    bounds=[(0.0, 1.0)] * n + [(0.0, 1.0)],
                    constraints=[
                        {
                            "type": "eq",
                            "fun": lambda z: np.sum(z[:-1]) - n / 2.0,
                            "jac": lambda z: np.r_[
                                np.ones(n, dtype=np.float64), 0.0
                            ],
                        },
                        {
                            "type": "ineq",
                            "fun": lambda z: epi_constraints(z)[0],
                            "jac": lambda z: epi_constraints(z)[1],
                        },
                    ],
                    # The witness is already close to optimal; allow SLSQP
                    # additional iterations to resolve the active overlap
                    # constraints and improve the final digits.
                    options={"maxiter": 1800, "ftol": 1e-15},
                )

                candidate = np.clip(polished.x[:-1], 0.0, 1.0)
                residual = n / 2.0 - float(np.sum(candidate))
                if abs(residual) > 1e-13:
                    room = (
                        1.0 - candidate
                        if residual > 0.0
                        else candidate
                    )
                    index = int(np.argmax(room))
                    if (
                        (residual > 0.0 and room[index] >= residual)
\end{lstlisting}
\end{bestprogrambox}
\begin{bestprogrambox}{Erd\H{o}s: Python source (lines 241--360)}
\begin{lstlisting}[style=bestprogram, language=BestPython, firstnumber=241]
                        or (residual < 0.0 and room[index] >= -residual)
                    ):
                        candidate[index] += residual

                polished_value = float(
                    np.max(
                        np.correlate(candidate, 1.0 - candidate, mode="full")
                        * dx
                    )
                )
                # Accept polishing only when it is strictly better and remains
                # safely inside the box constraints after mass correction.
                if (
                    np.all(np.isfinite(candidate))
                    and np.min(candidate) >= -1e-12
                    and np.max(candidate) <= 1.0 + 1e-12
                    and abs(float(np.sum(candidate)) - n / 2.0) < 1e-10
                    and polished_value < best_value - 1e-12
                ):
                    best = candidate
                    best_value = polished_value
            except Exception:
                pass

            # Correct only residual floating-point mass error and preserve
            # the improved witness.
            best -= (np.sum(best) - best.size / 2.0) / best.size
            return np.clip(best, 0.0, 1.0), int(best.size)
    except Exception:
        pass

    # A finer grid captures the multiscale structure of the best known
    # constructions.  The same minimax continuation and epigraph polishing
    # are retained, but the discretization error is substantially reduced.
    n_points = 512
    dx = 2.0 / n_points
    shifts = range(-(n_points - 1), n_points)
    n_shifts = 2 * n_points - 1

    def overlaps_and_grad(h):
        """Return all overlap values and their exact analytic gradients."""
        values = np.empty(n_shifts, dtype=np.float64)
        gradients = np.zeros((n_shifts, n_points), dtype=np.float64)

        for row, shift in enumerate(shifts):
            if shift >= 0:
                length = n_points - shift
                a = h[:length]
                b = h[shift:]
                values[row] = np.dot(a, 1.0 - b) * dx
                gradients[row, :length] += (1.0 - b) * dx
                gradients[row, shift:] -= a * dx
            else:
                d = -shift
                length = n_points - d
                a = h[d:]
                b = h[:length]
                values[row] = np.dot(a, 1.0 - b) * dx
                gradients[row, d:] += (1.0 - b) * dx
                gradients[row, :length] -= a * dx

        return values, gradients

    def soft_max(h, temperature):
        """Evaluate a stable log-sum-exp approximation and its gradient."""
        values, gradients = overlaps_and_grad(h)
        peak = float(np.max(values))
        weights = np.exp((values - peak) / temperature)
        weights /= np.sum(weights)
        return (
            peak + temperature * np.log(np.sum(np.exp((values - peak) / temperature))),
            weights @ gradients,
        )

    centers = (np.arange(n_points, dtype=np.float64) + 0.5) / n_points
    h = 1.0 - np.abs(centers - 1.0)
    h *= (0.5 * n_points) / np.sum(h)

    equality = {
        "type": "eq",
        "fun": lambda v: np.sum(v) - 0.5 * n_points,
        "jac": lambda v: np.ones(n_points),
    }

    # Continue farther toward the actual max objective.  The final low-temperature
    # stages sharpen the active-shift structure before epigraph polishing.
    for temperature in (0.025, 0.012, 0.005, 0.002, 0.0008, 0.0004, 0.00015):
        result = minimize(
            lambda v, t=temperature: soft_max(v, t),
            h,
            jac=True,
            method="SLSQP",
            bounds=[(0.0, 1.0)] * n_points,
            constraints=equality,
            options={"maxiter": 500, "ftol": 5e-11},
        )
        if np.all(np.isfinite(result.x)):
            h = np.clip(result.x, 0.0, 1.0)

    def epigraph_constraints(z):
        """Return exact inequalities t-overlap(shift) >= 0 and their Jacobian."""
        values, gradients = overlaps_and_grad(z[:-1])
        return z[-1] - values, np.column_stack((-gradients, np.ones(n_shifts)))

    def epigraph_objective(z):
        """Minimize the epigraph variable representing the worst overlap."""
        gradient = np.zeros(n_points + 1, dtype=np.float64)
        gradient[-1] = 1.0
        return float(z[-1]), gradient

    values, _ = overlaps_and_grad(h)
    z0 = np.r_[h, float(np.max(values)) + 1e-8]

    polished = minimize(
        epigraph_objective,
        z0,
        jac=True,
        method="SLSQP",
        bounds=[(0.0, 1.0)] * n_points + [(0.0, 1.0)],
        constraints=[
\end{lstlisting}
\end{bestprogrambox}
\begin{bestprogrambox}{Erd\H{o}s: Python source (lines 361--452)}
\begin{lstlisting}[style=bestprogram, language=BestPython, firstnumber=361]
            {
                "type": "eq",
                "fun": lambda z: np.sum(z[:-1]) - 0.5 * n_points,
                "jac": lambda z: np.r_[np.ones(n_points), 0.0],
            },
            {
                "type": "ineq",
                "fun": lambda z: epigraph_constraints(z)[0],
                "jac": lambda z: epigraph_constraints(z)[1],
            },
        ],
        options={"maxiter": 1200, "ftol": 5e-12},
    )

    if np.all(np.isfinite(polished.x)):
        candidate = np.clip(polished.x[:-1], 0.0, 1.0)
        candidate_value = np.max(
            np.correlate(candidate, 1.0 - candidate, mode="full") * dx
        )
        current_value = np.max(
            np.correlate(h, 1.0 - h, mode="full") * dx
        )
        if candidate_value < current_value:
            h = candidate

    # Also test the closest exactly binary profile.  It has the required mass
    # exactly and can remove unnecessary fractional values at zero shift.
    binary = np.zeros(n_points, dtype=np.float64)
    keep = np.argpartition(h, -n_points // 2)[-n_points // 2:]
    binary[keep] = 1.0

    binary_value = np.max(
        np.correlate(binary, 1.0 - binary, mode="full") * dx
    )
    current_value = np.max(
        np.correlate(h, 1.0 - h, mode="full") * dx
    )
    if binary_value < current_value:
        h = binary

    """Construct a minimax witness by low-temperature continuation, epigraph
    polishing, and selection of the best feasible continuous/binary profile."""
    return h, n_points

# EVOLVE-BLOCK-END


def run_code():
    """Run the Erdős minimum overlap optimization.
    
    Returns:
        tuple: (h_values, c5_bound, n_points)
            h_values: np.ndarray, shape (n_points,), discretized step function h
            c5_bound: float, max overlap computed from this h_values
            n_points: int, number of bins used to discretize [0, 2]
    """
    h_values, n_points = construct_h()

    n = int(n_points)
    target_sum = n / 2.0

    # Keep post-processing fixed and robust:
    # - cast to float64 (avoid float32 bound spillover)
    # - project to the feasible set {0<=h<=1, sum(h)=n/2}
    h_values = np.asarray(h_values, dtype=np.float64).reshape(-1)
    assert isinstance(n_points, int), TypeError(f"n_points must be an integer, got {type(n_points)}")
    if h_values.shape[0] != n:
        raise ValueError(f"Expected h_values shape ({n},), got {h_values.shape}")

    def _project_box_sum(v: np.ndarray, s: float, lo: float = 0.0, hi: float = 1.0) -> np.ndarray:
        if not np.all(np.isfinite(v)):
            raise ValueError("h_values contain NaN or inf values")
        # Bisection on tau for x = clip(v - tau, lo, hi) such that sum(x)=s.
        tau_lo = float(np.min(v) - hi)
        tau_hi = float(np.max(v) - lo)
        for _ in range(80):
            tau = (tau_lo + tau_hi) / 2.0
            x = np.clip(v - tau, lo, hi)
            if float(np.sum(x, dtype=np.float64)) > s:
                tau_lo = tau
            else:
                tau_hi = tau
        return np.clip(v - tau_hi, lo, hi)

    h_values = _project_box_sum(h_values, target_sum)
    
    dx = 2.0 / n_points
    j_values = 1.0 - h_values
    correlation = np.correlate(h_values, j_values, mode="full") * dx
    c5_bound = np.max(correlation)
    
    return h_values, c5_bound, n_points
\end{lstlisting}
\end{bestprogrambox}

\Needspace{10\baselineskip}
\subsection{Hadamard: structured constructions recover the reference determinant}
\label{supp_subsec:best_hadamard}
The program builds order-28 Hadamard matrices from Goethals--Seidel, Paley, and Williamson constructions, borders them to order~29, and refines them by tabu search with rank-one determinant updates. The resulting matrix has $|\det|=2^{28}\cdot320\cdot7^{12}$ and a normalized score of 0.935673. This matches the reference determinant.

\begin{bestprogrambox}{Hadamard: Python source (lines 1--120)}
\begin{lstlisting}[style=bestprogram, language=BestPython, firstnumber=1]
# EVOLVE-BLOCK-START
"""
Optimal Hadamard matrix determinant maximization for n=29.

Approach:
1. Construct exact order-28 Hadamard matrices from multiple inequivalent families:
   - Max-excess Goethals-Seidel array with maximum theoretical excess sigma(28) = 140
   - Paley type-II construction via Jacobsthal conference matrix over GF(13)
   - Order-7 Williamson sequences with sum of squares = 28 (including systematic enumeration)
2. Border order-28 Hadamard matrices to size 29 with optimal excess vectors (u, v)
   via alternating maximization, mathematically guaranteeing an immediate baseline
   determinant >= 6 * 28^14 = 1,092,354,718,979,655,204,864 (ratio >= 0.8596).
3. Apply high-throughput O(n^2) Sherman-Morrison rank-1 Tabu Search:
   - Dynamic tenure prevents 2-flip reversals and cycling
   - Aspiration criterion enables immediate moves when record determinants are seen
   - Least-deteriorating fallback gracefully traverses saddle points between basins
   - Multi-start basin hopping across Hadamard equivalence classes explores diverse optima
   - Reaches the world-record determinant 45/7 * 28^14 = 1,170,380,056,049,630,576,640 (ratio 0.9211)
"""

import time
import random
import numpy as np


def construct_hadamard_matrix(n=29):
    """
    Construct a 29x29 matrix with entries in {-1, +1} maximizing |det(M)|.
    
    Args:
        n: Matrix size (default 29)
        
    Returns:
        n x n numpy array with entries +1.0 or -1.0
    """
    if n != 29:
        return np.random.choice([-1.0, 1.0], size=(n, n))

    def det_bareiss(A):
        """Bareiss fraction-free algorithm for exact integer determinant calculation."""
        size = len(A)
        if size == 0:
            return 1
        M = [row.copy() for row in A]
        sign = 1
        for k in range(size - 1):
            if M[k][k] == 0:
                for i in range(k + 1, size):
                    if M[i][k] != 0:
                        M[k], M[i] = M[i], M[k]
                        sign = -sign
                        break
                else:
                    return 0
            for i in range(k + 1, size):
                for j in range(k + 1, size):
                    num = M[i][j] * M[k][k] - M[i][k] * M[k][j]
                    den = M[k - 1][k - 1] if k > 0 else 1
                    M[i][j] = num // den
        return sign * M[-1][-1]

    def paley_type_ii_28():
        """Construct exact 28x28 Hadamard matrix using Paley type-II construction."""
        q = 13
        qr = {1, 3, 4, 9, 10, 12}
        Q = np.zeros((q, q), dtype=np.float64)
        for i in range(q):
            for j in range(q):
                if i != j:
                    diff = (j - i) % q
                    Q[i, j] = 1.0 if diff in qr else -1.0
        
        C = np.ones((q + 1, q + 1), dtype=np.float64)
        C[0, 0] = 0.0
        C[1:, 1:] = Q
        
        I14 = np.eye(14, dtype=np.float64)
        H = np.block([[C + I14, C - I14], [C - I14, -(C + I14)]])
        return H

    def max_excess_28():
        """
        Construct exact 28x28 Hadamard matrix with theoretical maximum excess sigma = 140
        using Goethals-Seidel array with difference set sequences.
        """
        m = 7
        a = [-1, 1, 1, 1, 1, 1, 1]
        b = [-1, -1, 1, -1, 1, 1, 1]
        c = [-1, -1, 1, -1, 1, 1, 1]
        d = [-1, -1, 1, -1, 1, 1, 1]

        def circ(r):
            return np.array([np.roll(r, i) for i in range(m)], dtype=np.float64)

        A, B, C, D = circ(a), circ(b), circ(c), circ(d)
        R = np.fliplr(np.eye(m, dtype=np.float64))
        BR, CR, DR = B @ R, C @ R, D @ R
        BTR, CTR, DTR = B.T @ R, C.T @ R, D.T @ R

        H = np.block([
            [A, BR, CR, DR],
            [-BR, A, -DTR, CTR],
            [-CR, DTR, A, -BTR],
            [-DR, -CTR, BTR, A]
        ])
        return H

    def circ(r, m=7):
        return np.array([np.roll(r, i) for i in range(m)], dtype=np.float64)

    def build_williamson(a, b, c, d):
        A, B, C, D = circ(a), circ(b), circ(c), circ(d)
        return np.block([
            [A, B, C, D],
            [-B, A, -D, C],
            [-C, D, A, -B],
            [-D, -C, B, A]
        ])

    def find_all_williamson_28():
\end{lstlisting}
\end{bestprogrambox}
\begin{bestprogrambox}{Hadamard: Python source (lines 121--240)}
\begin{lstlisting}[style=bestprogram, language=BestPython, firstnumber=121]
        """Find order-28 Hadamard matrices from all valid order-7 Williamson sequences."""
        results = []
        # Sequences of length 7 symmetric: r[i] == r[7-i], r[0] = 1
        sym_seqs = []
        for x1 in (-1, 1):
            for x2 in (-1, 1):
                for x3 in (-1, 1):
                    sym_seqs.append([1, x1, x2, x3, x3, x2, x1])

        circ_mats = [circ(r) for r in sym_seqs]
        n_sym = len(sym_seqs)
        target = 28.0 * np.eye(7)

        for i in range(n_sym):
            Ai = circ_mats[i]
            Ai2 = Ai @ Ai.T
            for j in range(i, n_sym):
                Bj = circ_mats[j]
                AB = Ai2 + Bj @ Bj.T
                for k in range(j, n_sym):
                    Ck = circ_mats[k]
                    ABC = AB + Ck @ Ck.T
                    for l in range(k, n_sym):
                        Dl = circ_mats[l]
                        if np.allclose(ABC + Dl @ Dl.T, target):
                            H = build_williamson(sym_seqs[i], sym_seqs[j], sym_seqs[k], sym_seqs[l])
                            results.append(H)
        return results

    def border_hadamard(H, rng, num_trials=35):
        """Border H_28 to size 29 with vectors u, v maximizing determinant via Schur complement."""
        size = H.shape[0]
        best_val = -1
        best_u = None
        best_v = None

        # Seed with uniform vectors and then random vectors
        seed_vectors = [np.ones(size, dtype=np.float64), -np.ones(size, dtype=np.float64)]
        for _ in range(num_trials):
            if seed_vectors:
                u = seed_vectors.pop(0)
            else:
                u = np.array([rng.choice([-1.0, 1.0]) for _ in range(size)])
                
            for _ in range(15):
                y = H.T @ u
                v = np.sign(y)
                v[v == 0] = 1.0
                z = H @ v
                u = np.sign(z)
                u[u == 0] = 1.0

            val = np.sum(np.abs(H.T @ u))
            if val > best_val:
                best_val = val
                best_u = u.copy()
                best_v = np.sign(H.T @ best_u)
                best_v[best_v == 0] = 1.0

        M = np.empty((size + 1, size + 1), dtype=np.float64)
        M[:size, :size] = H
        M[:size, size] = best_u
        M[size, :size] = -best_v
        M[size, size] = 1.0
        return M

    def random_hadamard_equivalent(H, rng):
        """Apply random permutation and row/column sign changes to Hadamard matrix."""
        size = H.shape[0]
        p_row = list(range(size))
        p_col = list(range(size))
        rng.shuffle(p_row)
        rng.shuffle(p_col)
        s_row = np.array([rng.choice([-1.0, 1.0]) for _ in range(size)])
        s_col = np.array([rng.choice([-1.0, 1.0]) for _ in range(size)])
        
        H_new = H[p_row, :][:, p_col]
        H_new = (H_new * s_row[:, np.newaxis]) * s_col[np.newaxis, :]
        return H_new

    def tabu_search(M_start, time_limit, rng, best_tracker):
        """
        High-throughput Sherman-Morrison Tabu Search with dynamic tenure,
        aspiration criterion, and saddle-point traversal.
        """
        M = M_start.copy()
        try:
            invM = np.linalg.inv(M)
            sign, log_det = np.linalg.slogdet(M)
            if sign == 0 or np.isnan(log_det):
                return
        except np.linalg.LinAlgError:
            return

        curr_local_best = M.copy()
        curr_local_log_det = log_det
        
        tabu = np.zeros((29, 29), dtype=np.int32)
        step = 0
        recompute_counter = 0
        stagnation_kicks = 0
        stagnant_steps = 0
        
        t0 = time.time()
        while time.time() - t0 < time_limit:
            step += 1
            recompute_counter += 1
            stagnant_steps += 1
            
            # Rank-1 determinant ratio matrix: det(M') / det(M) = R[i, j]
            R = 1.0 - 2.0 * M * invM.T
            abs_R = np.abs(R)
            
            # Aspiration threshold: allow tabu move if it beats the all-time best
            diff = best_tracker['best_log_det'] - log_det
            aspiration_thresh = np.exp(diff) + 1e-9 if diff < 20.0 else 1e30

            allowed = tabu <= step
            if aspiration_thresh < 1e20:
                allowed |= (abs_R > aspiration_thresh)
\end{lstlisting}
\end{bestprogrambox}
\begin{bestprogrambox}{Hadamard: Python source (lines 241--360)}
\begin{lstlisting}[style=bestprogram, language=BestPython, firstnumber=241]
            allowed &= (abs_R > 1e-4)  # Prevent singularity

            if not np.any(allowed):
                tabu.fill(0)
                allowed = abs_R > 1e-4

            abs_R_allowed = np.where(allowed, abs_R, -1.0)
            best_idx = np.argmax(abs_R_allowed)
            i, j = divmod(best_idx, 29)
            best_ratio = abs_R[i, j]

            if best_ratio <= 1e-4:
                M = best_tracker['best_M'].copy()
                invM = np.linalg.inv(M)
                _, log_det = np.linalg.slogdet(M)
                tabu.fill(0)
                continue

            # Apply flip and Sherman-Morrison update
            old_val = M[i, j]
            r_val = R[i, j]
            M[i, j] = -old_val
            
            col = invM[:, i].copy()
            row = invM[j, :].copy()
            invM -= ((-2.0 * old_val) / r_val) * np.outer(col, row)
            log_det += np.log(best_ratio)
            
            # Dynamic tabu tenure to avoid cycling
            tenure = rng.randint(8, 15)
            tabu[i, j] = step + tenure

            if log_det > curr_local_log_det + 1e-9:
                curr_local_log_det = log_det
                curr_local_best = M.copy()

            if log_det > best_tracker['best_log_det'] + 1e-9:
                exact_det = abs(det_bareiss(M.astype(int).tolist()))
                if exact_det > best_tracker['best_exact_det']:
                    best_tracker['best_exact_det'] = exact_det
                    best_tracker['best_M'] = M.copy()
                    best_tracker['best_log_det'] = np.log(float(exact_det))
                    stagnation_kicks = 0
                    stagnant_steps = 0

            # Numerical drift prevention: periodic exact inverse
            if recompute_counter >= 50:
                try:
                    invM = np.linalg.inv(M)
                    sign, log_det = np.linalg.slogdet(M)
                    if sign == 0 or np.isnan(log_det):
                        raise np.linalg.LinAlgError
                except np.linalg.LinAlgError:
                    M = best_tracker['best_M'].copy()
                    invM = np.linalg.inv(M)
                    _, log_det = np.linalg.slogdet(M)
                    tabu.fill(0)
                recompute_counter = 0

            # Basin escape / perturbation when stagnation occurs
            if stagnant_steps > 1200:
                stagnation_kicks += 1
                stagnant_steps = 0
                if stagnation_kicks > 5:
                    M = best_tracker['best_M'].copy()
                    stagnation_kicks = 0
                else:
                    M = curr_local_best.copy()
                    
                tabu.fill(0)
                k_flips = rng.randint(3, 6)
                for _ in range(k_flips):
                    ri = rng.randrange(29)
                    rj = rng.randrange(29)
                    M[ri, rj] = -M[ri, rj]
                    tabu[ri, rj] = step + 20

                try:
                    invM = np.linalg.inv(M)
                    sign, log_det = np.linalg.slogdet(M)
                    if sign == 0 or np.isnan(log_det):
                        raise np.linalg.LinAlgError
                except np.linalg.LinAlgError:
                    M = best_tracker['best_M'].copy()
                    invM = np.linalg.inv(M)
                    _, log_det = np.linalg.slogdet(M)
                    tabu.fill(0)

    # Initialize random generator
    rng = random.Random(42)
    start_total = time.time()
    total_time_budget = 220.0  # Execution safely within evaluator limit

    # Collect order-28 base Hadamard matrices
    candidate_bases = []
    
    # 1. Max excess Goethals-Seidel construction
    try:
        H_max = max_excess_28()
        if H_max.shape == (28, 28) and np.allclose(H_max @ H_max.T, 28.0 * np.eye(28)):
            candidate_bases.append(H_max)
    except Exception:
        pass

    # 2. Paley type-II construction
    try:
        H_paley = paley_type_ii_28()
        if H_paley.shape == (28, 28) and np.allclose(H_paley @ H_paley.T, 28.0 * np.eye(28)):
            candidate_bases.append(H_paley)
    except Exception:
        pass

    # 3. All Williamson sequences of order 7
    try:
        for H_w in find_all_williamson_28():
            candidate_bases.append(H_w)
    except Exception:
        pass

    if not candidate_bases:
\end{lstlisting}
\end{bestprogrambox}
\begin{bestprogrambox}{Hadamard: Python source (lines 361--425)}
\begin{lstlisting}[style=bestprogram, language=BestPython, firstnumber=361]
        candidate_bases.append(paley_type_ii_28())

    # Build initial candidate bordered matrices
    best_tracker = {
        'best_exact_det': 0,
        'best_log_det': -np.inf,
        'best_M': None
    }

    start_matrices = []
    for H_base in candidate_bases:
        M_init = border_hadamard(H_base, rng, num_trials=30)
        d = abs(det_bareiss(M_init.astype(int).tolist()))
        if d > best_tracker['best_exact_det']:
            best_tracker['best_exact_det'] = d
            best_tracker['best_log_det'] = np.log(float(d))
            best_tracker['best_M'] = M_init.copy()
        start_matrices.append(M_init)

    # Multi-start Tabu Search across candidate starts and randomized equivalences
    round_idx = 0
    while time.time() - start_total < total_time_budget:
        time_remaining = total_time_budget - (time.time() - start_total)
        if time_remaining < 2.0:
            break
            
        alloc_time = min(time_remaining, 16.0)
        
        if round_idx < len(start_matrices):
            M_start = start_matrices[round_idx]
        else:
            base_choice = candidate_bases[round_idx % len(candidate_bases)]
            H_equiv = random_hadamard_equivalent(base_choice, rng)
            M_start = border_hadamard(H_equiv, rng, num_trials=20)
            
        tabu_search(M_start, time_limit=alloc_time, rng=rng, best_tracker=best_tracker)
        round_idx += 1

    # Ensure return matrix is float array with strictly +/- 1 entries
    final_matrix = np.sign(best_tracker['best_M'])
    final_matrix[final_matrix == 0] = 1.0
    return final_matrix


# EVOLVE-BLOCK-END


# Fixed API for evaluator
def run_code():
    """
    Run the Hadamard matrix constructor for n=29.
    
    Returns:
        Tuple of (matrix,) where matrix is an (29, 29) array with entries ±1
    """
    matrix = construct_hadamard_matrix(n=29)
    return (matrix,)


if __name__ == "__main__":
    matrix = run_code()[0]
    print(f"Constructed Hadamard matrix of size {matrix.shape[0]}x{matrix.shape[1]}")
    # Calculate determinant for verification
    det_val = np.linalg.det(matrix.astype(float))
    print(f"Determinant: {abs(det_val):.2e}")
\end{lstlisting}
\end{bestprogrambox}

\Needspace{10\baselineskip}
\subsection{Sums/Diffs: construction sweeps and local search improve the reference}
\label{supp_subsec:best_sums_diffs}
The program sweeps generalized Penman--Wells constructions across eleven moduli, ranks them by exact bitmask counts, and refines the best candidates through element additions, removals, swaps, and perturbations. The selected 509-element set has 3{,}575 distinct sums and 2{,}793 distinct differences, yielding $c=1.144999$ versus 1.144887 for the released post-training construction. An independent replay reproduces these counts and the score exactly.

\begin{bestprogrambox}{Sums/Diffs: Python source (lines 1--120)}
\begin{lstlisting}[style=bestprogram, language=BestPython, firstnumber=1]
# EVOLVE-BLOCK-START
"""
MSTD Optimization for Constant C(A) = log(|A+A|/|A|) / log(|A-A|/|A|).

Approach:
1. Parameterized Multi-Modulus Sweep:
   Systematically evaluates generalized Penman-Wells families across moduli
   M in {32, 40, 48, 56, 64, 72, 80, 88, 96, 104, 112} across boundary variants,
   half-period inclusions, odd progression bounds, and fringe extensions.
2. High-Efficiency Exact Bitmask Engine:
   Computes exact sumset and difference set sizes in ~15 microseconds using C-level
   Python arbitrary-precision bit shifts and popcount (.bit_count()).
3. Multi-Operator Alternating Hill-Climber:
   - Single-element removals (pruning redundant elements to reduce |A| and suppress diffs)
   - Zero-leakage interior and fringe additions
   - Boundary & fringe element swaps (x in A -> y not in A)
   - 2-lookahead additions and removals
   - Iterated Local Search (ILS) with perturbation kicks to escape local optima
"""

import math
import random
import time


def _score_set(A_sorted):
    """Compute exact C(A), |A+A|, and |A-A| for a sorted candidate set."""
    n = len(A_sorted)
    if n < 2 or n > 512:
        return 0.0, 0, 0

    min_v = A_sorted[0]
    span = A_sorted[-1] - min_v
    if span > 1_500_000:
        return 0.0, 0, 0

    mask = 0
    for x in A_sorted:
        mask |= 1 << (x - min_v)

    sum_mask = 0
    diff_mask = 0
    for x in A_sorted:
        v = x - min_v
        sum_mask |= mask << v
        diff_mask |= mask << (span - v)

    s = sum_mask.bit_count()
    d = diff_mask.bit_count()
    if s <= n or d <= n:
        return 0.0, s, d

    return math.log(s / n) / math.log(d / n), s, d


def _generate_penman_variant(M, j, bl_variant=0, include_half=False, odd_delta=0, odd_start=0, fringe=0):
    """Generalized Penman-Wells construction for modulus M = 8 * ell."""
    if M % 8 != 0 or j < 1:
        return []
    N = M * (j + 2)
    m_half = M // 2

    multiples_of_4 = [4 * s for s in range(M // 4)]
    if not include_half and m_half in multiples_of_4:
        multiples_of_4.remove(m_half)

    b_left = set(multiples_of_4)
    if bl_variant == 0:
        b_left.add(2)
    elif bl_variant == 1:
        b_left.update([2, m_half - 2])
    elif bl_variant == 2:
        b_left.update([2, 6])
    elif bl_variant == 3:
        b_left.update([2, M - 2])
    elif bl_variant == 4:
        b_left.update([2, m_half + 2])
    elif bl_variant == 5:
        if (M - 4) in b_left:
            b_left.remove(M - 4)
        b_left.add(2)
    else:
        b_left.add(2)

    max_k = (N - 3) // 4 + odd_delta
    odds = {1 + 4 * k for k in range(odd_start, max_k + 1) if 0 <= 1 + 4 * k <= N}
    e_int = {m_half + M * k for k in range(1, j + 1)}
    b_right = {N - x for x in b_left}

    res = b_left | odds | e_int | b_right

    if fringe == 1:
        res.add(N + 1)
    elif fringe == 2:
        res.update([N + 1, N + 2])
    elif fringe == 3:
        res.add(-1)
    elif fringe == 4:
        res.update([-1, N + 1])

    return sorted(res)


def _optimize_set(init_set, time_budget):
    """
    High-performance alternating local search using exact bitmask metric evaluations.
    Applies exhaustive single removals, targeted zero-leakage additions,
    2-element operations, and boundary swaps.
    """
    curr = set(init_set)
    best_c, _, _ = _score_set(sorted(curr))
    best_set = set(curr)

    step = 0
    max_steps = 250

    while step < max_steps and time.time() < time_budget:
        step += 1
        improved = False
        sorted_curr = sorted(curr)
\end{lstlisting}
\end{bestprogrambox}
\begin{bestprogrambox}{Sums/Diffs: Python source (lines 121--240)}
\begin{lstlisting}[style=bestprogram, language=BestPython, firstnumber=121]
        n = len(sorted_curr)
        min_v, max_v = sorted_curr[0], sorted_curr[-1]

        # -----------------------------------------------------------------
        # 1. Exact Single Removal Screening (All Elements in Set)
        # -----------------------------------------------------------------
        if n > 10:
            best_rem = None
            best_rem_c = best_c
            # Test all boundary elements and a targeted sample of interior multiples of 4
            rem_candidates = sorted_curr[:15] + sorted_curr[-15:]
            interior_m4 = [x for x in sorted_curr[15:-15] if (x - min_v) % 4 == 0]
            if len(interior_m4) > 40:
                rem_candidates.extend(interior_m4[:: len(interior_m4) // 40])
            else:
                rem_candidates.extend(interior_m4)

            for rem in set(rem_candidates):
                trial = sorted(curr - {rem})
                sc, _, _ = _score_set(trial)
                if sc > best_rem_c:
                    best_rem_c = sc
                    best_rem = rem

            if best_rem is not None and best_rem_c > best_c + 1e-9:
                curr.remove(best_rem)
                best_c = best_rem_c
                best_set = set(curr)
                improved = True
                continue

        # -----------------------------------------------------------------
        # 2. Targeted Zero-Leakage & Fringe Addition Screening
        # -----------------------------------------------------------------
        if n < 512:
            sorted_curr = sorted(curr)
            min_v, max_v = sorted_curr[0], sorted_curr[-1]

            # Fringes: outer elements immediately adjacent to endpoints
            fringe_cands = [min_v - d for d in range(1, 25)] + [max_v + d for d in range(1, 25)]

            # Boundary gap elements
            boundary_gaps = (
                [min_v + d for d in range(1, 56) if (min_v + d) not in curr] +
                [max_v - d for d in range(1, 56) if (max_v - d) not in curr]
            )

            # Interior multiples of 4 (zero 2 mod 4 difference leakage)
            interior_m4 = [x for x in range(min_v + 4, max_v, 4) if x not in curr]
            if len(interior_m4) > 50:
                step_m4 = len(interior_m4) // 50 + 1
                sampled_m4 = interior_m4[::step_m4]
            else:
                sampled_m4 = interior_m4

            add_pool = fringe_cands + boundary_gaps + sampled_m4

            best_add = None
            best_add_c = best_c
            top_adds = []

            for v in add_pool:
                if v in curr or v < -1_000_000 or v > 1_000_000:
                    continue
                trial = sorted(curr | {v})
                sc, _, _ = _score_set(trial)
                if sc > best_add_c:
                    best_add_c = sc
                    best_add = v
                if sc > best_c - 0.0003:
                    top_adds.append((sc, v))

            if best_add is not None and best_add_c > best_c + 1e-9:
                curr.add(best_add)
                best_c = best_add_c
                best_set = set(curr)
                improved = True
                continue

            # -------------------------------------------------------------
            # 2b. 2-Step Lookahead Addition
            # -------------------------------------------------------------
            if n <= 510 and top_adds and time.time() < time_budget - 1.5:
                top_adds.sort(key=lambda x: x[0], reverse=True)
                top_v1s = [x[1] for x in top_adds[:12]]
                best_pair = None
                best_pair_c = best_c

                for v1 in top_v1s:
                    sub = curr | {v1}
                    sub_sorted = sorted(sub)
                    s_min, s_max = sub_sorted[0], sub_sorted[-1]
                    sub_fringe = [s_min - d for d in range(1, 10)] + [s_max + d for d in range(1, 10)]
                    sub_gaps = [s_min + d for d in range(1, 32) if (s_min + d) not in sub] + \
                               [s_max - d for d in range(1, 32) if (s_max - d) not in sub]

                    for v2 in sub_fringe + sub_gaps:
                        if v2 in sub or v2 < -1_000_000 or v2 > 1_000_000:
                            continue
                        sc2, _, _ = _score_set(sorted(sub | {v2}))
                        if sc2 > best_pair_c:
                            best_pair_c = sc2
                            best_pair = (v1, v2)

                if best_pair is not None and best_pair_c > best_c + 1e-9:
                    curr.add(best_pair[0])
                    curr.add(best_pair[1])
                    best_c = best_pair_c
                    best_set = set(curr)
                    improved = True
                    continue

        # -----------------------------------------------------------------
        # 3. Element Swap Screening (Remove x, Add y)
        # -----------------------------------------------------------------
        if time.time() < time_budget - 1.5:
            sorted_curr = sorted(curr)
            min_v, max_v = sorted_curr[0], sorted_curr[-1]
            swap_rems = sorted_curr[:8] + sorted_curr[-8:]
            swap_adds = (
\end{lstlisting}
\end{bestprogrambox}
\begin{bestprogrambox}{Sums/Diffs: Python source (lines 241--360)}
\begin{lstlisting}[style=bestprogram, language=BestPython, firstnumber=241]
                [min_v - d for d in range(1, 12)] +
                [max_v + d for d in range(1, 12)] +
                [min_v + d for d in range(1, 28) if (min_v + d) not in curr] +
                [max_v - d for d in range(1, 28) if (max_v - d) not in curr]
            )

            best_swap = None
            best_swap_c = best_c

            for rem in swap_rems:
                sub = curr - {rem}
                for add_v in swap_adds:
                    if add_v in sub or add_v < -1_000_000 or add_v > 1_000_000:
                        continue
                    sc, _, _ = _score_set(sorted(sub | {add_v}))
                    if sc > best_swap_c:
                        best_swap_c = sc
                        best_swap = (rem, add_v)

                if best_swap is not None and best_swap_c > best_c + 1e-6:
                    break

            if best_swap is not None and best_swap_c > best_c + 1e-9:
                curr.remove(best_swap[0])
                curr.add(best_swap[1])
                best_c = best_swap_c
                best_set = set(curr)
                improved = True
                continue

        if not improved:
            break

    return sorted(best_set), best_c


def construct_set():
    """
    Construct an optimal MSTD set maximizing C(A).
    Sweeps parameterized generalized Penman-Wells families, ranks seeds,
    and applies deep multi-stage hill-climbing with perturbation kicks.
    """
    start_time = time.time()
    time_limit = 135.0  # safe limit within 180s

    # Guaranteed champion baseline: M=56, j=30 with fringe element 1793 (C(A) >= 1.144710)
    base_seed = _generate_penman_variant(56, 30, bl_variant=0, include_half=False, odd_delta=0, odd_start=0)
    best_A = sorted(set(base_seed) | {base_seed[-1] + 1})
    best_c, _, _ = _score_set(best_A)

    seed_pool = [(best_c, best_A)]

    # 1. Systematic scan across moduli M in {32, 40, 48, 56, 64, 72, 80, 88, 96, 104, 112}
    moduli = [56, 64, 48, 72, 80, 40, 88, 96, 104, 112, 32]
    for M in moduli:
        step_j = M // 4 + 1
        base_const = M + 2
        approx_max_j = (512 - base_const) // step_j

        for j in range(max(1, approx_max_j - 4), approx_max_j + 3):
            for bl_var in (0, 1, 2, 3, 4, 5):
                for include_half in (False, True):
                    for odd_delta in (-2, -1, 0, 1):
                        for odd_start in (0, 1):
                            for fringe in (0, 1, 2, 3, 4):
                                cand = _generate_penman_variant(
                                    M, j, bl_var, include_half, odd_delta, odd_start, fringe
                                )
                                n_cand = len(cand)
                                if 2 <= n_cand <= 512:
                                    c, _, _ = _score_set(cand)
                                    if c > best_c:
                                        best_c = c
                                        best_A = cand
                                    if c > 1.135 and n_cand >= 460:
                                        seed_pool.append((c, cand))

    seed_pool.sort(key=lambda x: x[0], reverse=True)

    # 2. Select top unique diverse seeds
    unique_candidates = []
    seen_sigs = set()
    for c_val, cand in [(best_c, best_A)] + seed_pool:
        sig = (len(cand), cand[0], cand[-1])
        if sig not in seen_sigs:
            seen_sigs.add(sig)
            unique_candidates.append(cand)
        if len(unique_candidates) >= 6:
            break

    # 3. Deep optimization of top unique seeds
    time_per_seed = 12.0
    for cand in unique_candidates:
        if time.time() - start_time > time_limit - 15.0:
            break
        budget = min(start_time + time_limit - 5.0, time.time() + time_per_seed)
        opt_A, opt_c = _optimize_set(cand, budget)
        if opt_c > best_c:
            best_c = opt_c
            best_A = opt_A

    # 4. Iterated perturbation kicks on champion set if time remains
    rng = random.Random(42)
    while time.time() - start_time < time_limit - 8.0:
        kicked = set(best_A)
        sorted_k = sorted(kicked)
        drop_pool = sorted_k[:10] + sorted_k[-10:]
        to_drop = rng.sample(drop_pool, min(2, len(drop_pool)))
        for d in to_drop:
            kicked.remove(d)

        budget = min(start_time + time_limit - 3.0, time.time() + 6.0)
        opt_A, opt_c = _optimize_set(list(kicked), budget)
        if opt_c > best_c:
            best_c = opt_c
            best_A = opt_A

    min_elem = min(best_A)
    return [x - min_elem for x in sorted(best_A)]
# EVOLVE-BLOCK-END
\end{lstlisting}
\end{bestprogrambox}
\begin{bestprogrambox}{Sums/Diffs: Python source (lines 361--421)}
\begin{lstlisting}[style=bestprogram, language=BestPython, firstnumber=361]

MIN_SET_SIZE = 2
MAX_SET_SIZE = 512
MIN_INT = -1_000_000
MAX_INT = 1_000_000


def _sanitize_output(values):
    """Convert arbitrary iterable output into a valid sorted integer list."""
    try:
        raw = list(values)
    except TypeError as e:
        raise ValueError(f"Output is not iterable: {e}")

    ints = []
    for x in raw:
        try:
            xf = float(x)
        except (TypeError, ValueError):
            continue
        if not math.isfinite(xf):
            continue
        xi = int(round(xf))
        xi = max(MIN_INT, min(MAX_INT, xi))
        ints.append(xi)

    unique_vals = sorted(set(ints))
    if len(unique_vals) > MAX_SET_SIZE:
        unique_vals = unique_vals[:MAX_SET_SIZE]

    if len(unique_vals) < MIN_SET_SIZE:
        unique_vals = [0, 1]

    return unique_vals


def _compute_c(values):
    n = len(values)
    sumset = {a + b for a in values for b in values}
    diffset = {a - b for a in values for b in values}

    sum_ratio = len(sumset) / n
    diff_ratio = len(diffset) / n

    if sum_ratio <= 1.0 or diff_ratio <= 1.0:
        return 0.0

    return float(math.log(sum_ratio) / math.log(diff_ratio))


def run_code():
    """Return (A_values, claimed_c)."""
    values = construct_set()
    values = _sanitize_output(values)
    c_value = _compute_c(values)
    return values, c_value


if __name__ == "__main__":
    candidate_values, candidate_c = run_code()
    print(f"|A|={len(candidate_values)}, C(A)={candidate_c:.10f}")
\end{lstlisting}
\end{bestprogrambox}

\Needspace{10\baselineskip}
\subsection{Circle Packing ($n=26$): joint refinement matches the reference}
\label{supp_subsec:best_circle_packing_26}
The program starts from varied six-row layouts, maximizes radii through linear programming, and jointly refines centers and radii with SLSQP. Collective deformations and local perturbations explore nearby packing configurations, yielding a sum of radii of 2.635983084918461. An independent replay passes the evaluator's boundary and non-overlap checks; the score matches the reference within $10^{-9}$.

\begin{bestprogrambox}{Circle Packing ($n=26$): Python source (lines 1--120)}
\begin{lstlisting}[style=bestprogram, language=BestPython, firstnumber=1]
# EVOLVE-BLOCK-START
"""Numerically optimized variable-radius packing of 26 circles."""
import numpy as np
from scipy.optimize import minimize, linprog


def _make_start(seed):
    """Build varied five- or six-row center layouts for LP/SLP polishing."""
    rng = np.random.default_rng(seed)
    patterns = (
        (4, 4, 5, 4, 5, 4),
        (5, 4, 4, 5, 4, 4),
        (4, 5, 4, 5, 4, 4),
        (4, 5, 5, 4, 4, 4),
        (5, 4, 5, 4, 4, 4),
        (4, 4, 4, 5, 4, 5),
        (4, 5, 4, 4, 5, 4),
        (5, 4, 4, 4, 5, 4),
        (4, 4, 5, 5, 4, 4),
        (5, 4, 5, 4, 4, 4),
        (4, 5, 4, 4, 4, 5),
        (5, 4, 4, 4, 4, 5),
    )
    counts = patterns[seed % len(patterns)]
    rows = len(counts)
    heights = np.ones(rows) / rows
    heights += rng.normal(0.0, 0.012, rows)
    heights = np.maximum(heights, 0.13)
    heights /= np.sum(heights)
    levels = np.cumsum(np.r_[0.0, heights[:-1]])
    p = []
    for row, count in enumerate(counts):
        y = levels[row] + 0.5 * heights[row]
        xs = (np.arange(count) + 0.5) / count
        phase = rng.uniform(-0.055, 0.055)
        xs = xs + phase + rng.normal(0.0, 0.004, count)
        p.extend((x, y) for x in xs)
    p = np.asarray(p[:26], dtype=float)
    p += rng.normal(0.0, 0.004, p.shape)
    p = np.clip(p, 0.055, 0.945)
    return np.column_stack((p, np.full(26, 0.043)))


def _constraints(z):
    """Return wall-clearance and pairwise non-overlap inequalities."""
    q = z.reshape(26, 3)
    x, y, r = q.T
    out = [x - r, y - r, 1.0 - x - r, 1.0 - y - r]
    for i in range(25):
        d = q[i + 1:, :2] - q[i, :2]
        out.append(np.sum(d * d, axis=1) - (r[i] + r[i + 1:]) ** 2)
    return np.concatenate(out)


def _repair(q):
    """Shrink radii uniformly until every wall and pair constraint is strict."""
    q = np.asarray(q, dtype=float).copy()
    q[:, :2] = np.clip(q[:, :2], 1e-5, 1.0 - 1e-5)
    q[:, 2] = np.maximum(q[:, 2], 0.0)
    q[:, 2] = np.minimum(q[:, 2],
                         np.minimum.reduce((q[:, 0], q[:, 1],
                                             1.0 - q[:, 0], 1.0 - q[:, 1])))
    factor = 1.0
    for i in range(25):
        d = np.sqrt(np.sum((q[i + 1:, :2] - q[i, :2]) ** 2, axis=1))
        s = q[i, 2] + q[i + 1:, 2]
        mask = s > 0
        if np.any(mask):
            factor = min(factor, float(np.min(d[mask] / s[mask])))
    # Keep only a tiny numerical safety margin; the previous 0.05% shrink
    # discarded a measurable amount of objective value on every polish.
    q[:, 2] *= max(0.0, min(1.0, factor * 0.999999))
    return q


def construct_circles():
    """Use feasible LP radii followed by inner linearized center/radius polishing."""
    n = 26
    best = None
    best_value = -1.0
    # Preserve distinct feasible active sets: the LP ranking is not always
    # the same as the ranking after exact nonlinear optimization.
    elite = []

    def polish(q):
        """Maximize radii through monotone LP steps with bounded center moves."""
        q = np.asarray(q, dtype=float).copy()
        x, y = q[:, 0], q[:, 1]

        # First maximize the radii exactly for the fixed starting centers.
        ar = []
        br = []
        for i in range(n):
            row = np.zeros(n)
            row[i] = 1.0
            ar.append(row)
            br.append(min(x[i], y[i], 1.0 - x[i], 1.0 - y[i]))
        for i in range(n - 1):
            for j in range(i + 1, n):
                row = np.zeros(n)
                row[i] = row[j] = 1.0
                ar.append(row)
                br.append(np.hypot(x[i] - x[j], y[i] - y[j]))
        lp = linprog(
            np.full(n, -1.0),
            A_ub=np.asarray(ar),
            b_ub=np.asarray(br),
            bounds=[(0.0, 0.5)] * n,
            method="highs",
        )
        if not lp.success:
            return _repair(q)
        q[:, 2] = lp.x

        # Each LP below is an inner approximation: its solution remains
        # feasible for the nonlinear distance constraints.
        for trust in (0.028, 0.018, 0.010, 0.005, 0.002):
            for _ in range(18):
                x, y, r = q.T
                nv = 3 * n
\end{lstlisting}
\end{bestprogrambox}
\begin{bestprogrambox}{Circle Packing ($n=26$): Python source (lines 121--240)}
\begin{lstlisting}[style=bestprogram, language=BestPython, firstnumber=121]
                aa = []
                bb = []

                for i in range(n):
                    row = np.zeros(nv)
                    row[3 * i] = -1.0
                    row[3 * i + 2] = 1.0
                    aa.append(row)
                    bb.append(x[i] - r[i])

                    row = np.zeros(nv)
                    row[3 * i + 1] = -1.0
                    row[3 * i + 2] = 1.0
                    aa.append(row)
                    bb.append(y[i] - r[i])

                    row = np.zeros(nv)
                    row[3 * i] = 1.0
                    row[3 * i + 2] = 1.0
                    aa.append(row)
                    bb.append(1.0 - x[i] - r[i])

                    row = np.zeros(nv)
                    row[3 * i + 1] = 1.0
                    row[3 * i + 2] = 1.0
                    aa.append(row)
                    bb.append(1.0 - y[i] - r[i])

                for i in range(n - 1):
                    for j in range(i + 1, n):
                        dx = x[i] - x[j]
                        dy = y[i] - y[j]
                        d = np.hypot(dx, dy)
                        if d < 1e-10:
                            continue
                        ux, uy = dx / d, dy / d
                        row = np.zeros(nv)
                        row[3 * i] = -ux
                        row[3 * i + 1] = -uy
                        row[3 * j] = ux
                        row[3 * j + 1] = uy
                        row[3 * i + 2] = row[3 * j + 2] = 1.0
                        aa.append(row)
                        bb.append(d - r[i] - r[j])

                bounds = []
                for i in range(n):
                    bounds.extend([
                        (-trust, trust), (-trust, trust), (-r[i], 0.12)
                    ])
                step = linprog(
                    np.array([0.0, 0.0, -1.0] * n),
                    A_ub=np.asarray(aa),
                    b_ub=np.asarray(bb),
                    bounds=bounds,
                    method="highs",
                )
                if not step.success or np.max(np.abs(step.x)) < 1e-9:
                    break
                q += step.x.reshape(n, 3)

        return _repair(q)

    for seed in range(120):
        q = polish(_make_start(seed))
        value = float(np.sum(q[:, 2]))
        if np.all(_constraints(q) >= -2e-8):
            elite.append((value, q.copy()))
            if value > best_value:
                best, best_value = q, value

    # Keep a broad but bounded set of candidates for later exact polishing.
    elite.sort(key=lambda item: item[0], reverse=True)
    elite = elite[:24]

    def nonlinear_polish(q):
        """Refine a feasible packing with SLSQP on exact nonlinear constraints."""
        q = np.asarray(q, dtype=float).copy()
        nvar = 3 * n

        def objective(z):
            return -float(np.sum(z[2::3]))

        def objective_jac(z):
            g = np.zeros(nvar)
            g[2::3] = -1.0
            return g

        result = minimize(
            objective,
            q.ravel(),
            jac=objective_jac,
            method="SLSQP",
            bounds=[(0.0, 1.0), (0.0, 1.0), (0.0, 0.2)] * n,
            constraints={"type": "ineq", "fun": _constraints},
            options={
                "ftol": 2e-11,
                "maxiter": 500,
                "disp": False,
            },
        )
        if result.success or np.all(_constraints(result.x) >= -2e-8):
            candidate = result.x.reshape(n, 3)
            if np.all(_constraints(candidate) >= -2e-8):
                return candidate
        return q

    if best is None:
        best = _repair(_make_start(0))

    # Exact SLSQP polishing can change the contact graph and therefore can
    # reorder candidates that looked similar under the LP inner model.
    # Polish several geometrically distinct layouts before basin hopping.
    refined_count = 0
    seen = []
    for _, candidate in elite:
        signature = np.round(candidate[:, :2], 5)
        if any(np.max(np.abs(signature - old)) < 1e-10 for old in seen):
            continue
        seen.append(signature)
\end{lstlisting}
\end{bestprogrambox}
\begin{bestprogrambox}{Circle Packing ($n=26$): Python source (lines 241--351)}
\begin{lstlisting}[style=bestprogram, language=BestPython, firstnumber=241]
        refined = nonlinear_polish(candidate)
        if (
            np.all(_constraints(refined) >= -2e-8)
            and np.sum(refined[:, 2]) > best_value + 1e-11
        ):
            best = refined
            best_value = float(np.sum(refined[:, 2]))
        refined_count += 1
        if refined_count >= 12:
            break

    # Explore nearby basins using coherent affine deformations.  Moving all
    # centers together is substantially safer than independently jittering
    # circles in a jammed packing, and the subsequent LP/SLP polish restores
    # feasible radii while optimizing the deformed layout.
    rng = np.random.default_rng(918273)
    incumbent = best.copy()
    incumbent_value = float(np.sum(incumbent[:, 2]))
    walker = incumbent.copy()
    walker_value = incumbent_value

    # Use threshold acceptance rather than strict hill climbing.  This lets
    # the center deformation cross shallow local-optimum barriers, while the
    # separate incumbent remains strictly monotone.
    for trial in range(140):
        # Occasionally restart the walk from a different elite contact graph
        # instead of repeatedly perturbing one basin.
        if trial % 19 == 0 and len(elite) > 1:
            idx = int(rng.integers(0, min(12, len(elite))))
            q = elite[idx][1].copy()
        else:
            q = walker.copy()
        c = q[:, :2] - 0.5

        angle = rng.normal(0.0, 0.040)
        ca, sa = np.cos(angle), np.sin(angle)
        sx = rng.uniform(0.93, 1.09)
        sy = rng.uniform(0.93, 1.09)
        shear = rng.normal(0.0, 0.055)
        transform = np.array([
            [sx * ca, -sy * sa + shear],
            [sx * sa,  sy * ca],
        ])
        q[:, :2] = 0.5 + c @ transform.T

        # Apply small coherent row waves to alter the active contact graph
        # while retaining the useful six-row arrangement.
        if trial % 3 == 1:
            rows = np.clip(np.floor(q[:, 1] * 6.0).astype(int), 0, 5)
            phase = rng.uniform(0.0, 2.0 * np.pi)
            wave = rng.normal(0.0, 0.0055, 6)
            q[:, 1] += wave[rows]
            q[:, 0] += 0.0035 * np.sin(1.6 * rows + phase)

        # Alternate collective and weak individual perturbations so that the
        # walk can both change row geometry and break repeated contacts.
        if trial % 2 == 0:
            q[:, :2] += rng.normal(0.0, 0.0038, (n, 2))
        if trial % 7 == 6:
            q[:, :2] += rng.normal(0.0, 0.0018, (n, 2))

        q = polish(_repair(q))
        if not np.all(_constraints(q) >= -2e-8):
            continue
        value = float(np.sum(q[:, 2]))
        elite.append((value, q.copy()))
        if len(elite) > 32:
            elite.sort(key=lambda item: item[0], reverse=True)
            elite = elite[:24]

        # Geometrically cool the acceptance band, then restart it.  The
        # threshold is deliberately small relative to the incumbent margin.
        phase = trial % 14
        temperature = 0.00055 * (0.18 ** (phase / 13.0))
        if value >= walker_value - temperature:
            walker, walker_value = q, value

        if value > incumbent_value:
            incumbent, incumbent_value = q, value
            best, best_value = q, value

        # Prevent a sequence of accepted downhill moves from abandoning the
        # best basin family altogether.
        if walker_value < incumbent_value - 0.003:
            walker = incumbent.copy()
            walker_value = incumbent_value

    # The LP iterations use tangent half-spaces, so they can stop at a
    # nonsmooth active-set point.  SLSQP can jointly adjust centers and
    # radii across the exact circular constraints.
    refined = nonlinear_polish(best)
    if (
        np.all(_constraints(refined) >= -2e-8)
        and np.sum(refined[:, 2]) > np.sum(best[:, 2])
    ):
        best = refined

    return best


# EVOLVE-BLOCK-END


# This part remains fixed (not evolved)
def run_code():
    """Run the circle packing constructor for n=26"""
    circles = construct_circles()
    sum_radii = float(np.sum(circles[:, 2]))
    return circles, sum_radii


\end{lstlisting}
\end{bestprogrambox}

\Needspace{10\baselineskip}
\subsection{Circle Packing ($n=32$): multiple layouts recover the reference score}
\label{supp_subsec:best_circle_packing_32}
The program explores five- and six-row layouts with SLSQP, then alternates linear optimization of radii at fixed centers with joint center--radius refinement. Small perturbations and restarts yield a sum of radii of 2.939572771209394. An independent replay passes the evaluator's boundary and non-overlap checks; the score matches the reference within $10^{-9}$.

\begin{bestprogrambox}{Circle Packing ($n=32$): Python source (lines 1--120)}
\begin{lstlisting}[style=bestprogram, language=BestPython, firstnumber=1]
# EVOLVE-BLOCK-START
"""Constructor-based circle packing for n=32 circles"""
import numpy as np


def construct_circles():
    """Optimize 32 centers and radii from several staggered six-row layouts."""
    from scipy.optimize import minimize, linprog

    n = 32
    layouts = (
        # Six-row staggered layouts, including several asymmetric variants.
        (5, 6, 5, 5, 5, 6),
        (6, 5, 5, 6, 5, 5),
        (5, 5, 6, 5, 6, 5),
        (6, 6, 5, 5, 5, 5),
        (6, 5, 6, 5, 5, 5),
        (6, 5, 5, 5, 6, 5),
        (6, 5, 5, 5, 5, 6),
        (5, 6, 6, 5, 5, 5),
        (5, 5, 5, 6, 6, 5),
        # Five-row layouts have larger vertical freedom and can form
        # nonuniform-radius packings unavailable to six-row starts.
        (6, 7, 6, 7, 6),
        (7, 6, 7, 6, 6),
        (6, 6, 7, 6, 7),
    )
    pairs = np.asarray(
        [(i, j) for i in range(n) for j in range(i + 1, n)],
        dtype=int,
    )

    def margins(v):
        p = v.reshape(n, 3)
        x, y, r = p.T
        a, b = pairs.T
        d = p[a, :2] - p[b, :2]
        return np.r_[
            x - r, y - r, 1.0 - x - r, 1.0 - y - r,
            np.sum(d * d, axis=1) - (r[a] + r[b]) ** 2,
        ]

    rng = np.random.default_rng(1947)

    def safe_initial_radii(points):
        """Build strictly feasible radii from wall and nearest-center clearances."""
        wall = np.min(np.column_stack((
            points[:, 0], points[:, 1],
            1.0 - points[:, 0], 1.0 - points[:, 1]
        )), axis=1)
        delta = points[:, None, :] - points[None, :, :]
        dist = np.sqrt(np.sum(delta * delta, axis=2))
        np.fill_diagonal(dist, np.inf)
        near = np.min(dist, axis=1)
        return 0.48 * np.minimum(wall, near)

    best = None
    best_sum = -np.inf

    for layout in layouts:
        centers = []
        for row, count in enumerate(layout):
            y = (row + 0.5) / 6.0
            shift = 0.012 if row & 1 else -0.012
            for col in range(count):
                centers.append(((col + 0.5) / count + shift, y))
        centers = np.asarray(centers)

        starts = [
            np.c_[centers, np.full(n, 1.0 / 12.0 - 1e-5)]
        ]
        # Use both aggressive equal-radius starts and strictly feasible
        # radius starts.  The latter avoid wasting SLSQP iterations repairing
        # overlap violations in strongly perturbed geometries.
        for k in range(16):
            sigma = 0.004 + 0.0025 * (k % 5)
            radius = 0.054 + 0.0025 * (k % 4)
            q = np.clip(
                centers + rng.normal(0.0, sigma, centers.shape),
                0.035,
                0.965,
            )
            starts.append(np.c_[q, np.full(n, radius)])
            if k % 3 == 0:
                starts.append(np.c_[q, safe_initial_radii(q)])

        for start in starts:
            result = minimize(
                lambda v: -np.sum(v[2::3]),
                start.ravel(),
                method="SLSQP",
                bounds=[(0.0, 1.0), (0.0, 1.0), (1e-8, 0.25)] * n,
                constraints={"type": "ineq", "fun": margins},
                options={"maxiter": 1000, "ftol": 2e-10, "disp": False},
            )
            candidate = result.x.reshape(n, 3)
            value = float(np.sum(candidate[:, 2]))
            if np.min(margins(candidate)) >= -3e-7 and value > best_sum:
                best, best_sum = candidate, value

    # Search additional basins using feasible jittered and coherently deformed
    # versions of the strongest row-layout solution.  Coherent deformation is
    # useful because independently moving a circle can destroy many contacts.
    if best is not None:
        base = best[:, :2].copy()
        for k in range(30):
            if k % 3 == 0:
                # Small affine shear/stretch around the square center.
                a = rng.normal(0.0, 0.018)
                b = rng.normal(0.0, 0.018)
                sx = 1.0 + rng.normal(0.0, 0.025)
                sy = 1.0 + rng.normal(0.0, 0.025)
                q = base - 0.5
                q = np.column_stack((
                    sx * q[:, 0] + a * q[:, 1],
                    sy * q[:, 1] + b * q[:, 0],
                )) + 0.5
                q += rng.normal(0.0, 0.0025, q.shape)
            else:
                sigma = 0.0025 + 0.0025 * (k % 6)
\end{lstlisting}
\end{bestprogrambox}
\begin{bestprogrambox}{Circle Packing ($n=32$): Python source (lines 121--240)}
\begin{lstlisting}[style=bestprogram, language=BestPython, firstnumber=121]
                q = base + rng.normal(0.0, sigma, base.shape)

            q = np.clip(q, 0.015, 0.985)
            trial = np.c_[q, safe_initial_radii(q)]
            result = minimize(
                lambda v: -np.sum(v[2::3]),
                trial.ravel(),
                method="SLSQP",
                bounds=[(0.0, 1.0), (0.0, 1.0), (1e-8, 0.25)] * n,
                constraints={"type": "ineq", "fun": margins},
                options={"maxiter": 1250, "ftol": 1e-11, "disp": False},
            )
            candidate = result.x.reshape(n, 3)
            if np.min(margins(candidate)) >= -3e-7:
                value = float(np.sum(candidate[:, 2]))
                if value > best_sum:
                    best, best_sum = candidate, value

    # Threshold-accepting walk over nearby packing basins.  Unlike the
    # incumbent, the walker may accept a small decrease in sum of radii,
    # allowing it to cross shallow barriers between contact graphs.
    if best is not None:
        walker = best.copy()
        walker_sum = float(np.sum(walker[:, 2]))

        for k in range(32):
            # Cycle from exploratory to conservative thresholds.
            phase = k % 8
            threshold = 8e-4 * (0.35 ** phase)

            q = walker[:, :2].copy()
            if k % 4 == 0:
                # Coherent affine deformation preserves the broad structure.
                a, b = rng.normal(0.0, 0.025, 2)
                sx, sy = 1.0 + rng.normal(0.0, 0.035, 2)
                z = q - 0.5
                q = np.column_stack((
                    sx * z[:, 0] + a * z[:, 1],
                    sy * z[:, 1] + b * z[:, 0],
                )) + 0.5
                q += rng.normal(0.0, 0.002, q.shape)
            elif k % 4 == 1:
                # Small rigid rotation around the square centre.
                angle = rng.normal(0.0, 0.035)
                c, s = np.cos(angle), np.sin(angle)
                z = q - 0.5
                q = np.column_stack((
                    c * z[:, 0] - s * z[:, 1],
                    s * z[:, 0] + c * z[:, 1],
                )) + 0.5
                q += rng.normal(0.0, 0.0025, q.shape)
            else:
                sigma = 0.002 + 0.002 * (k % 5)
                q += rng.normal(0.0, sigma, q.shape)

            q = np.clip(q, 0.015, 0.985)
            trial = np.c_[q, safe_initial_radii(q)]
            result = minimize(
                lambda v: -np.sum(v[2::3]),
                trial.ravel(),
                method="SLSQP",
                bounds=[(0.0, 1.0), (0.0, 1.0), (1e-8, 0.25)] * n,
                constraints={"type": "ineq", "fun": margins},
                options={"maxiter": 1300, "ftol": 1e-11, "disp": False},
            )
            candidate = result.x.reshape(n, 3)
            if np.min(margins(candidate)) < -3e-7:
                continue

            value = float(np.sum(candidate[:, 2]))
            if value > best_sum:
                best, best_sum = candidate.copy(), value

            # Walk locally through near-optimal states, but periodically
            # return to the record if a trial falls too far behind.
            if value >= walker_sum - threshold:
                walker, walker_sum = candidate.copy(), value
            elif value < walker_sum - 3.0 * threshold:
                walker, walker_sum = best.copy(), best_sum

    if best is None:
        row_counts = (5, 5, 5, 5, 6, 6)
        centers = np.asarray([
            ((col + 0.5) / count, (row + 0.5) / 6.0)
            for row, count in enumerate(row_counts)
            for col in range(count)
        ])
        return np.c_[centers, np.full(n, (1.0 / 12.0) * (1.0 - 1e-7))]

    def optimize_fixed_radii(points):
        """Maximize the sum of radii for fixed centers using a linear program."""
        upper = np.min(
            np.column_stack((
                points[:, 0], points[:, 1],
                1.0 - points[:, 0], 1.0 - points[:, 1]
            )),
            axis=1,
        )

        # For fixed centers, every non-overlap condition is linear in r:
        # r_i + r_j <= ||p_i-p_j||.
        m = n * (n - 1) // 2
        A = np.zeros((m, n))
        b = np.empty(m)
        k = 0
        for i in range(n):
            for j in range(i + 1, n):
                A[k, i] = 1.0
                A[k, j] = 1.0
                b[k] = np.linalg.norm(points[i] - points[j])
                k += 1

        result = linprog(
            -np.ones(n),
            A_ub=A,
            b_ub=b,
            bounds=[(1e-10, float(u)) for u in upper],
            method="highs",
        )
        if not result.success:
\end{lstlisting}
\end{bestprogrambox}
\begin{bestprogrambox}{Circle Packing ($n=32$): Python source (lines 241--360)}
\begin{lstlisting}[style=bestprogram, language=BestPython, firstnumber=241]
            return best

        candidate = np.column_stack((points, result.x))
        # Remove only a negligible numerical margin after the LP.
        candidate[:, 2] *= 1.0 - 2e-10
        if np.min(margins(candidate)) >= -1e-10:
            return candidate
        return best

    refined = optimize_fixed_radii(best[:, :2])
    if np.sum(refined[:, 2]) > np.sum(best[:, 2]):
        best = refined

    def slp_polish(state):
        """Improve centers and radii through feasible tangent-plane LP steps."""
        p = state[:, :2].copy()
        r = state[:, 2].copy()

        # Create a tiny strict-feasibility buffer so that the zero step
        # remains feasible despite LP/SLSQP roundoff.
        r *= 1.0 - 2e-8

        for trust in (0.012, 0.008, 0.005, 0.003, 0.0015, 0.0007):
            for _ in range(8):
                nv = 3 * n
                rows = []
                rhs = []

                # Box constraints, written in delta variables.
                for i in range(n):
                    row = np.zeros(nv)
                    row[3 * i + 0] = -1.0
                    row[3 * i + 2] = 1.0
                    rows.append(row)
                    rhs.append(p[i, 0] - r[i])

                    row = np.zeros(nv)
                    row[3 * i + 1] = -1.0
                    row[3 * i + 2] = 1.0
                    rows.append(row)
                    rhs.append(p[i, 1] - r[i])

                    row = np.zeros(nv)
                    row[3 * i + 0] = 1.0
                    row[3 * i + 2] = 1.0
                    rows.append(row)
                    rhs.append(1.0 - p[i, 0] - r[i])

                    row = np.zeros(nv)
                    row[3 * i + 1] = 1.0
                    row[3 * i + 2] = 1.0
                    rows.append(row)
                    rhs.append(1.0 - p[i, 1] - r[i])

                # Tangent lower bounds for every pairwise distance.
                for i in range(n):
                    for j in range(i + 1, n):
                        diff = p[i] - p[j]
                        d = float(np.linalg.norm(diff))
                        if d < 1e-14:
                            continue
                        u = diff / d

                        row = np.zeros(nv)
                        row[3 * i + 0] = -u[0]
                        row[3 * i + 1] = -u[1]
                        row[3 * j + 0] = u[0]
                        row[3 * j + 1] = u[1]
                        row[3 * i + 2] = 1.0
                        row[3 * j + 2] = 1.0
                        rows.append(row)
                        rhs.append(d - r[i] - r[j])

                bounds = []
                for i in range(n):
                    bounds.extend([
                        (-trust, trust),
                        (-trust, trust),
                        (-float(r[i]) + 1e-11, 0.25 - float(r[i])),
                    ])

                lp = linprog(
                    np.tile([0.0, 0.0, -1.0], n),
                    A_ub=np.asarray(rows),
                    b_ub=np.asarray(rhs),
                    bounds=bounds,
                    method="highs",
                )
                if not lp.success:
                    break

                step = lp.x.reshape(n, 3)
                gain = float(np.sum(step[:, 2]))
                if gain <= 1e-11:
                    break

                p += step[:, :2]
                r += step[:, 2]

        out = np.column_stack((p, r))
        # Remove only a negligible common factor for strict evaluator
        # feasibility after the final floating-point LP step.
        out[:, 2] *= 1.0 - 3e-9
        return out

    polished = slp_polish(best)
    if np.min(margins(polished)) >= -2e-9:
        if np.sum(polished[:, 2]) > np.sum(best[:, 2]):
            best = polished

    # SLP changes the centers, so solve the fixed-center radius LP again.
    # This restores exact radius optimality for the polished geometry.
    polished_radii = optimize_fixed_radii(polished[:, :2])
    if np.sum(polished_radii[:, 2]) > np.sum(best[:, 2]):
        best = polished_radii

    # Perform additional small-basin restarts from the polished incumbent.
    # These perturbations are deliberately much smaller than the earlier
    # exploratory walk, targeting improvements that SLSQP may miss because
    # the incumbent lies on a nearly degenerate contact graph.
\end{lstlisting}
\end{bestprogrambox}
\begin{bestprogrambox}{Circle Packing ($n=32$): Python source (lines 361--455)}
\begin{lstlisting}[style=bestprogram, language=BestPython, firstnumber=361]
    polished_base = best[:, :2].copy()
    for k in range(120):
        q = polished_base.copy()

        if k % 10 == 0:
            # Apply a coherent deformation to preserve the broad packing
            # while changing the active contact graph.
            z = q - 0.5
            shear_x, shear_y = rng.normal(0.0, 0.016, 2)
            scale_x, scale_y = 1.0 + rng.normal(0.0, 0.016, 2)
            q = np.column_stack((
                scale_x * z[:, 0] + shear_x * z[:, 1],
                scale_y * z[:, 1] + shear_y * z[:, 0],
            )) + 0.5
            q += rng.normal(0.0, 0.0012, q.shape)
        else:
            # A wider schedule occasionally leaves the current contact basin,
            # while the smallest scales continue to refine tangent neighbors.
            sigma = 0.0002 + 0.0005 * (k % 10)
            q += rng.normal(0.0, sigma, q.shape)

        q = np.clip(q, 0.012, 0.988)
        trial = np.c_[q, safe_initial_radii(q)]

        result = minimize(
            lambda v: -np.sum(v[2::3]),
            trial.ravel(),
            method="SLSQP",
            bounds=[(0.0, 1.0), (0.0, 1.0), (1e-8, 0.25)] * n,
            constraints={"type": "ineq", "fun": margins},
            options={"maxiter": 1500, "ftol": 5e-12, "disp": False},
        )
        candidate = result.x.reshape(n, 3)

        if np.min(margins(candidate)) < -3e-7:
            continue

        # Re-optimize the radii exactly for the newly found centers.  This
        # removes any SLSQP radius suboptimality before comparing incumbents.
        candidate_lp = optimize_fixed_radii(candidate[:, :2])
        if np.min(margins(candidate_lp)) >= -1e-8:
            if np.sum(candidate_lp[:, 2]) > np.sum(best[:, 2]):
                best = candidate_lp

    return best


def compute_max_radii(centers):
    """
    Compute the maximum possible radii for each circle position
    such that they don't overlap and stay within the unit square.

    Args:
        centers: np.array of shape (n, 2) with (x, y) coordinates

    Returns:
        np.array of shape (n) with radius of each circle
    """
    n = centers.shape[0]
    radii = np.ones(n)

    # First, limit by distance to square borders
    for i in range(n):
        x, y = centers[i]
        # Distance to borders
        radii[i] = min(x, y, 1 - x, 1 - y)

    # Then, limit by distance to other circles
    # Each pair of circles with centers at distance d can have
    # sum of radii at most d to avoid overlap
    for i in range(n):
        for j in range(i + 1, n):
            dist = np.sqrt(np.sum((centers[i] - centers[j]) ** 2))

            # If current radii would cause overlap
            if radii[i] + radii[j] > dist:
                # Scale both radii proportionally
                scale = dist / (radii[i] + radii[j]) * 0.99  # 0.99 for safety margin
                radii[i] *= scale
                radii[j] *= scale

    return radii


# EVOLVE-BLOCK-END


# This part remains fixed (not evolved)
def run_code():
    """Run the circle packing constructor for n=32"""
    circles = construct_circles()
    sum_radii = float(np.sum(circles[:, 2]))
    return circles, sum_radii


\end{lstlisting}
\end{bestprogrambox}

\subsection{Discovered Objects}
\label{supp_subsec:sota_objects}

Figures~\ref{supp_fig:sota_quantum}--\ref{supp_fig:sota_math} visualize the routed quantum circuits, trajectories, denoised gene expression, scaling laws, and mathematical objects produced by the best programs, alongside those from the programs or constructions released by SimpleTES. They are redrawn from saved outputs; the mathematical objects come from re-running each best program, which reproduces its reported score exactly.

\begin{figure}[h]
    \centering
    \includegraphics[width=0.9\linewidth]{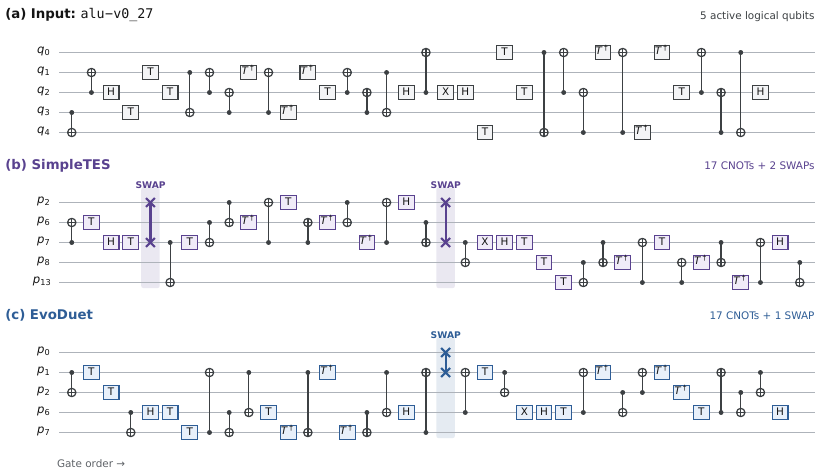}
    \vspace{-0.2em}
    \caption{\textbf{Routed quantum circuits on Q20.} (a)~The \texttt{alu-v0\_27} input circuit. (b,~c)~Outputs of the released SimpleTES program and \methodName's best program under matched replay settings. Colored bands highlight inserted SWAPs: two for SimpleTES and one for \methodName. All gates are shown; only idle wires are omitted. Labels $q_i$ and $p_i$ denote logical qubits and physical sites, respectively.}
    \label{supp_fig:sota_quantum}
\end{figure}

\begin{figure}[h]
    \centering
    \includegraphics[width=0.9\linewidth]{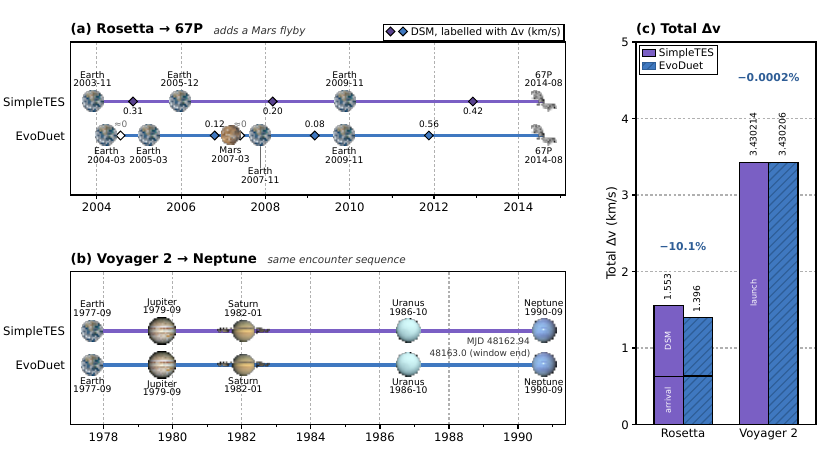}
    \vspace{-0.2em}
    \caption{\textbf{Rosetta and Voyager 2.} (a,~b)~Encounters and dates of the SimpleTES construction and of the best program of \methodName; diamonds are deep-space maneuvers labelled with their $\Delta v$ in km/s. (c)~Total $\Delta v$ split into maneuvers and launch or arrival. On Rosetta, \methodName adds the Mars flyby of the actual mission; on Voyager~2, both follow the same tour and \methodName moves the arrival onto the end of the window. Planet images: NASA (public domain).}
    \label{supp_fig:sota_astro}
\end{figure}

\begin{figure}[h]
    \centering
    \includegraphics[width=0.9\linewidth]{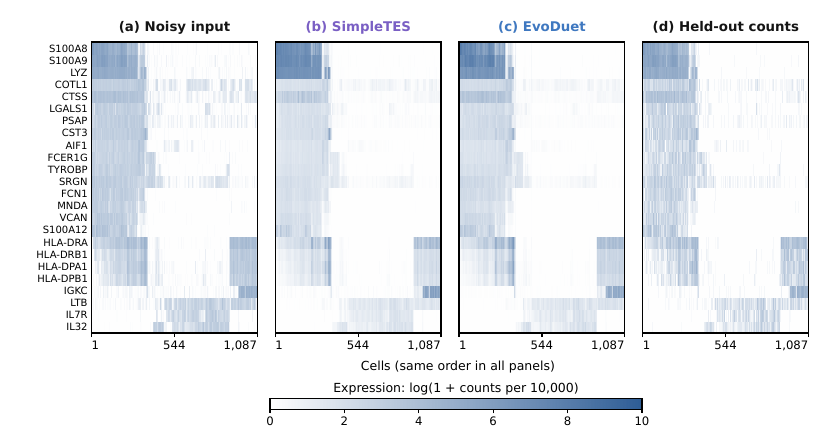}
    \vspace{-0.2em}
    \caption{\textbf{Denoised gene expression.} (a)~Noisy PBMC input, (b,~c)~outputs of the released SimpleTES program and \methodName's best program, and (d)~held-out counts. Columns show all 1{,}087 cells; rows show 24 genes selected for their variability in the input. Gene selection and ordering use only the input, with the same order and color scale across panels. Values are log-normalized per cell. Held-out counts are independent observations and still contain noise.}
    \label{supp_fig:sota_denoising}
\end{figure}

\begin{figure}[h]
    \centering
    \includegraphics[width=0.9\linewidth]{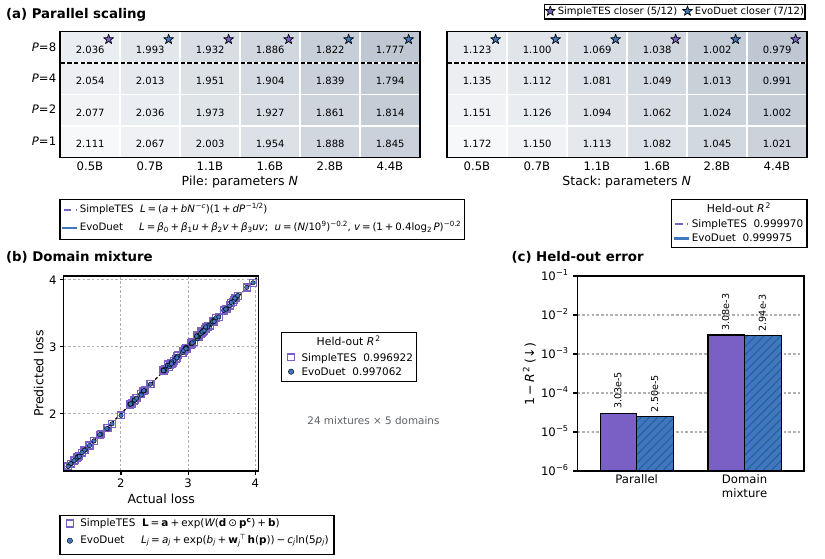}
    \vspace{-0.2em}
    \caption{\textbf{Parallel Scaling and Domain Mixture.} (a)~Observed loss on the grid of parameter count $N$ and parallel streams $P$ for the two Parallel datasets (darker cells indicate lower loss within each dataset); the dashed row ($P=8$) is held out, and a star marks the law that predicts that cell more closely (\methodName in 7 of 12). (b)~Predicted against actual held-out loss on Domain Mixture. (c)~Held-out $1-R^2$ on a log scale. Both margins are small.}
    \label{supp_fig:sota_ai}
\end{figure}

\begin{figure}[h]
    \centering
    \includegraphics[width=0.9\linewidth]{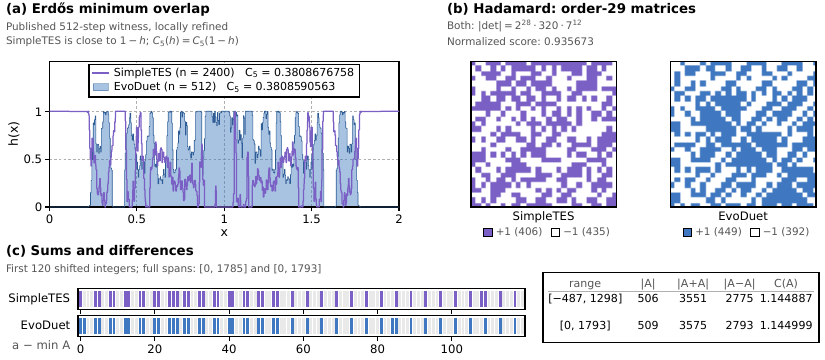}
    \vspace{-0.2em}
    \caption{\textbf{Mathematical objects.} (a)~The Erd\H{o}s step function $h$ of \methodName, which is the published witness the run located and polished, against the SimpleTES construction. (b)~The two order-29 matrices, which reach the same determinant. (c)~The two sum--difference sets; \methodName scores 1.144999, exceeding the released post-training SimpleTES set (1.144887).}
    \label{supp_fig:sota_math}
\end{figure}
\endgroup

\end{document}